\documentclass{article}
\PassOptionsToPackage{square,comma,numbers,compress}{natbib}

\usepackage[preprint]{neurips_2023}

\usepackage[utf8]{inputenc} %
\usepackage[T1]{fontenc}    %
\usepackage[hidelinks,colorlinks=true]{hyperref}       %
\usepackage[dvipsnames]{xcolor}
\usepackage{url}            %
\usepackage{booktabs}       %
\usepackage{amsfonts}       %
\usepackage{nicefrac}       %
\usepackage{microtype}      %
\usepackage{xcolor}         %
\usepackage{float}
\usepackage{graphicx}
\graphicspath{ {figures/} }
\usepackage{array}
\usepackage{times,graphicx,tabulary,multirow,xspace}
\usepackage{subfig}
\usepackage{changepage}
\usepackage{color, colortbl}

\usepackage{amsmath,amssymb,amsthm}
\usepackage{geometry}
\usepackage{mathtools}

\definecolor{LightGray}{rgb}{0.92,0.92,0.92}
\definecolor{Gray1}{rgb}{0.95,0.95,0.95}
\definecolor{Gray2}{rgb}{0.9,0.9,0.9}

\definecolor{redhl}{HTML}{ea9999}
\definecolor{greenhl}{HTML}{d9ead3}
\definecolor{bluehl}{HTML}{c9daf8}
\definecolor{yellowhl}{HTML}{fff2cc}
\makeatletter
\DeclareRobustCommand\onedot{\futurelet\@let@token\@onedot}
\def\@onedot{\ifx\@let@token.\else.\null\fi\xspace}

\def\ie{\emph{i.e}\onedot} 
 
\def\etc{\emph{etc}\onedot}

\makeatother

\newcommand{\modelname}{GPT-4o\xspace}
\newcommand{\modelnamefull}{GPT-4o(mni)\xspace}

\title{Preliminary Explorations with \modelnamefull \\ Native Image Generation}

\author{
{\bf Pu Cao$^{\dagger}$, Feng Zhou$^{*}$, Junyi Ji$^{*}$, Qingye Kong$^{*}$, Zhixiang Lv$^{*}$,Mingjian Zhang$^{*}$} \\ 
\textbf{Xuekun Zhao$^{*}$, Siqi Wu, Yinghui Lin, Qing Song, Lu Yang$^{\spadesuit,\dagger}$} \\
\and
Beijing University of Posts and Telecommunications\\
\and
\footnotesize{
$^*$~Equally Contribution\;
$^{\dagger}$~Project Leader\;
$^{\spadesuit}$~Correspondence Author \;
}\\
\and
\small \texttt{\{caopu,soeaver\}@bupt.edu.cn}
}

\begin{document}

\maketitle

\vspace{20pt}
\begin{abstract}
Recently, the visual generation ability by \modelnamefull has been unlocked by OpenAI\footnote{\url{https://openai.com/index/introducing-4o-image-generation/}}. It demonstrates a very remarkable generation capability with excellent multimodal condition understanding and varied task instructions.
In this paper, we aim to explore the capabilities of GPT-4o across various tasks. Inspired by previous study\citep{yang2023dawn}, we constructed a task taxonomy along with a carefully curated set of test samples to conduct a comprehensive qualitative test. Benefiting from GPT-4o's powerful multimodal comprehension, its image-generation process demonstrates abilities surpassing those of traditional image-generation tasks. Thus, regarding the dimensions of model capabilities, we evaluate its performance across six task categories: traditional image generation tasks, discriminative tasks, knowledge-based generation, commonsense-based generation, spatially-aware image generation, and temporally-aware image generation. These tasks not only assess the quality and conditional alignment of the model's outputs but also probe deeper into \modelname's understanding of real-world concepts. 
Our results reveal that GPT-4o performs impressively well in general-purpose synthesis tasks, showing strong capabilities in text-to-image generation, visual stylization, and low-level image processing. However, significant limitations remain in its ability to perform precise spatial reasoning, instruction-grounded generation, and consistent temporal prediction. Furthermore, when faced with knowledge-intensive or domain-specific scenarios, such as scientific illustrations or mathematical plots, the model often exhibits hallucinations, factual errors, or structural inconsistencies. These findings suggest that while \modelname marks a substantial advancement in unified multimodal generation, there is still a long way to go before it can be reliably applied to professional or safety-critical domains.

\end{abstract}
% \vspace{20pt}

{
  \hypersetup{linkcolor=black}
  \tableofcontents
  \label{sec:toc}
}

\clearpage
{
\hypersetup{linkcolor=black}
\addcontentsline{toc}{section}{List of Figures}
\listoffigures
\label{sec:lof}
}
\clearpage

%\input{01-intro}
%\input{02-method}
%\clearpage
%\input{03-VL}
%\clearpage
%\input{04-language}
%\clearpage
%\input{05-pointing}
%\clearpage
%\input{06-temporal}
%\clearpage
%\input{07-IQ}
%\clearpage
%\input{08-EQ}
%\clearpage
%\input{09-app}
%\clearpage
%\input{10-future}
%\input{11-limitations-conclusions}

\section{Introduction}
\label{sec:intro}

OpenAI's latest image generation model, native image generation mode of \modelnamefull\footnote{\url{https://openai.com/index/introducing-4o-image-generation/}}, has demonstrated remarkable generative capabilities, attracting widespread attention in both the research community and social media. Distinct from previous generative models, GPT-4o exhibits exceptional task generalization ability. Benefiting from being embedded within a large language model (LLM), it also demonstrates strong information comprehension capabilities. Therefore, inspired by GPT4-V(ision) exploration study\cite{yang2023dawn}, conducting extensive and systematic evaluations of this model is highly beneficial. Such evaluations will help precisely delineate the model's strengths and limitations and provide insights for guiding future technological advancements.

Our exploration of \modelname is guided by the following perspectives:
\begin{itemize} 
\item \textbf{Overall Characteristics of \modelname's Image Generation.} In Sec.~\ref{sec:overall}, we first investigate the overall characteristics of \modelname's image generation capabilities, including generative resolution, aspect ratio, numerical constraints, and input limitations, \etc. \modelname does not perform well in generating images with specific resolutions, which consequently impacts the control over aspect ratios.

\item \textbf{Visual Synthesis Quality and Conditional Alignment.} These two dimensions are critically important in image generation tasks. In Sec.~\ref{sec:tradition}, we conduct extensive experiments on various traditional image-generation scenarios, including text-conditioned image generation, multimodal-conditioned image generation, and even low-level image processing tasks, \etc. This comprehensive analysis allows us to thoroughly assess the synthesis quality and conditional alignment capabilities of the model. \modelname performs well in most of these tasks.

\item \textbf{Visual Understanding.} Leveraging its intrinsic multimodal comprehension capability~\cite{yang2023dawn}, \modelname can effectively handle discriminative tasks by directly visualizing task outputs on images. In Sec.~\ref{sec:discrimative}, we explore its performance across diverse discriminative scenarios and observe that visual and textual outputs exhibit distinct performance patterns across different tasks. In most discriminative tasks, \modelname relies primarily on global semantic cues from the image rather than precise visual reasoning or task-specific interpretation.

\item \textbf{Knowledge and Commonsense-based Generation.} In Sec.\ref{sec:knowledge} and Sec.\ref{sec:commonsense}, we employ carefully designed prompts that require a model's understanding of domain knowledge and commonsense reasoning, aiming to evaluate whether \modelname can effectively leverage such knowledge during image generation. This capability also serves as a critical indicator of whether the model can be regarded as a world model. Unfortunately, we find that \modelname struggles when reasoning about complex real-world scenarios.

\item \textbf{Spatial Reasoning. } In Sec.~\ref{sec:spatial}, we investigate the spatially-aware image generation capabilities of GPT-4o, including its understanding of viewpoints,  positions, and the logical relationships between entities in space. Our tests vary across different object types and scene contexts to examine the boundaries of the model’s spatial understanding capabilities. GPT-4o exhibits generally acceptable performance across most tests, though its spatial precision remains limited.

\item \textbf{Temporal Reasoning. } In Sec.~\ref{sec:temporal}, we explored GPT-4o’s temporal reasoning capabilities, primarily by prompting it to predict the outcome of specific frames within a video sequence. While GPT-4o demonstrated some ability in temporal reasoning, there remains significant room for improvement.

\end{itemize}

Based on the aforementioned analyses and experimental results, we further discuss the current limitations encountered by the image generation model in Sec.~\ref{sec:sc}. Based on the aforementioned analyses and experimental results, we further discuss the current limitations encountered by the image generation model in Sec.~\ref{sec:sc}. Specifically, we identify four major limitations: 
(1) \textbf{Inadequate real-world modeling}, which manifests in the model’s inability to reliably reflect physical laws, commonsense knowledge, or temporal continuity—indicating a clear gap between \modelname and a true world model;
(2)\textbf{Weak generation process control}, where the model lacks fine-grained control over resolution, aspect ratio, and pixel-level numerical properties, limiting its suitability for applications requiring structural precision;
(3) \textbf{Lack of spatial alignment}, where \modelname fails to generate outputs that adhere to spatial constraints or reference structures, especially in layout-controlled, segmentation, and pose-conditioned tasks; 
and (4)\textbf{Instruction misalignment}, as the model often fails to correctly interpret or follow task definitions and textual instructions, even when aided by in-context examples; 
These findings highlight fundamental limitations in the current design of general-purpose vision-language models and point to critical directions for future improvements.

It is important to note that this report primarily focuses on a qualitative exploration of the diverse capabilities exhibited by \modelname, rather than providing rigorous quantitative performance metrics~\cite{yan2025gpt} or focusing comparisons with other models~\cite{chen2025empirical} as done in prior studies. For each experimental task, we have manually curated a representative set of instruction prompts to demonstrate the effectiveness and generalization ability of the model. Some of these prompts were carefully designed or automatically generated with the assistance of ChatGPT to ensure diversity and comprehensive coverage. Nevertheless, during our tests, we observed occasional instability and randomness, including instances where \modelname may unexpectedly fail or explicitly refuse to generate the desired images.

\section{Overall Characteristics of \modelname Image Generation}
\label{sec:overall}

In this section, we begin by examining the overall characteristics of \modelname in the context of image generation. Specifically, we analyze key factors such as the default and maximum supported image resolution, the model's ability to handle and preserve various aspect ratios, as well as its behavior in response to numerically constrained inputs. These foundational properties provide essential insights into the model’s generative framework and help establish a baseline understanding of its strengths and limitations in producing structurally consistent and resolution-aware visual content.

\subsection{Image Resolution}
\label{sec:res}

\begin{figure}[h]
    \centering
	\includegraphics[width=1.0\linewidth]{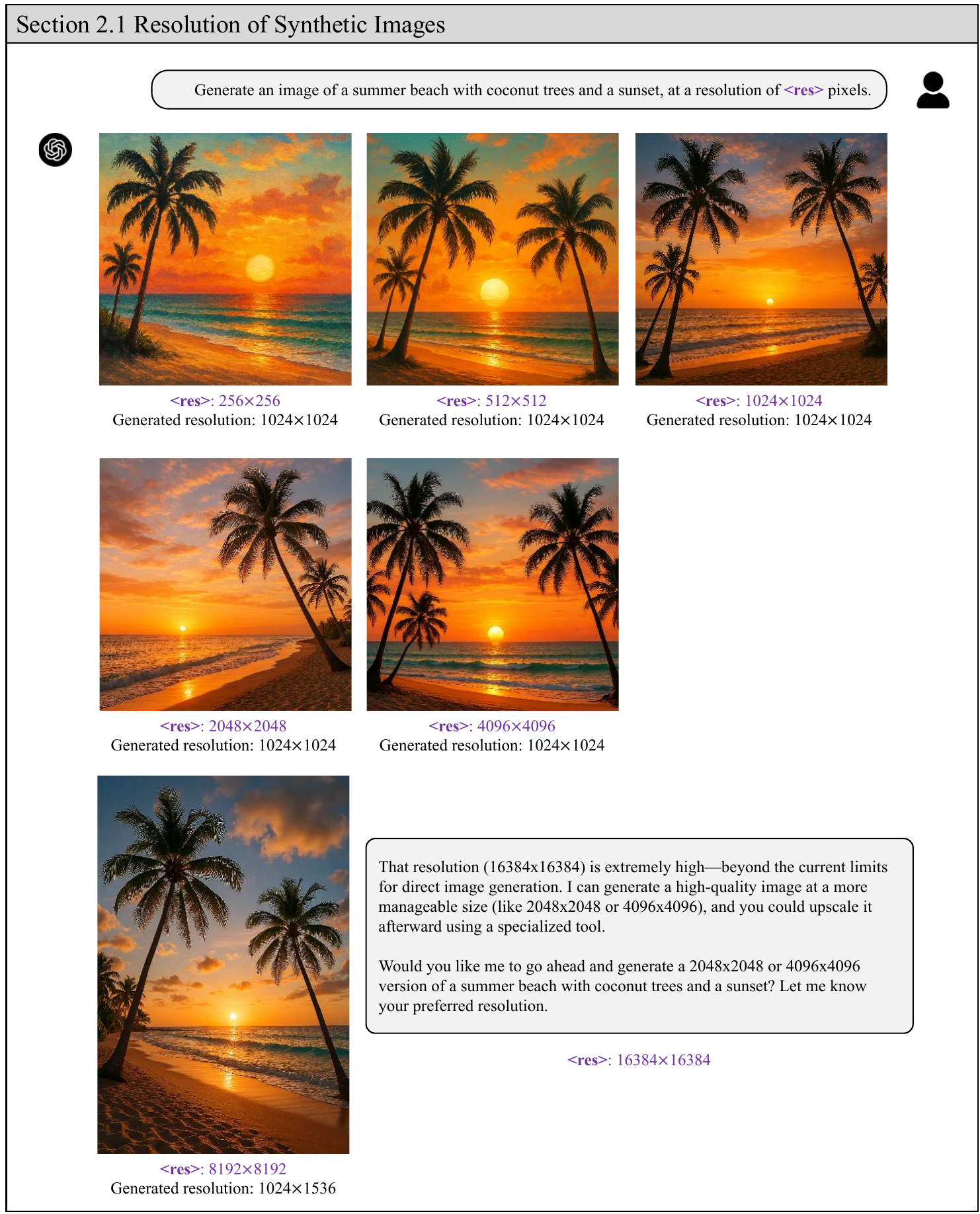}
	\caption[Section~\ref{sec:res}: Image Resolution]{Despite resolution-specific prompts, \modelname consistently outputs images at $1024\times 1024$. In cases of extremely high-resolution requests (\ie, $16384\times 16384$), the model refuses to generate any result, indicating a limitation in resolution controllability.}
	\label{fig:resolution}
\end{figure}

In Fig.~\ref{fig:resolution}, we prompt \modelname to generate identical content across a range of target resolutions, including $256\times 256$, $512\times 512$, $1024\times 1024$, $2048\times 2048$, $4096\times 4096$, $8192\times 8192$, and $16384\times 16384$. However, we observe that \textbf{\modelname fails to precisely control image resolution} and can only produce outputs in three fixed sizes: $1024\times 1024$, $1024\times 1536$, and $1536\times 1024$.

Specifically, for any resolution prompt below $4096\times 4096$, \modelname consistently outputs images in $1024\times 1024$. When higher resolutions such as $8192\times 8192$ are requested, the model instead returns images in either $1024\times 1536$ or $1536\times 1024$. For extremely high-resolution prompts (\ie, $16384\times 16384$), \modelname directly refuses to generate any result. This limitation is further confirmed in subsequent experiments and highlights a significant constraint in resolution control within the current version of the model.

Although \modelname struggles with explicit resolution control, we observe that the richness and scale of the generated content tend to improve when higher-resolution prompts are provided. For instance, as shown in Fig.~\ref{fig:resolution}, the image generated with a $256\times 256$ prompt contains fewer scene elements and presents a narrower field of view. In contrast, the prompts with $512\times 512$ and $1024\times 1024$ result in more detailed scenes with broader perspectives. This suggests that, despite its inability to strictly adhere to resolution constraints, \modelname is capable of interpreting the semantic meaning of resolution specified in the prompt and adjusting the scale and content density of the generated image accordingly.

\subsection{Aspect Ratio}
\label{sec:aspect}

\begin{figure}[h]
    \centering
	\includegraphics[width=1.0\linewidth]{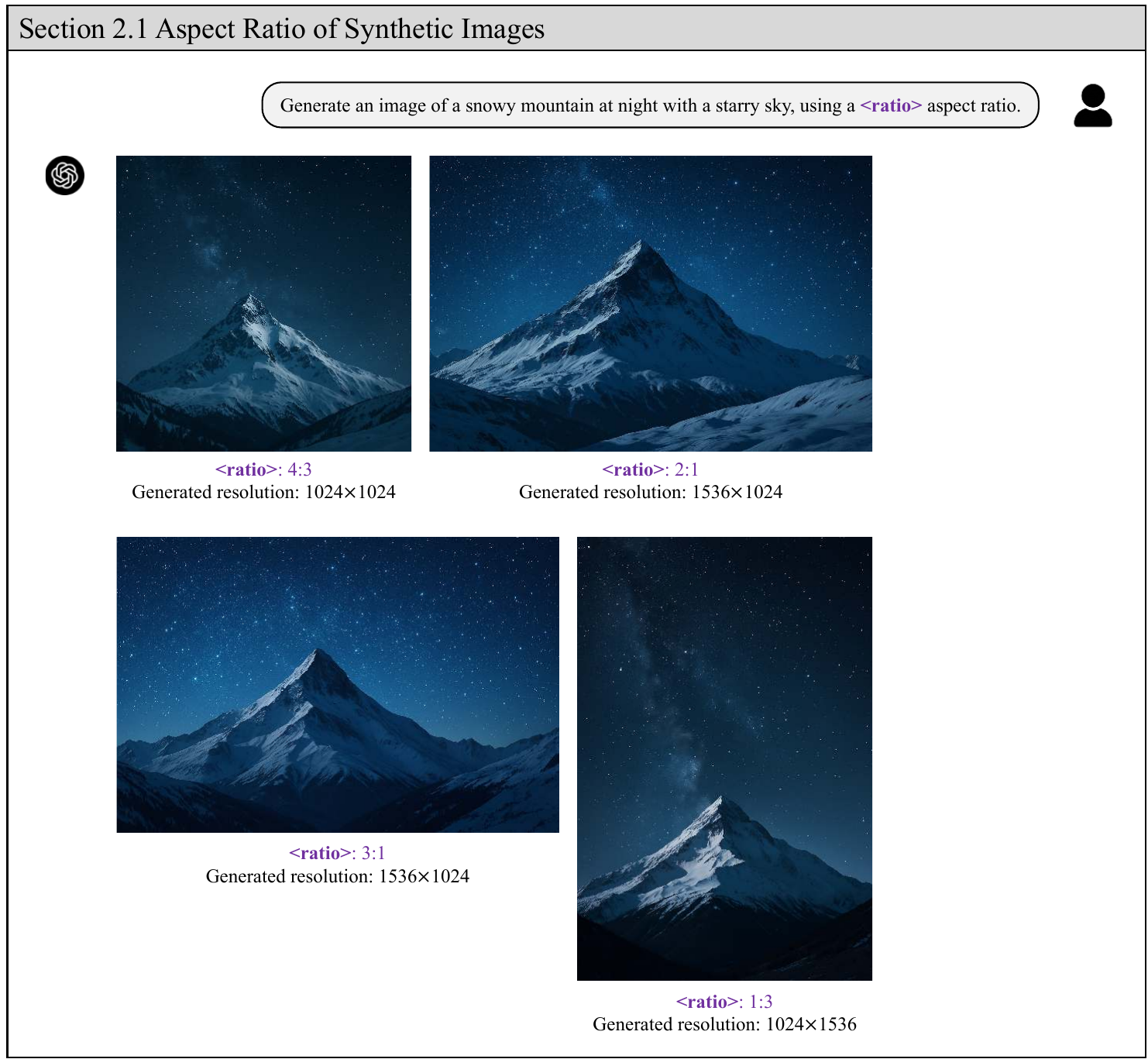}
	\caption[Section~\ref{sec:aspect}: Aspect Ratio]{\modelname is unable to strictly follow aspect ratio prompts and instead selects from a limited set of resolutions. For instance, it outputs $1024\times 1024$ for a 4:3 prompt, and $1536\times 1024$ for wider ratios such as 2:1 or 3:1, approximating the requested aspect ratios using the closest available resolution.}
	\label{fig:aspect}
\end{figure}

Due to the fact that \modelname can only generate images in three fixed resolutions, it is also unable to directly control the aspect ratio as specified in the prompt. Instead, \textbf{\modelname tends to select the closest available resolution that approximates the desired aspect ratio}, as shown in Fig~\ref{fig:aspect}. For example, when a 4:3 aspect ratio is requested, the model outputs an image at $1024\times 1024$. In contrast, for aspect ratios such as 2:1 or 3:1, the model typically generates images at $1536\times 1024$.

\subsection{Numerical Property}
\label{sec:numerical}

\begin{figure}[h]
    \centering
	\includegraphics[width=1.0\linewidth]{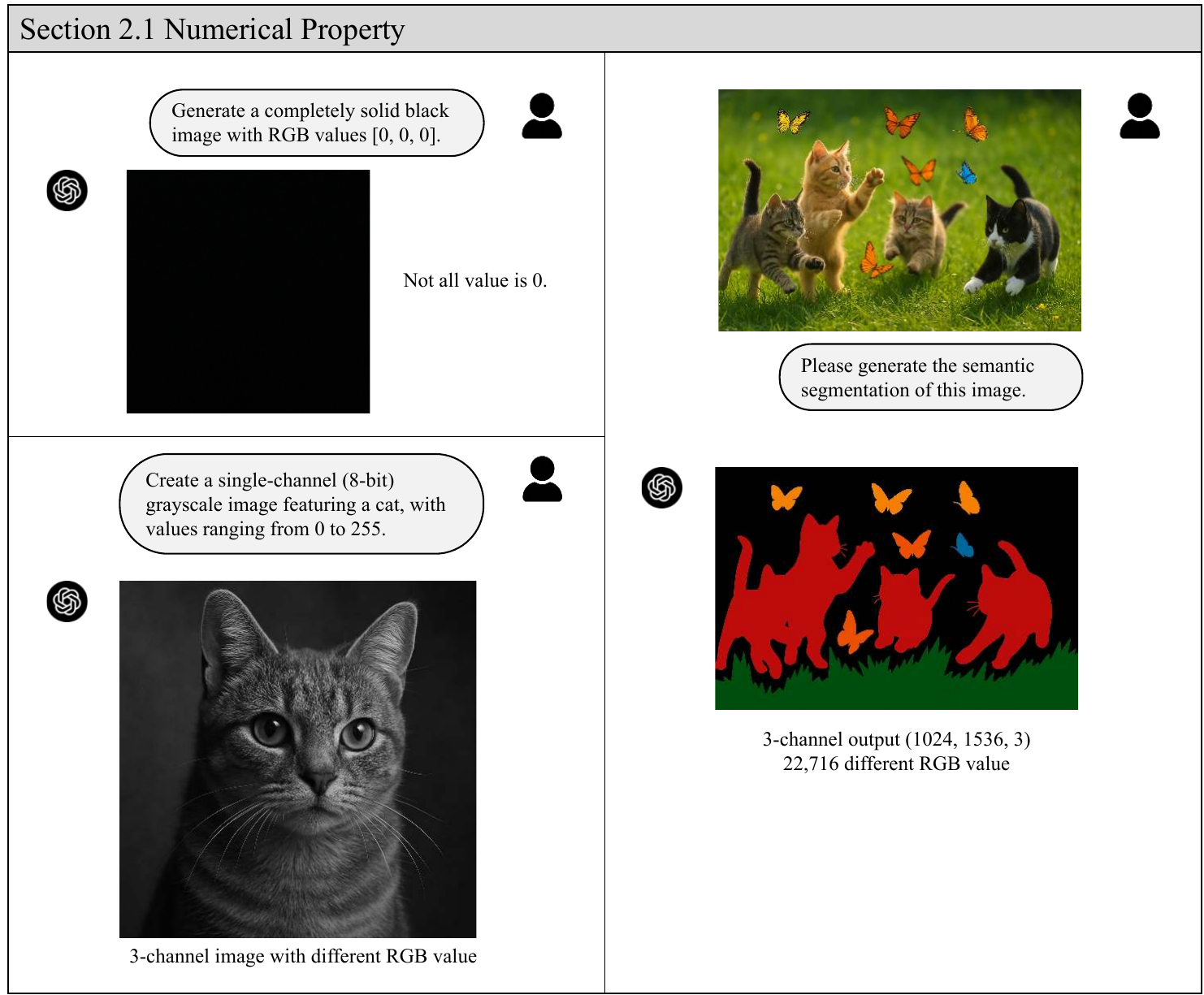}
	\caption[Section~\ref{sec:numerical}: Numerical Property]{}
	\label{fig:numerical}
\end{figure}

In Fig.~\ref{fig:numerical}, we attempt to control the numerical properties of the output images. Across all three scenarios, \textbf{\modelname fails to generate outputs with precise numerical values}, as verified using Python-based analysis.

For instance, when prompted to generate a completely black image with RGB values of $[0,0,0]$, the resulting image appears visually black but still contains a range of low non-zero RGB values. Similarly, when tasked with generating grayscale images, \modelname produces standard three-channel RGB images where the channel values are not equal, thereby deviating from true grayscale.

Additionally, when we prompt the model to generate segmentation masks with five distinct regions, the output appears to contain five visually distinguishable colors. However, a pixel-wise analysis reveals a total of 22,716 unique RGB values across the $1024\times1536 = 1,572,864$ pixels. This suggests that \modelname generates results aligned with human visual perception, rather than maintaining strict pixel-level numerical accuracy.

\clearpage

\clearpage

\section{Traditional Image Generation Tasks}
\label{sec:tradition}

\subsection{Text-Conditioned Image Generation}
\label{sec:t2i}
Text-conditioned image generation is a fundamental capability of multimodal generative models, where the model is expected to synthesize high-quality images that align closely with the semantics and visual cues described in a given prompt. In this section, we evaluate \modelname's performance across three representative tasks, including text-to-image synthesis, text rendering, and document image generation. Together, these tasks probe both the general semantic alignment and the fine-grained spatial and linguistic understanding of \modelname.

\subsubsection{Text-to-Image Generation}
\label{sec:basict2i}
Text-to-image generation has become one of the most representative benchmarks in multimodal learning, reflecting a model's ability to ground linguistic semantics into coherent visual outputs. With the emergence of diffusion-based\cite{nichol2021glide,razzhigaev2023kandinsky,kim2023diffblender,saharia2022photorealistic,rombach2022high,podell2023sdxl} and autoregressive-based\cite{han2024infinity,tschannen2024jetformer,gu2024dart,yu2022parti,wang2024emu3} generative models, recent works have demonstrated remarkable progress in both visual fidelity and semantic alignment. However, challenges remain in faithfully capturing fine-grained textual details, resolving compositional complexity, and generalizing to open-domain prompts. We evaluate \modelname on general text-to-image generation tasks, where the prompts range from simple object-centric descriptions to imaginative and surreal scenes\cite{zhang2024itercomp}.

\begin{figure}[h]
    \centering
    \includegraphics[width=1.0\linewidth]{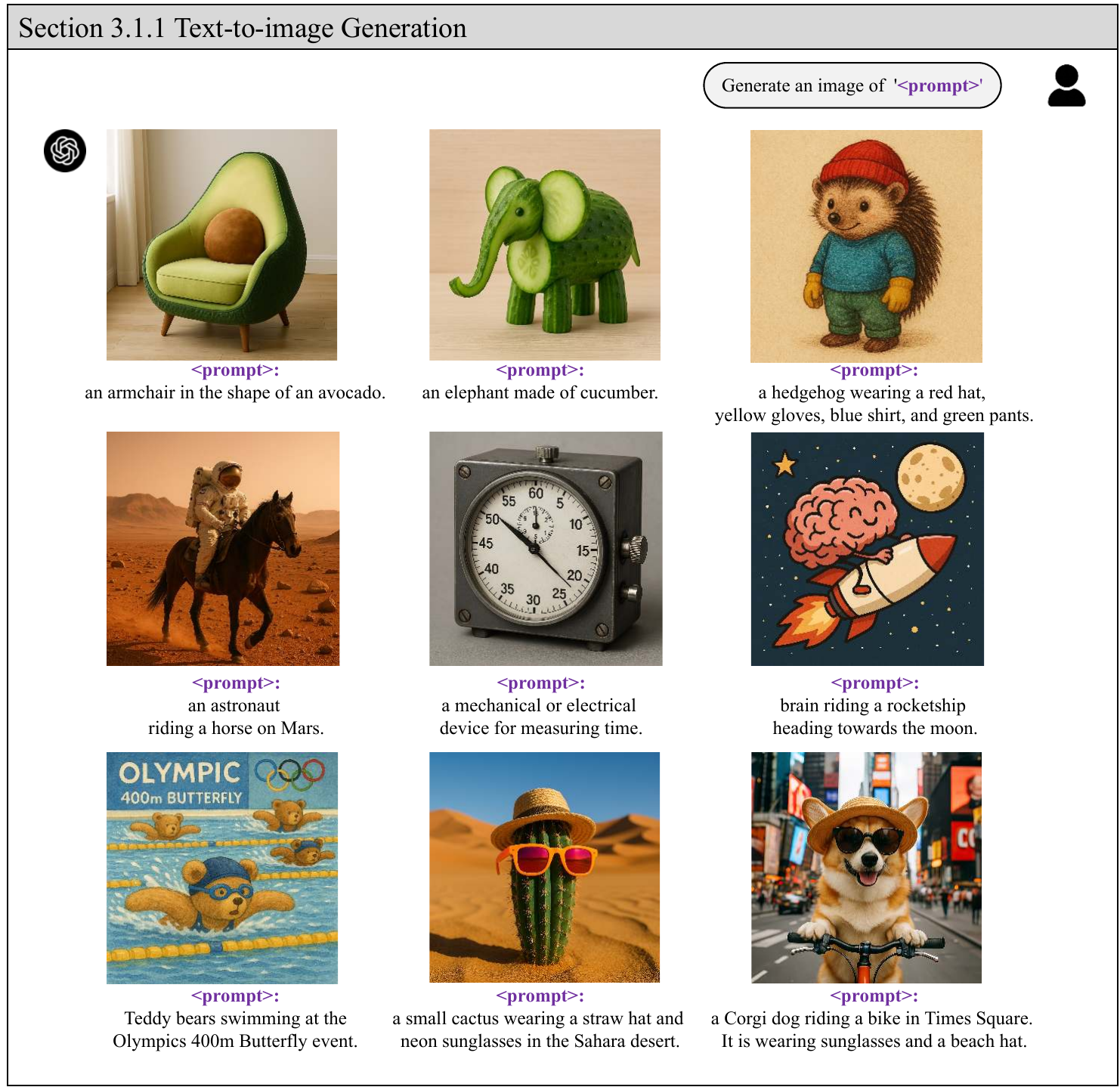}
    \caption[Sec~\ref{sec:basict2i}: Text-to-image Generation]{Examples of text-to-image generation results by \modelname.}
    \label{fig:t2i}
\end{figure}

\begin{figure}[h]
    \centering
    \includegraphics[width=1.0\linewidth]{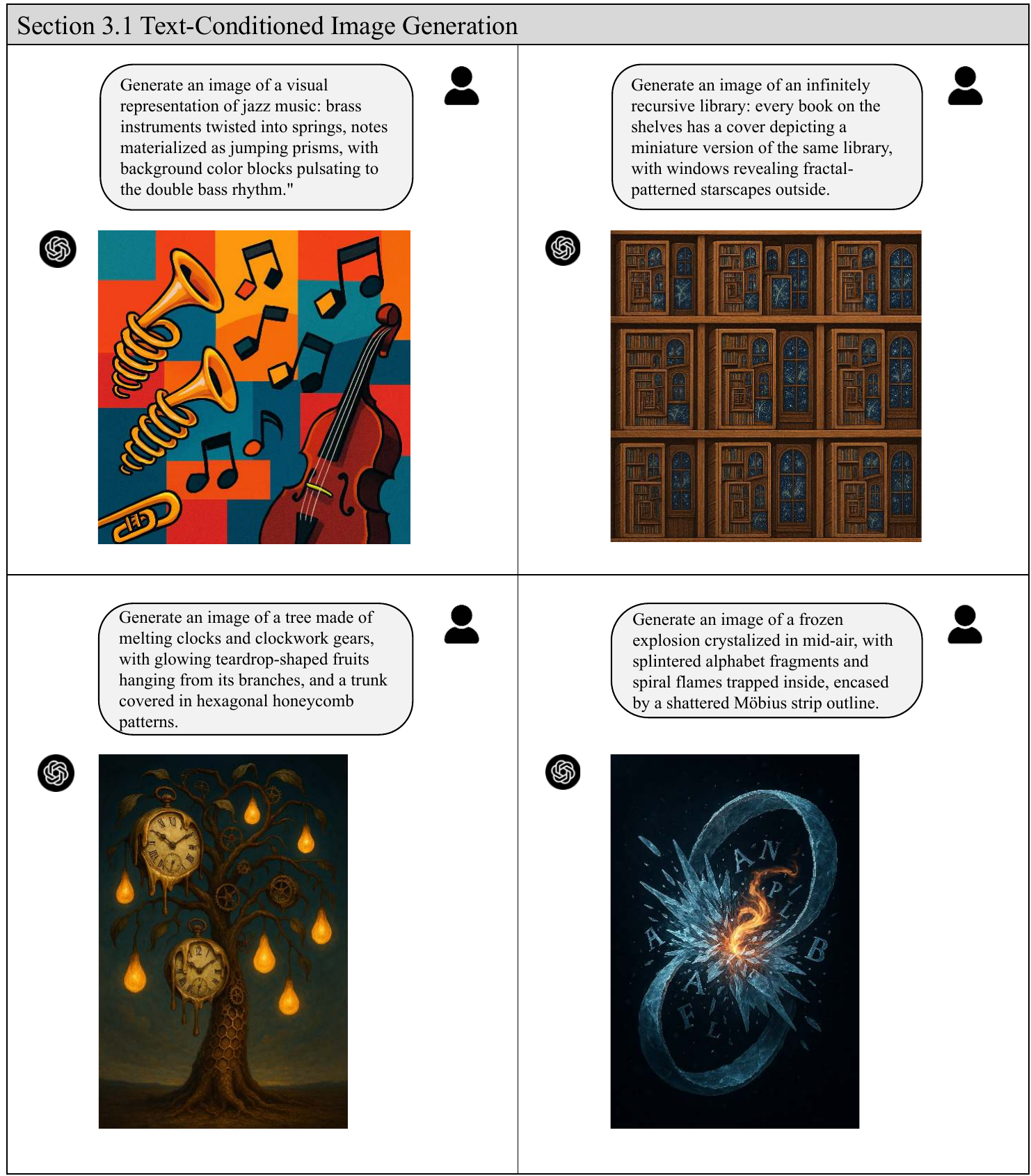}
    \caption[Sec~\ref{sec:basict2i}: Abstract Text-to-image Generation]{Examples of abstract text-to-image generation results by \modelname.}
    \label{fig:abstract-t2i}
\end{figure}

% TODO
As shown in Fig.~\ref{fig:t2i} and Fig.~\ref{fig:abstract-t2i}, \modelname demonstrates strong capabilities in generating visually coherent, vivid, and semantically aligned images across a wide range of creative prompts. It correctly grounds object attributes (e.g., “a dragon fruit wearing a karate belt in the snow”), scene settings (e.g., “a corgi dog riding a bike in Times Square”), and compositional instructions (e.g., “a brain riding a rocketship”). These results indicate that the model has a strong grasp of object composition, spatial reasoning, and multimodal imagination.

\begin{figure}[h]
    \centering
    \includegraphics[width=1.0\linewidth]{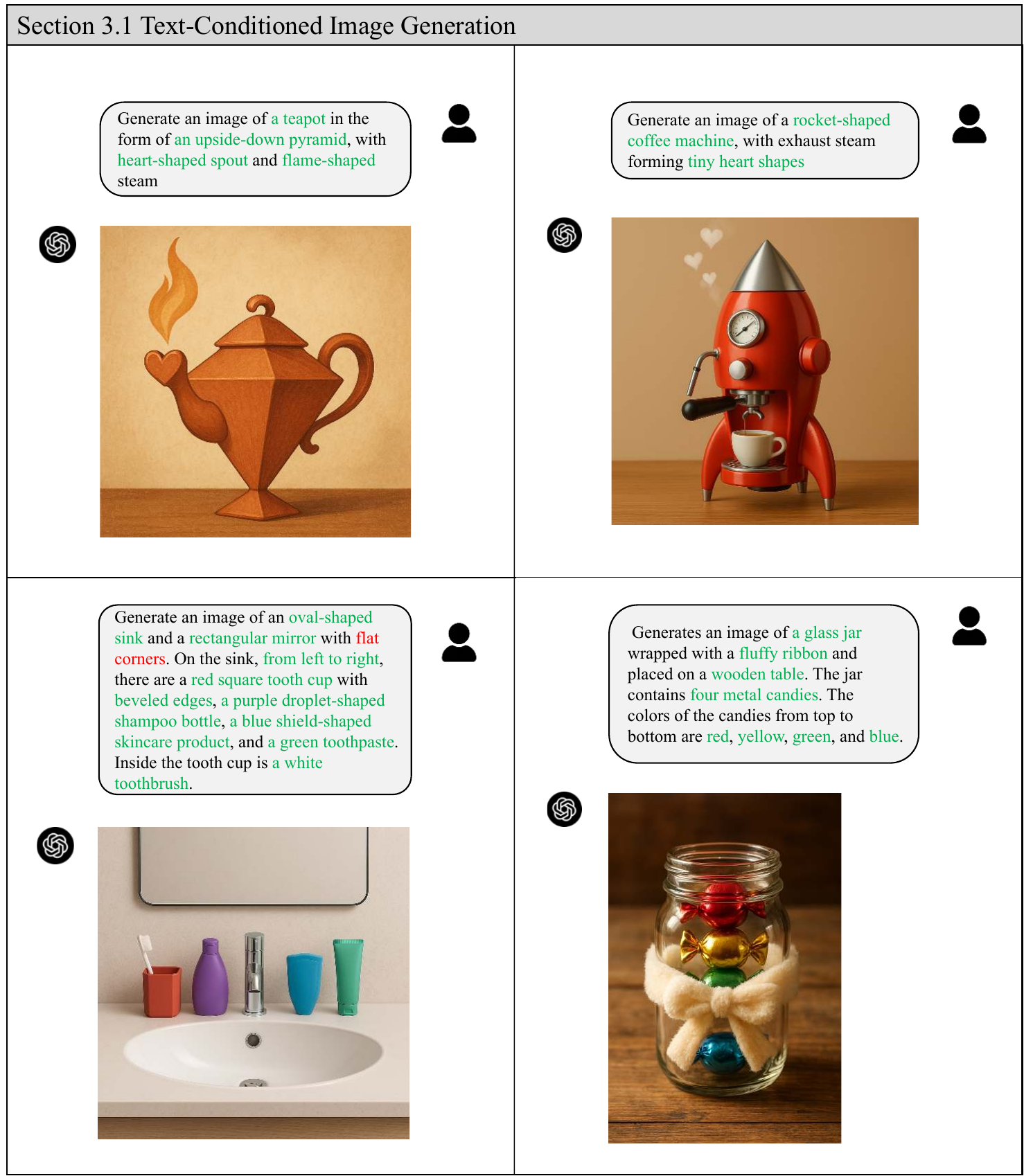}
    \caption[Sec~\ref{sec:basict2i}: Complex Text-to-image Generation]{Examples of complex text-to-image generation results by \modelname.}
    \label{fig:complextext}
\end{figure}

To further probe the model's fine-grained understanding of compositional instructions, we test it with more complex prompts involving multiple objects, attributes, spatial relations, and fine structure. As illustrated in Fig.~\ref{fig:complextext}, the prompts include multi-part visual arrangements such as detailed sink setups and customized product placement. While \modelname generally captures the scene layout and major object categories, we observe occasional failures in exact spatial arrangement and object details. For example, the flat corner of the mirror is underrepresented in Fig.~\ref{fig:complextext}. These results reveal that while the model performs well on general descriptive prompts, it still faces challenges in faithfully capturing intricate compositional constraints.

\clearpage

\clearpage

\subsubsection{Text Rendering}
\label{sec:textrendering}
Text rendering is a challenging subtask of text-conditioned image generation, where the model is required not only to synthesize visual content, but also to accurately render specified textual content as an integral part of the image\cite{Ramesh2022HierarchicalTI,liu2022character,ma2023glyphdraw,chen2023textdiffuser,tuo2023anytext,chen2023textdiffuser2,zhao2023udifftext}. This task tests the model's fine-grained spatial control, font fidelity, multilingual capability, and ability to balance visual aesthetics with text legibility.

\begin{figure}[h]
    \centering
    \includegraphics[width=1.0\linewidth]{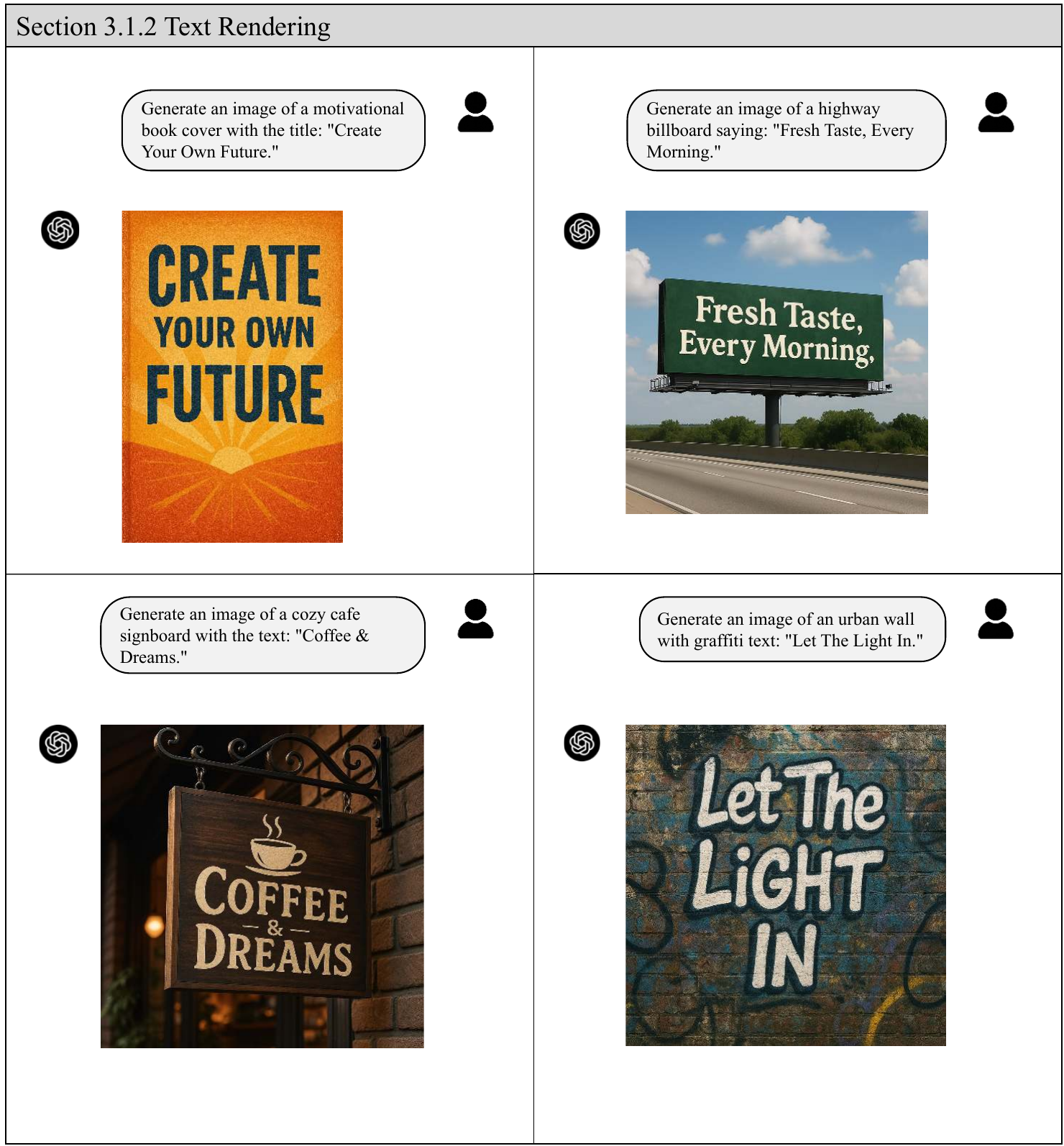}
    \caption[Sec~\ref{sec:textrendering}: Text Rendering]{Examples of text rendering in stylized scenes. \modelname embeds short texts into posters, signs, and graffiti with moderate accuracy.}
    \label{fig:textrendering}
\end{figure}

As shown in Fig.~\ref{fig:textrendering}, \modelname is able to generate plausible images that embed short, stylized texts across a variety of contexts such as signage, graffiti, posters, and book covers. The generated results generally follow the given prompts, with appropriate font placement and contextual consistency. However, minor issues remain in font clarity and character shape consistency, especially in artistic scenes.

\begin{figure}[h]
    \centering
    \includegraphics[width=1.0\linewidth]{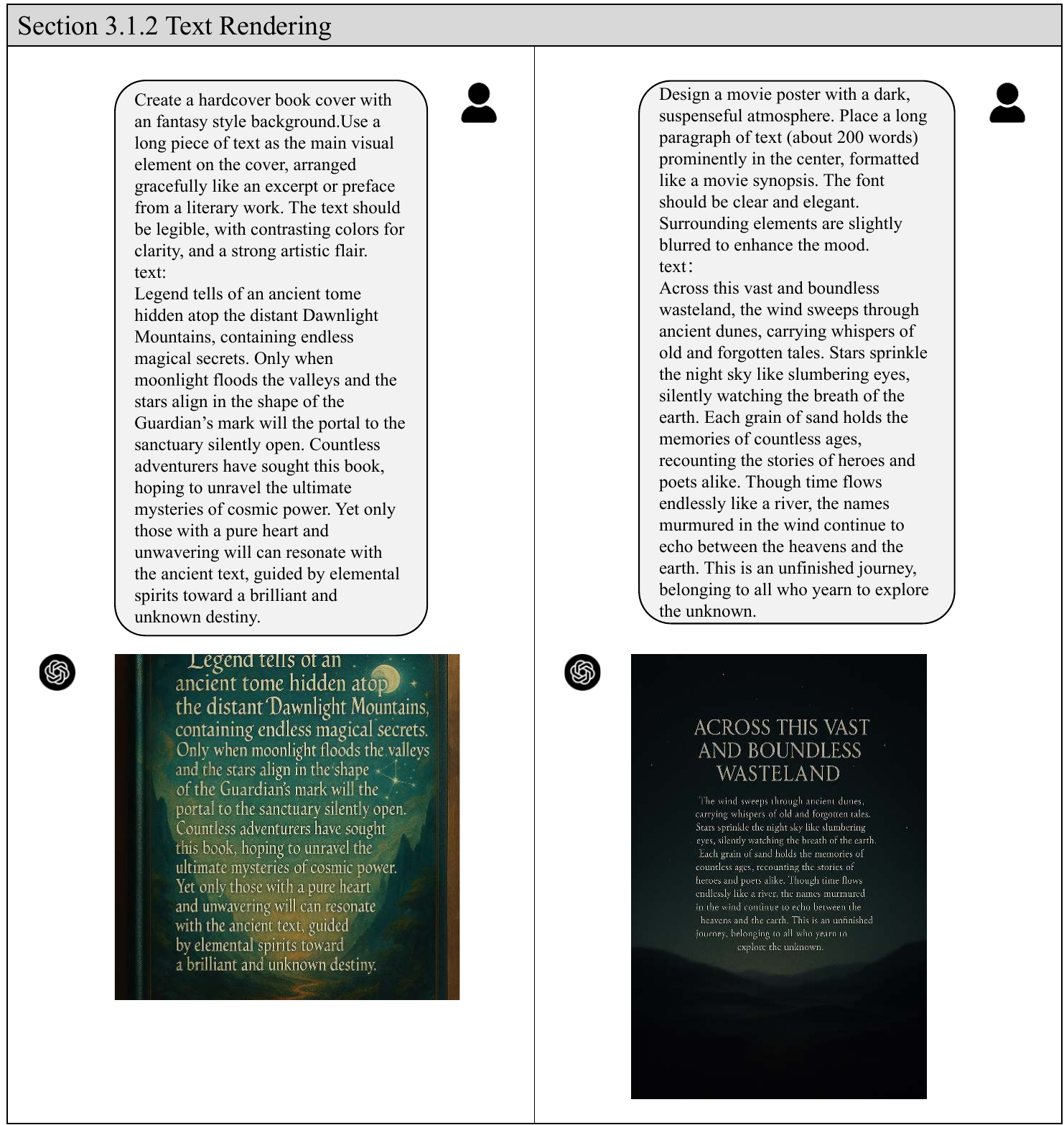}
    \caption[Sec~\ref{sec:textrendering}: Long Text Rendering]{Examples of long text rendering tasks, including book cover paragraphs and movie synopses.}
    \label{fig:textrendering-long}
\end{figure}

To further test the model’s capacity for handling extended textual content, we introduce long paragraph prompts including literary excerpts and movie synopses. As shown in Fig.~\ref{fig:textrendering-long}, \modelname demonstrates basic formatting capability, generating legible long-form text with consistent line layout and suitable contrast. 

\begin{figure}[h]
    \centering
    \includegraphics[width=1.0\linewidth]{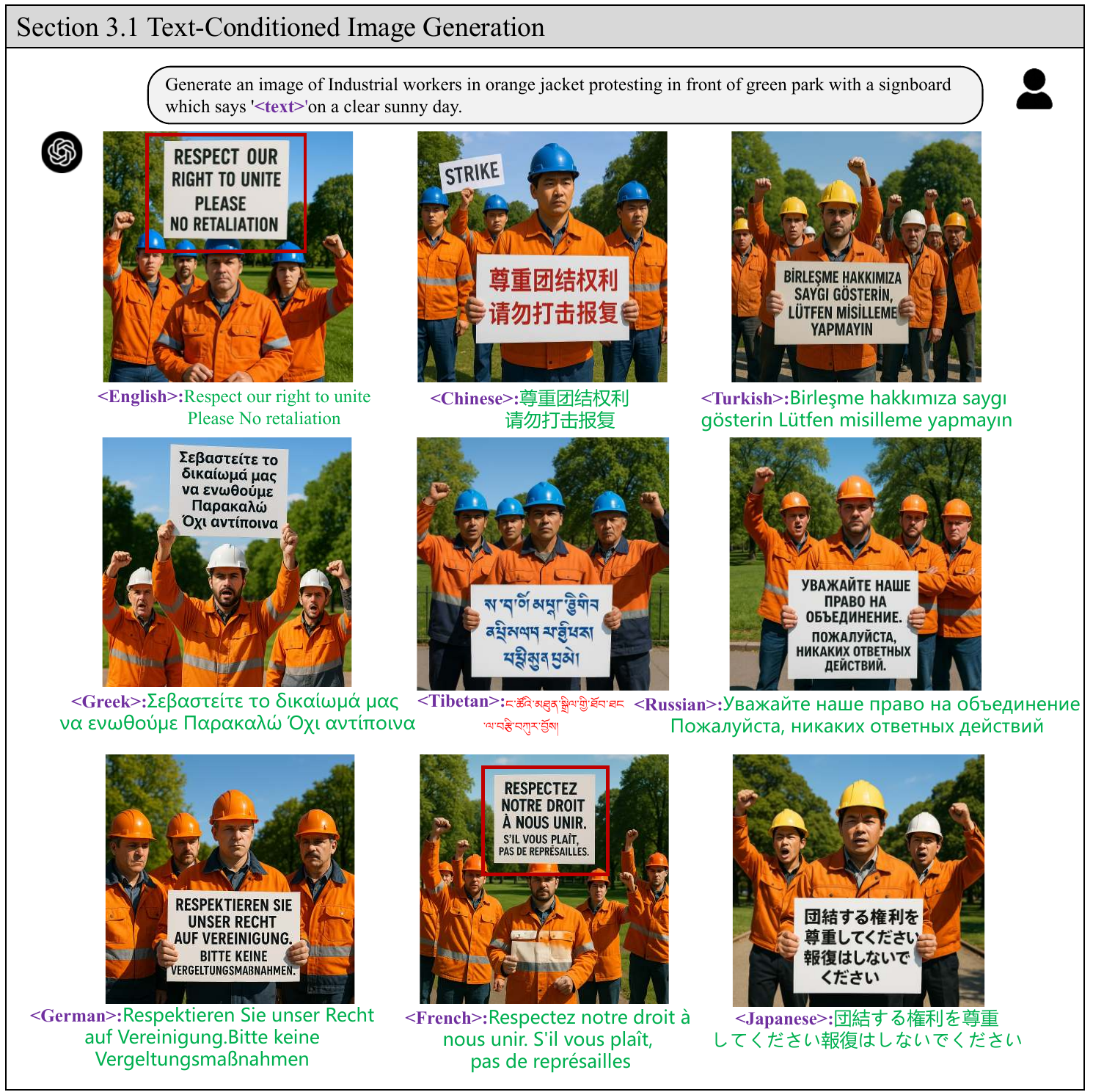}
    \caption[Sec~\ref{sec:textrendering}: Multilingual Text Rendering]{Multilingual text rendering across nine languages. Non-Latin scripts often show rendering issues.}
    \label{fig:textrendering-multilingual}
\end{figure}

Additionally, we examine the model’s ability to render multilingual content by prompting it with a variety of languages including English, Chinese, Turkish, Tibetan, Russian, and Japanese (Fig.~\ref{fig:textrendering-multilingual}). While \modelname performs reasonably well on most of the scripts, it struggles with Tibetan, which may be caused by its limited data. In addition, the generated images occasionally contain semantic or contextual errors (e.g., a sign appearing to float instead of being held). These issues can arise even in high-frequency languages such as English.

Overall, \modelname demonstrates a promising ability in text rendering tasks, especially for short English phrases in stylistic contexts. However, its robustness in multilingual, long-form, or precise layout-sensitive scenarios still requires significant improvement.

\clearpage

\clearpage

\subsubsection{Document Image Generation}
\label{sec:document}
Document image generation refers to the task of producing structured visual outputs, such as documents, tables, and charts—based on textual specifications. This task requires not only semantic understanding and layout composition, but also spatial reasoning, formatting alignment, and numerical precision. We evaluate \modelname across four representative subtasks: textual document layout generation, table generation, catalog composition, and chart visualization.

\begin{figure}[h]
    \centering
    \includegraphics[width=1.0\linewidth]{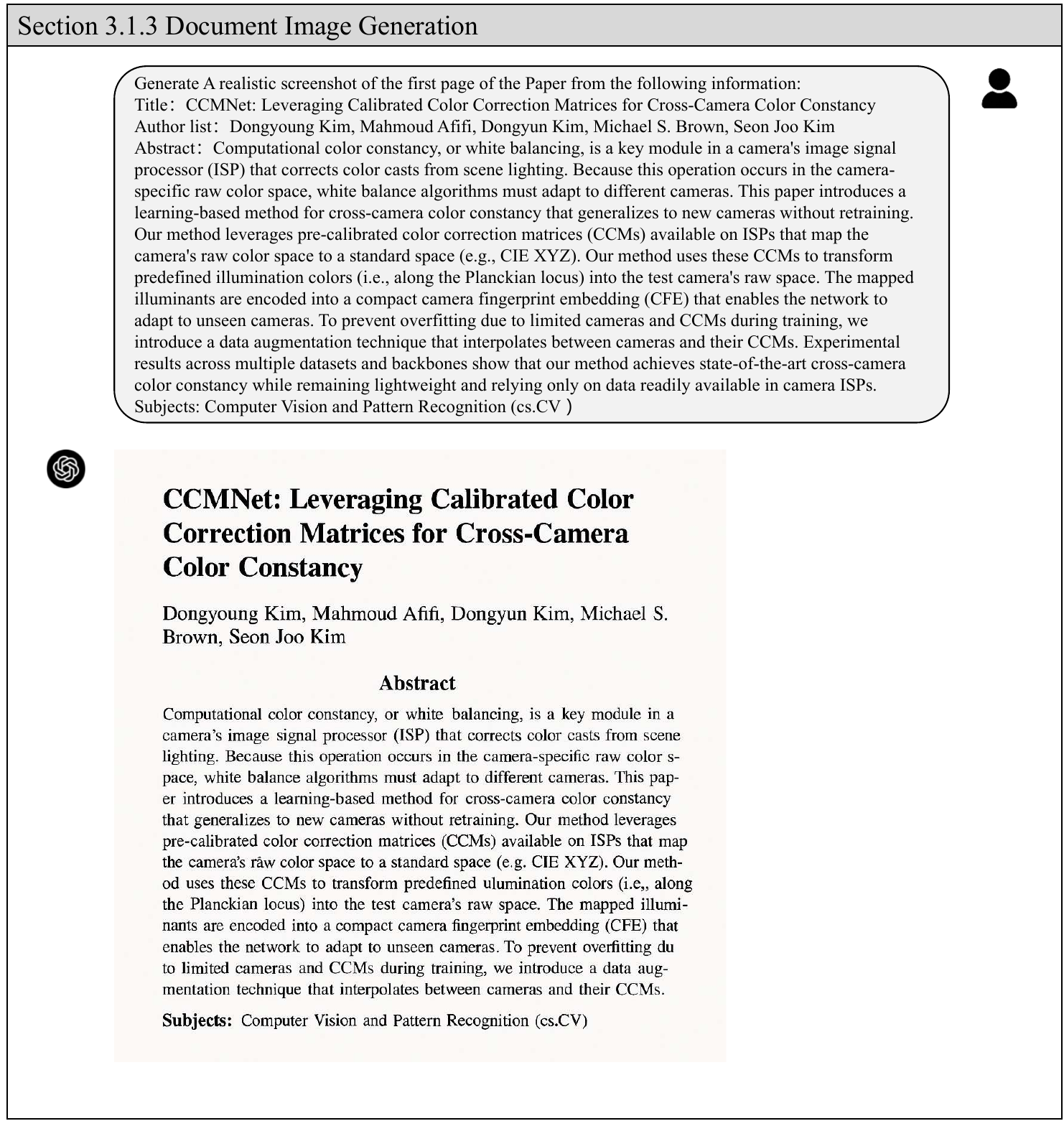}
    \caption[Sec~\ref{sec:document}: Textual Document Image Generation]{Examples of textual document image generation by \modelname.}
    \label{fig:doctext}
\end{figure}

As shown in Fig.~\ref{fig:doctext}, \modelname can synthesize realistic visual representations of academic documents when given metadata like title, author list, and abstract. The layout structure is relatively well-formed and the image resembles typical paper screenshots, although minor spacing and text overflow issues may occur.

\begin{figure}[h]
    \centering
    \includegraphics[width=1.0\linewidth]{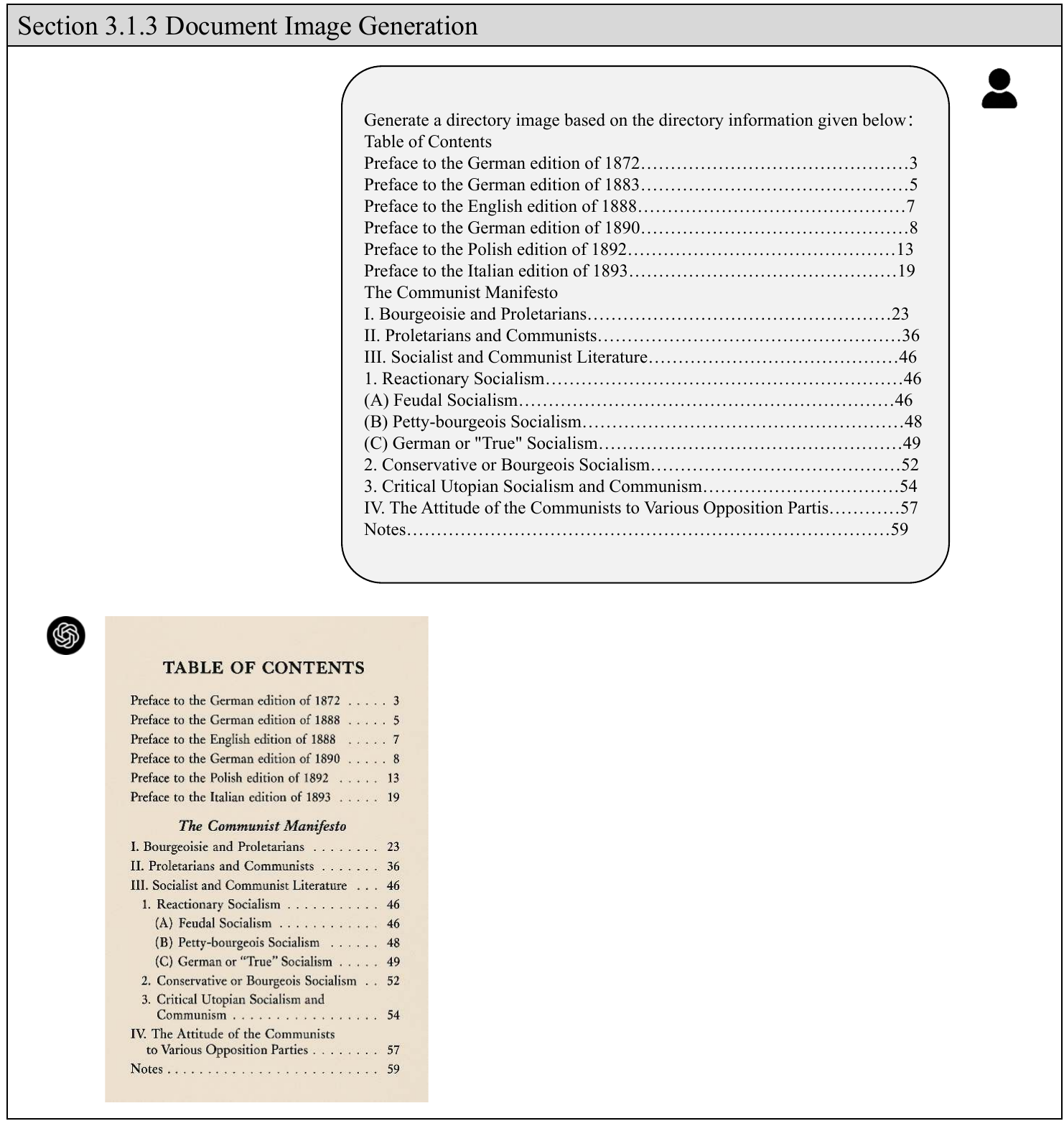}
    \caption[Sec~\ref{sec:document}: Catalog Image Generation]{Examples of catalog image generation by \modelname.}
    \label{fig:catalog}
\end{figure}

In Fig.~\ref{fig:catalog}, we test the generation of structured table-of-contents pages. \modelname succeeds in capturing the multi-level hierarchy and spacing of entries, but occasional alignment drift and symbol inconsistency are observed.

\begin{figure}[h]
    \centering
    \includegraphics[width=1.0\linewidth]{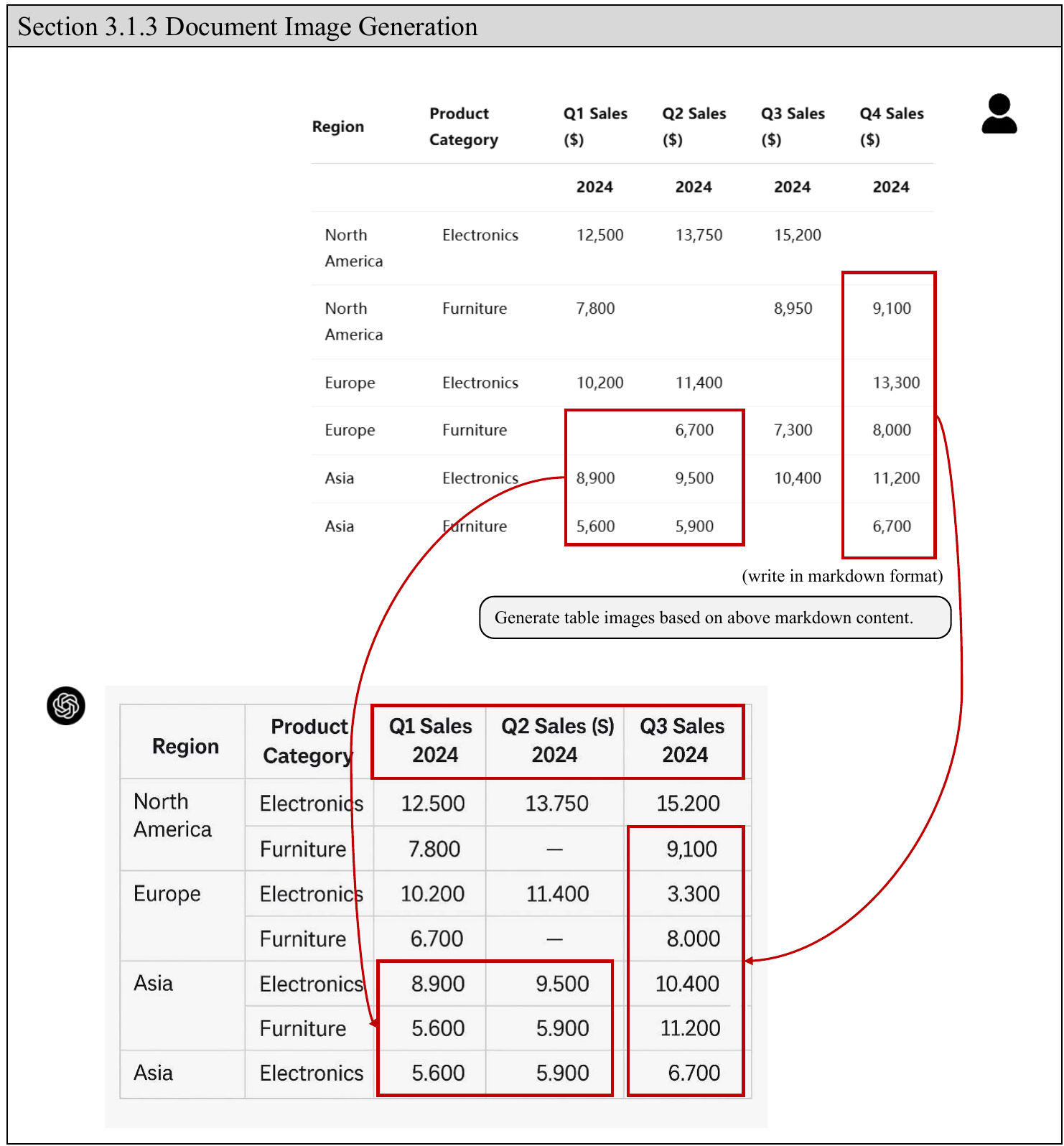}
    \caption[Sec~\ref{sec:document}: Markdown-to-Table Generation]{Examples of markdown-to-table generation by \modelname.}
    \label{fig:doctable}
\end{figure}

For tabular generation (Fig.~\ref{fig:doctable}), the model is given structured markdown input and expected to generate a rendered table. The results exhibit promising alignment and visual styling, but some text wrapping and boundary precision issues remain.

\begin{figure}[h]
    \centering
    \includegraphics[width=1.0\linewidth]{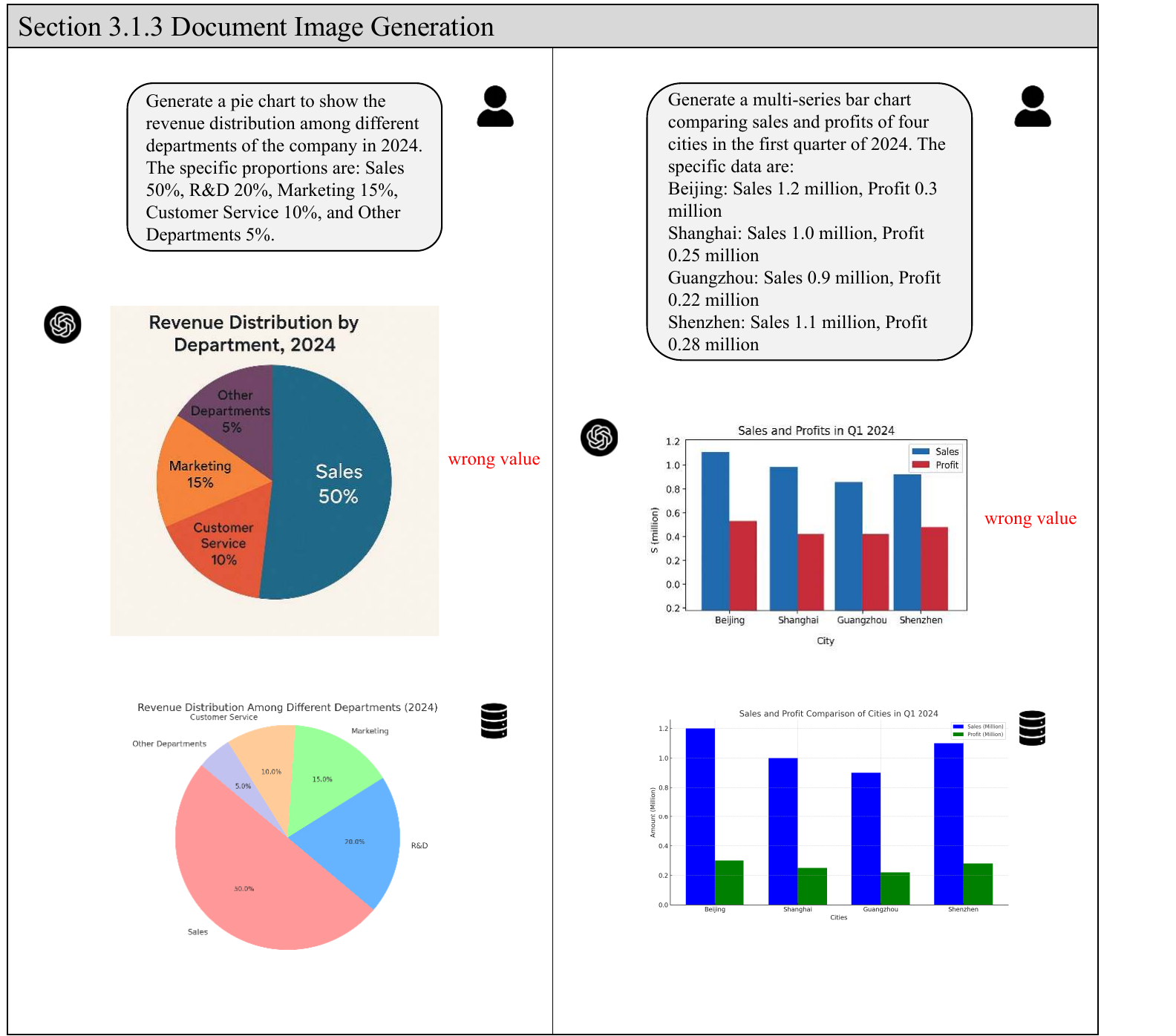}
    \caption[Sec~\ref{sec:document}: Chart Generation]{Examples of chart generation by \modelname.}
    \label{fig:docchart}
\end{figure}

\begin{figure}[h]
    \centering
    \includegraphics[width=1.0\linewidth]{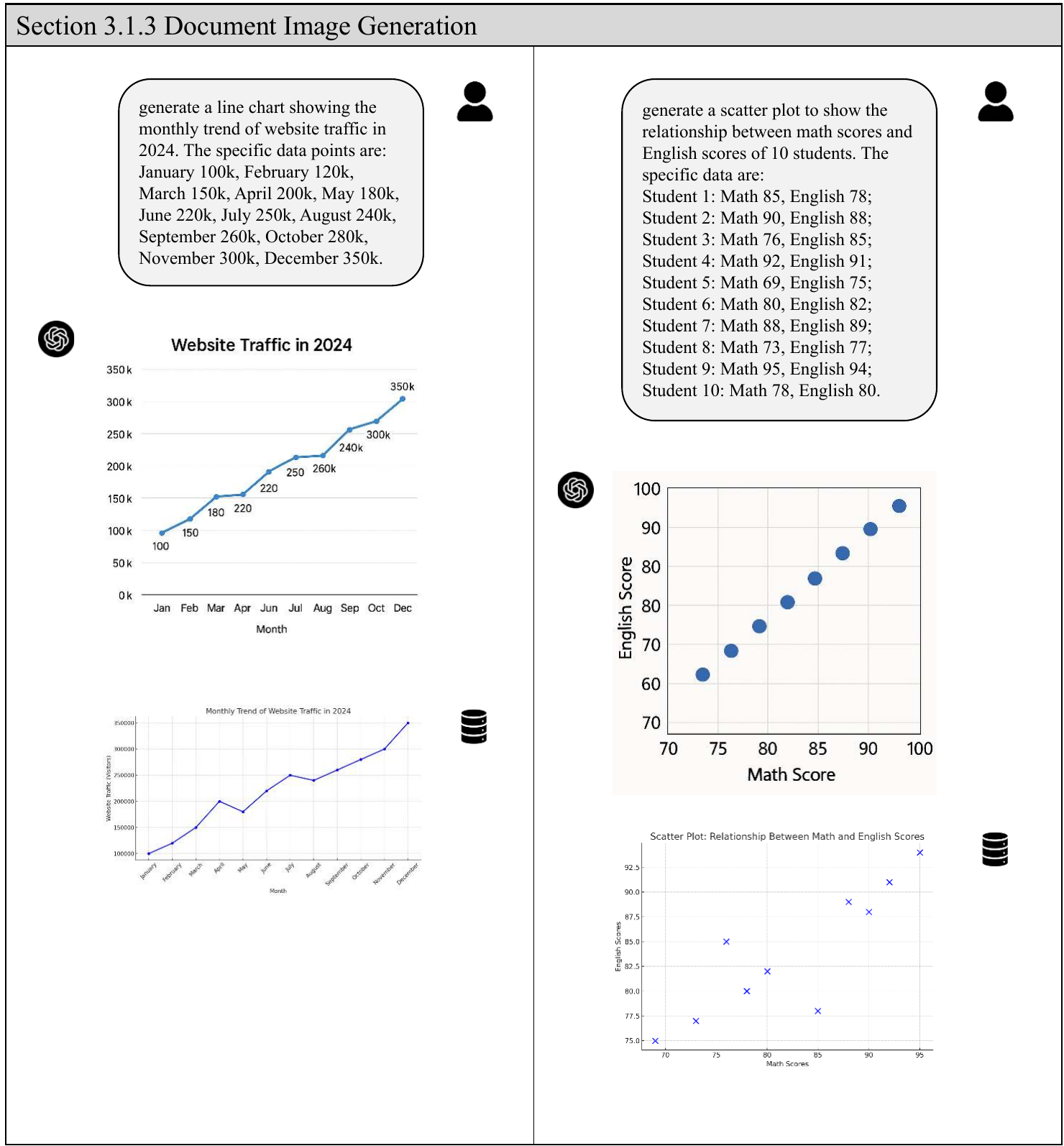}
    \caption[Sec~\ref{sec:document}: Chart Generation]{Additional examples of chart generation by \modelname.}
    \label{fig:docchart2}
\end{figure}

More challenging are chart-related tasks (Fig.~\ref{fig:docchart}, \ref{fig:docchart2}), where \modelname is prompted to visualize numerical data in bar, pie, line, and scatter plot formats. While the model captures general layout and chart types correctly, we observe frequent numerical inaccuracies, axis mislabeling, and data point misplacement. In some cases, the visual values do not match the specified ones, which poses concerns for trustworthy data communication.

Overall, \modelname demonstrates a basic ability to generate visually plausible document-style images. However, in tasks requiring precise control of structure, spatial arrangement, and quantitative correctness, significant limitations remain.

\clearpage

%
%\begin{figure}[h]
%    \centering
%    \includegraphics[width=1.0\linewidth]{figures/doc}
%    \caption[Sec~\ref{sec:document}: Document Image Generation]{}
%    \label{fig:doc}
%\end{figure}

\clearpage

\subsection{Multimodal-Conditioned Image Generaiton}
\label{sec:m2i}
Multimodal-conditioned image generation requires the model to integrate and respond to complex inputs, which may combine both textual and visual signals. This setting tests \modelname’s multimodal understanding and its ability to blend visual information with linguistic instructions in a coherent and contextually appropriate manner.

\subsubsection{Image Editing}
\label{sec:edit}
Image editing tasks evaluate whether \modelname can modify an existing image according to a specific instruction, such as changing object attributes, removing elements, or altering the scene style. This task reflects the model’s ability to comprehend visual content and apply precise, localized modifications while preserving the overall image consistency\cite{brooks2023instructpix2pix,Fu_Hu_Du_Yang_Gan_2023,Brack_Friedrich_Kornmeier_Tsaban_Schramowski_Kersting_Passos_2023,Zhang_Mo_Chen_Sun_Su_2023}.

\begin{figure}[h]
    \centering
    \includegraphics[width=1.0\linewidth]{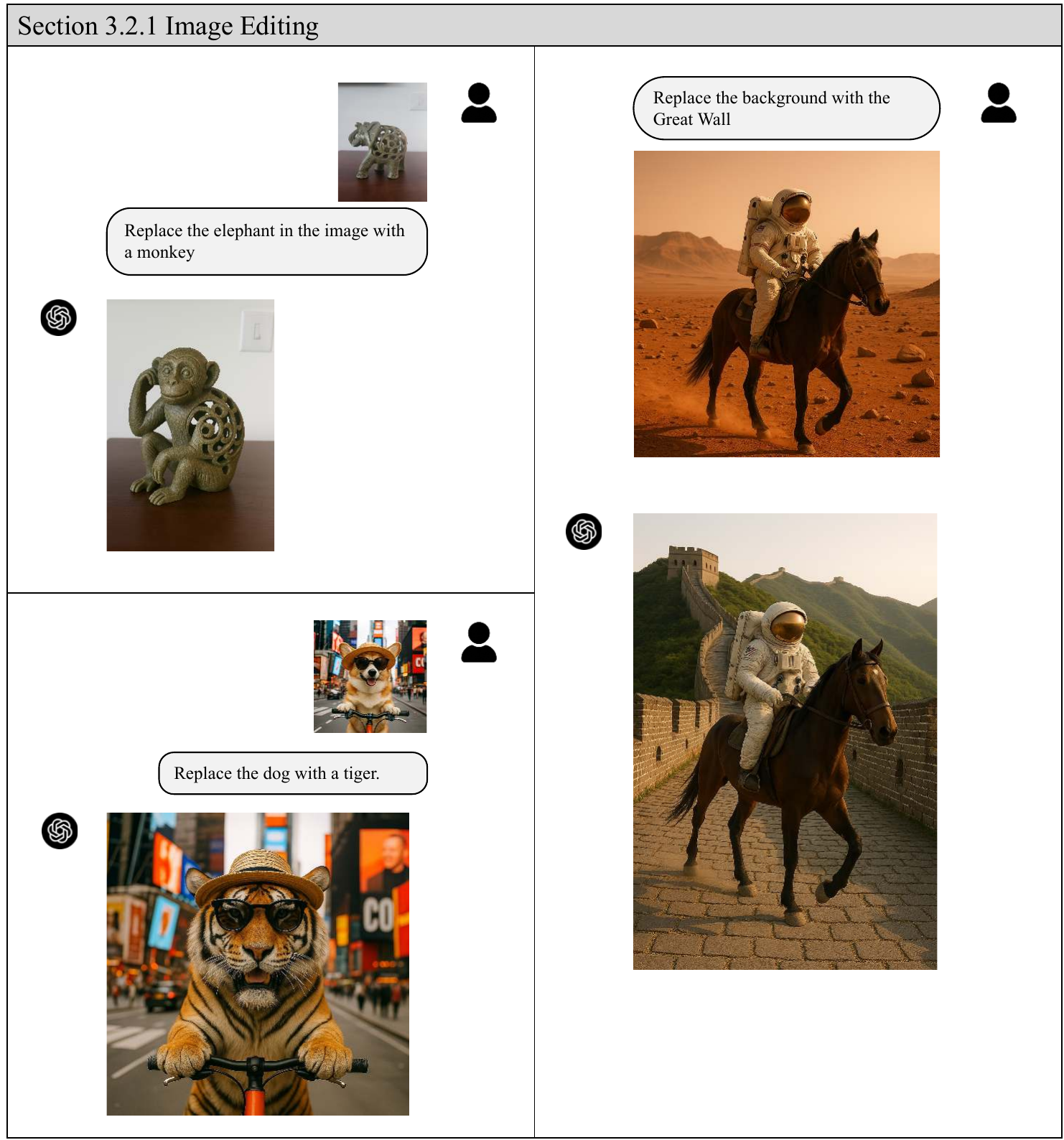}
    \caption[Sec~\ref{sec:edit}: Concept Replace]{Examples of concept replacement, where \modelname replaces target objects or backgrounds (e.g., animal species, scenes) while preserving the remaining content.}
    \label{fig:concept-replace}
\end{figure}

As shown in Fig.~\ref{fig:concept-replace}, we first evaluate concept replacement tasks, where the model is instructed to replace individual objects or backgrounds (e.g., "replace the elephant with a monkey", "replace the background with the Great Wall"). \modelname successfully captures the target concept in most cases, but occasionally fails to fully preserve structural context or scale, leading to semantic mismatches or blending issues.

\begin{figure}[h]
    \centering
    \includegraphics[width=1.0\linewidth]{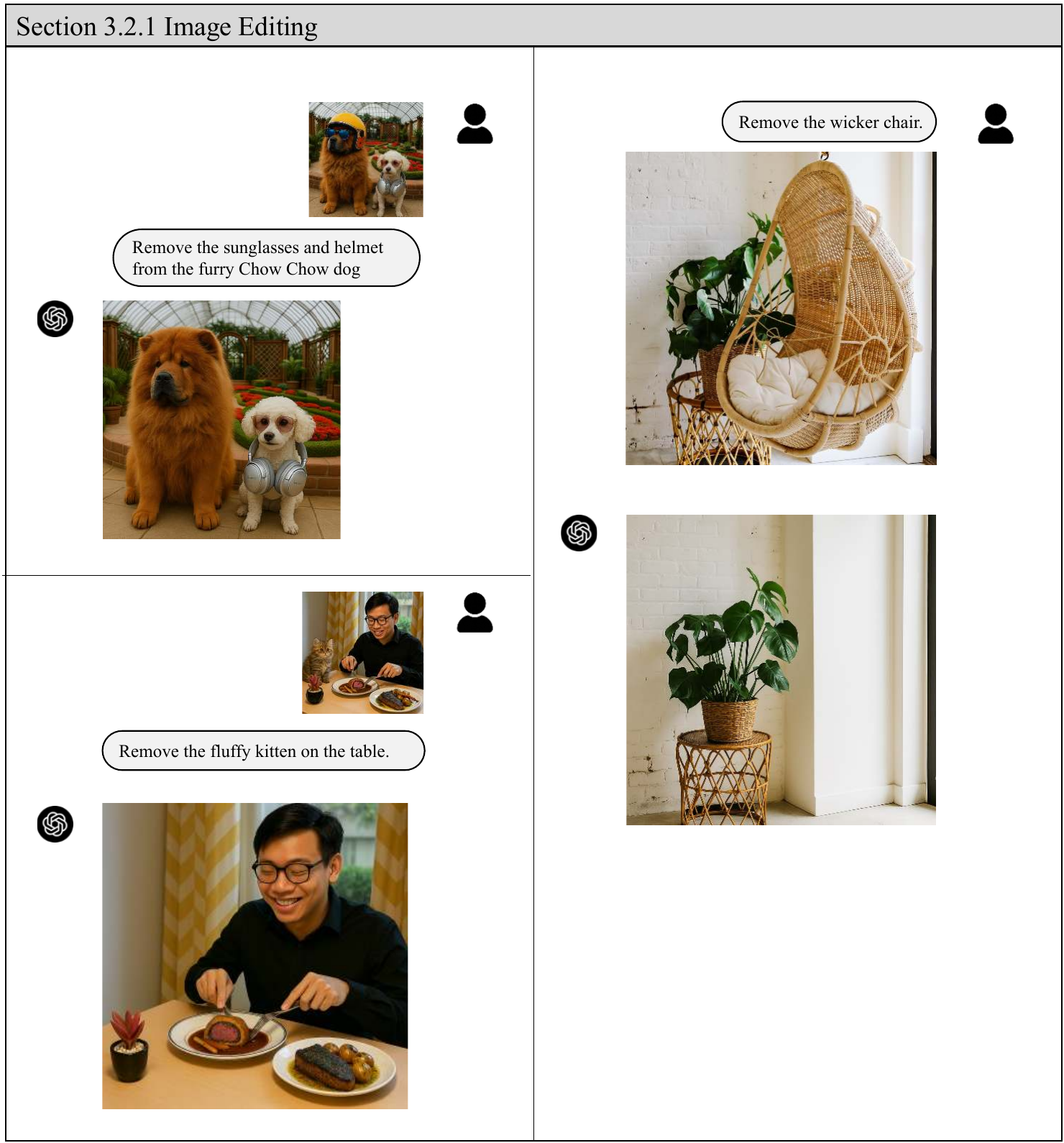}
    \caption[Sec~\ref{sec:edit}: Concept Removal]{Examples of concept removal tasks, where \modelname eliminates specific objects (e.g., accessories, animals) based on the instruction.}
    \label{fig:doc}
\end{figure}

In Fig.~\ref{fig:doc}, we evaluate object removal tasks such as "remove the sunglasses and helmet from the dog" or "remove the kitten from the table". The model performs reasonably well in eliminating specified targets while maintaining the surrounding content, although minor residue artifacts or unnatural texture patches may occur.

\begin{figure}[h]
    \centering
    \includegraphics[width=1.0\linewidth]{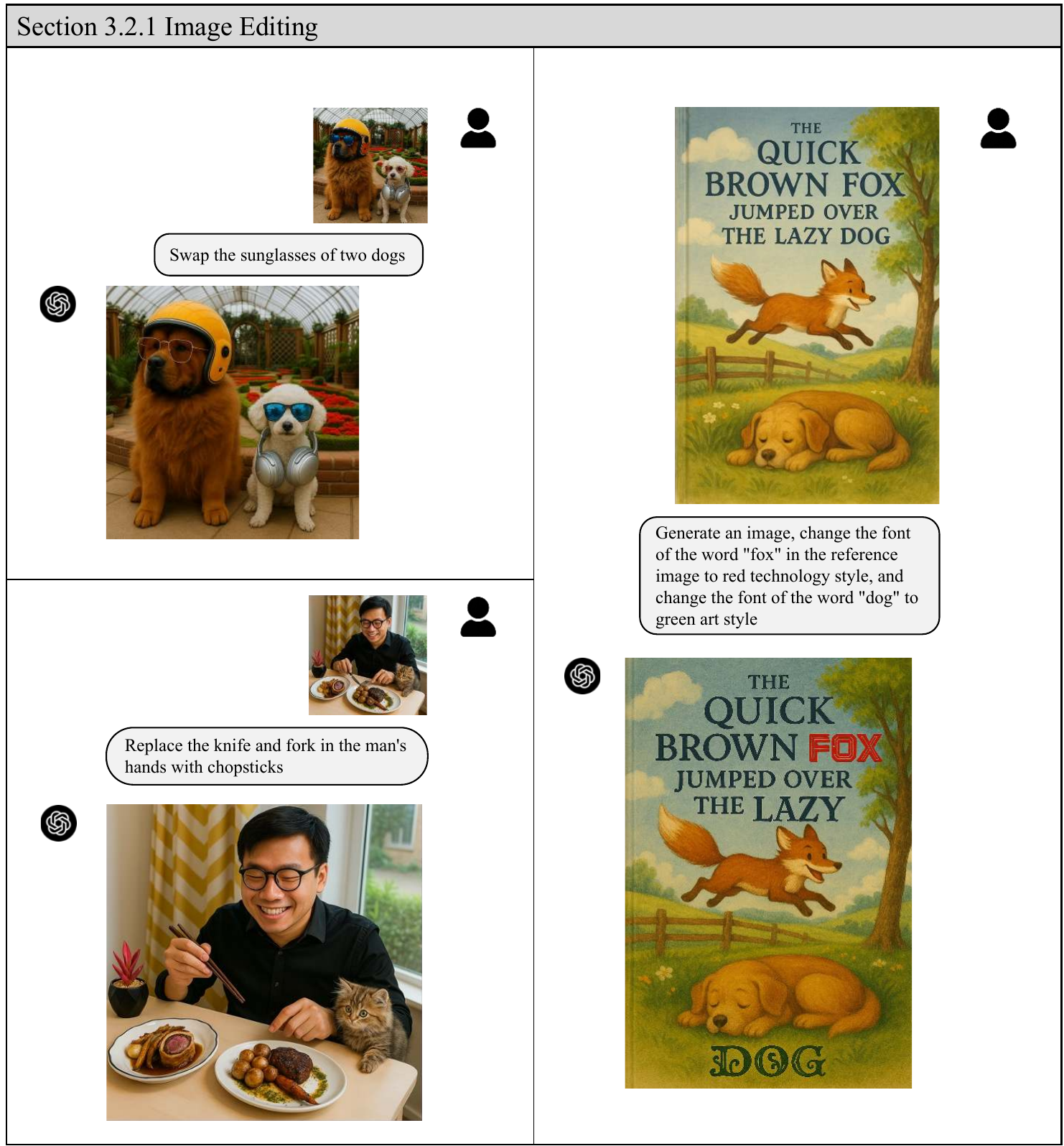}
    \caption[Sec~\ref{sec:edit}: Image Editing]{Complex image editing examples involving multiple changes, such as swapping accessories or modifying fonts and materials.}
    \label{fig:edit1}
\end{figure}

\begin{figure}[h]
    \centering
    \includegraphics[width=1.0\linewidth]{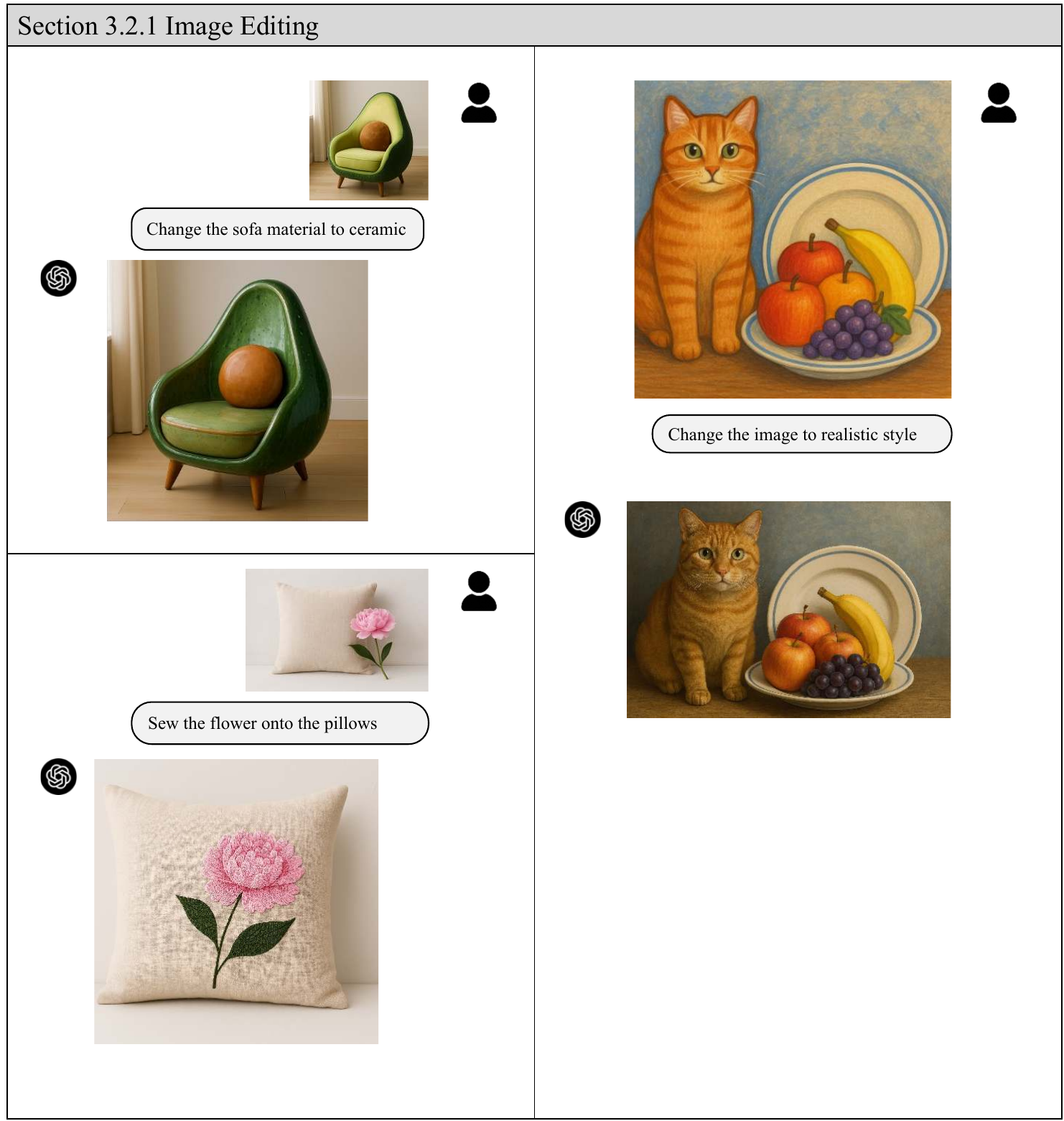}
    \caption[Sec~\ref{sec:edit}: Image Editing]{Additional examples of multi-step and stylistic edits, where \modelname is tasked with applying compositional instructions involving multiple transformations.}
    \label{fig:edit2}
\end{figure}

More challenging are complex multi-edit scenarios shown in Fig.~\ref{fig:edit1} and Fig.~\ref{fig:edit2}, including multiple simultaneous changes (e.g., swapping accessories between subjects, modifying font styles, or transforming materials). \modelname shows a basic ability to interpret compositional instructions and perform diverse edits.

Overall, while \modelname demonstrates promising performance in direct concept replacement.

\clearpage

\clearpage
\subsubsection{Personalized Image Generation}
\label{sec:personalization}
Personalization focuses on generating content tailored to specific entities, styles, or individuals, such as producing consistent images of a user-provided character or object\cite{cao2024controllable}. This task evaluates \modelname’s memory retention, visual consistency, and adaptability across multimodal conditions. In this section, we follow the evaluation protocols proposed in [A] and test a variety of personalization scenarios.

\begin{figure}[h]
    \centering
    \includegraphics[width=1.0\linewidth]{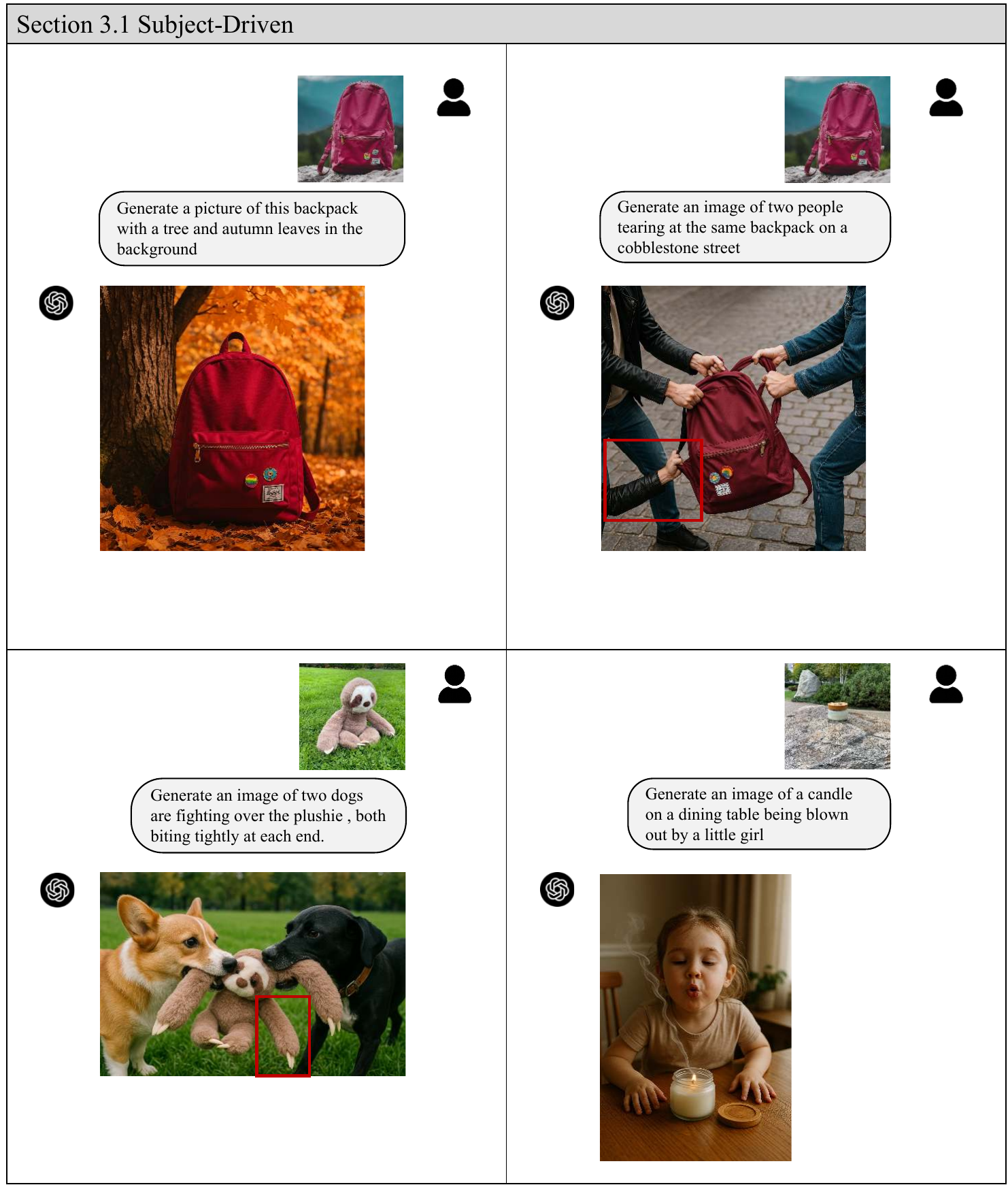}
    \caption[Sec~\ref{sec:personalization}: Subject-driven Image Generation]{Examples of subject-driven image generation by \modelname.}
    \label{fig:subject-driven}
\end{figure}

\textbf{Subject-driven generation.} For subject-driven generation~\cite{dong2022dreamartist,gal2022image,ruiz2023dreambooth,voynov2023p+,zhang2023prospect,xiao2023comcat}, \modelname is expected to synthesize new images while preserving the identity and attributes of the given subject, conditioned on new prompts. \modelname generally succeeds in capturing the overall appearance and theme of the subject described in the input image-text pair. However, certain structural inaccuracies still occur. For instance, in Fig.~\ref{fig:subject-driven}, the prompt “two people tearing” results in an image containing two figures, but with a total of five hands, revealing flaws in body-part reasoning. In another example, the model-generated version of a customized character includes an extra arm, suggesting inconsistencies in spatial anatomy and object integrity.

\begin{figure}[h]
    \centering
    \includegraphics[width=1.0\linewidth]{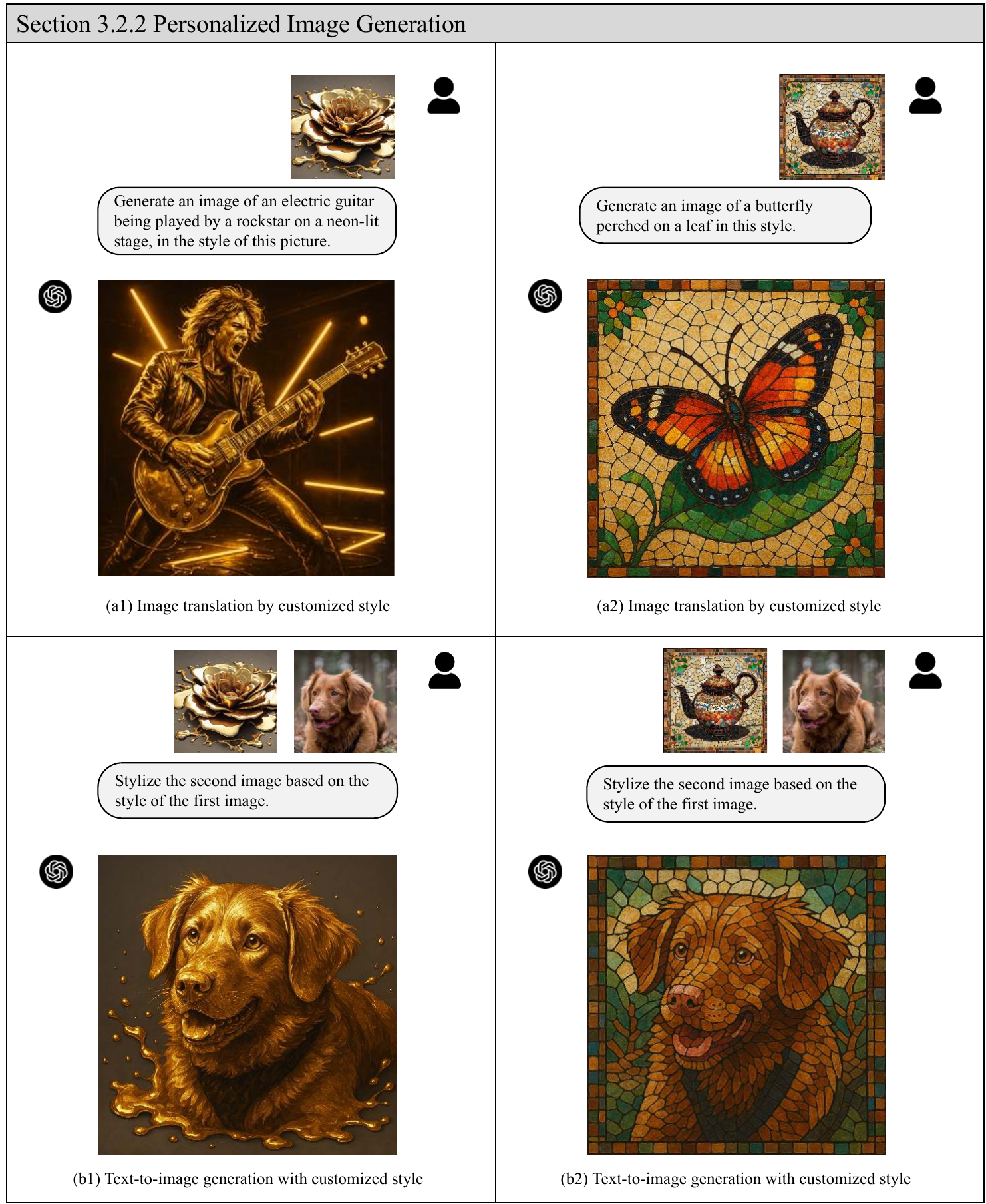}
    \caption[Sec~\ref{sec:personalization}: Style-driven Image Generation]{Examples of style-driven image generation by \modelname.}
    \label{fig:style-driven}
\end{figure}

\textbf{Style-driven generation.} For style-driven generation\cite{sohn2023styledrop,liu2023stylecrafter,chen2023artadapter}, where the goal is to transfer a visual style from a reference image to new content, the model is able to follow the overall aesthetic and object features with reasonable fidelity. However, we observe mismatches in tonal and stylistic coherence. As shown in Fig.~\ref{fig:style-driven}, the generated image appears notably darker than the reference, indicating that while the model retains high-level style cues, it may struggle with precise low-level appearance reproduction.

\begin{figure}[h]
    \centering
    \includegraphics[width=1.0\linewidth]{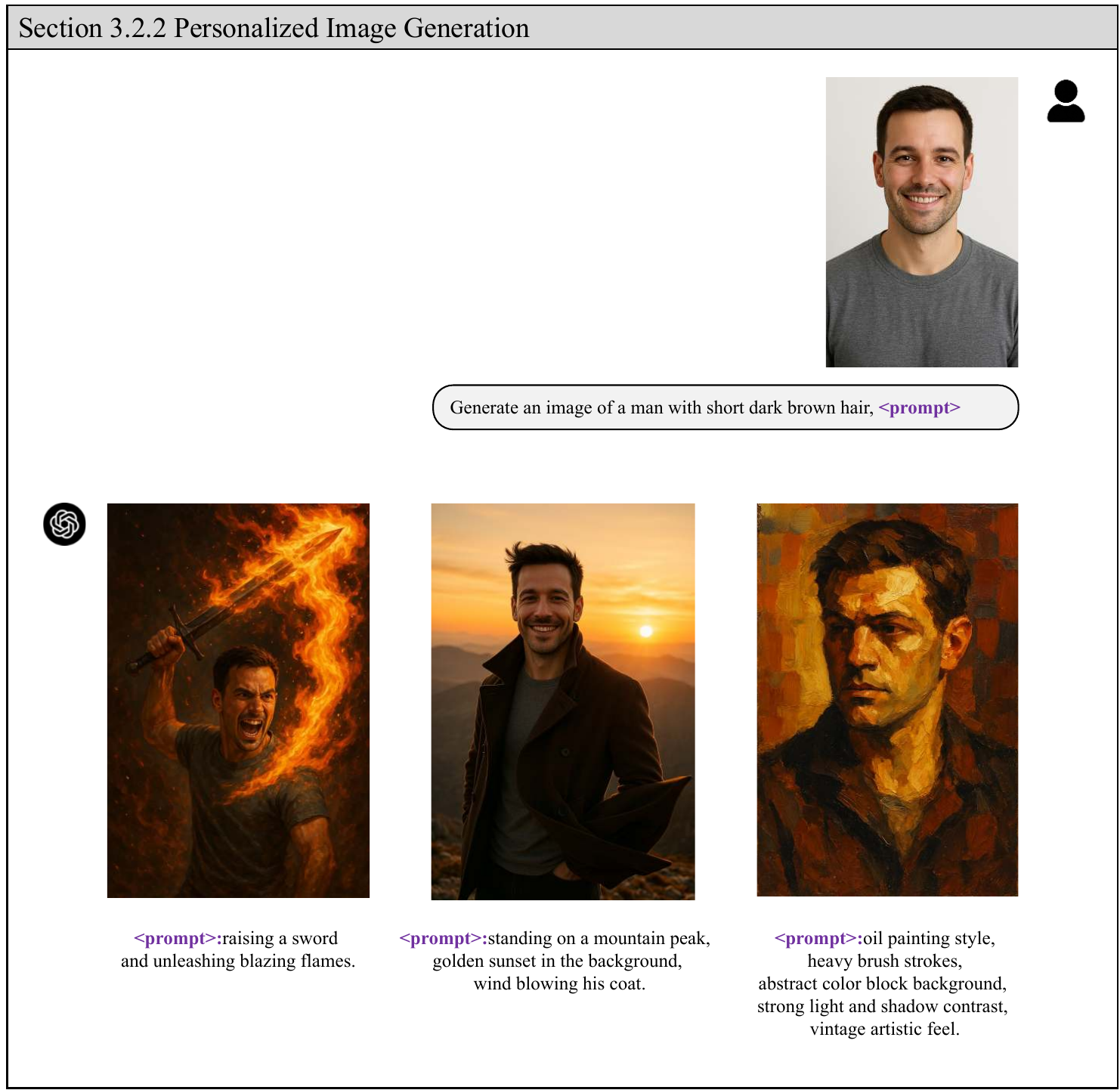}
    \caption[Sec~\ref{sec:personalization}: Person-driven Image Generation]{Examples of person-driven image generation by \modelname.}
    \label{fig:person-driven1}
\end{figure}

\begin{figure}[h]
    \centering
    \includegraphics[width=1.0\linewidth]{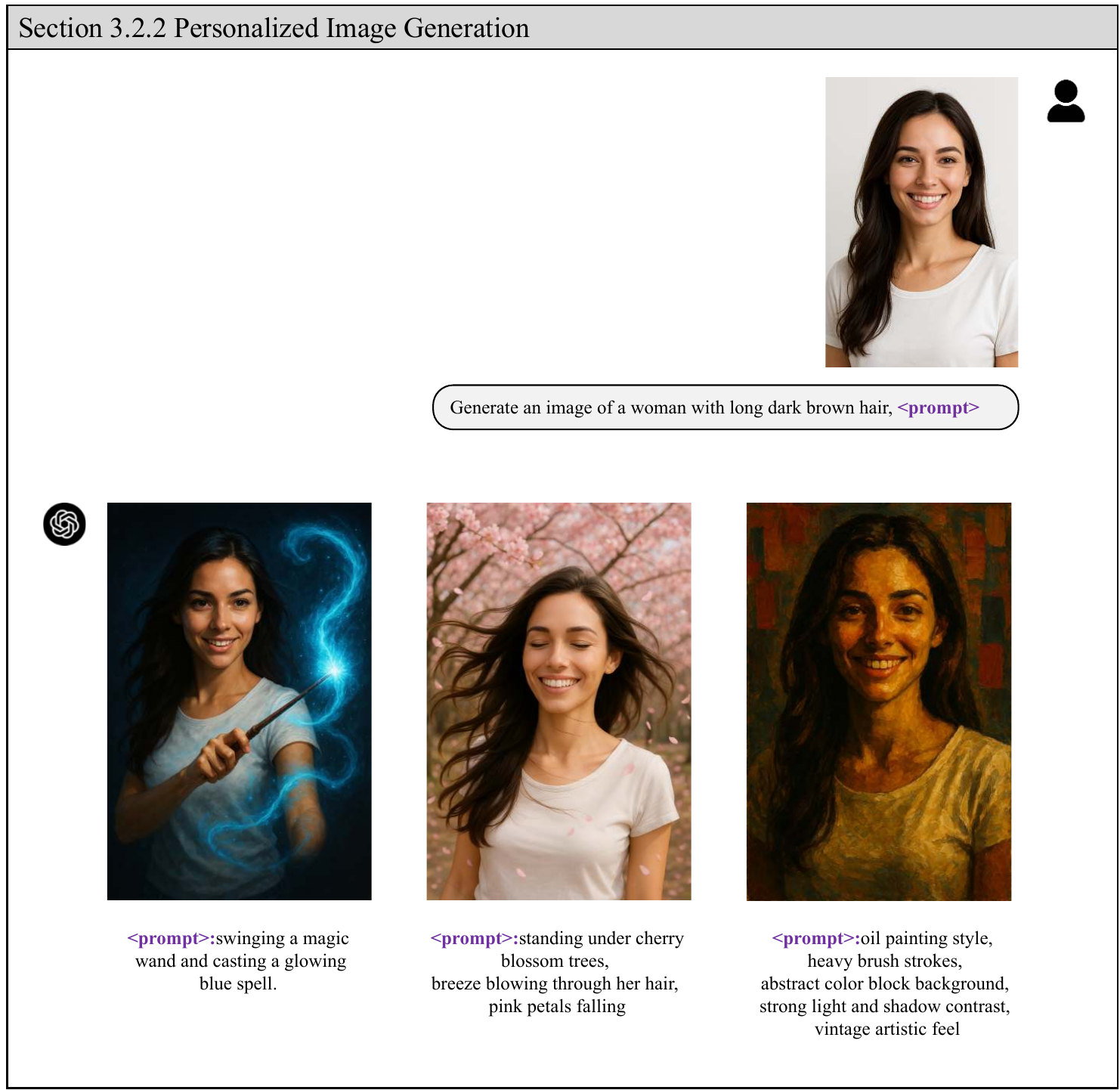}
    \caption[Sec~\ref{sec:personalization}: Person-driven Image Generation]{Additional examples of perosn-driven image generation by \modelname.}
    \label{fig:person-driven2}
\end{figure}

\textbf{Person-driven generation.} In person-driven generation\cite{xiao2023fastcomposer,valevski2023face0,chen2023dreamidentity}, the model is tasked with preserving the identity, appearance, and facial attributes of a reference individual across different prompts and scenarios. As shown in our results in Fig.~\ref{fig:person-driven1} and Fig.~\ref{fig:person-driven2}, \modelname is able to capture key visual characteristics such as hairstyle, clothing, and facial structure, and consistently reproduce them in novel contexts. The generated outputs exhibit strong coherence and personalization, even under varied instructions and poses.

\begin{figure}[h]
    \centering
    \includegraphics[width=1.0\linewidth]{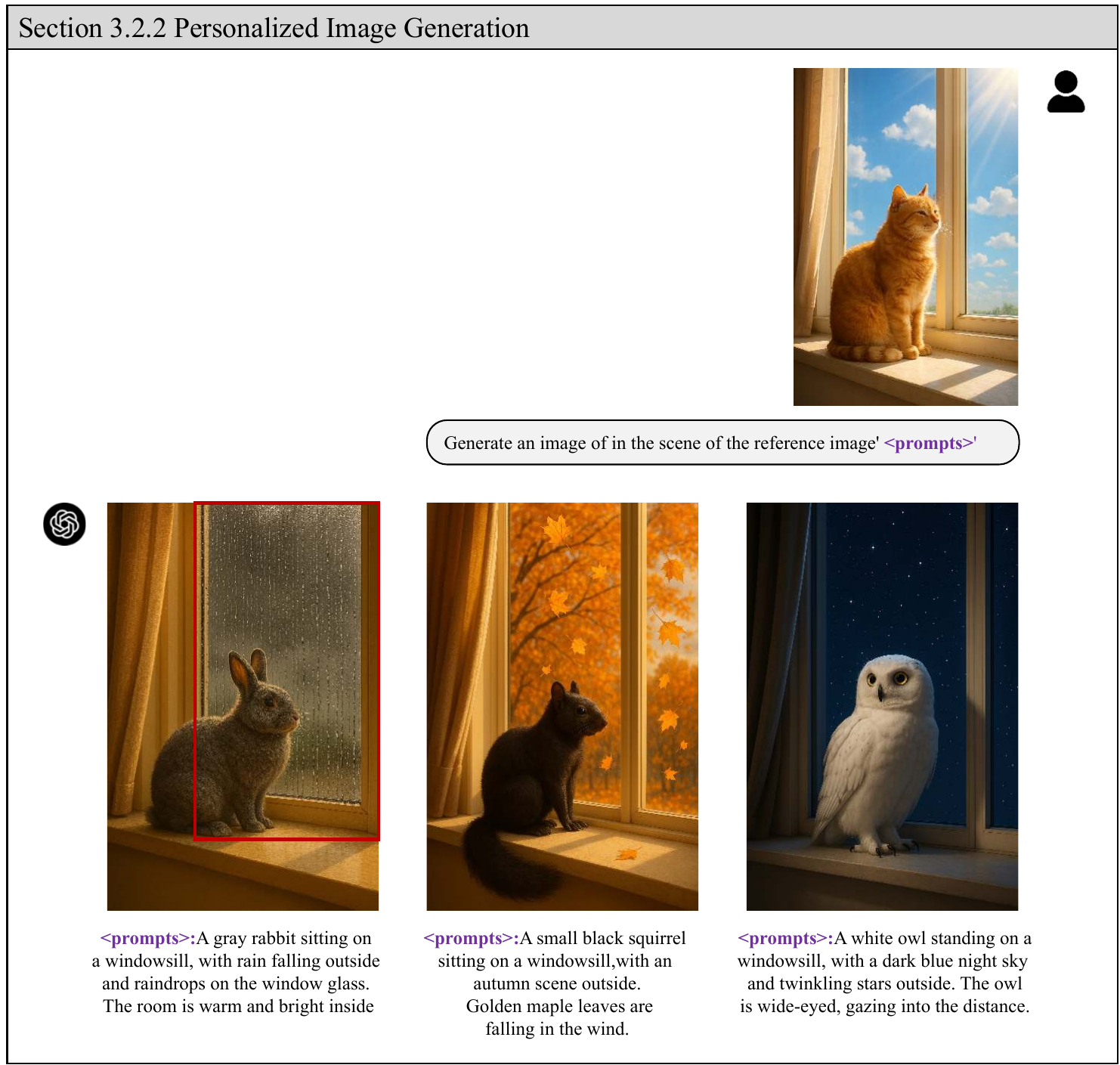}
    \caption[Sec~\ref{sec:personalization}: Scene-driven Image Generation]{Examples of scene-driven image generation by \modelname.}
    \label{fig:scene-driven}
\end{figure}

\textbf{Scene-driven generation.} For scene-driven generation, the goal is to retain the background layout, spatial structure, and environmental elements from the reference image while modifying the foreground or semantic content. While \modelname demonstrates a reasonable ability to extract and imitate overall scene information, some visual mismatches are still observed, as can be shown in Fig.~\ref{fig:scene-driven}. For example, in the leftmost image of Fig.~\ref{fig:scene-driven}, the window structure differs significantly from the reference, indicating that the model may struggle with fine-grained background alignment.

\begin{figure}[h]
    \centering
    \includegraphics[width=1.0\linewidth]{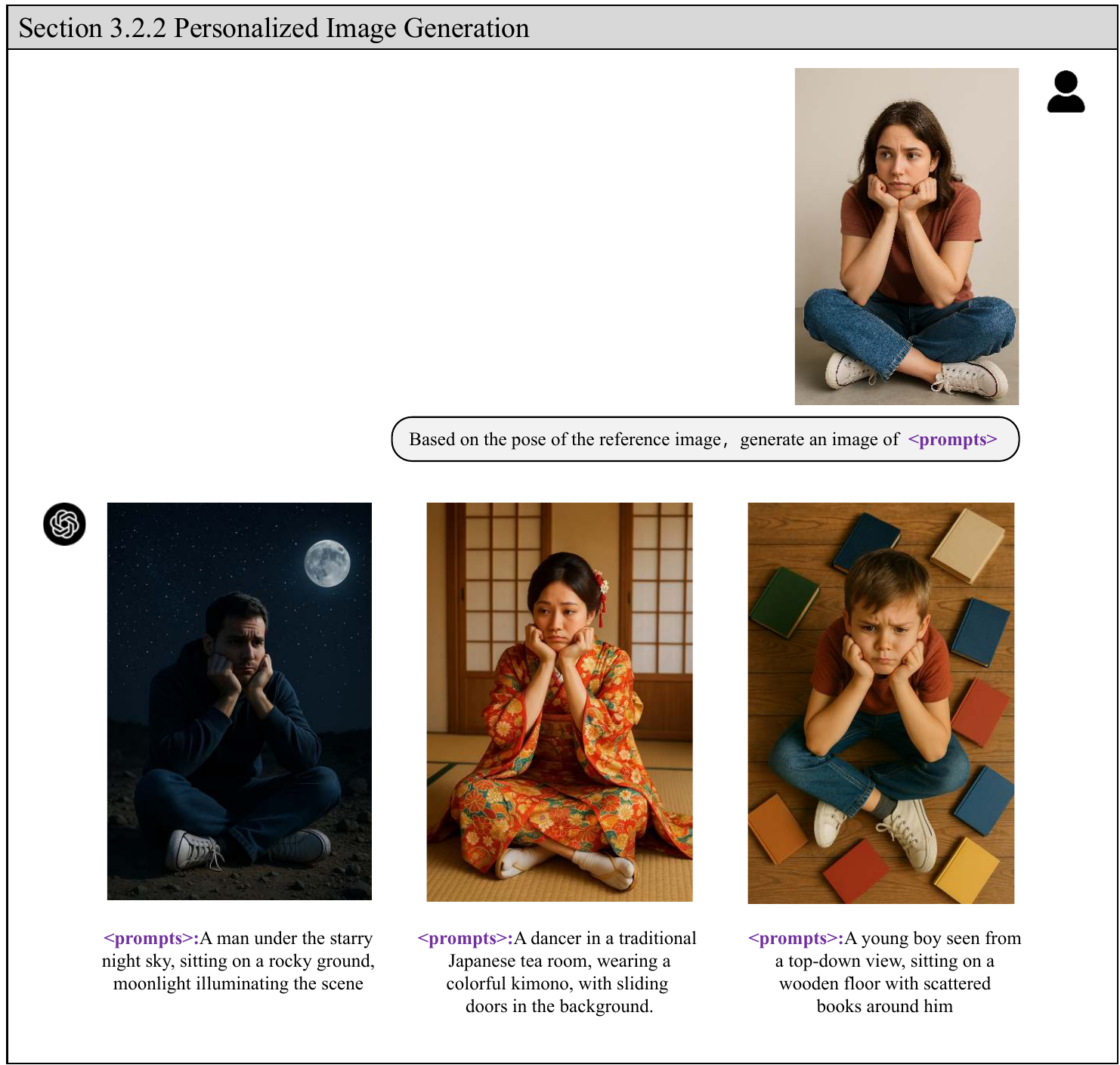}
    \caption[Sec~\ref{sec:personalization}: Pose-driven Image Generation]{Examples of pose-driven image generation by \modelname.}
    \label{fig:pose-driven}
\end{figure}

\textbf{Pose-driven generation.} In pose-driven generation, the model is provided with an input image showcasing a specific body posture and is expected to transfer that pose to a new subject or scene. In this task, \modelname shows robust performance (Fig.~\ref{fig:pose-driven}), successfully transferring pose configurations and generating visually plausible results, with correct limb orientation and body articulation preserved.

\begin{figure}[h]
    \centering
    \includegraphics[width=1.0\linewidth]{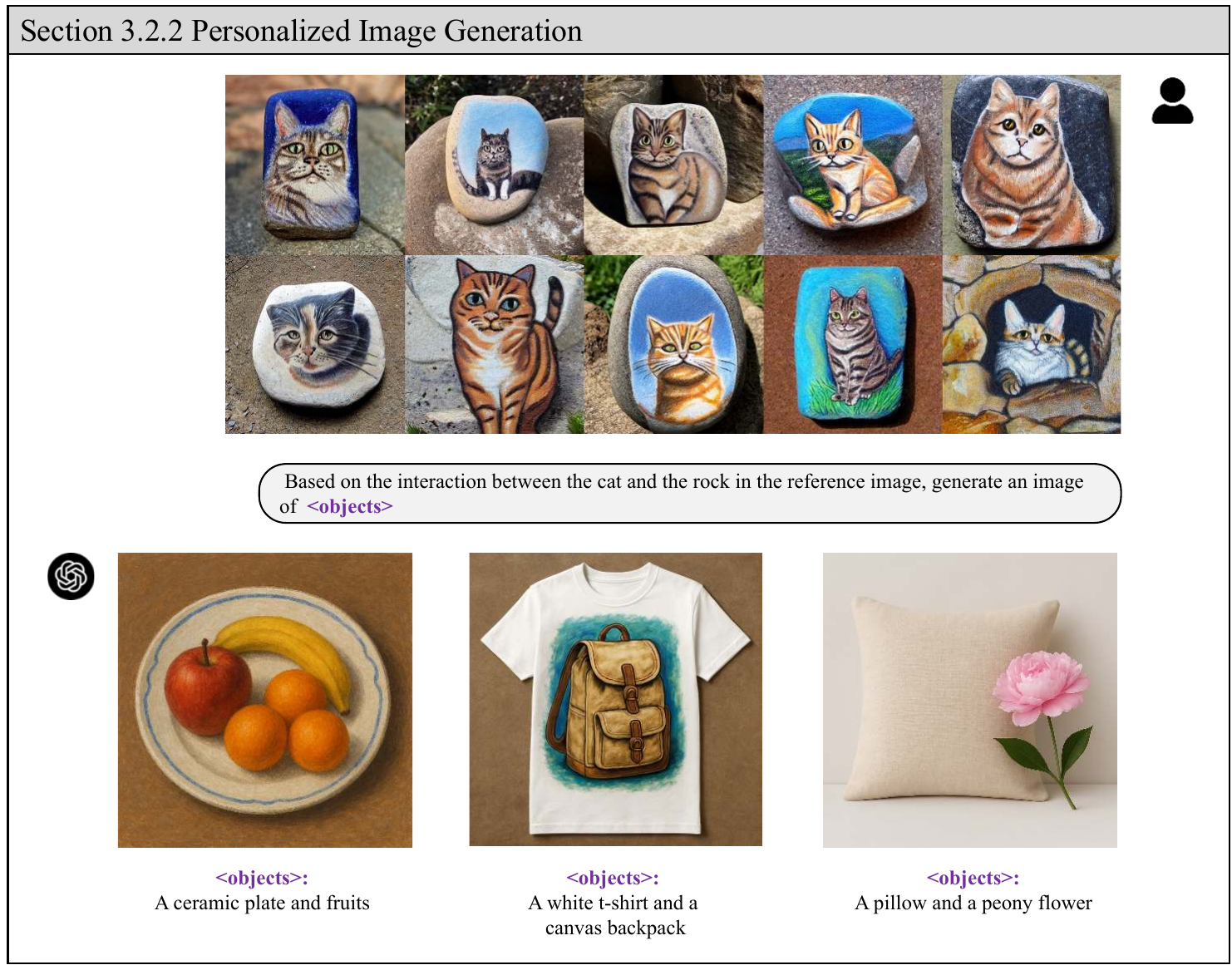}
    \caption[Sec~\ref{sec:personalization}: Interaction-driven Image Generation]{Examples of interaction-driven image generation by \modelname.}
    \label{fig:relation-driven}
\end{figure}

\textbf{Interaction-driven generation.} We also examine interaction-driven generation\cite{huang2023learning,huang2023reversion}, where the goal is to preserve the semantic or visual interaction between objects in the reference image, such as spatial relations, embedding, or stylistic blending across elements. In this task, \modelname shows clear limitations (Fig.~\ref{fig:relation-driven}). For example, in the reference image, a cat is drawn onto the surface of a rock using a blended style that resembles colored pencil and watercolor. Ideally, the model should preserve both the embedded positioning of the cat and the stylistic fusion with the rock texture. However, as illustrated in Fig.~X, only the second generated image shows a slight indication of this interaction. The other two outputs fail to reflect either the spatial composition or the intended artistic medium, producing results that are visually detached from the reference semantics. This suggests that while \modelname can recognize explicit content elements, it struggles to infer and reconstruct implicit relationships.

\begin{figure}[h]
    \centering
    \includegraphics[width=1.0\linewidth]{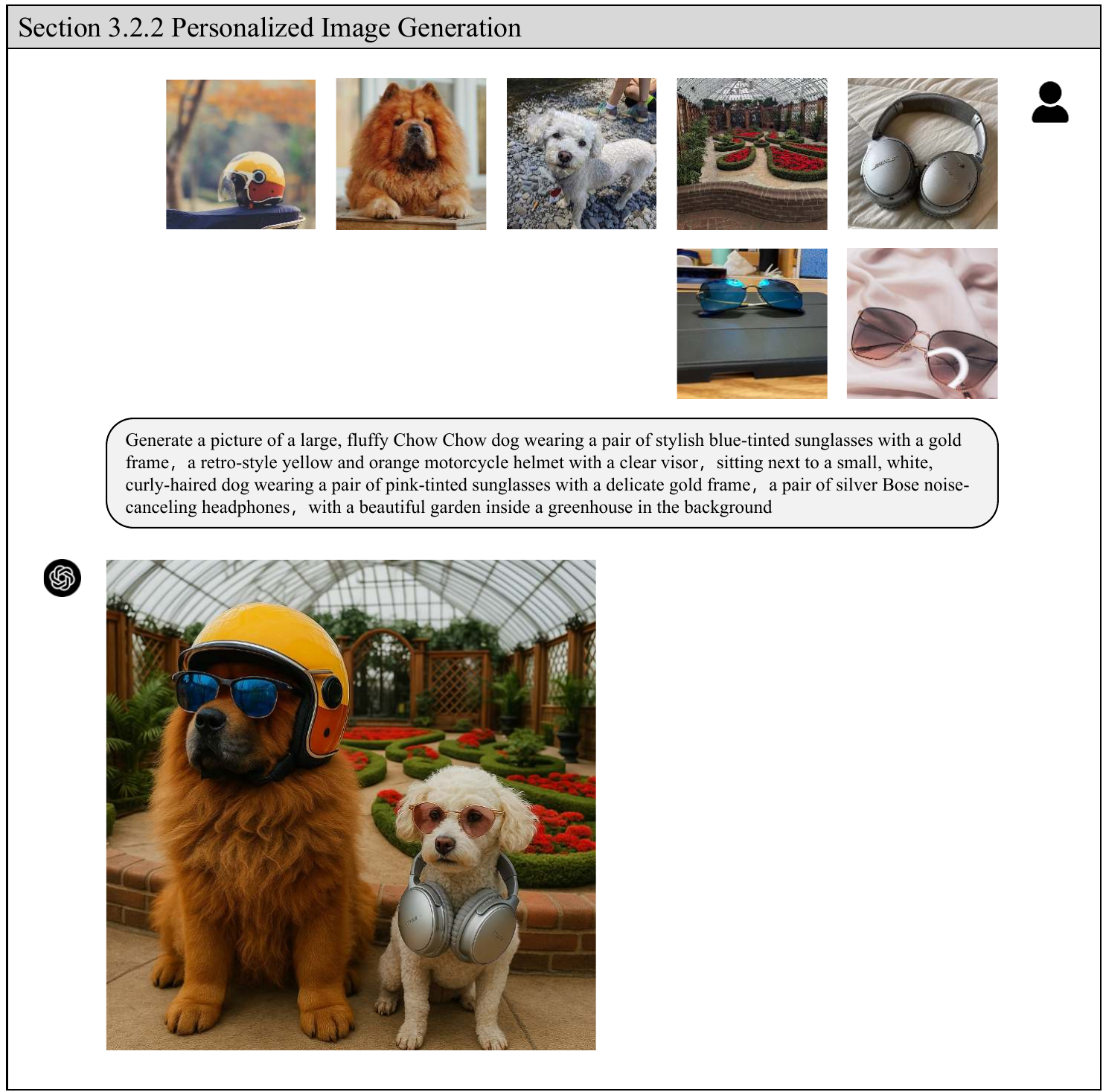}
    \caption[Sec~\ref{sec:personalization}: Multi-Concept Personalization]{Examples of multi-concept personalization results generated by \modelname.}
    \label{fig:multiconcept-describe}
\end{figure}

\begin{figure}[h]
    \centering
    \includegraphics[width=1.0\linewidth]{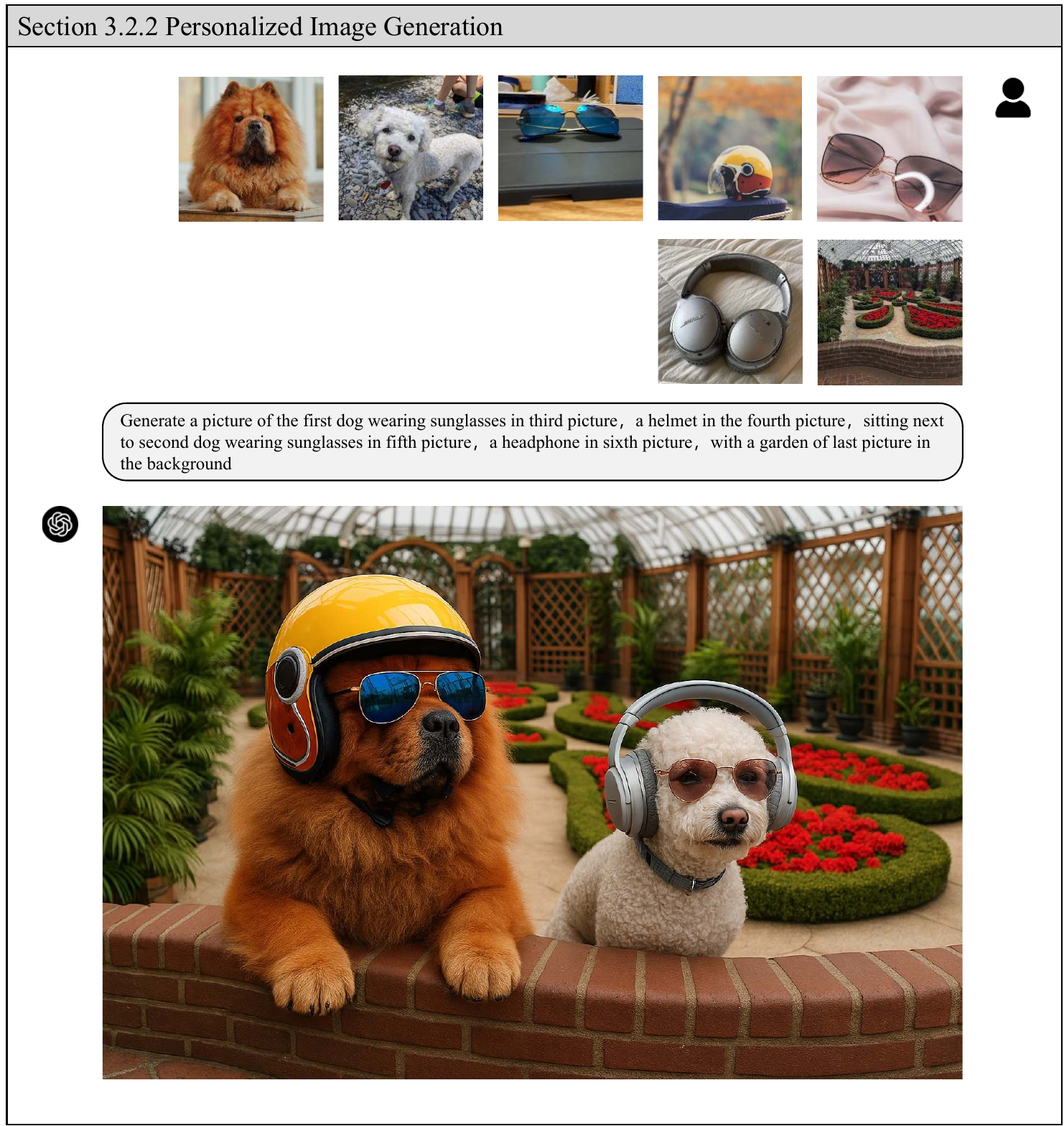}
    \caption[Sec~\ref{sec:personalization}: Multi-Concept Personalization]{Additional examples of multi-concept personalization results generated by \modelname.}
    \label{fig:multiconcept-number}
\end{figure}

\textbf{Multi-concept personalization.}  We also explore \modelname's ability in multi-concept personalization\cite{huang2023composer,hu2023cocktail,smith2023continual,liu2023cones,liu2023cones}, where multiple customized objects or entities are referenced simultaneously within a single prompt. This task is more demanding, as it requires the model to not only retain individual object characteristics but also correctly compose them within a coherent scene.
As illustrated in Fig.~\ref{fig:multiconcept-describe}, we test this setting by defining a prompt that combines seven distinct objects and attributes, such as specific sunglasses, helmets, headphones, and background scenes. \modelname demonstrates impressive performance in preserving and combining all the specified concepts into a well-structured image. The generated output accurately reflects the visual identity of each reference item and maintains proper spatial composition.
To further assess the model’s referential understanding, we experiment with compositional prompts using ordinal expressions and image-position references. For example, we ask the model to “use the sunglasses from the third image” or “place the first dog next to the second dog,” as shown in Fig.~\ref{fig:multiconcept-number}. \modelname is able to follow these instructions, indicating an emerging capability to link textual ordinal cues with visual memory from prior references.
Overall, these results suggest that \modelname is not only capable of handling multi-object conditioning, but also demonstrates a basic understanding of structured referential language in personalized generation scenarios.

\begin{figure}[h]
    \centering
    \includegraphics[width=1.0\linewidth]{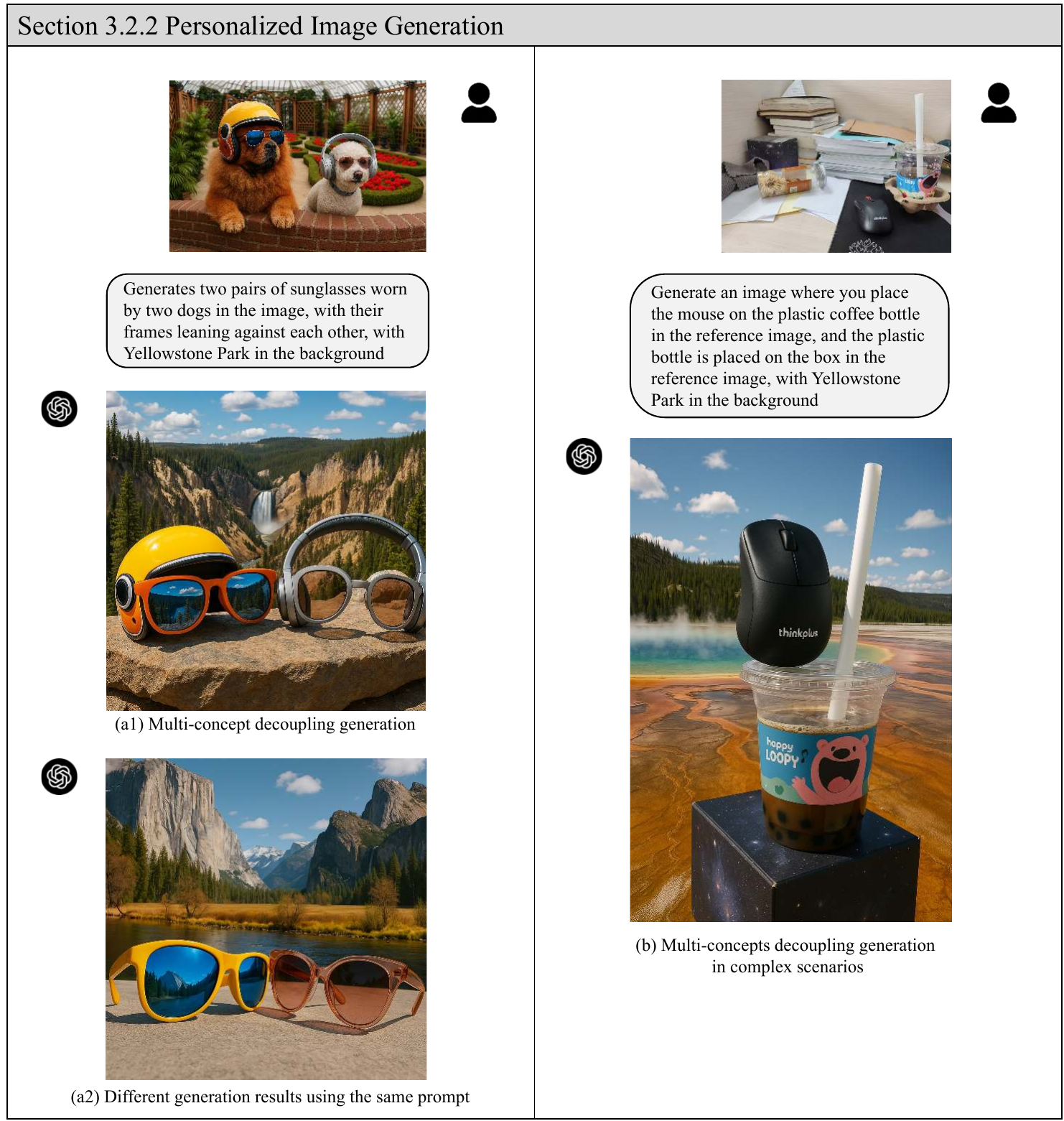}
    \caption[Sec~\ref{sec:personalization}: Fine-grained Decoupled personalization]{Examples of fine-grained decoupled personalization results generated by \modelname.}
    \label{fig:complex-customize}
\end{figure}

\textbf{Fine-grained decoupled personalization.}
We further evaluate \modelname's ability to perform fine-grained concept disentanglement\cite{chen2023disenbooth,cai2023decoupled,li2023generate,motamed2023lego}, where multiple attributes or objects within a reference image must be independently extracted and recombined based on textual instruction. This task is particularly challenging when the reference contains visually or semantically entangled elements.
As shown in Fig.~\ref{fig:complex-customize}, we observe that \modelname often struggles to decouple closely associated concepts, leading to semantic leakage or inconsistencies across generated results. In the left column (a1, a2), although the prompt is kept constant, the model incorrectly includes additional concepts in both outputs—such as generating a yellow helmet and a white pair of headphones not mentioned in the prompt, or using the wrong sunglasses color. This indicates that residual associations from the reference image influence the final output, violating prompt constraints.
In the right image (b), the prompt describes a multi-object spatial hierarchy: the mouse is placed on a plastic coffee bottle, which is placed on a box. While the model generally respects the described structure, the color of the mouse’s scroll wheel is altered in the generated result, indicating the model’s tendency to modify irrelevant attributes during complex recompositions.
These results suggest that while \modelname can handle multi-concept prompts at a high level, its ability to isolate and control fine-grained attributes in densely composed scenes is still limited.

Through our comprehensive evaluation of various personalization tasks, we find that \modelname demonstrates notable capabilities in identity preservation, stylistic transfer, and flexible compositional reasoning. In particular, the model performs well in scenarios where the personalized entity is visually distinct and the prompt is structurally simple.
However, several limitations persist. First, the model occasionally fails to maintain fine-grained consistency across attributes (e.g., color, accessories, pose details), especially under complex spatial compositions or multi-stage instructions. Second, when disentangling densely co-occurring elements within a single reference image, \modelname tends to produce semantic entanglement, leaking unintended concepts into the output. Finally, in cases requiring precise reference-based control (e.g., ordinal alignment, interaction binding), its understanding of prompt-referent mapping remains unstable.

Overall, \modelname shows strong potential in personalization-oriented generation, but its current performance is still bounded by limitations in structure disentanglement, attribute fidelity, and precise prompt grounding. Future improvements may require explicit memory modules, better concept binding mechanisms, or fine-tuned alignment between image and textual referents.

\clearpage

%\begin{figure}[h]
%    \centering
%    \includegraphics[width=1.0\linewidth]{figures/complex-customize2}
%    \caption[Sec~\ref{sec:personalization}: Complex Image Customization]{}
%    \label{fig:complex-customize2}
%\end{figure}

\clearpage

\subsubsection{Spatial Control}
\label{sec:spatial_control}
Spatial control tasks aim to guide image generation using explicit spatial constraints, such as layout, sketch, depth map, pose, and multiple-condition combination. These tasks evaluate the model's ability to understand spatial priors, align multimodal conditions, and preserve relative geometric structures in the generation process. We follow the general inference paradigm in recent controllable generation studies~\cite{zhang2023adding,kim2023diffblender,hu2023cocktail,qin2023unicontrol,mou2023t2i,balaji2022ediffi}, and systematically test \modelname across several typical spatial guidance modalities.

\begin{figure}[h]
    \centering
    \includegraphics[width=1.0\linewidth]{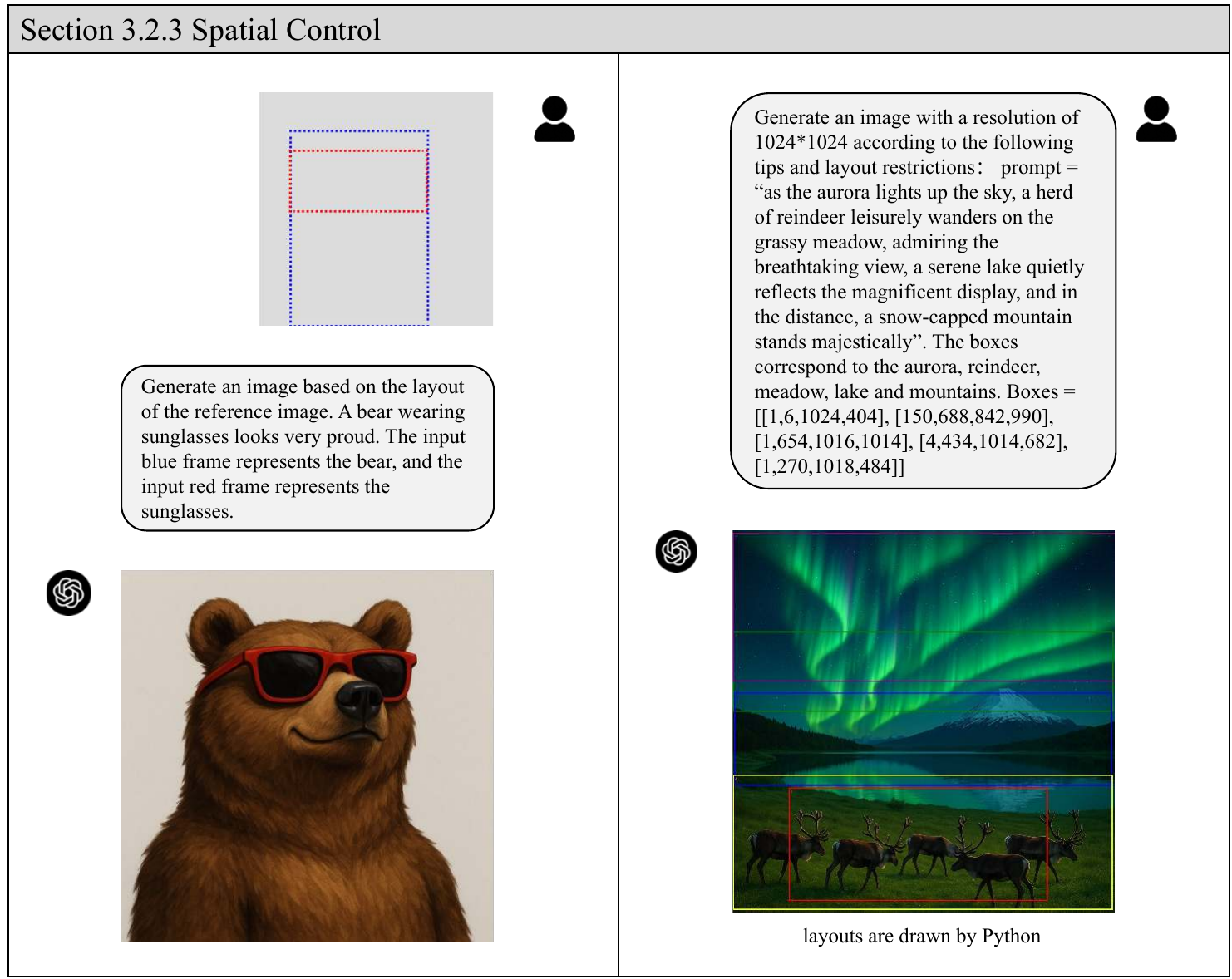}
    \caption[Sec~\ref{sec:spatial_control}: Layout-to-image with Visual and Textual Conditions]{Examples of layout-to-image results generated by \modelname.}
    \label{fig:layout-text}
\end{figure}

\textbf{Layout-to-image generation.}  
In layout-based spatial control~\cite{li2023gligen,ham2023modulating,xue2023freestyle,zheng2023layoutdiffusion,jia2023ssmg}, the model is required to generate an image conditioned on both bounding box layout and corresponding textual labels. As shown in Fig.~\ref{fig:layout-text}, \modelname is able to align objects with their designated positions and labels (e.g., “bear with sunglasses”) under simple geometric constraints. However, when handling complex scenes with textual conditions, positional drift can still be observed.

\begin{figure}[h]
    \centering
    \includegraphics[width=1.0\linewidth]{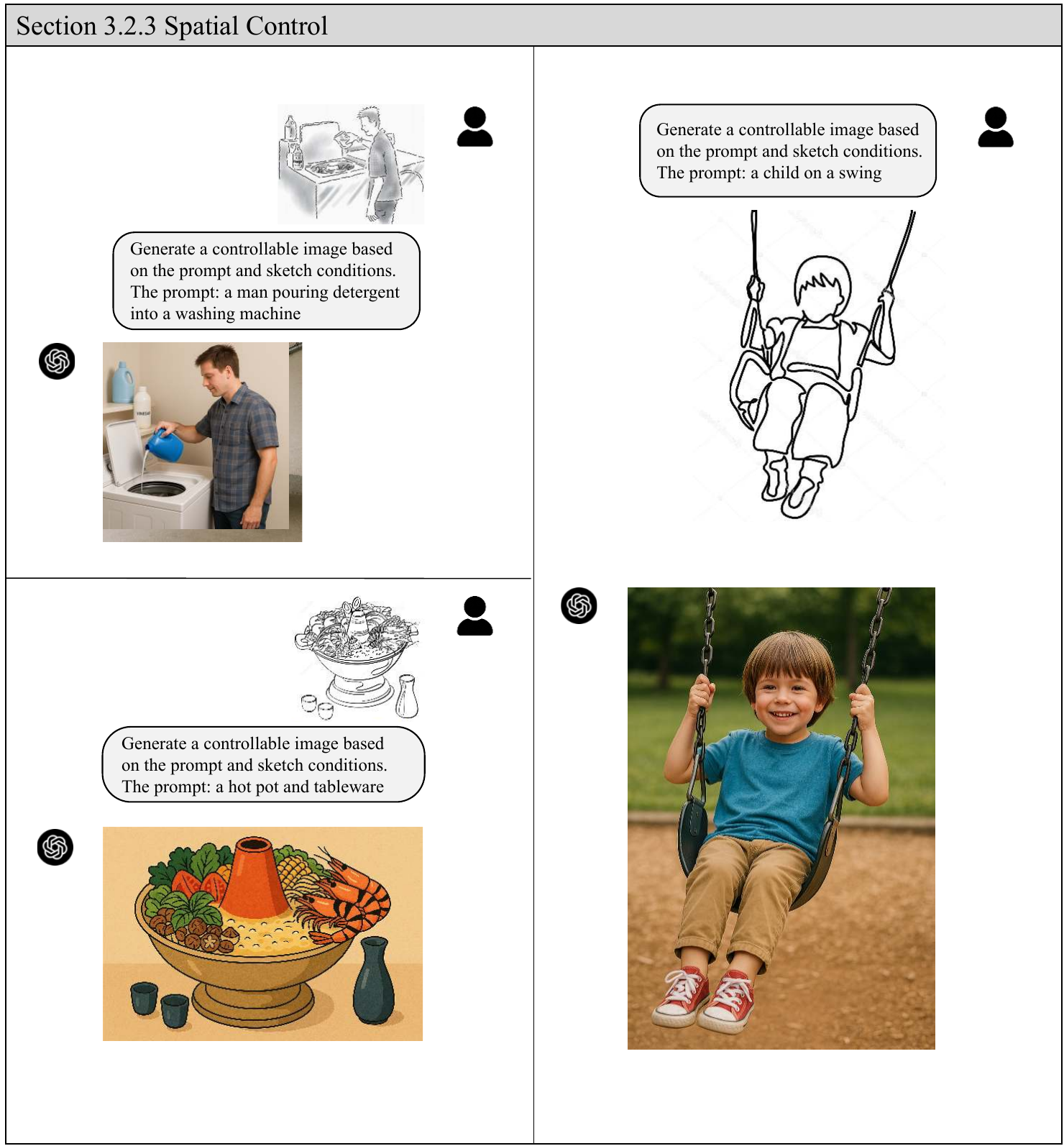}
    \caption[Sec~\ref{sec:spatial_control}: Sketch-to-image Generation]{Examples of sketch-to-image results generated by \modelname.}
    \label{fig:sketch}
\end{figure}

\textbf{Sketch-based generation.}  
In sketch-to-image generation, the model is guided by rough structural outlines, such as hand-drawn contours or simplified line maps, along with accompanying textual prompts. While \modelname demonstrates strong semantic understanding and is able to produce visually plausible and contextually relevant content, it often fails to maintain strict spatial alignment with the input sketch. For instance, in the upper-left example of Fig.~\ref{fig:sketch}, the generated washing machine deviates notably from the sketched shape, exhibiting different proportions and design. In the lower-left example, the orientation of the shrimp inside the pot is reversed compared to the sketch. These discrepancies suggest that although \modelname leverages sketch cues as high-level guidance, it lacks the geometric fidelity required for precise spatial control, especially for fine-grained structures.

\begin{figure}[h]
    \centering
    \includegraphics[width=1.0\linewidth]{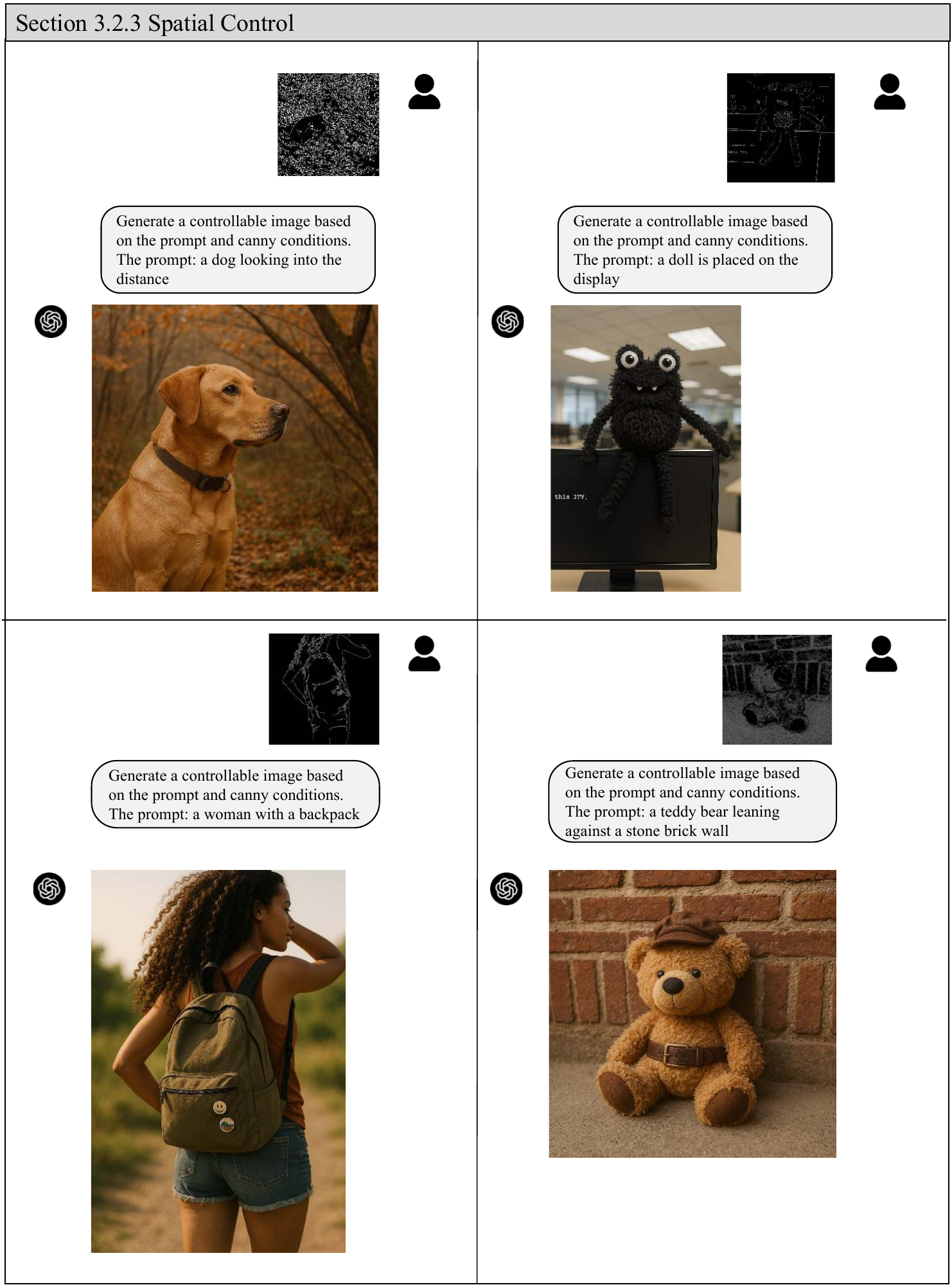}
    \caption[Sec~\ref{sec:spatial_control}: Canny-to-image Generation]{Examples of canny-to-image results generated by \modelname.}
    \label{fig:canny}
\end{figure}

\textbf{Canny-to-image generation.}  
Edge-based generation with canny maps is widely used to provide coarse structural control in synthesis tasks. In this setting, \modelname receives a low-level edge map alongside a textual prompt, and is expected to respect the contour constraints during image generation. As illustrated in Fig.~\ref{fig:canny}, \modelname captures the overall composition and generates semantically meaningful content, but frequently introduces elements not supported by the control input. For example, in the lower-left image, the generated figure includes a head that does not appear in the input canny sketch, and the orientation of the backpack deviates significantly from the edge shape. These observations suggest that while the model can use edge information as a loose guide, it tends to hallucinate plausible but uncontrolled details, revealing limited adherence to strict spatial boundaries under weak supervision.

\begin{figure}[h]
    \centering
    \includegraphics[width=1.0\linewidth]{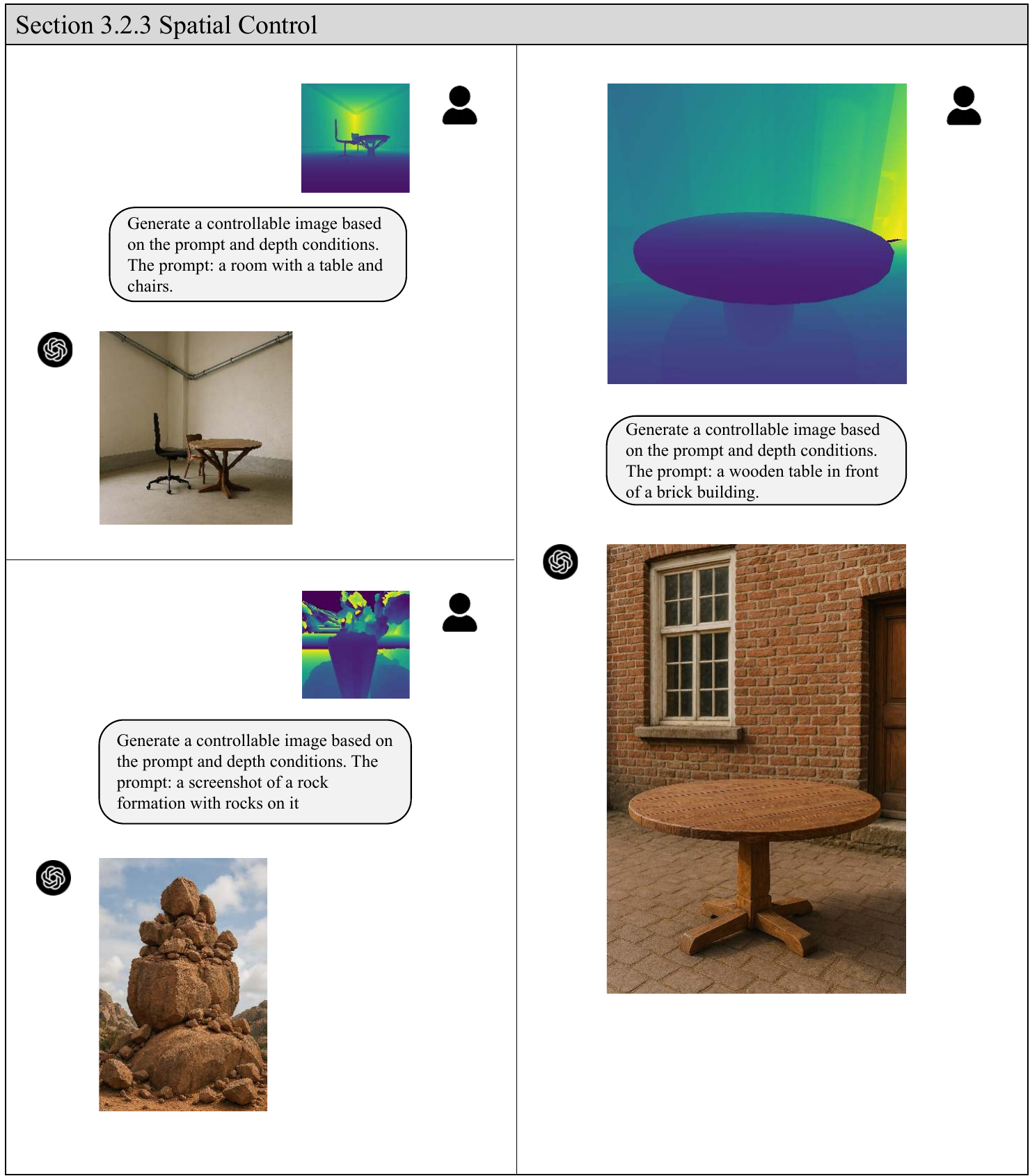}
    \caption[Sec~\ref{sec:spatial_control}: Depth-to-image Generation]{Examples of depth-to-image results generated by \modelname.}
    \label{fig:depth}
\end{figure}

\textbf{Depth-to-image generation.}  
Depth conditioning is designed to guide image generation with spatial hierarchy and foreground-background separation cues. Ideally, a model should interpret the depth map to preserve relative object positioning and scene structure. However, as shown in Fig.~\ref{fig:depth}, \modelname exhibits inconsistent performance. While the top-left example shows a reasonable understanding of depth—maintaining plausible layering and perspective—the other two results reveal clear mismatches. In particular, the bottom-left image fails to reflect the intended depth gradient entirely, with foreground and background elements appearing flattened or confused. These cases suggest that \modelname has limited ability to extract and utilize geometric depth relationships from the input, especially when the depth map lacks clear semantic context.

\begin{figure}[h]
    \centering
    \includegraphics[width=1.0\linewidth]{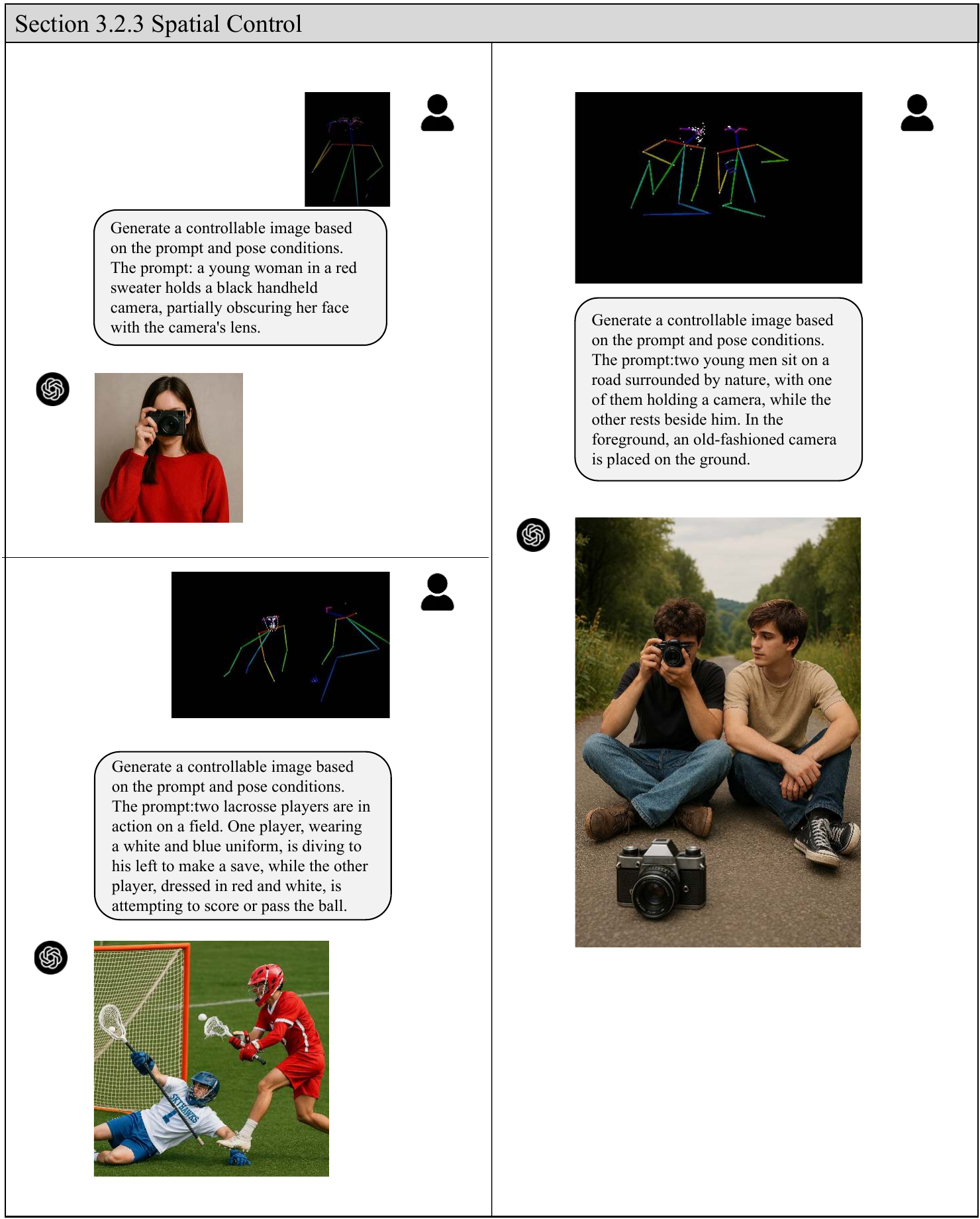}
    \caption[Sec~\ref{sec:spatial_control}: Pose-to-image Generation]{Examples of pose-to-image results generated by \modelname.}
    \label{fig:pose}
\end{figure}

\textbf{Pose-to-image generation.}  
In pose-conditioned generation, the model is expected to synthesize human figures that align closely with provided pose skeletons, capturing limb orientation, joint position, and overall body configuration. However, as shown in Fig.~\ref{fig:pose}, \modelname exhibits limited precision in this task. While it generally respects the number of people and the rough action described by the pose input, it frequently fails to accurately match the detailed body positions and limb articulations. Misalignments such as incorrect arm angles, swapped leg positions, or distorted body structures are commonly observed. These results suggest that although \modelname has basic awareness of pose input, it struggles to convert sparse keypoint guidance into spatially faithful and anatomically correct visual representations.

\begin{figure}[h]
    \centering
    \includegraphics[width=1.0\linewidth]{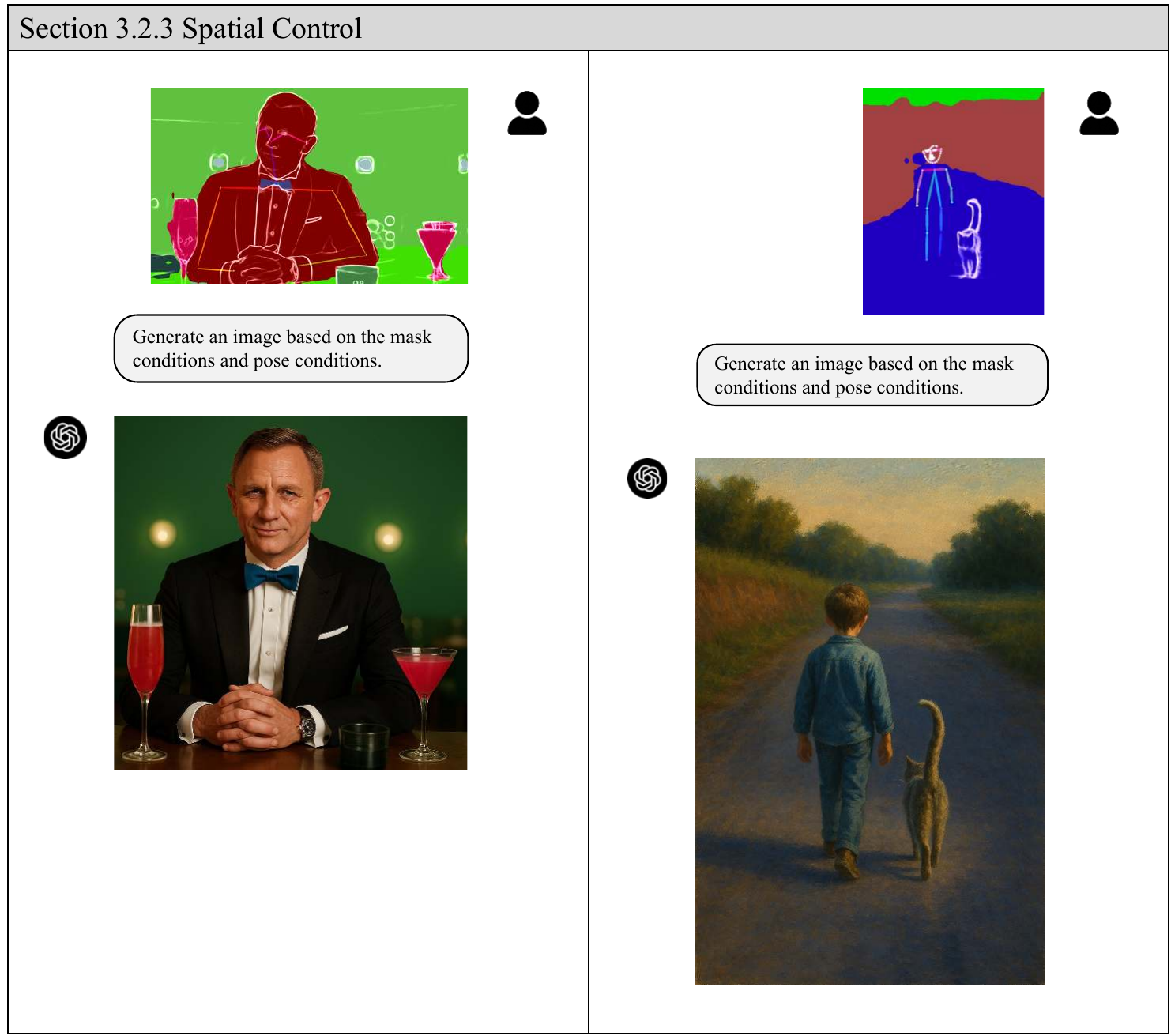}
    \caption[Sec~\ref{sec:spatial_control}: Spatial Control with Multiple Conditions]{Examples of multi-conditioned spatial control results generated by \modelname.}
    \label{fig:spatial-multiple}
\end{figure}

\textbf{Multi-condition spatial control.}  
In more advanced settings, we provide \modelname with multiple spatial conditions simultaneously. As shown in Fig.~\ref{fig:spatial-multiple}, the model can integrate these heterogeneous priors and produce spatially grounded compositions. While coarse alignment is preserved, fine-grained consistency between multiple guidance types is occasionally inconsistent, revealing limitations in fusion robustness.

Across the evaluated spatial control tasks, \modelname demonstrates a moderate ability to incorporate spatial conditions into image generation. It performs relatively well in simple layout-to-image scenarios and contour-based sketch generation, where structural prompts are clear and semantically aligned. However, in more structured or fine-grained tasks such as depth-based synthesis, pose replication, and edge-controlled generation, the model frequently fails to maintain strict spatial alignment. Common issues include hallucinated content not present in the control input, incorrect object orientation, and poor adherence to depth or joint structure. Additionally, in multi-modal control settings (e.g., pose + mask), the fusion of different spatial modalities remains unstable, leading to partial or conflicting alignment.
These findings suggest that \modelname treats spatial inputs more as soft guidance rather than strict geometric constraints. While it can use these signals to influence composition, it often prioritizes semantic plausibility or stylistic fluency over spatial fidelity.

\clearpage

\clearpage

\subsubsection{Image Inpainitng \& Outpainting}
\label{sec:inpainting}
Image inpainting\cite{Corneanu} and outpainting\cite{Liu_Wang_Qian_Wang_Rui_2024,Ai_Cao_Lu_Chen_Ma_Zhou_Kim_Hui_Wang_2024,Tang_2024} evaluate a model’s ability to fill in missing regions or extend image boundaries based on contextual understanding. These tasks require semantic coherence, texture consistency, and spatial continuity, especially when dealing with large or complex masked areas.

\begin{figure}[h]
    \centering
    \includegraphics[width=1.0\linewidth]{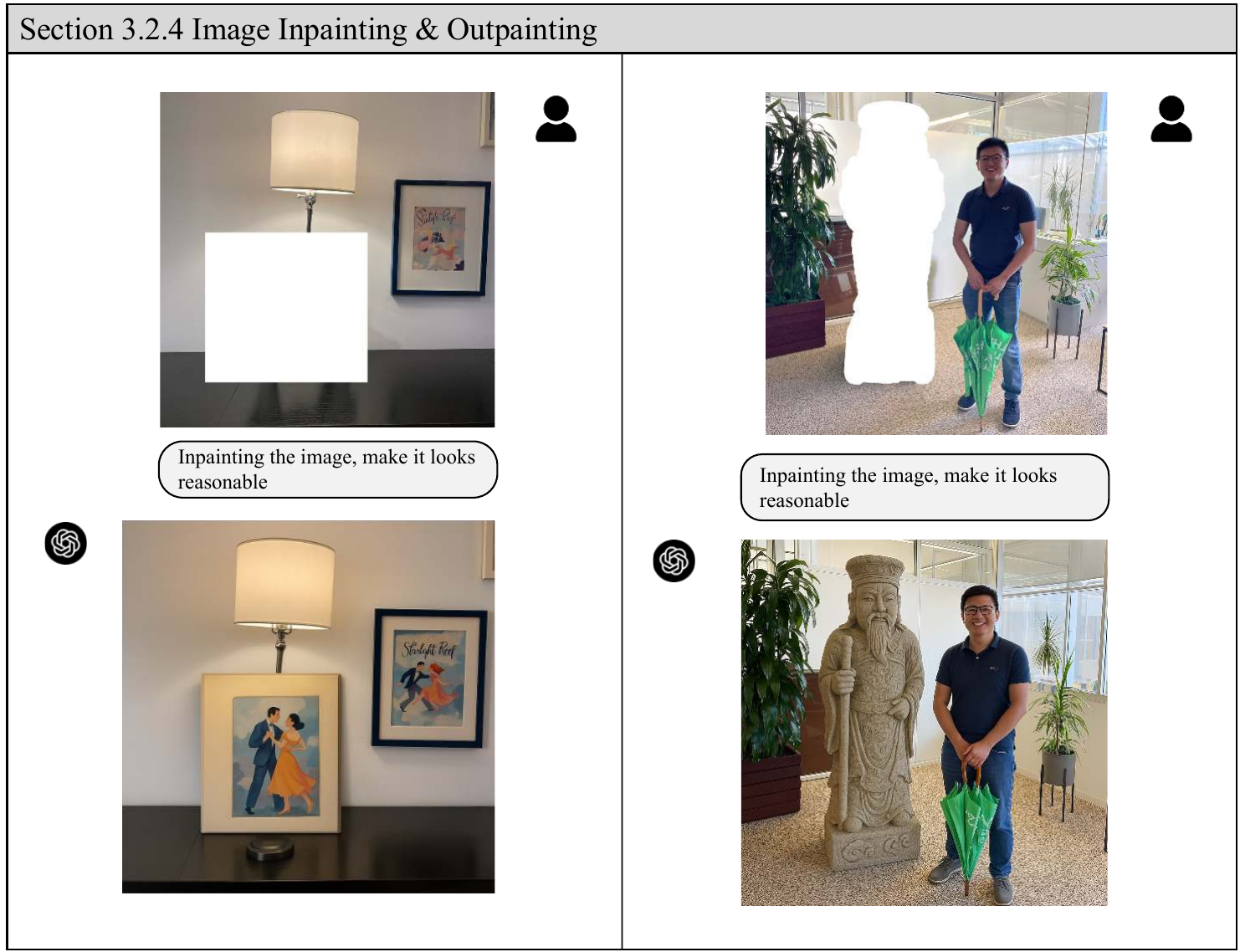}
    \caption[Sec~\ref{sec:inpainting}: Image Inpainting]{Examples of image inpainting generated by \modelname.}
    \label{fig:inpainting}
\end{figure}

\textbf{Image Inpainting.}  
As shown in Fig.~\ref{fig:inpainting}, \modelname demonstrates the ability to fill masked regions with plausible content. However, we observe that it tends to alter the original image content outside the masked area, leading to structural and textural inconsistencies. In the left example, the lighting across the entire image changes, and the picture frame on the right wall is replaced with a different image. Similarly, in the right example, both the human subject and the background are modified beyond the inpainted region. This suggests that while \modelname understands the global context well, it lacks fine-grained control to preserve the unmasked areas faithfully during inpainting.

\begin{figure}[h]
    \centering
    \includegraphics[width=1.0\linewidth]{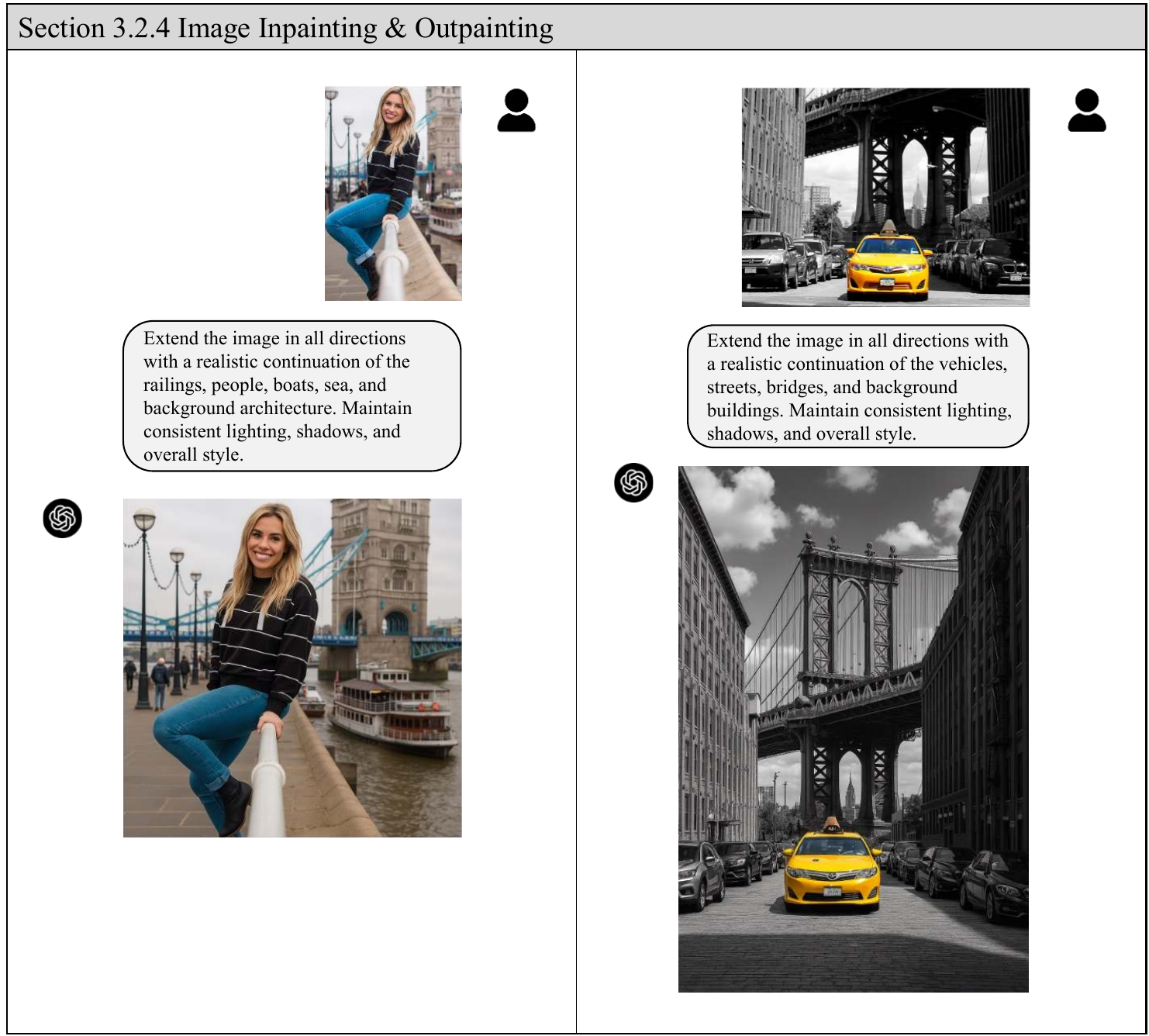}
    \caption[Sec~\ref{sec:inpainting}: Image Outpainting]{Examples of image outpainting generated by \modelname.}
    \label{fig:outpainting}
\end{figure}

\textbf{Image Outpainting.}  
For outpainting (Fig.~\ref{fig:outpainting}), \modelname shows an impressive ability to extend images beyond their original boundaries while maintaining global style, color tone, and lighting conditions. The extended content is contextually coherent and stylistically aligned. However, inconsistencies in texture resolution and object detail are still apparent. The generated regions may look visually convincing at first glance but fail to maintain strict alignment in material or surface continuity, revealing limitations in local-level synthesis fidelity.

Overall, \modelname handles global structure and visual harmony well in both tasks but struggles with strict preservation of content and fine texture matching, particularly in inpainting.

\clearpage

\clearpage

\subsubsection{Story Visualization}
\label{sec:story}
Story visualization tasks require the model to generate a coherent sequence of images based on a narrative input, maintaining consistency in character appearance, scene context, and temporal progression\cite{zhou2024storydiffusion,he2025dreamstoryopendomainstoryvisualization,chen2024mangadiffusion,wu2024diffsensei,gong2023talecrafter}. These tasks are inherently multimodal and test a model's ability to track evolving information while producing temporally aligned visual outputs.

\begin{figure}[h]
    \centering
    \includegraphics[width=1.0\linewidth]{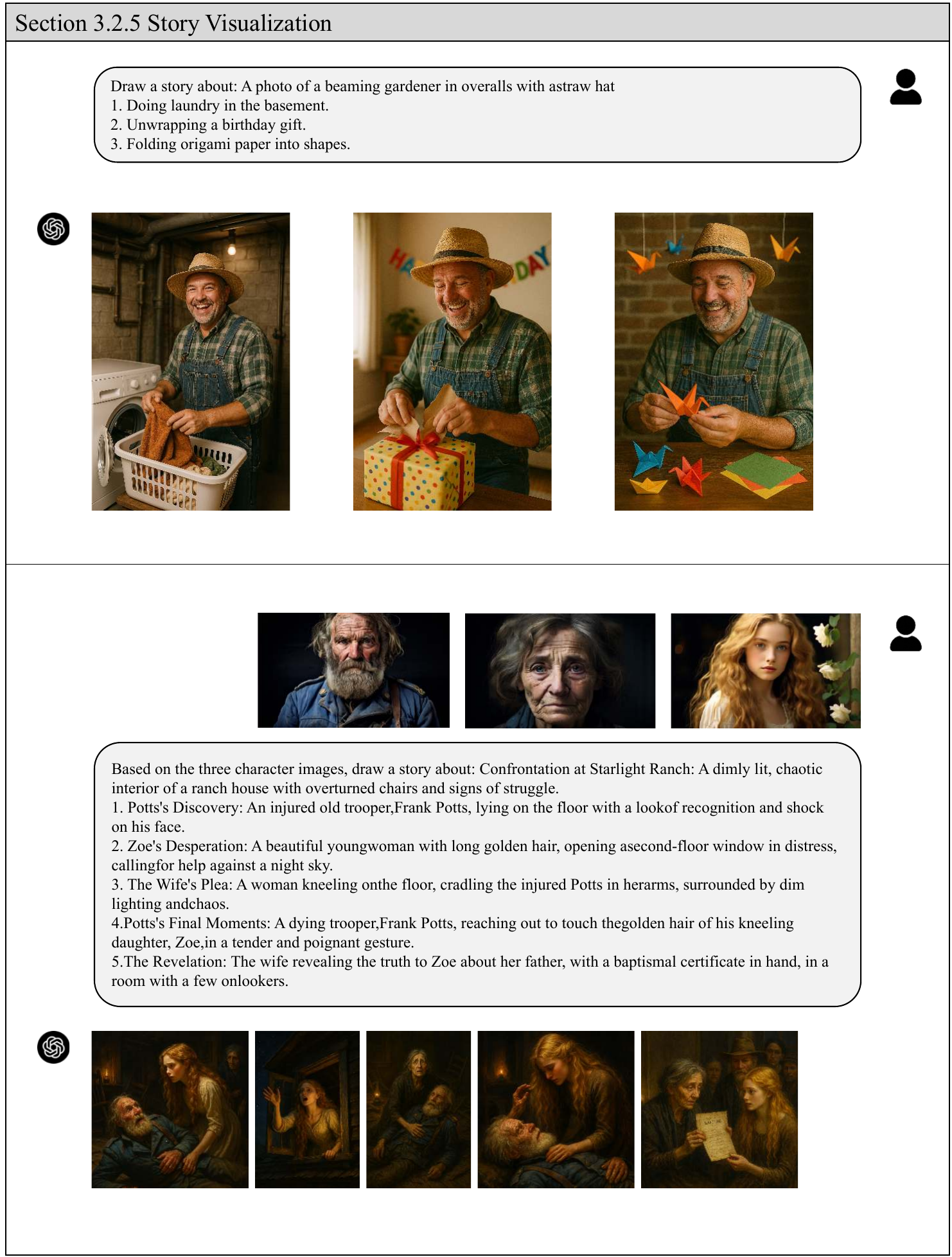}
    \caption[Sec~\ref{sec:story}: Story Visualization]{Examples of story visualization generated by \modelname.}
    \label{fig:story1}
\end{figure}

In Fig.~\ref{fig:story1}, \modelname is able to follow the story flow and generate a sequence that aligns with the described events. Character identity and scene continuity are reasonably well maintained across the visual sequence, suggesting strong support for short-range consistency when reference images are available.

\begin{figure}[h]
    \centering
    \includegraphics[width=1.0\linewidth]{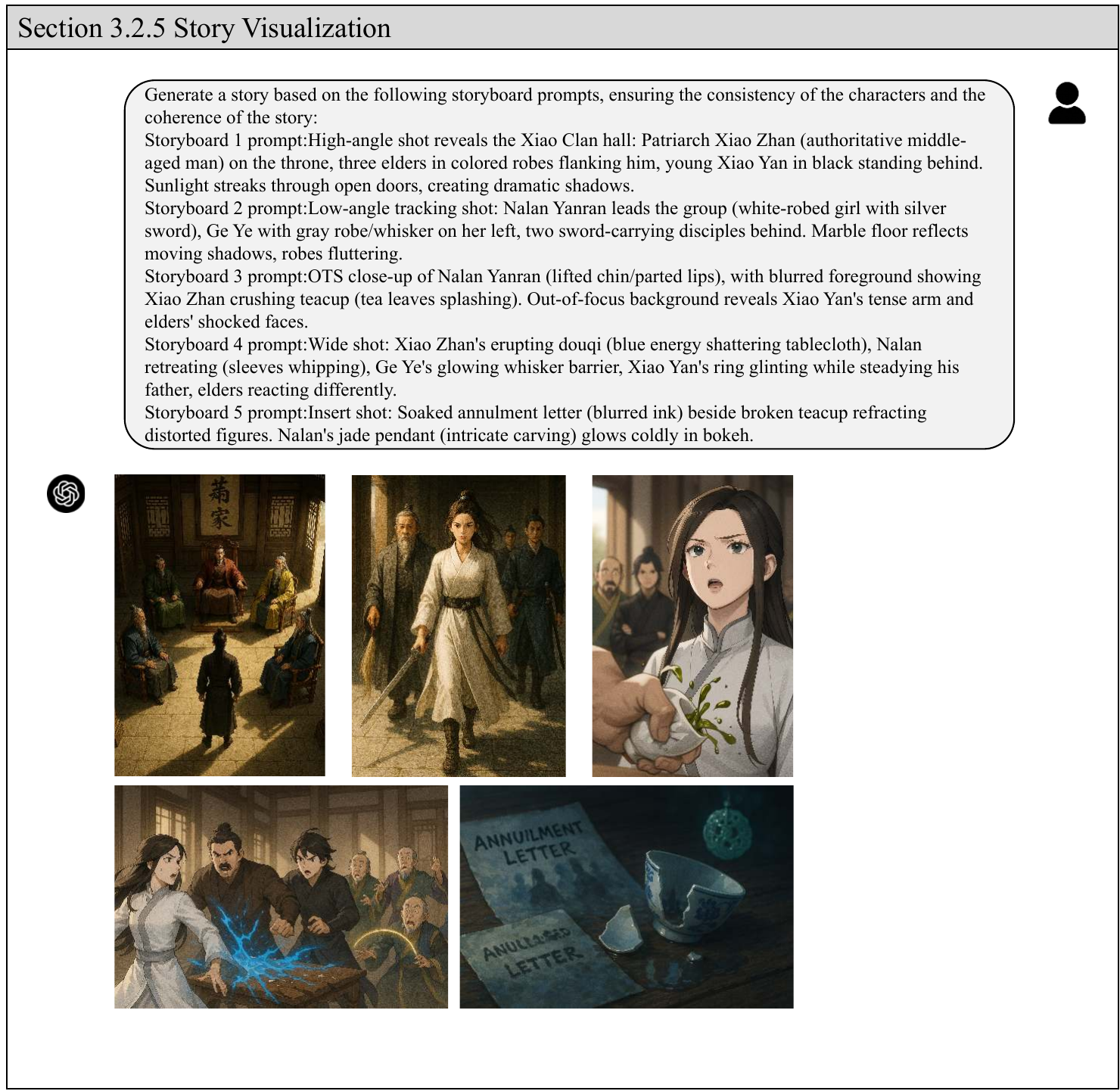}
    \caption[Sec~\ref{sec:story}: Story Visualization]{Additional examples of story visualization generated by \modelname.}
    \label{fig:story2}
\end{figure}

In Fig.~\ref{fig:story2}, \modelname still performs well in aligning generated content with story semantics. However, inconsistencies begin to emerge, particularly in visual style. As shown in the third and fourth frames, the visual style changes from the other frames. This indicates limitations in long-range style memory and visual coherence across temporally adjacent images.

Overall, \modelname demonstrates strong capability in aligning visual content to story instructions and preserving local character continuity. Nonetheless, consistent style retention and stable progression across multiple frames remain challenging, especially when no visual guidance is provided.

\clearpage

\clearpage

\subsubsection{Virtual Try-on}
\label{sec:vt}
Virtual try-on aims to synthesize realistic images of a target person wearing garments from a reference image\cite{zhu2023tryondiffusion,han2018viton,wang2018cpvton,han2021vitonhd,he2022styleposegan,pisanello2020dresscode}. This task evaluates the model’s ability to disentangle appearance from clothing, retain person identity, and accurately transfer fine-grained garment attributes such as texture, color, and structure.

\begin{figure}[h]
    \centering
    \includegraphics[width=1.0\linewidth]{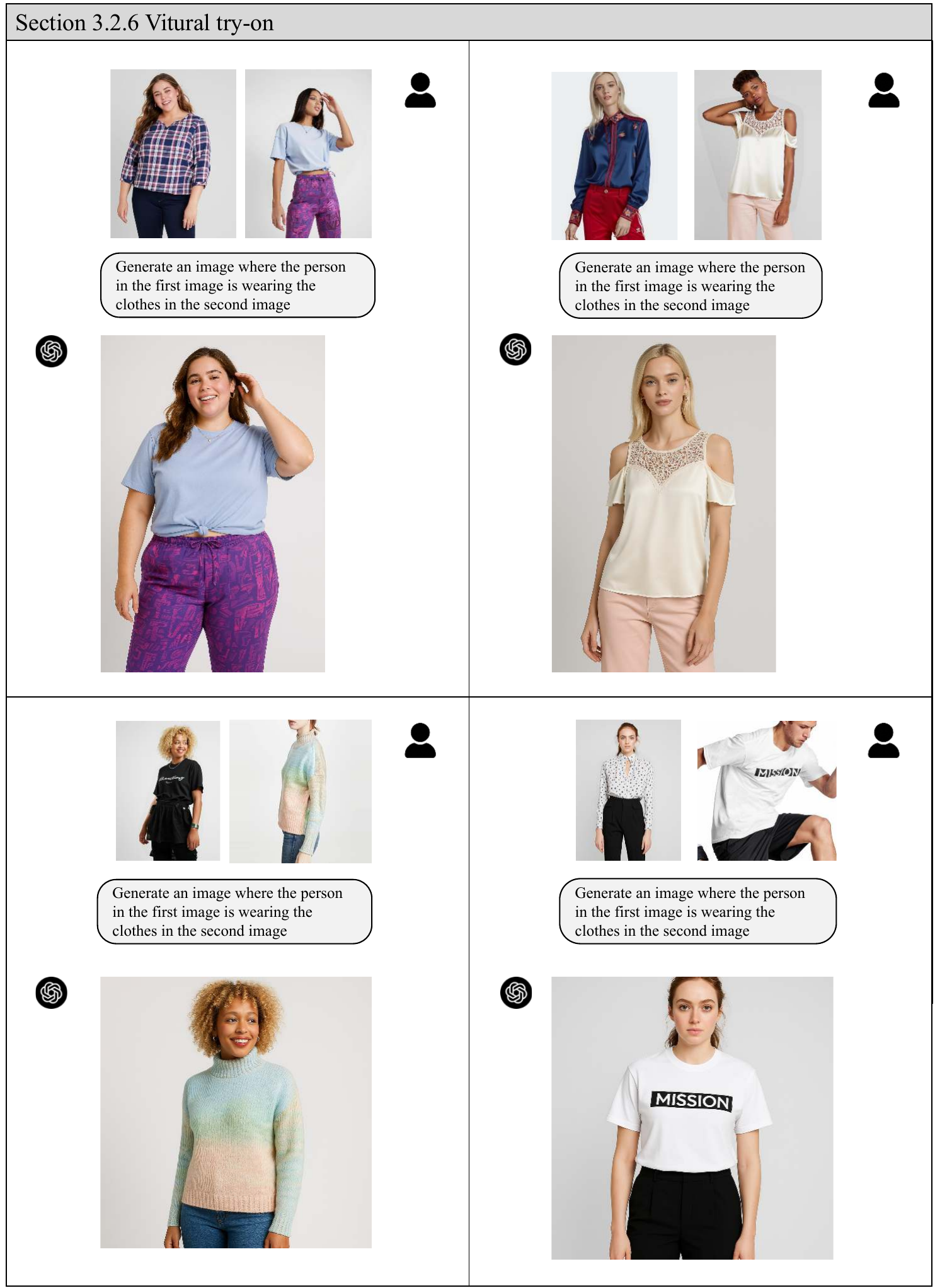}
    \caption[Sec~\ref{sec:vt}: Virtual Try-on]{Examples of virtual try-on generated by \modelname.}
    \label{fig:tryon}
\end{figure}

As shown in Fig.~\ref{fig:tryon}, \modelname demonstrates strong performance in this task. The generated outputs not only preserve the identity and pose of the target individual but also successfully transfer the visual features of the garments, including sleeve length, neckline design, printed logos, and material patterns. Even subtle stylistic details are faithfully reproduced, such as the gradient texture in sweaters and brand-specific typography on shirts.

Overall, these results indicate that \modelname exhibits a strong capability in clothing transfer, producing realistic and identity-preserving virtual try-on results with high visual fidelity.

\clearpage

\clearpage

\clearpage
\subsection{Low-level Image Generation}
\label{sec:ir}
Low-level image generation tasks focus on enhancing or restoring the quality of images at the pixel level. These tasks test \modelname’s capability in fine-grained visual synthesis, including recovery of detail and color fidelity in degraded images.

\subsubsection{Image Super-Resolution}
\label{sec:sr}
Image super-resolution (SR) aims to reconstruct high-resolution (HR) images from low-resolution (LR) inputs, enhancing visual quality by improving sharpness, restoring details, and generating realistic textures\cite{agustsson2017ntire,martin2001database,liang2021swinir,wang2018esrgan}. This task is fundamental in image enhancement and widely applicable in real-world scenarios such as photography, surveillance, and remote sensing.

\begin{figure}[h]
    \centering
    \includegraphics[width=1.0\linewidth]{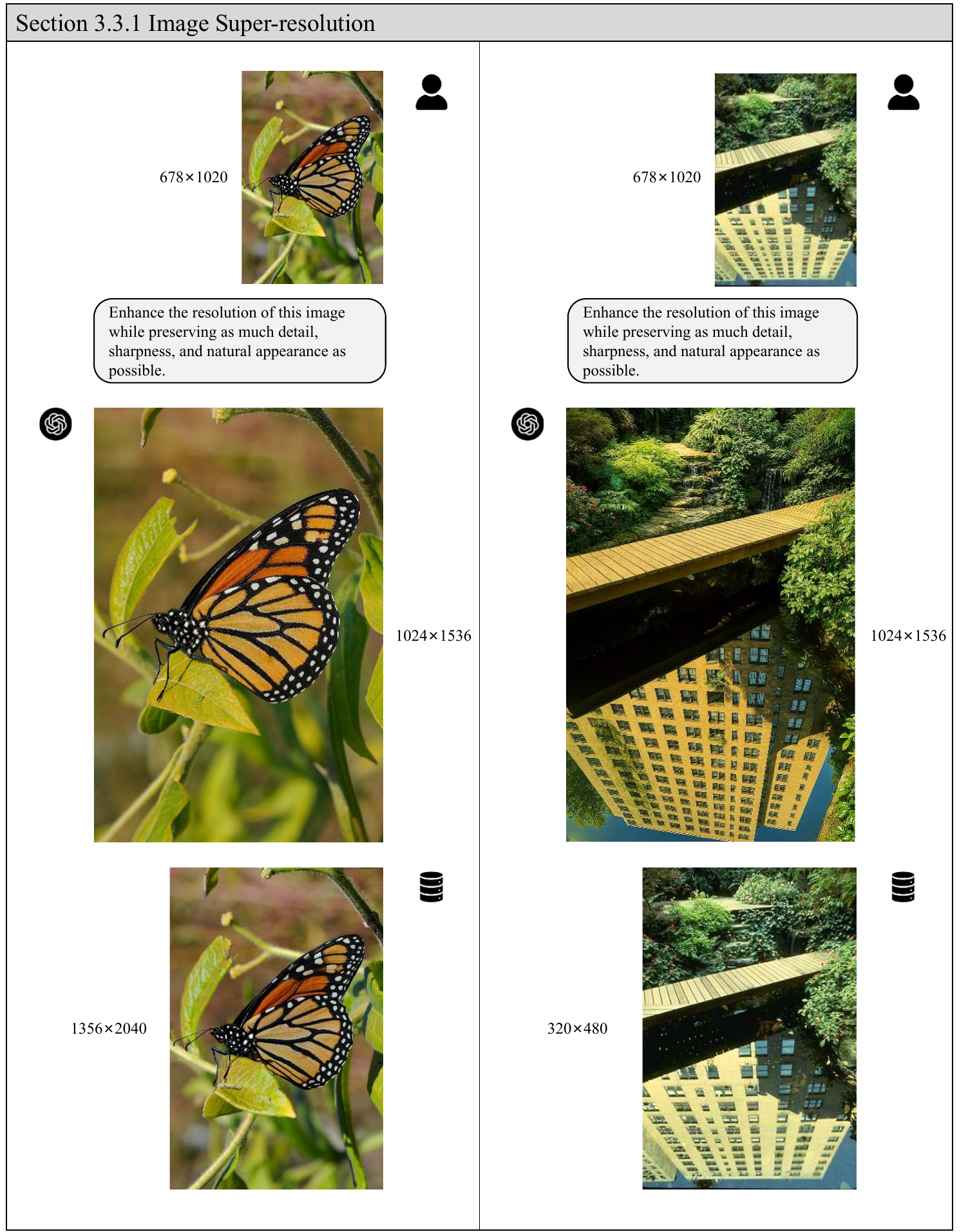}
    \caption[Sec~\ref{sec:sr}: Image Super-Resolution]{Examples of image super-resolution results generated by \modelname. The model can enhance low-resolution images with improved sharpness, detail recovery, and natural appearance. However, the maximum output resolution is constrained by the model's generation limits.}
    \label{fig:sr1}
\end{figure}

\begin{figure}[h]
    \centering
    \includegraphics[width=1.0\linewidth]{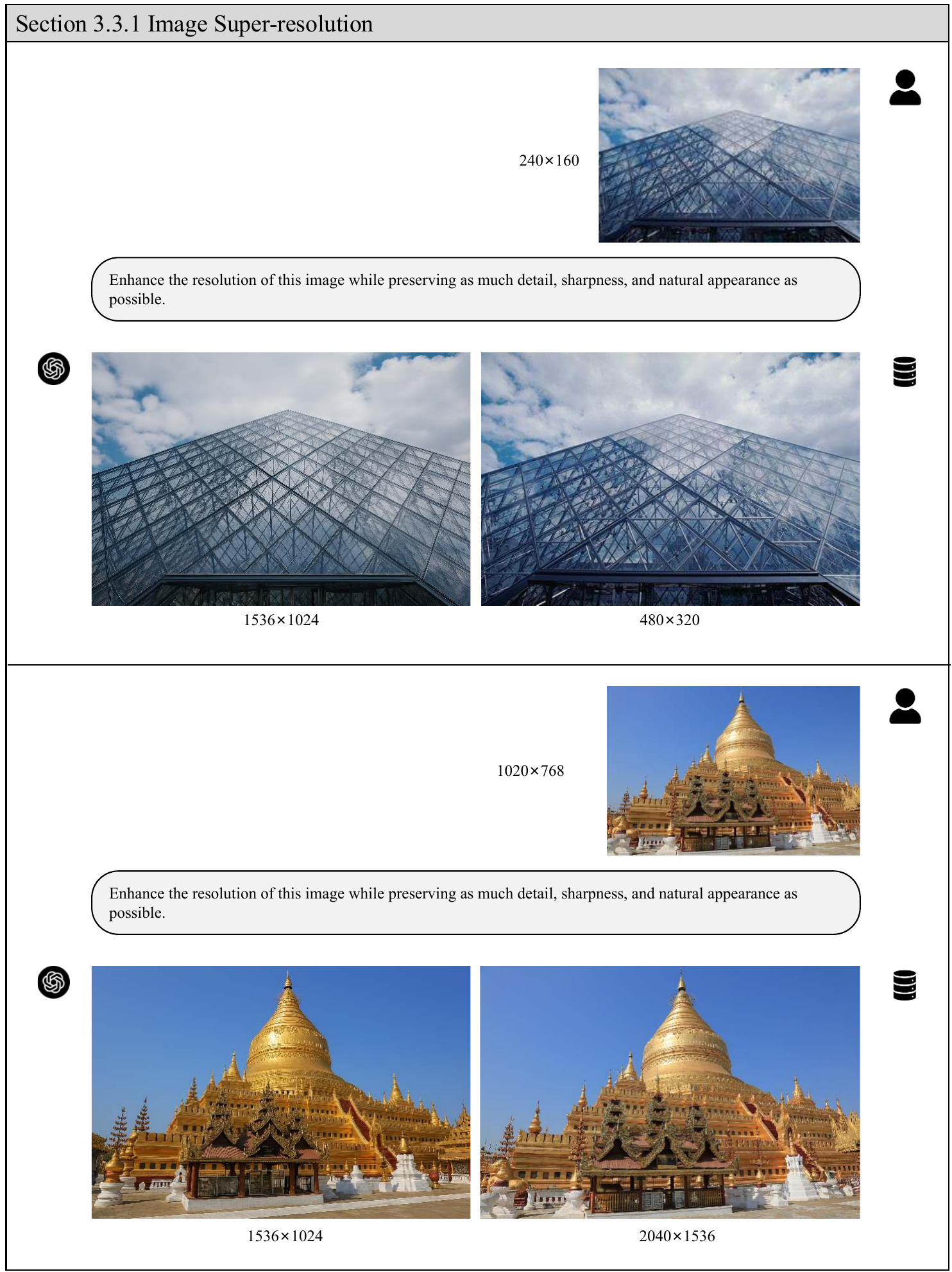}
    \caption[Sec~\ref{sec:sr}: Image Super-Resolution]{Additional examples of image super-resolution results generated by \modelname.}
    \label{fig:sr2}
\end{figure}

As shown in Fig.\ref{fig:sr1} and Fig.\ref{fig:sr2}, \modelname demonstrates good capability in image super-resolution. The model can effectively enhance the clarity of LR images and recover fine details, making the visual results more appealing and closer to natural images.

However, the generation process is still constrained by the model's maximum output resolution (as mentioned in Sec~\ref{sec:res}), which is limited to a few fixed sizes (e.g., $1024\times1536$, $1536\times1024$). While the model is able to improve image quality within this resolution boundary, it cannot directly produce ultra-high-resolution images beyond its generation limit.

Overall, \modelname shows promising performance in image super-resolution tasks, striking a balance between detail restoration, sharpness enhancement, and content consistency under resolution constraints.

\clearpage

\clearpage
\subsubsection{Image Colorization}
\label{sec:color}
Image colorization aims to convert grayscale images into plausible and aesthetically pleasing color versions\cite{zhang2016colorful,vzeger2021grayscale,nazeri2018image}, often conditioned on either general priors or user-specified color styles. This task evaluates the model’s ability to infer semantic content, preserve structure, and apply style-aware color mappings.

\begin{figure}[h]
    \centering
    \includegraphics[width=1.0\linewidth]{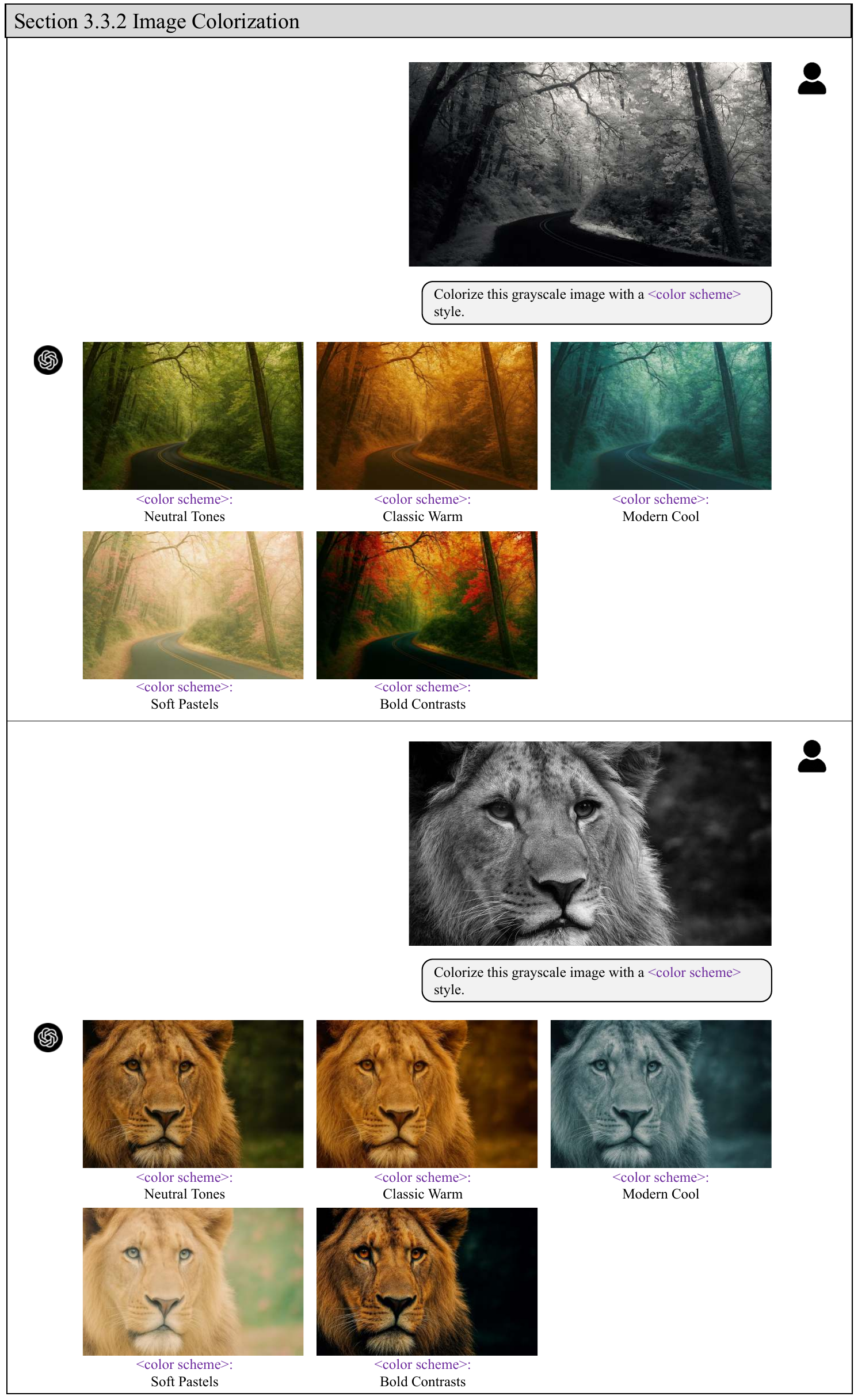}
    \caption[Sec~\ref{sec:color}: Image Colorization]{Exampes of image colorization generated by \modelname.}
    \label{fig:color1}
\end{figure}

\begin{figure}[h]
    \centering
    \includegraphics[width=1.0\linewidth]{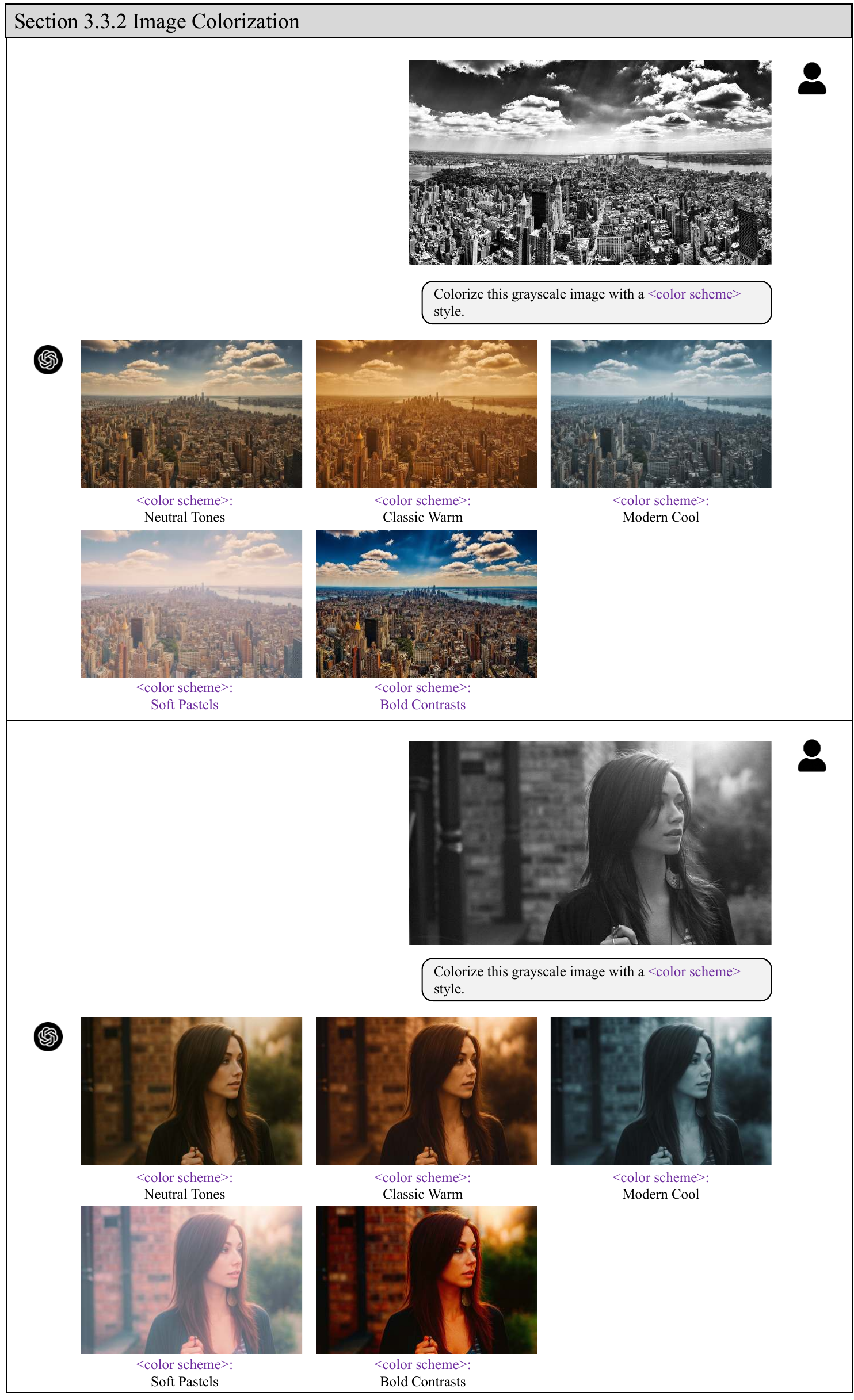}
    \caption[Sec~\ref{sec:color}: Image Colorization]{Additional examples of image colorization generated by \modelname.}
    \label{fig:color2}
\end{figure}

As shown in Fig.~\ref{fig:color1} and Fig.~\ref{fig:color2}, \modelname produces highly convincing colorized results under various stylistic instructions. Given different color schemes—such as “Neutral Tones,” “Classic Warm,” “Modern Cool,” “Soft Pastels,” and “Bold Contrasts”—the model consistently generates images with accurate semantic interpretations and clear stylistic alignment. For instance, warm tones are correctly applied to skin and wood materials under the “Classic Warm” style, while cooler tones are used for metallic or urban structures in the “Modern Cool” setting.
The model also demonstrates a strong capability to preserve spatial structures and avoid color bleeding or semantic mismatches. Across all evaluated results, \modelname exhibits excellent performance in aligning global color tone with regional object categories and overall visual harmony.

Overall, these results indicate that \modelname handles colorization tasks with both semantic understanding and stylistic control, making it a strong candidate for controllable grayscale-to-color image synthesis.

\clearpage

\clearpage

\subsubsection{Image Restoration}
\label{sec:image_restoration}
Image restoration includes removing noise, blur, or other artifacts from corrupted images. This task measures the model's ability to recover clean, high-quality visuals from degraded inputs. \modelname demonstrates a certain level of capability in image restoration tasks, effectively removing various degradations such as noise, blur, rain, or haze. However, we observe that the restored results may not always strictly align with the original input in terms of fine-grained details or textures. This indicates that the model tends to focus more on generating perceptually clean and plausible results rather than strictly preserving pixel-level accuracy.

\begin{figure}[h]
    \centering
	\includegraphics[width=1.0\linewidth]{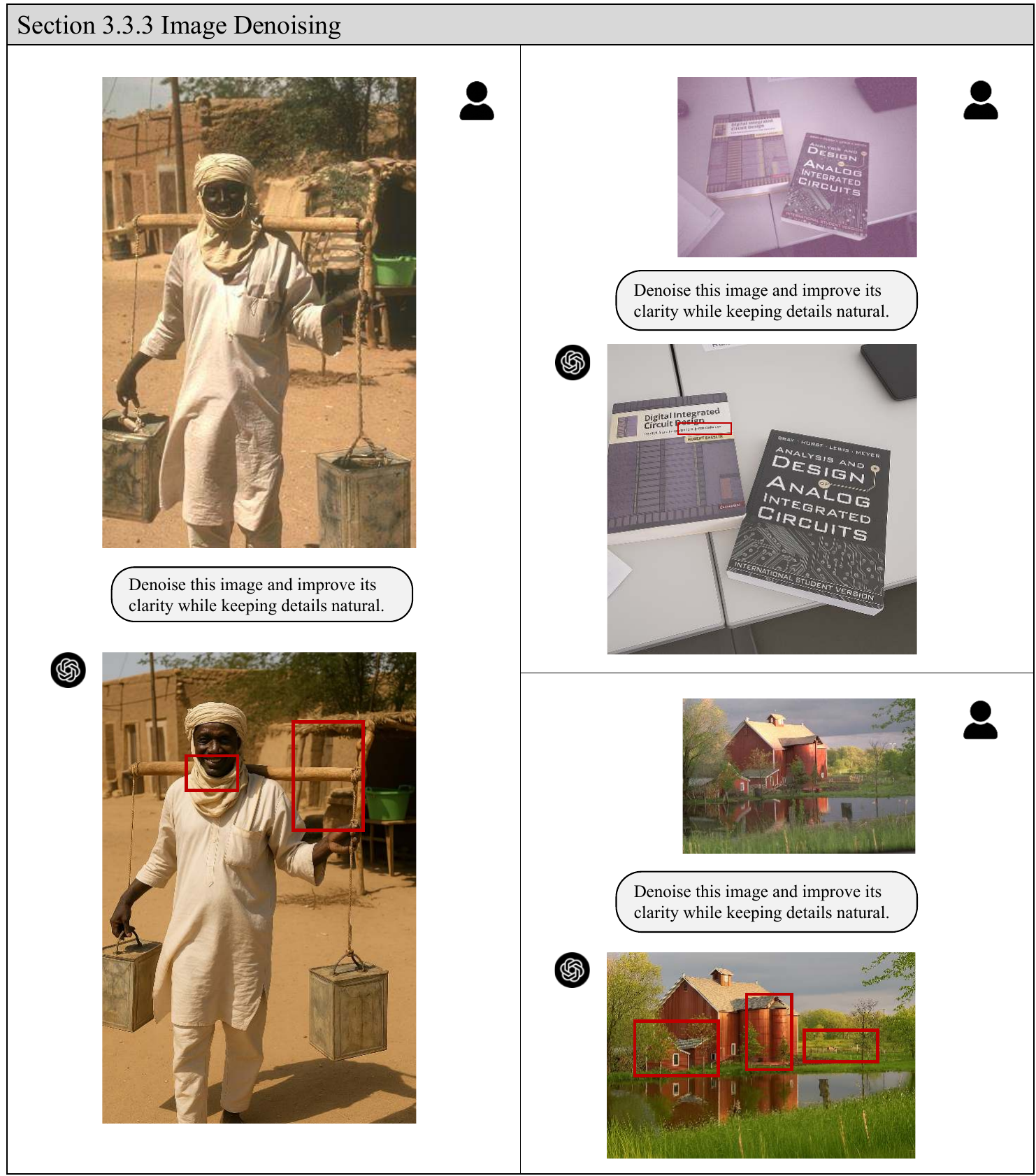}
%	\vspace{-1mm}
	\caption[Sec~\ref{sec:image_restoration}: Image Denoising]{Examples of \modelname performing image denoising. While the model is able to reduce noise and improve overall image clarity, it often alters fine-grained details or introduces slight distortions, as highlighted in the red boxes. This indicates a trade-off between denoising effectiveness and structural fidelity.}
	\label{fig:denoising}
\end{figure}

\textbf{Image Denoising} Image denoising refers to the task of removing random noise from corrupted images while preserving the underlying structure and visual details\cite{martin2001database}. Effective denoising demonstrates a model's ability to distinguish signal from noise and recover clean visual content with high fidelity\cite{marinescu2020bayesian,cai2021learning,ghasemabadi2024cascadedgaze}. Figure~\ref{fig:denoising} presents several representative examples of \modelname's performance on the image denoising task. The model is prompted to remove noise while preserving natural details across a variety of scenes, including portraits, documents, and landscapes. Overall, \modelname demonstrates a reasonable ability to suppress visual noise and enhance global clarity. For instance, in the case of the noisy book cover, the text becomes significantly more legible after denoising, indicating that the model successfully restores high-frequency details.

However, a closer inspection reveals that \modelname often alters certain local structures in the process. As highlighted by the red boxes, some fine-grained textures and object boundaries are unintentionally smoothed or distorted. In the portrait example (left), facial details and background elements deviate noticeably from the original, suggesting a degradation in structural fidelity. Similarly, in the landscape scene, the architectural components of the house exhibit shape inconsistencies post-denoising.

These observations suggest that while \modelname is capable of producing visually cleaner images, it does so by prioritizing perceptual enhancement over pixel-level accuracy. This behavior is consistent with its design as a vision-language model optimized for human-aligned generation, rather than precise signal restoration.

\textbf{Image Deblurring}
Image deblurring is a classical low-level vision task that aims to restore clear and sharp images from blurry inputs\cite{nah2017deep,rim2020real}. In this part, we evaluate \modelname on two typical types of deblurring tasks: motion deblurring\cite{li2023real,liu2022application,zhong2023blur} and defocus deblurring\cite{quan2021gaussian,quan2024deep,abuolaim2020defocus,abuolaim2021learning}. Motion deblurring targets images degraded due to object or camera movement, while defocus deblurring focuses on images blurred by the depth of field limitations or lens defocus.

\begin{figure}[h]
    \centering
    \includegraphics[width=1.0\linewidth]{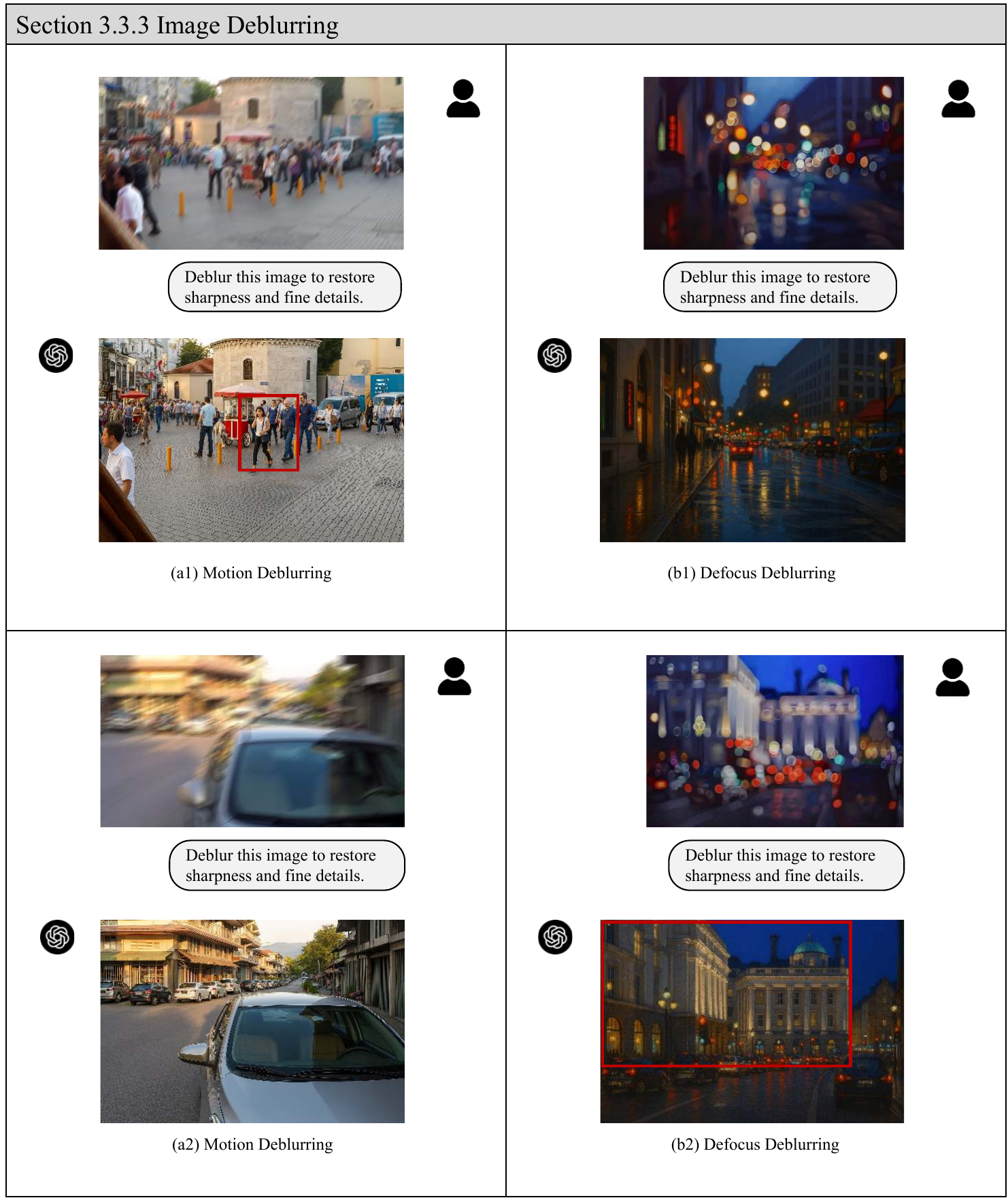}
    \caption[Sec~\ref{sec:image_restoration}: Image Deblurring]{Examples of image deblurring results generated by \modelname. The model is able to restore the overall structure and sharpness of the input images for both motion deblurring and defocus deblurring tasks. However, certain color deviations (e.g., building colors in b2) and fine detail inconsistencies (e.g., human appearance in a1) are observed compared to the original content.}
    \label{fig:deblurring}
\end{figure}

As shown in Fig.~\ref{fig:deblurring}, \modelname demonstrates a promising ability to recover the overall structure, sharpness, and scene layout of the blurred input images. For motion deblurring (examples (a1) and (a2)), the model effectively restores the main objects and background clarity. Similarly, for defocus deblurring (examples (b1) and (b2)), the scene composition is generally well-aligned with the input, and the synthesized results present improved sharpness.

However, we also observe noticeable limitations. In some cases, the fine-grained details are not faithfully preserved. For example, in (a1), the restored human figures exhibit appearance inconsistencies compared to the original. Additionally, for defocus deblurring, although the scene structure is preserved, certain color deviations occur, as seen in (b2), where the building color differs significantly from the input image. These observations suggest that while \modelname is capable of perceptually restoring sharp images, its deblurring process tends to prioritize visual plausibility over strict content fidelity.

\textbf{Image Dehazing} 
Image dehazing aims to remove haze or fog from images, enhancing scene visibility while preserving natural colors and structural details. This task is crucial for improving image quality in low-visibility conditions, such as in autonomous driving, remote sensing, and outdoor photography\cite{liu2021synthetic,zheng2023curricular,wu2021contrastive,qin2020ffa}.

\begin{figure}[h]
    \centering
    \includegraphics[width=1.0\linewidth]{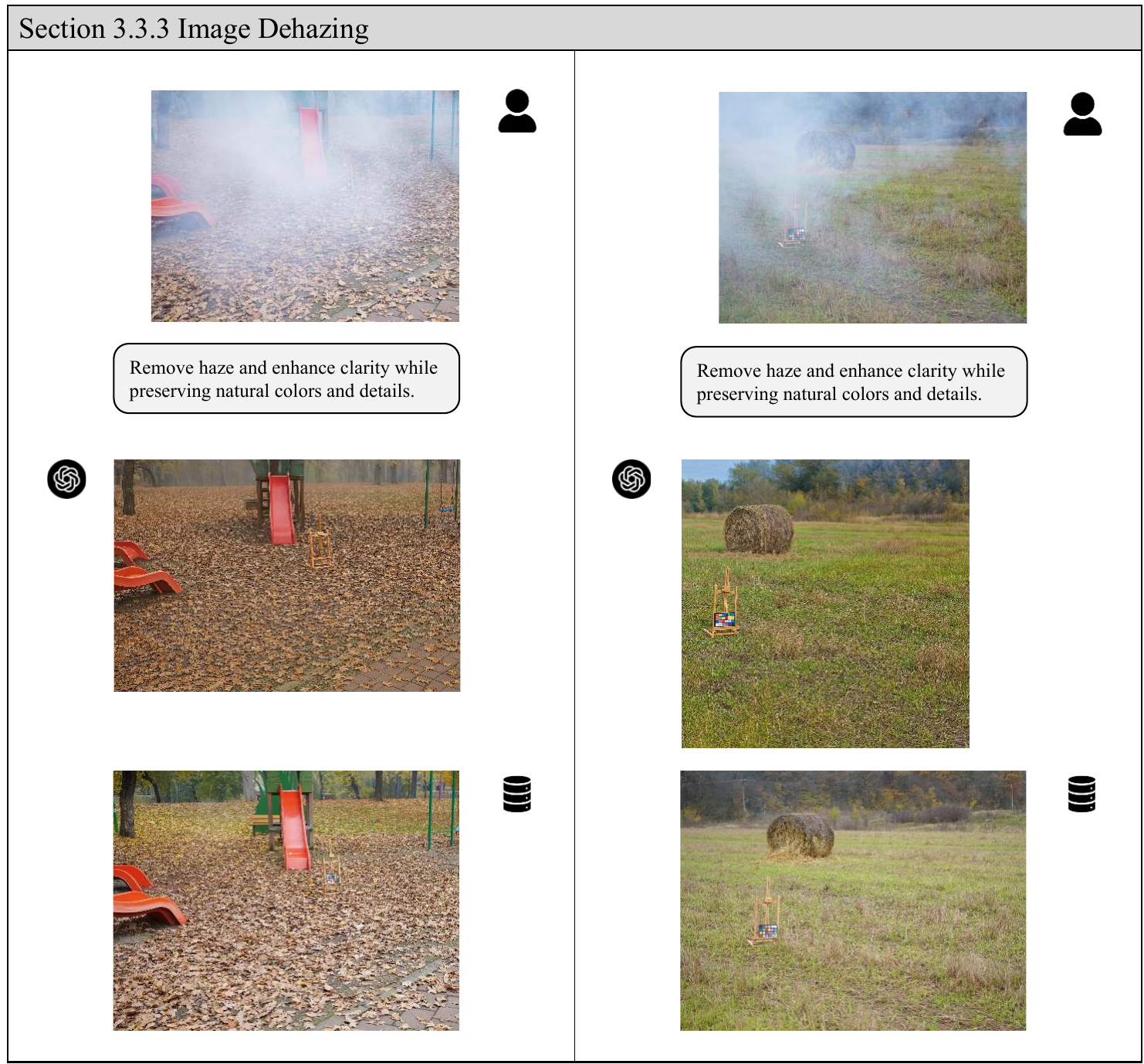}
    \caption[Sec~\ref{sec:image_restoration}: Image Dehazing]{Examples of image dehazing results generated by \modelname. The model effectively removes haze and restores scene clarity while preserving natural colors and structural details.}
    \label{fig:dehazing}
\end{figure}

\begin{figure}[h]
    \centering
    \includegraphics[width=1.0\linewidth]{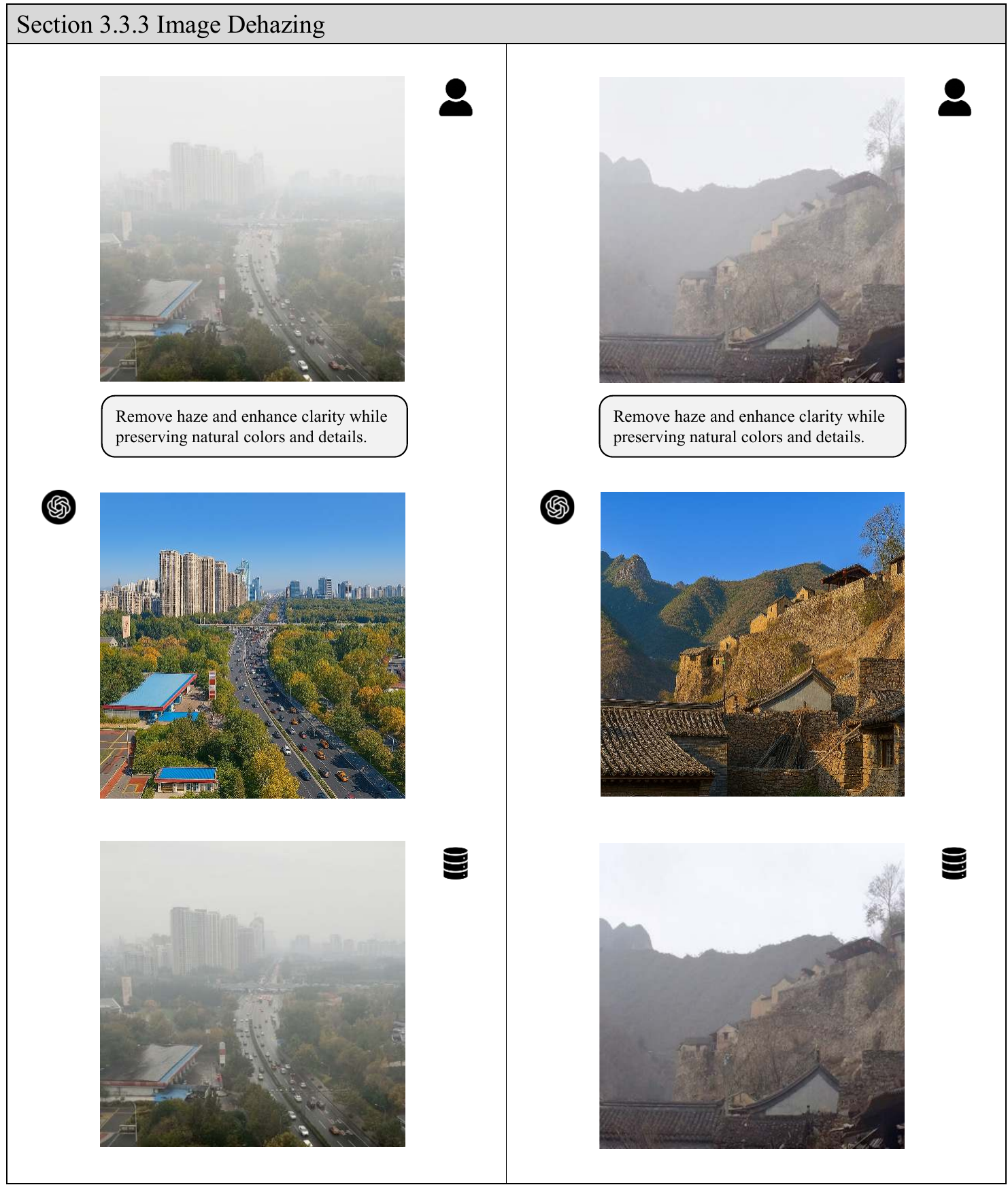}
    \caption[Sec~\ref{sec:image_restoration}: Image Dehazing]{Additional examples of image dehazing results generated by \modelname. The model not only removes haze and restores scene clarity but also tends to enhance the overall color saturation and vividness of the image. This indicates that \modelname performs dehazing in a perceptual manner, focusing on both visibility improvement and aesthetic enhancement.}
    \label{fig:dehazing1}
\end{figure}

As shown in Fig.\ref{fig:dehazing} and Fig.\ref{fig:dehazing1}, \modelname demonstrates strong performance in the image dehazing task. The model can effectively remove haze from various scenes, recovering clear and sharp visual content. In addition to haze removal, we observe that \modelname tends to enhance the image's color saturation and vividness. For example, the sky becomes bluer, and the vegetation appears greener after dehazing, making the scene more visually appealing.

Such results indicate that \modelname performs dehazing in a perceptual and generative manner. While the visibility improvement is significant, the model also enriches the visual effects beyond the original content, reflecting its tendency to optimize for human aesthetic preferences.

Overall, \modelname achieves a good balance between haze removal, detail preservation, and visual enhancement, making it suitable for real-world outdoor image enhancement applications.

\textbf{Image Deraining} Image deraining is a crucial low-level vision task that aims to remove rain-related degradations from images while preserving scene content and enhancing visibility. In real-world scenarios, rain artifacts can be categorized into two types: \textit{rain streaks}\cite{yang2017deep,zhang2019image,qin2019nasnet} that appear in the air, and \textit{raindrops}\cite{qian2018attentive,kwon2023raindrop,quan2021removing} that attach to the camera lens or window surfaces. Both types significantly affect the quality of images and hinder the performance of downstream vision tasks.

\begin{figure}[h]
    \centering
    \includegraphics[width=1.0\linewidth]{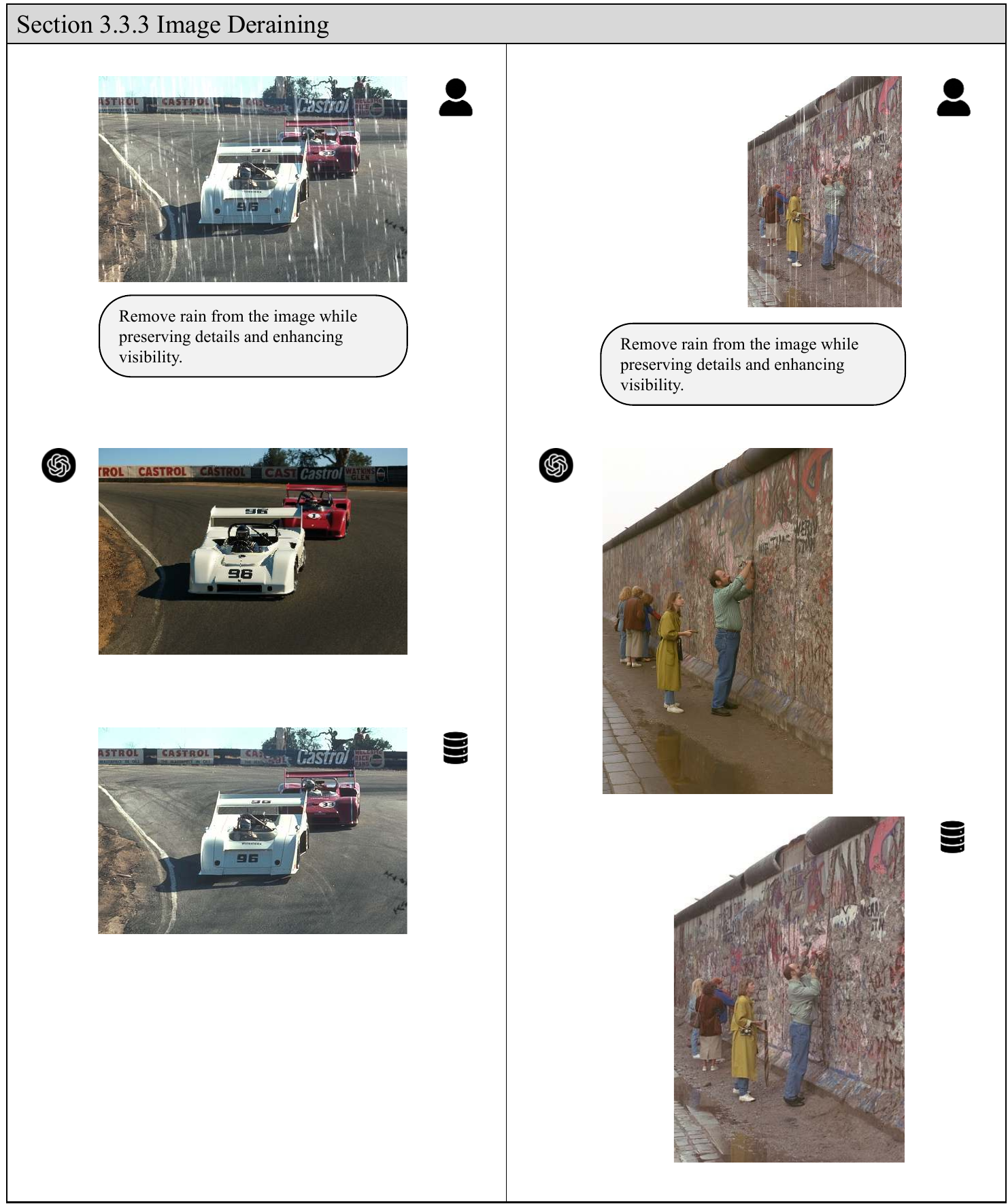}
    \caption[Sec~\ref{sec:image_restoration}: Image Deraining]{Examples of image deraining results generated by \modelname. The model can effectively remove rain streaks and improve overall visibility. However, it often leads to the loss of fine-grained details, such as road textures (left) and wall paintings (right), indicating a trade-off between rain removal and content preservation.}
    \label{fig:deraining}
\end{figure}

\begin{figure}[h]
    \centering
    \includegraphics[width=1.0\linewidth]{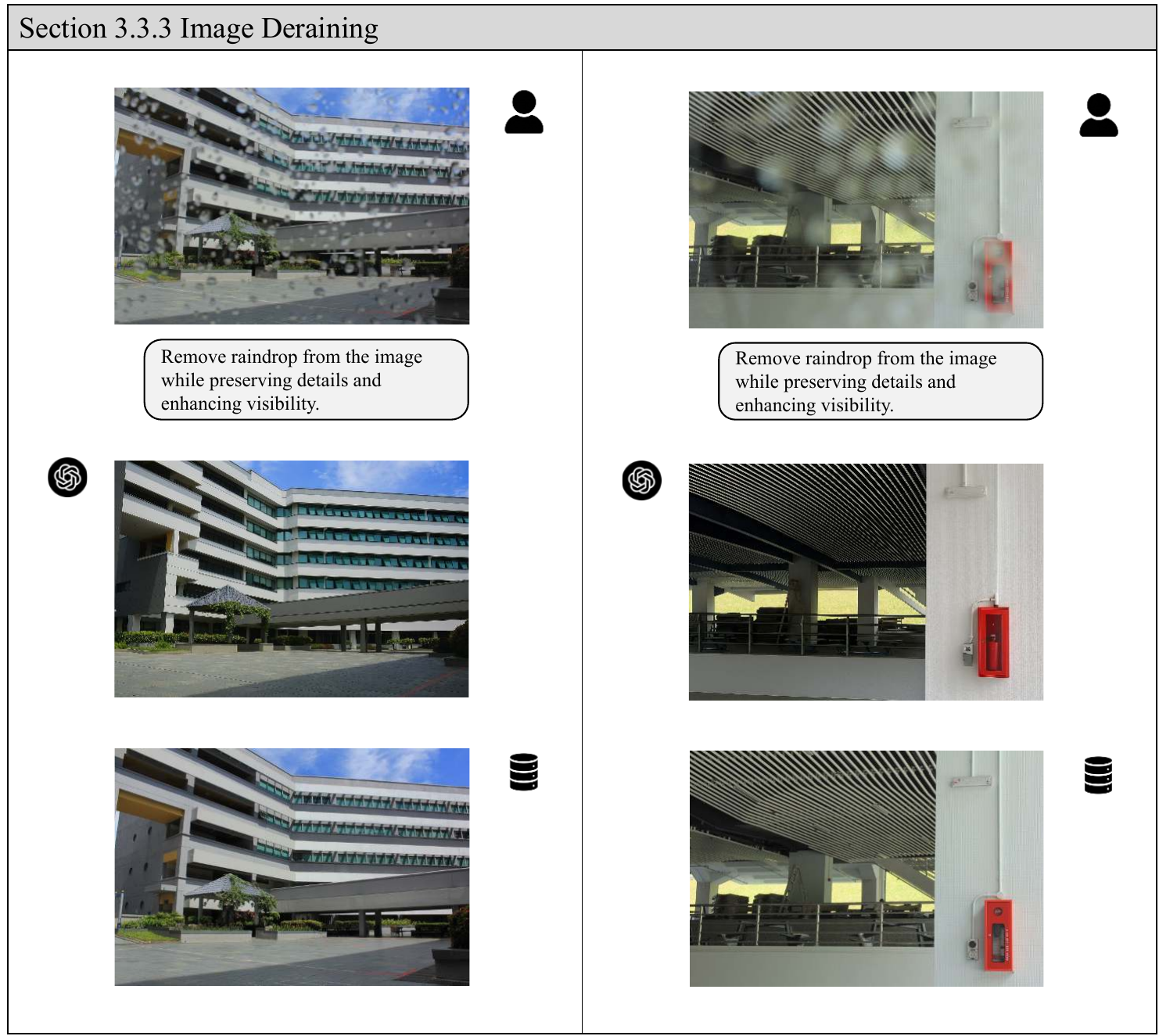}
    \caption[Sec~\ref{sec:image_restoration}: Image Deraining]{Examples of raindrop removal results generated by \modelname. The model effectively removes raindrops from the image surface and restores the occluded regions. However, some subtle texture inconsistency or content hallucination may occur in the recovered areas. Overall, the visibility and clarity of the scene are significantly improved.}
    \label{fig:raindrop}
\end{figure}

As shown in Fig.\ref{fig:deraining} and Fig.\ref{fig:raindrop}, \modelname demonstrates promising capabilities in handling both rain streak removal and raindrop removal tasks. For rain streak removal, the model can effectively eliminate rain patterns from various scenes, including dynamic driving environments and urban streets. The visibility and clarity of the derained results are greatly improved.
For raindrop removal, \modelname is able to restore the occluded content behind the raindrops and recover the overall structure of the scene. The generated images exhibit clean surfaces and improved visual quality.

However, we also observe some limitations in both tasks. For rain streak removal, the model tends to over-smooth the textures of the original scene. For instance, in the driving scene, road textures and tire marks are partially lost. In the wall scene, graffiti details are blurred after deraining. In the case of raindrop removal, although the global structure is restored, there are occasional inconsistencies or hallucinated content in the recovered areas, especially when the occluded regions are complex.

These observations suggest that while \modelname shows strong rain removal ability and achieves satisfactory perceptual quality, there is still room for improvement in preserving fine-grained details and ensuring content consistency in heavily degraded regions.

\textbf{Image Desnowing}
Image desnowing aims to remove snow particles or snow cover from images and restore the original clean scene while preserving fine details and color consistency\cite{chen2021all,cheng2023snow,jaw2020desnowgan}. This task is particularly useful in outdoor image enhancement and pre-processing for downstream vision applications in winter conditions.

\begin{figure}[h]
    \centering
    \includegraphics[width=1.0\linewidth]{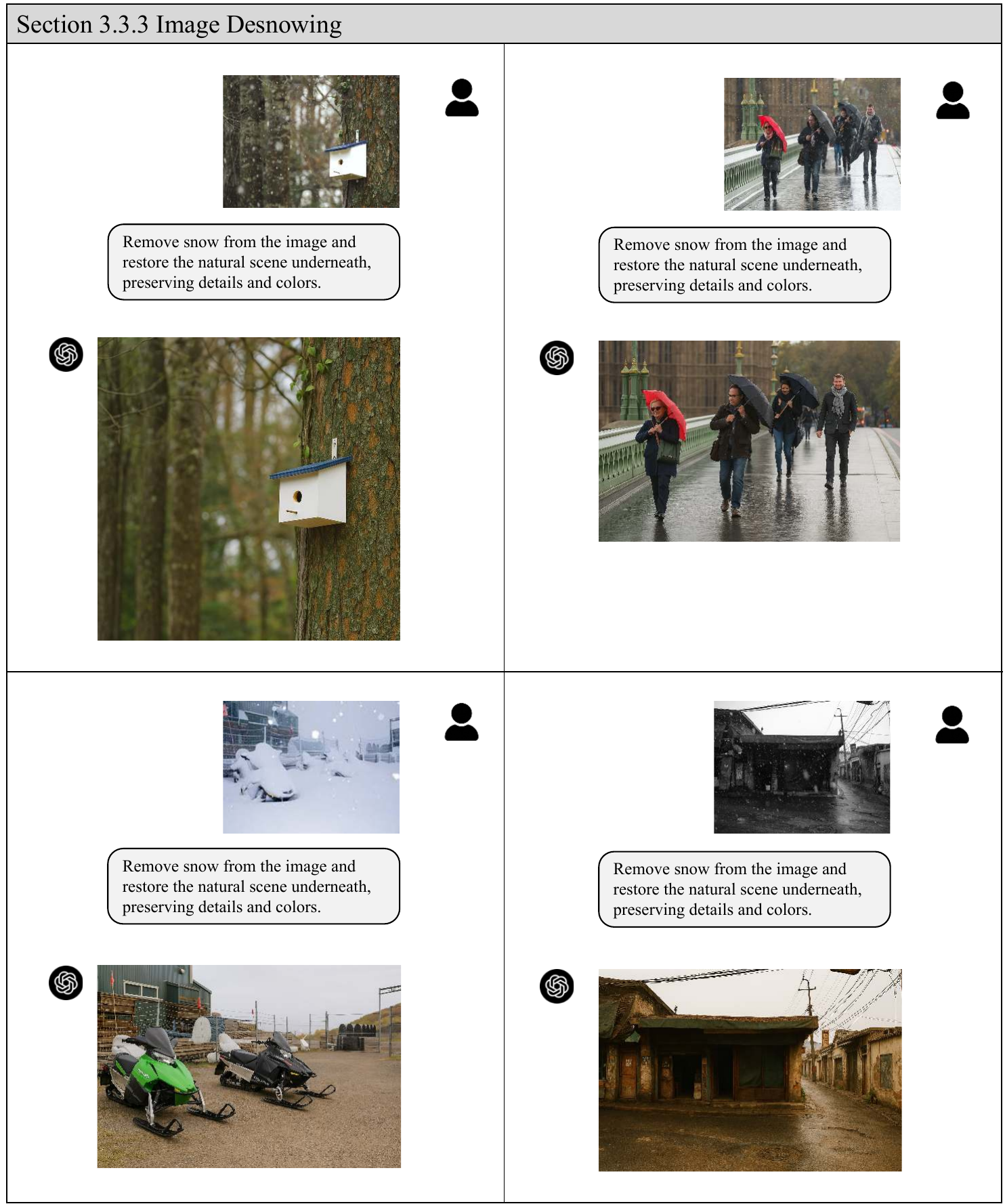}
    \caption[Sec~\ref{sec:image_restoration}: Image Desnowing]{Examples of image desnowing results generated by \modelname. The model effectively removes snow from various scenes and restores the underlying content with good visual consistency. The generated images preserve the natural colors and structural details of the original scenes, demonstrating strong desnowing capability.}
    \label{fig:desnowing}
\end{figure}

As shown in Fig.~\ref{fig:desnowing}, \modelname achieves impressive performance on the desnowing task across a variety of scenes, including forests, urban streets, and snow-covered objects. The model is capable of effectively removing both falling snow and accumulated snow, restoring the original appearance of the scene.

\textbf{Low-light Image Enhancement}
Low-light image enhancement aims to improve the visual quality of images captured under poor lighting conditions. The task focuses on increasing brightness and contrast while maintaining the original scene's colors and details, enabling better visibility for both human perception and downstream vision tasks.

\begin{figure}[h]
    \centering
    \includegraphics[width=1.0\linewidth]{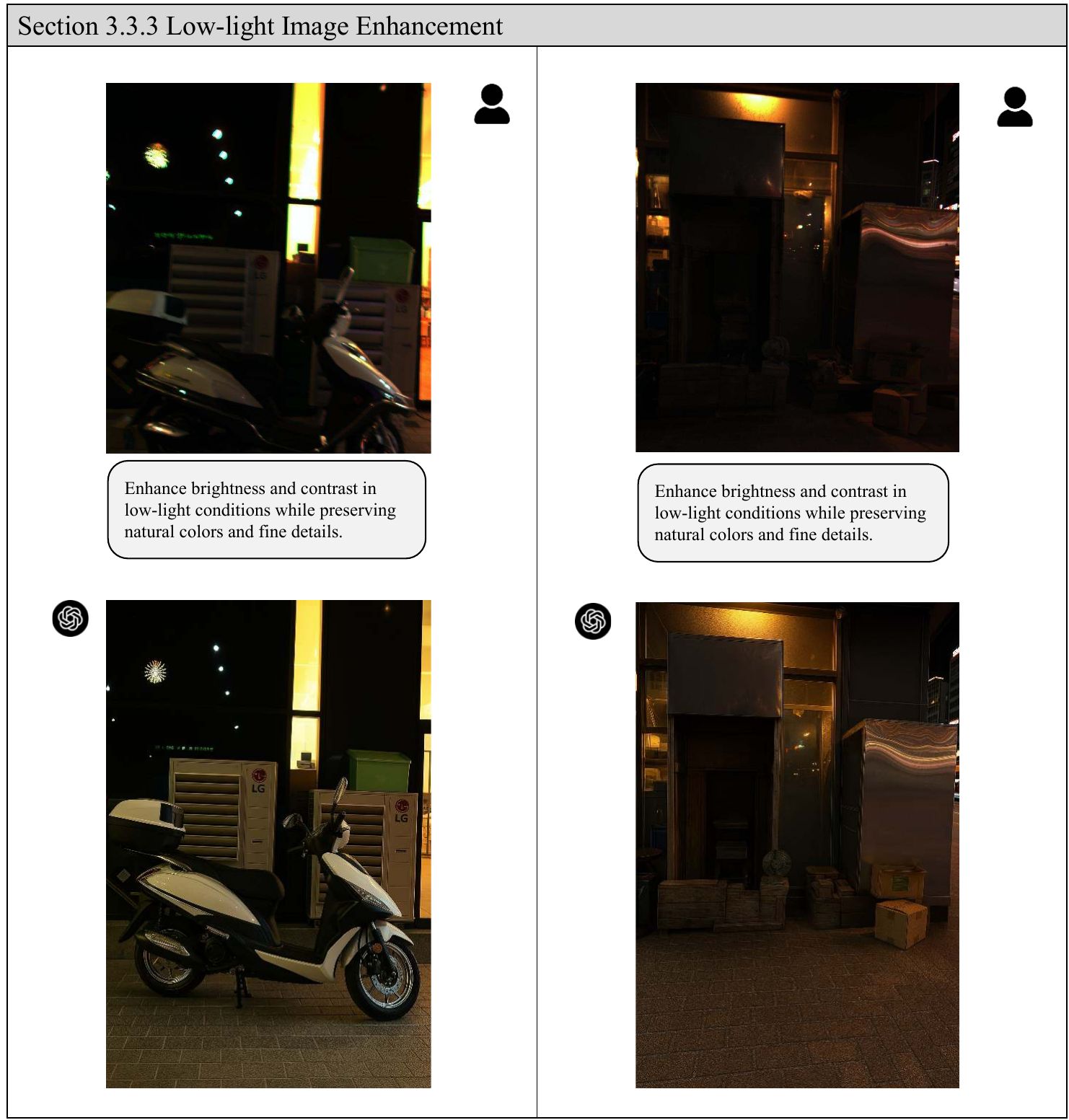}
    \caption[Sec~\ref{sec:image_restoration}: Low-light Image Enhancement]{Examples of low-light image enhancement results generated by \modelname. The model effectively improves brightness and contrast in dark scenes while preserving natural colors and structural details. The enhanced images exhibit better visibility and a more realistic appearance compared to the original low-light inputs.}
    \label{fig:lowlight}
\end{figure}

As shown in Fig.~\ref{fig:lowlight}, \modelname demonstrates promising performance in low-light image enhancement. The model can significantly brighten dark regions and restore scene visibility, making previously invisible content clearly observable. Meanwhile, the model preserves the natural appearance of the scene without introducing severe color distortions.

\textbf{Old Photo Restoration}
Old photo restoration aims to repair and enhance old, degraded, or damaged images, improving their clarity and visual quality while preserving the original style or vintage characteristics\cite{wan2022old}. This task often requires complex reasoning to remove artifacts, restore missing content, and even colorize black-and-white photos when desired.

\begin{figure}[h]
    \centering
    \includegraphics[width=1.0\linewidth]{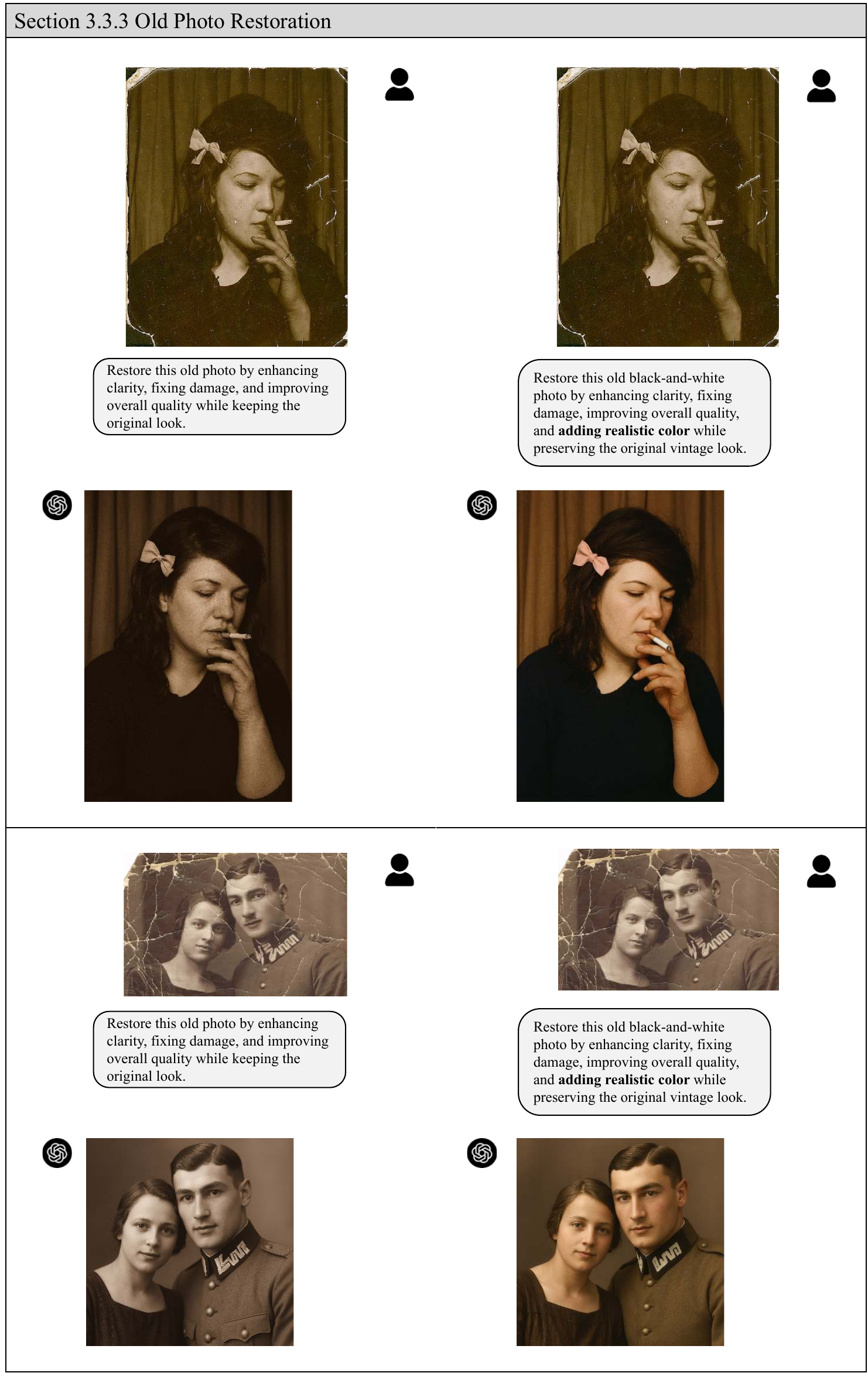}
    \caption[Sec~\ref{sec:image_restoration}: Old Photo Restoration]{Examples of old photo restoration results generated by \modelname. The model effectively enhances image clarity, repairs damaged regions, and restores overall quality. }
    \label{fig:oldphoto1}
\end{figure}

\begin{figure}[h]
    \centering
    \includegraphics[width=1.0\linewidth]{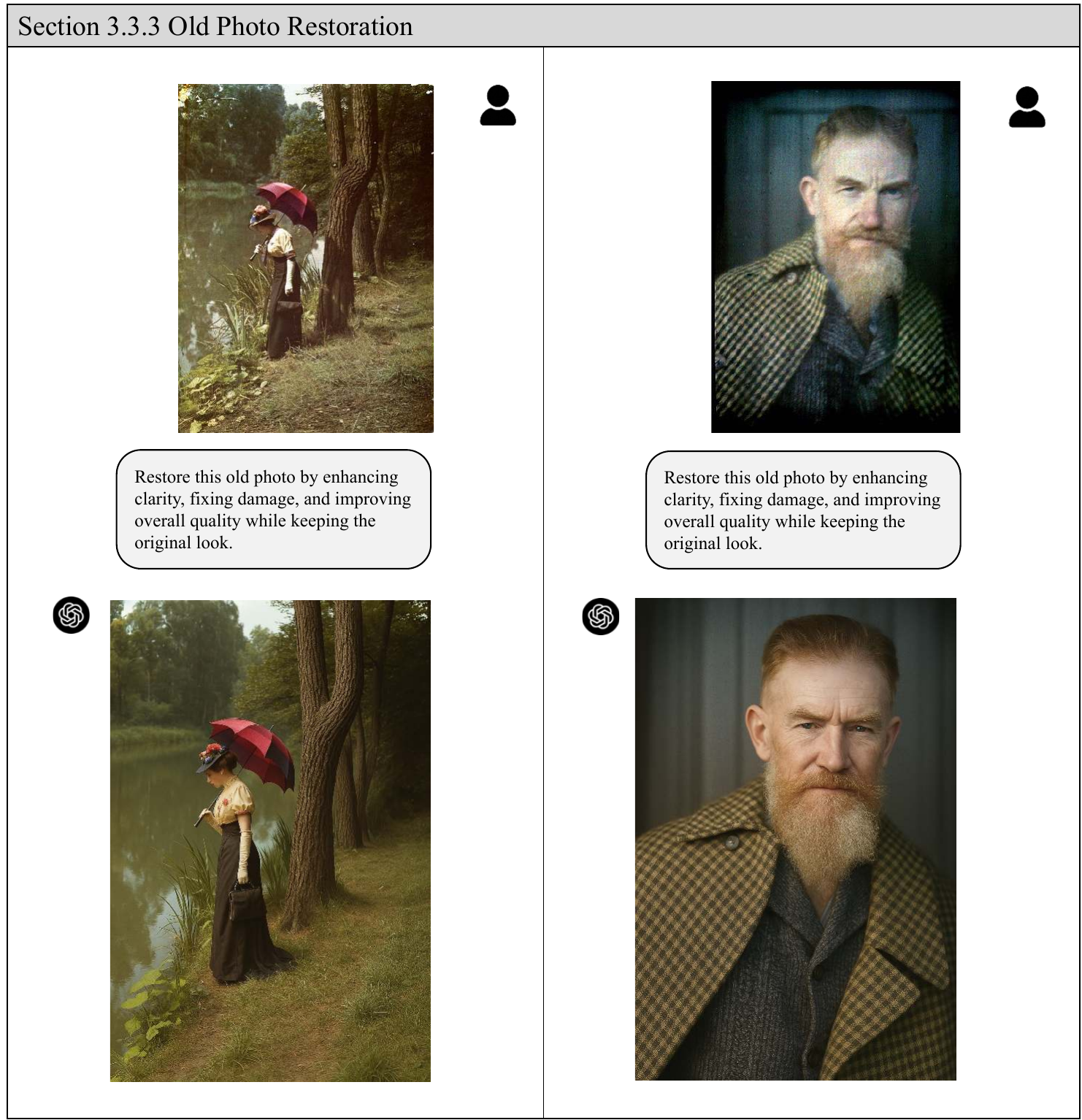}
    \caption[Sec~\ref{sec:image_restoration}: Old Photo Restoration]{Additional examples of old photo restoration results generated by \modelname.}
    \label{fig:oldphoto2}
\end{figure}

As shown in Fig.\ref{fig:oldphoto1} and Fig.\ref{fig:oldphoto2}, \modelname demonstrates remarkable capabilities in old photo restoration. The model not only improves sharpness and removes artifacts but also exhibits controllable generation ability. Specifically, given different prompts, the model can either maintain the original black-and-white or vintage style or add realistic colors to the restored image.

\clearpage

\clearpage

\clearpage
\subsubsection{Shadow Removal}
\label{sec:shadow}
Shadow removal aims to eliminate shadows from images while preserving the original content and structure of the scene\cite{wang2018stacked,qu2017deshadownet,le2021physics}. This task is essential for improving image aesthetics and supporting downstream vision tasks such as segmentation and detection.

\begin{figure}[h]
    \centering
    \includegraphics[width=1.0\linewidth]{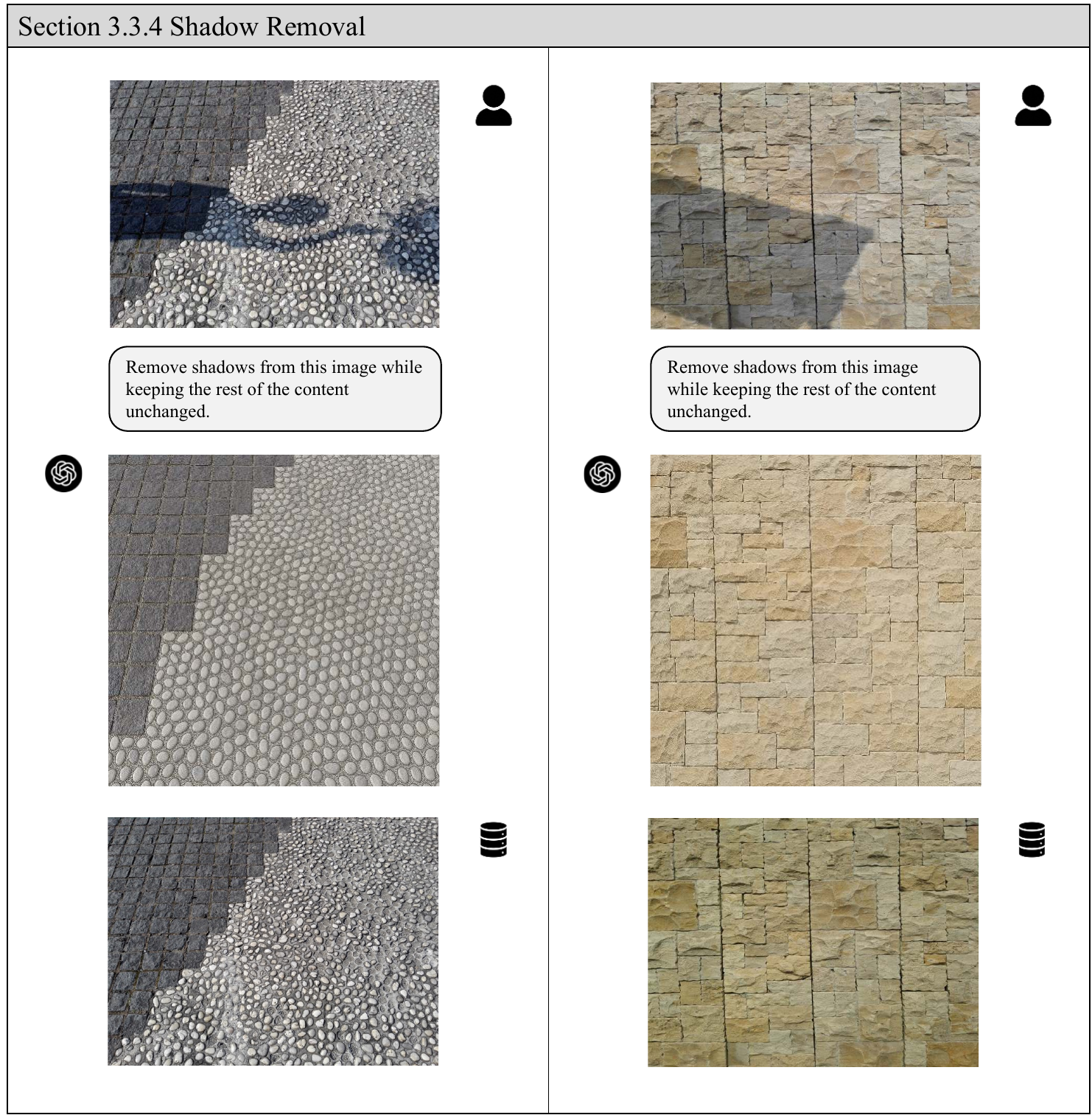}
    \caption[Sec~\ref{sec:shadow}: Shasow Removal]{Examples of shadow removal results generated by \modelname. The model effectively removes shadows from the images and restores the overall appearance of the scene. However, slight detail loss can be observed in the restored regions, such as the shape and arrangement of pebbles in the left example.}
    \label{fig:shadow}
\end{figure}

As shown in Fig.~\ref{fig:shadow}, \modelname demonstrates a strong ability to remove shadows from different textures and scenes. The model can effectively restore the occluded regions and generate consistent and visually pleasing results.
However, we observe that shadow removal may lead to slight detail loss in the restored areas. For example, in the left case, the arrangement and shape of the pebbles deviate slightly from the original image after shadow removal. This indicates that while \modelname focuses on generating perceptually consistent outputs, fine-grained structure preservation remains a challenge.

\clearpage
\clearpage

\subsubsection{Reflection Removal}
\label{sec:reflect}
Reflection removal aims to eliminate unwanted reflections from glass or smooth surfaces while preserving the original scene content, texture, and lighting\cite{li2013exploiting,amanlou2022single}. This task is particularly challenging when reflection and transmission layers are entangled or semantically similar.

\begin{figure}[h]
    \centering
    \includegraphics[width=1.0\linewidth]{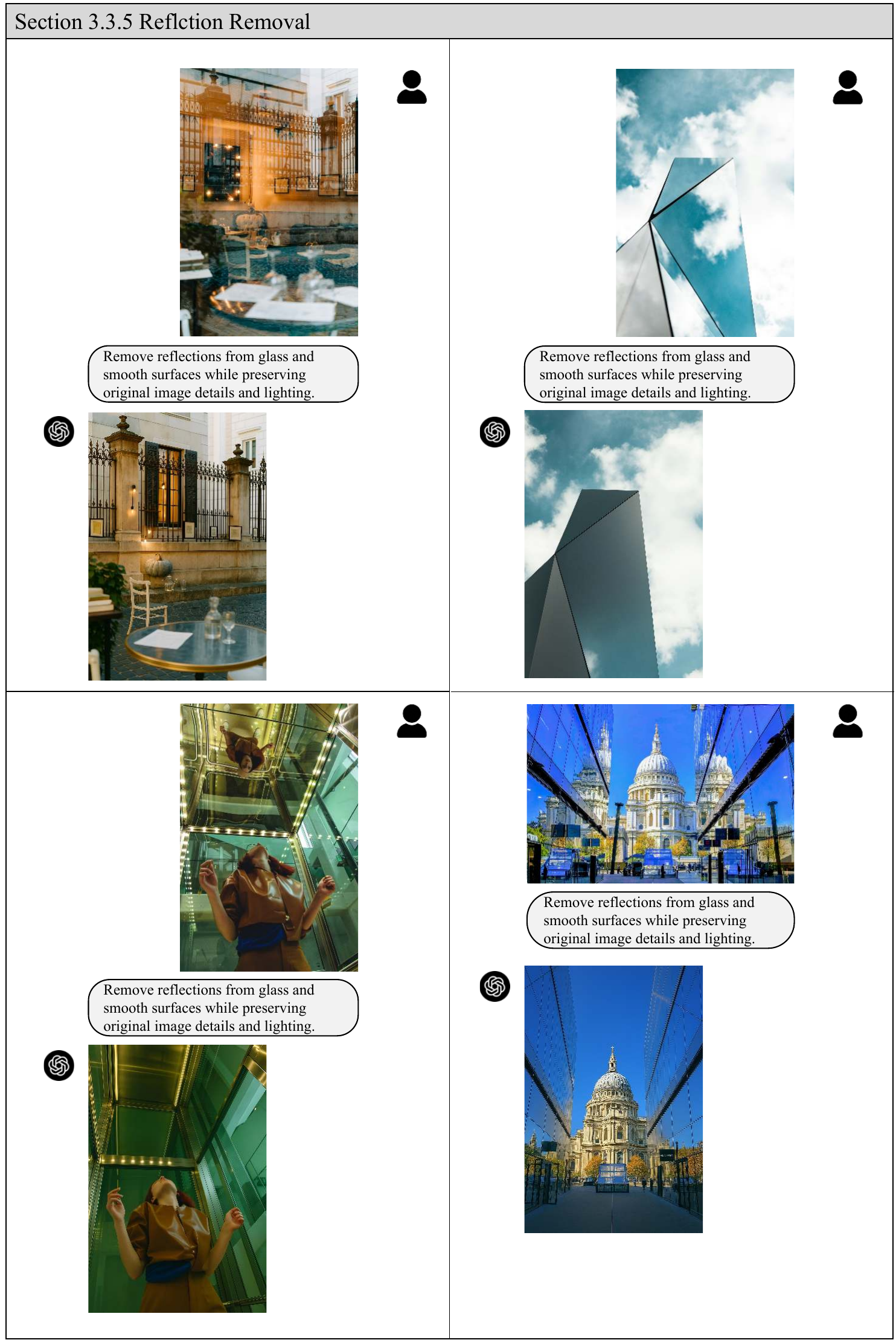}
    \caption[Sec~\ref{sec:reflect}: Reflection Removal]{Examples of reflection removal results generated by \modelname.}
    \label{fig:reflect}
\end{figure}

As shown in Fig.~\ref{fig:reflect}, \modelname demonstrates the ability to remove prominent reflections, producing visually cleaner results. However, the model also introduces significant content alterations beyond the reflected regions. In the upper-left example, several objects on the table are removed or modified, indicating a loss of fine-grained detail. In the upper-right case, the structure of the background building has been changed, with architectural elements inconsistently reconstructed. Similarly, in the lower-left image, the eyeglasses worn by the person are completely removed, despite being part of the original non-reflective content.

These results suggest that \modelname treats reflection removal as a global editing task rather than a localized layer separation process. While it succeeds in suppressing reflections, it lacks the fine discrimination needed to preserve all true scene content. Future improvements may require explicit reflection segmentation or layer decomposition to better disentangle reflection and background information.

\clearpage
\clearpage

\subsubsection{Image Relighting}
\label{sec:relight}
Image relighting refers to the process of modifying the lighting conditions of a given image while maintaining the original content and geometry\cite{zhu2022designing,zhou2019deep}. It evaluates the model’s ability to understand light direction, intensity, and color temperature, and apply photorealistic transformations guided by either textual instructions or visual references.

\begin{figure}[h]
    \centering
    \includegraphics[width=1.0\linewidth]{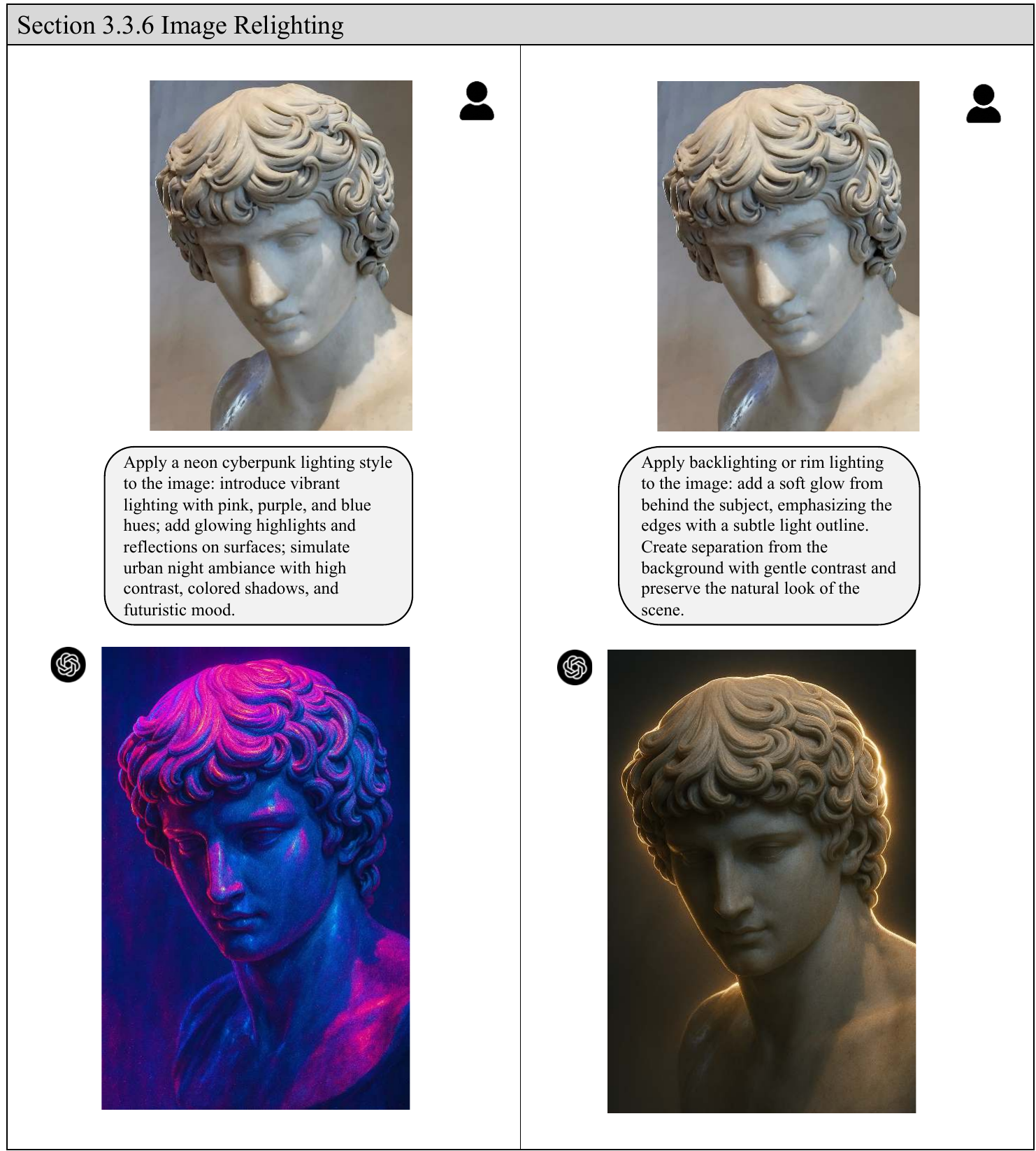}
    \caption[Sec~\ref{sec:relight}: Image Relighting with Textual Prompts]{Examples of image relighting with textual prompts generated by \modelname.}
    \label{fig:relight-text}
\end{figure}

\textbf{Text-based relighting.}  
As shown in Fig.~\ref{fig:relight-text}, \modelname effectively interprets text prompts describing various lighting styles, such as "neon cyberpunk" or "rim lighting." The model successfully applies vibrant lighting with colored shadows and glowing highlights in cyberpunk settings, and soft backlighting in natural rim light scenarios. It demonstrates precise semantic understanding and strong control over global illumination effects.

\begin{figure}[h]
    \centering
    \includegraphics[width=1.0\linewidth]{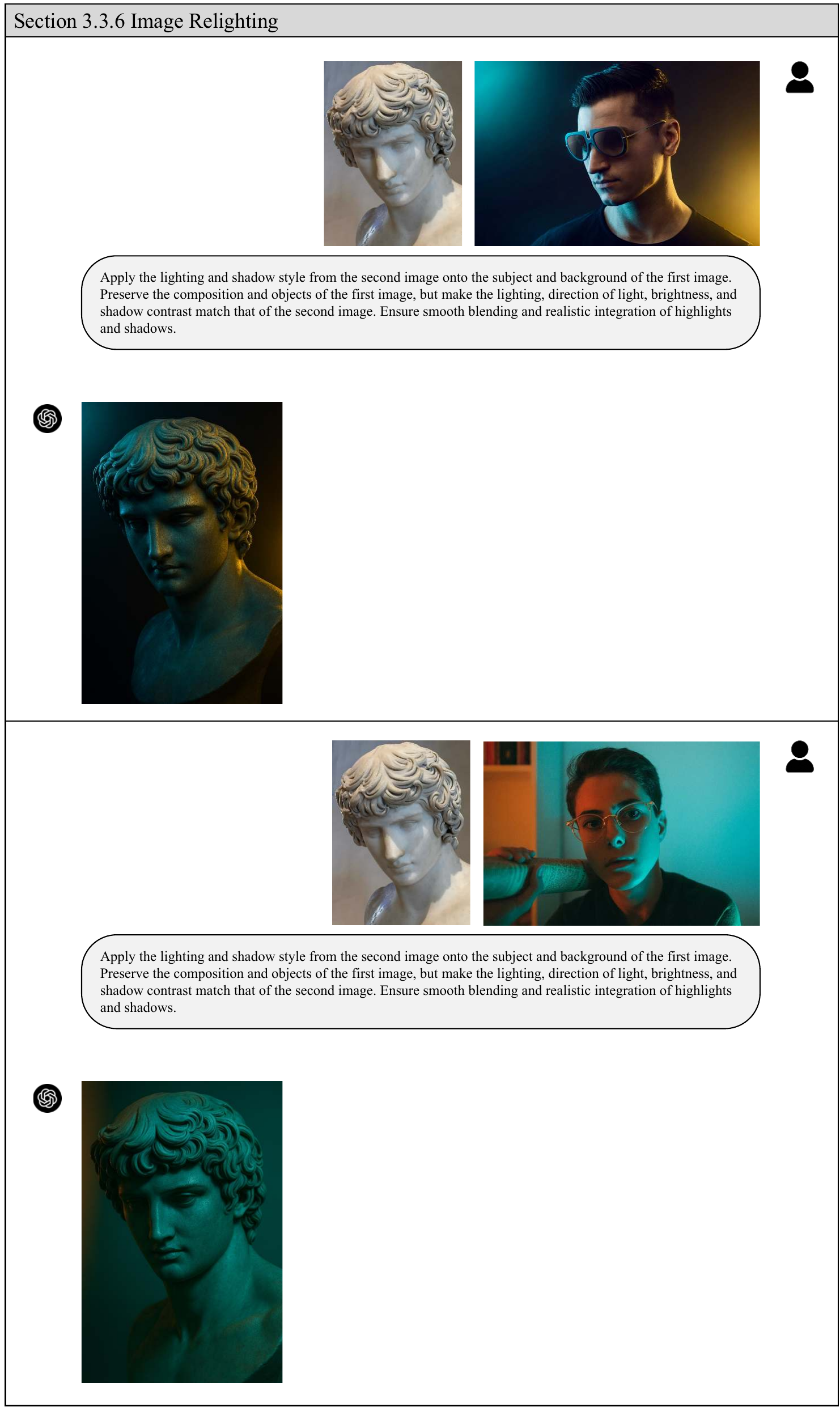}
    \caption[Sec~\ref{sec:relight}: Image Relighting with Referenced Images]{Examples of image relighting with reference images generated by \modelname.}
    \label{fig:relight-image}
\end{figure}

\textbf{Reference-based relighting.}  
In Fig.~\ref{fig:relight-image}, the model is conditioned on a visual reference that provides lighting cues such as direction, intensity, and mood. The results show that \modelname is able to mimic the lighting setup from the reference image, adjusting contrast, highlight positioning, and overall ambiance accordingly. The subject and background are consistently lit in a visually coherent manner.

Overall, \modelname performs well in both text- and reference-guided relighting. The lighting is adjusted naturally without distorting the subject geometry or textures, indicating a strong capability for controllable and photorealistic light manipulation.

\clearpage

\clearpage

\subsubsection{Underwater Image Enhancement}
\label{sec:underwater}
Underwater image enhancement aims to improve the visual quality of images captured in underwater environments\cite{li2019underwater,raveendran2021underwater,zhang2023underwater}, which often suffer from color distortion, low contrast, and poor visibility due to light absorption and scattering in water. The task focuses on restoring natural color balance, enhancing image clarity, and increasing scene visibility.

\begin{figure}[h]
    \centering
    \includegraphics[width=1.0\linewidth]{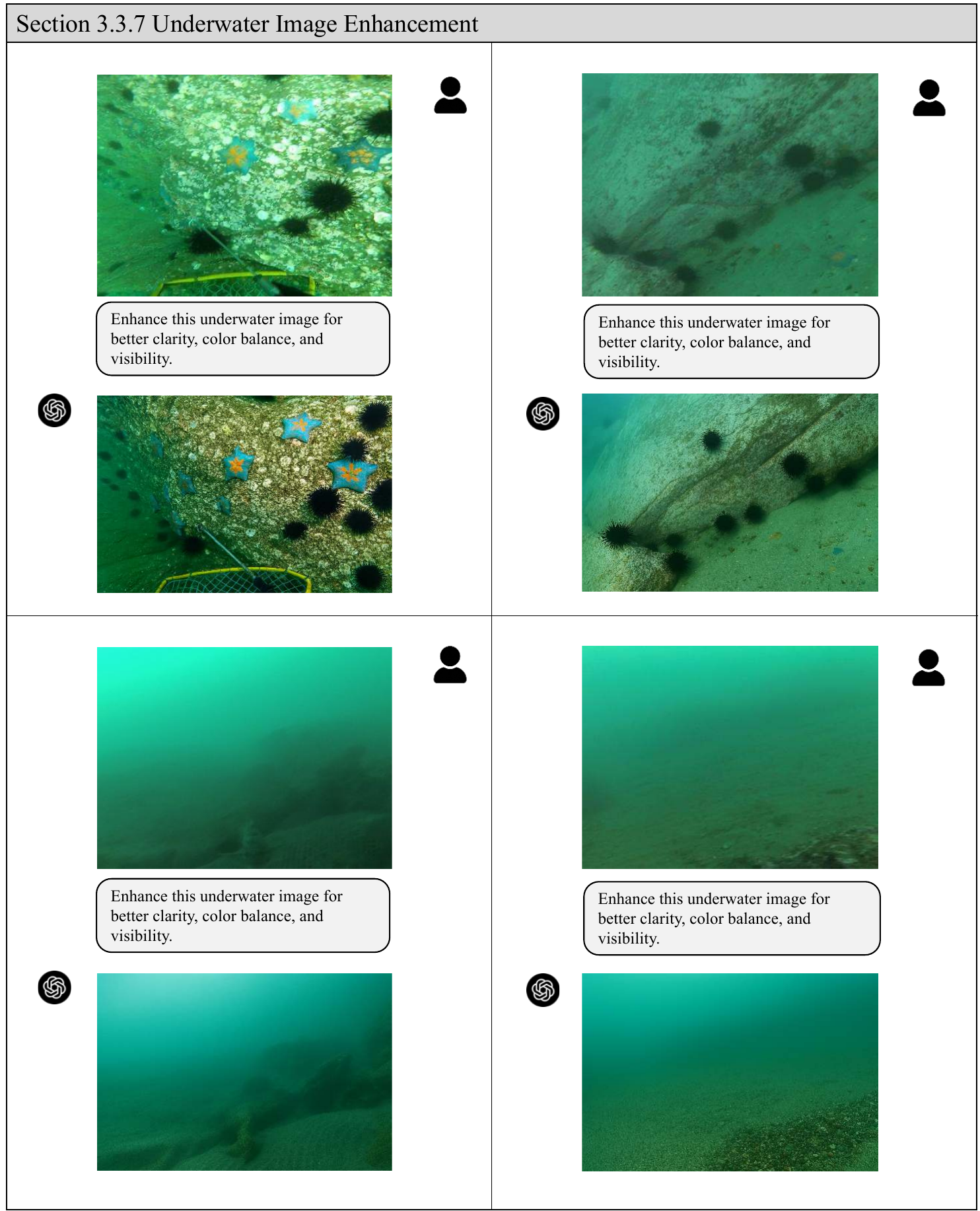}
    \caption[Sec~\ref{sec:underwater}: Underwater Image Enhancement]{Examples of underwater image enhancement results generated by \modelname. The model shows the ability to enhance underwater images by improving clarity, color balance, and visibility. However, due to the challenging nature of underwater environments, the enhancement results vary across different scenes. Some examples demonstrate promising improvements, while others still suffer from limited enhancement effects.}
    \label{fig:underwater}
\end{figure}

As shown in Fig.~\ref{fig:underwater}, \modelname demonstrates a certain capability in underwater image enhancement. In some relatively clear scenes, the model is able to effectively enhance details, correct color bias, and significantly improve the overall visibility of the image. However, underwater environments present a more challenging scenario compared to other enhancement tasks, especially when the input image suffers from severe degradation.

We observe that the model's enhancement performance varies depending on the input image quality and scene complexity. While some examples achieve promising results with clearer textures and better color restoration, other cases remain limited in improvement, showing only slight changes after enhancement.
These results indicate that \modelname has the potential to handle underwater image enhancement but still faces challenges in generalizing to highly degraded or extremely blurry underwater scenes.

\clearpage
\clearpage

\subsubsection{Low-level Data Synthesis}
\label{sec:lowdata}
Low-level data synthesis aims to generate degraded images by applying specific visual artifacts or environmental conditions to clean images\cite{}. This task plays an important role in evaluating a model's fine-grained control ability in image generation and its potential application in data augmentation for low-level vision tasks, such as image restoration, enhancement, and robustness testing.

\begin{figure}[h]
    \centering
    \includegraphics[width=1.0\linewidth]{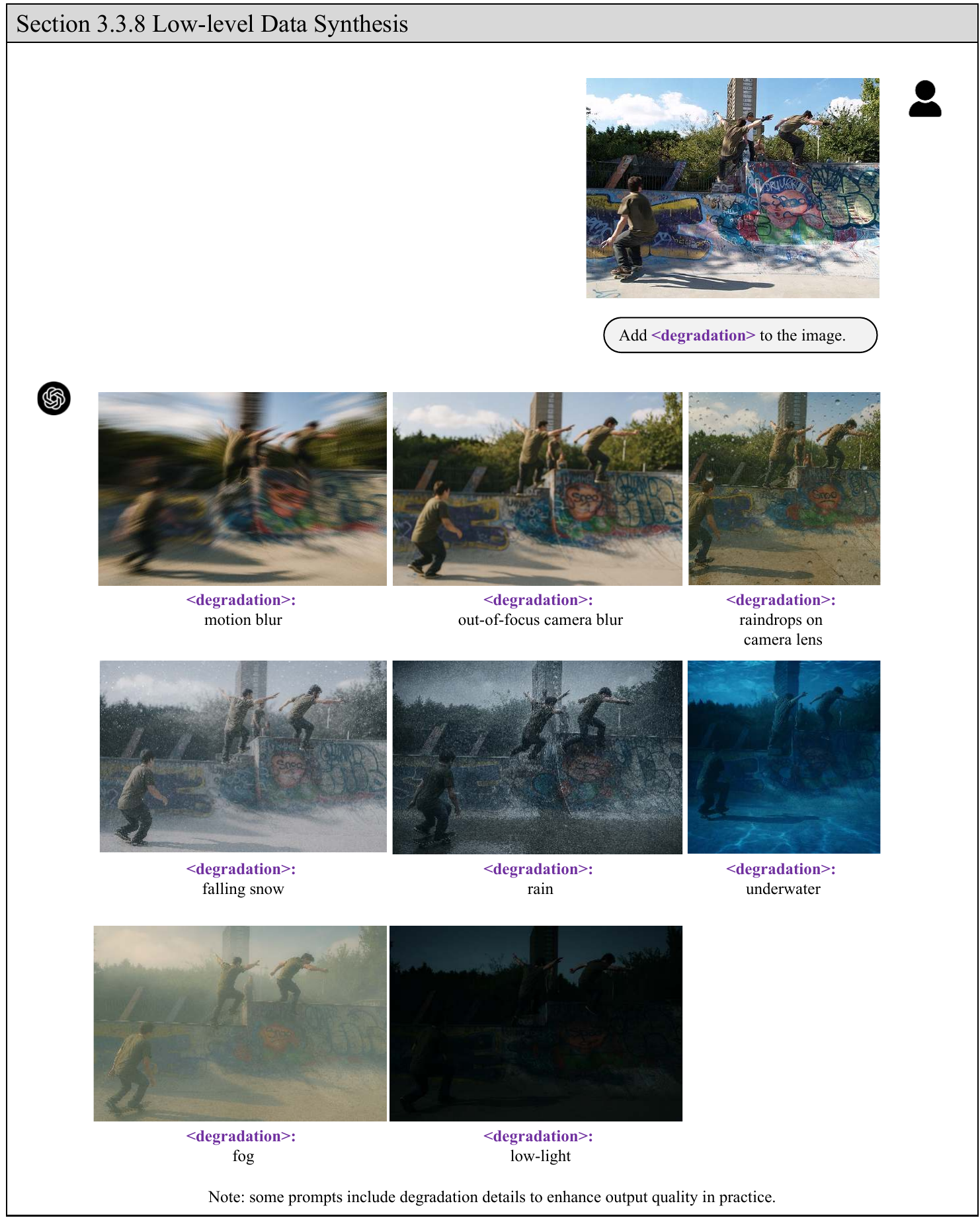}
    \caption[Sec~\ref{sec:lowdata}: Low-level Data Synthesis]{Examples of low-level data synthesis results generated by \modelname for a skateboard scene. The model applies various degradation types to a clean image, including motion blur, camera blur, raindrops, snow, rain, underwater, fog, and low-light.}
    \label{fig:lowlevelsyn}
\end{figure}

\begin{figure}[h]
    \centering
    \includegraphics[width=1.0\linewidth]{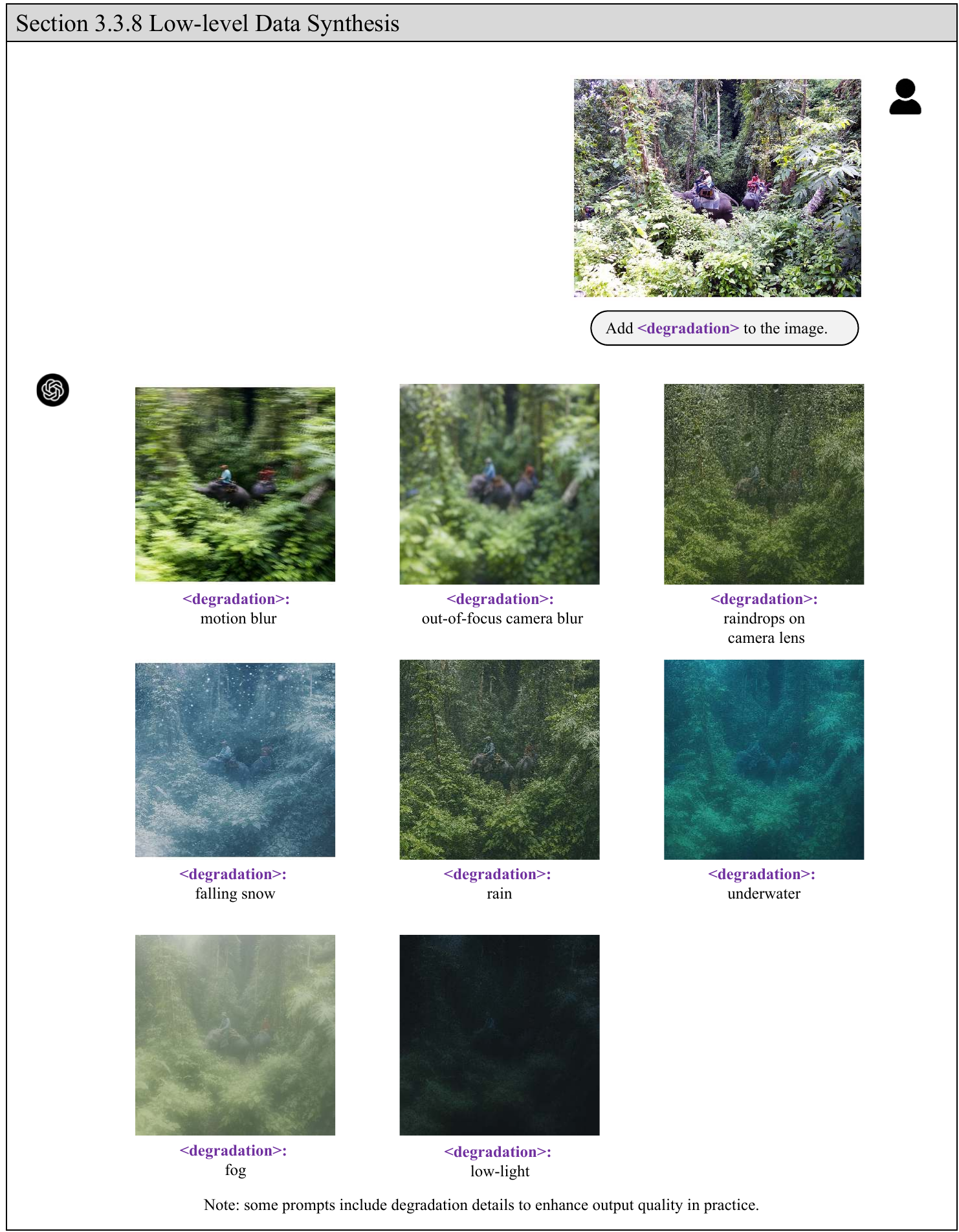}
    \caption[Sec~\ref{sec:lowdata}: Low-level Data Synthesis]{Examples of low-level data synthesis results generated by \modelname for a forest scene. The model successfully simulates diverse degradation effects while maintaining the consistency of scene structure.}
    \label{fig:lowlevelsyn2}
\end{figure}

As shown in Fig.\ref{fig:lowlevelsyn} and Fig.\ref{fig:lowlevelsyn2}, we evaluate \modelname's capability in synthesizing various degradation types, including motion blur, out-of-focus blur, raindrops on camera lens, snow, rain, underwater effect, fog, and low-light conditions. The model generates images with these degradation effects by following explicit text instructions.

Overall, \modelname demonstrates strong generation capability in this task. The synthesized degradations are visually consistent with human perception and exhibit diverse patterns across different scenes. However, the degradation details are sometimes exaggerated or stylized, rather than strictly following real-world physical rules. This observation indicates that the current model generation is more perceptual-driven than physically accurate.

\clearpage

\clearpage

\clearpage
\section{Discriminative Image Generation}
\label{sec:discrimative}

\subsection{Detection}
\label{sec:det}

\subsubsection{General Object Detection}
\label{sec:obj}
General object detection is a fundamental and widely studied task in computer vision\cite{lin2014microsoft,pascal-voc-2012,pascal-voc-2007,carion2020end,ren2015faster,girshick2015fast,girshick2014rich,liu2016ssd}, aiming to identify and localize objects of interest within natural images. Specifically, the task requires detecting various categories of objects and providing their corresponding bounding boxes.
This task serves as an important benchmark to evaluate the model's basic visual understanding and spatial localization ability. Unlike traditional object detection models that are trained with large-scale annotated datasets, \modelname performs detection in a zero-shot manner, relying on its general visual reasoning capabilities without specific detection head tuning.

In this work, we test \modelname in the general object detection scenario, covering typical scenes such as animals, daily objects, and natural environments. The goal is to explore whether the model can accurately recognize object instances, understand object boundaries, and output structured detection results in both visual and textual modalities. The evaluation results also help reveal the current limitations of multimodal large models in structured perception tasks.

\textbf{Visual outputs v.s. Textual outputs.}
Since \modelname supports both visual and textual outputs, we conduct a comparative analysis of these two modalities in the general object detection task. For textual outputs, we use Python to visualize the detection results based on the returned bounding box coordinates.

\begin{figure}[h]
    \centering
    \includegraphics[width=1.0\linewidth]{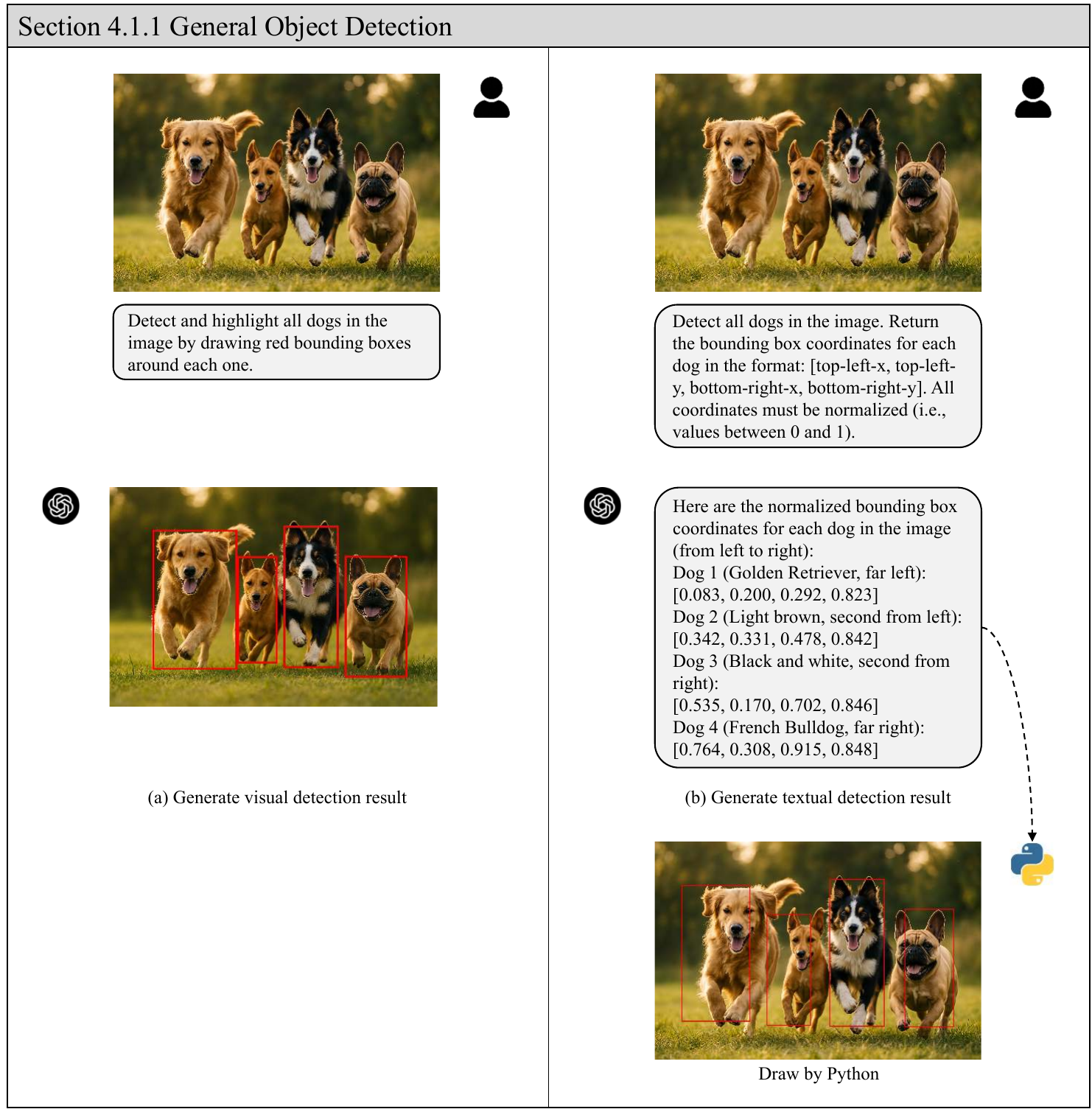}
    \caption[Sec~\ref{sec:obj}: Visual and textual detection results]{Examples of general object detection results generated by \modelname for detecting dogs. The model supports both visual outputs and textual outputs, which can be visualized by external tools. In this example, the visual output shows better accuracy and tighter bounding boxes compared to the textual output.}
    \label{fig:det-multimodal1}
\end{figure}

\begin{figure}[h]
    \centering
    \includegraphics[width=1.0\linewidth]{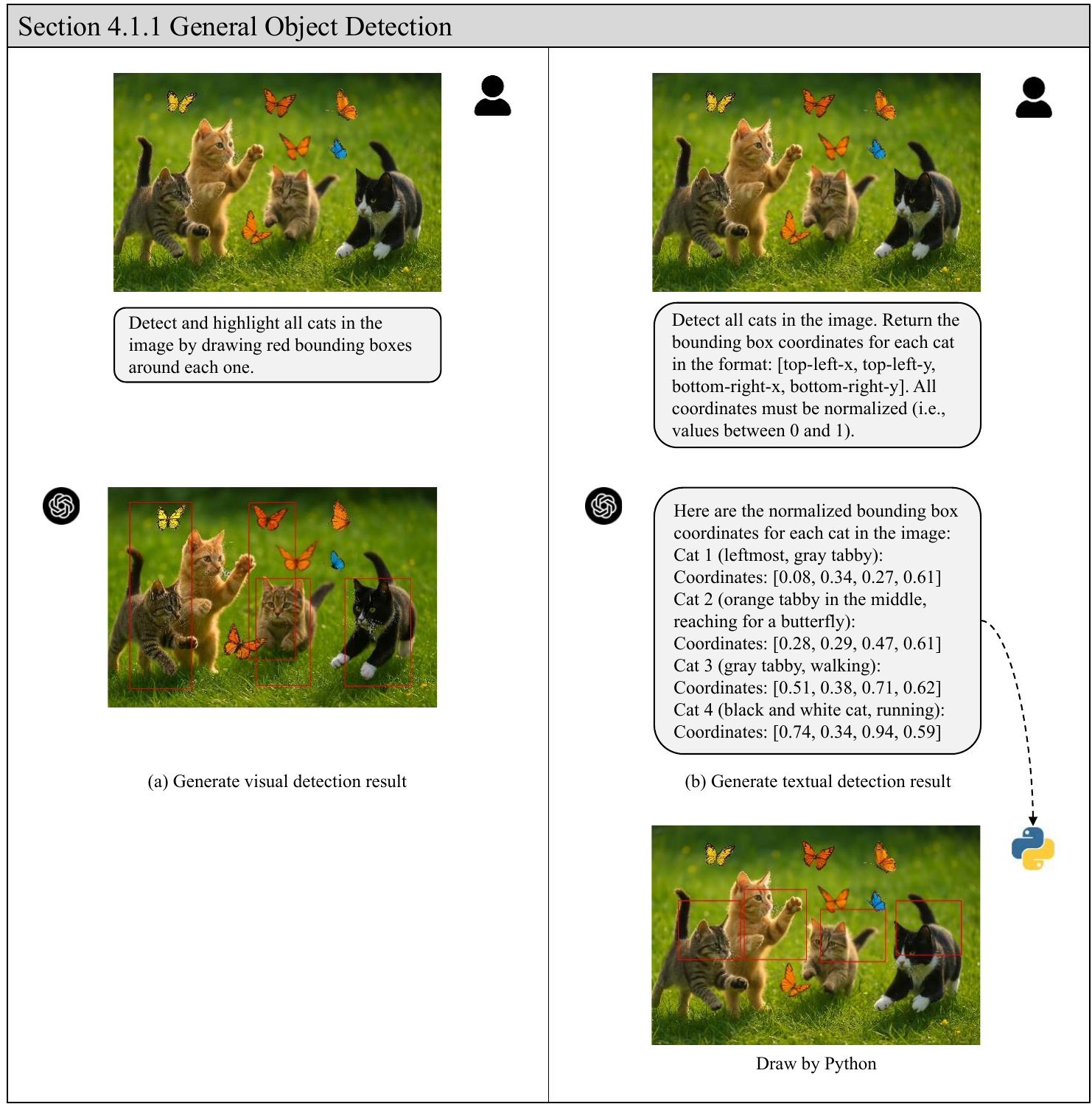}
    \caption[Sec~\ref{sec:obj}: Visual and textual detection results]{Examples of general object detection results generated by \modelname for detecting cats. In this case, both visual and textual outputs show limited accuracy, and the detection results are not ideal. However, the visual output still performs slightly better, especially in locating the two cats on the right.}
    \label{fig:det-multimodal2}
\end{figure}

As shown in Fig.~\ref{fig:det-multimodal1}, the visual output achieves better performance than the textual output. The generated bounding boxes in the visual output are more accurate, with tighter boundaries that better align with the objects. However, as we discussed earlier, the object detection accuracy of \modelname is not stable.
In Fig.~\ref{fig:det-multimodal2}, neither modality produces satisfactory detection results. Both visual and textual outputs show obvious inaccuracies, and several objects are not well localized. Nevertheless, the visual output still presents a slight advantage over the textual output, especially in accurately detecting the two cats on the right side of the image.

These results indicate that while \modelname possesses certain object detection capabilities, it still struggles with precise localization, and its detection accuracy varies across different scenarios.

\clearpage

\clearpage
\subsubsection{Object Detection within Satellite Imagery}
\label{sec:satellite}

Object detection within satellite imagery is a particularly challenging task due to the complex background, small target size, and dense object distribution\cite{xia2018dota}. Unlike natural scene images, satellite images often contain a large number of small objects (e.g., vehicles) with low contrast to the background, requiring models to possess strong localization and fine-grained recognition capabilities.

\begin{figure}[h]
    \centering
    \includegraphics[width=1.0\linewidth]{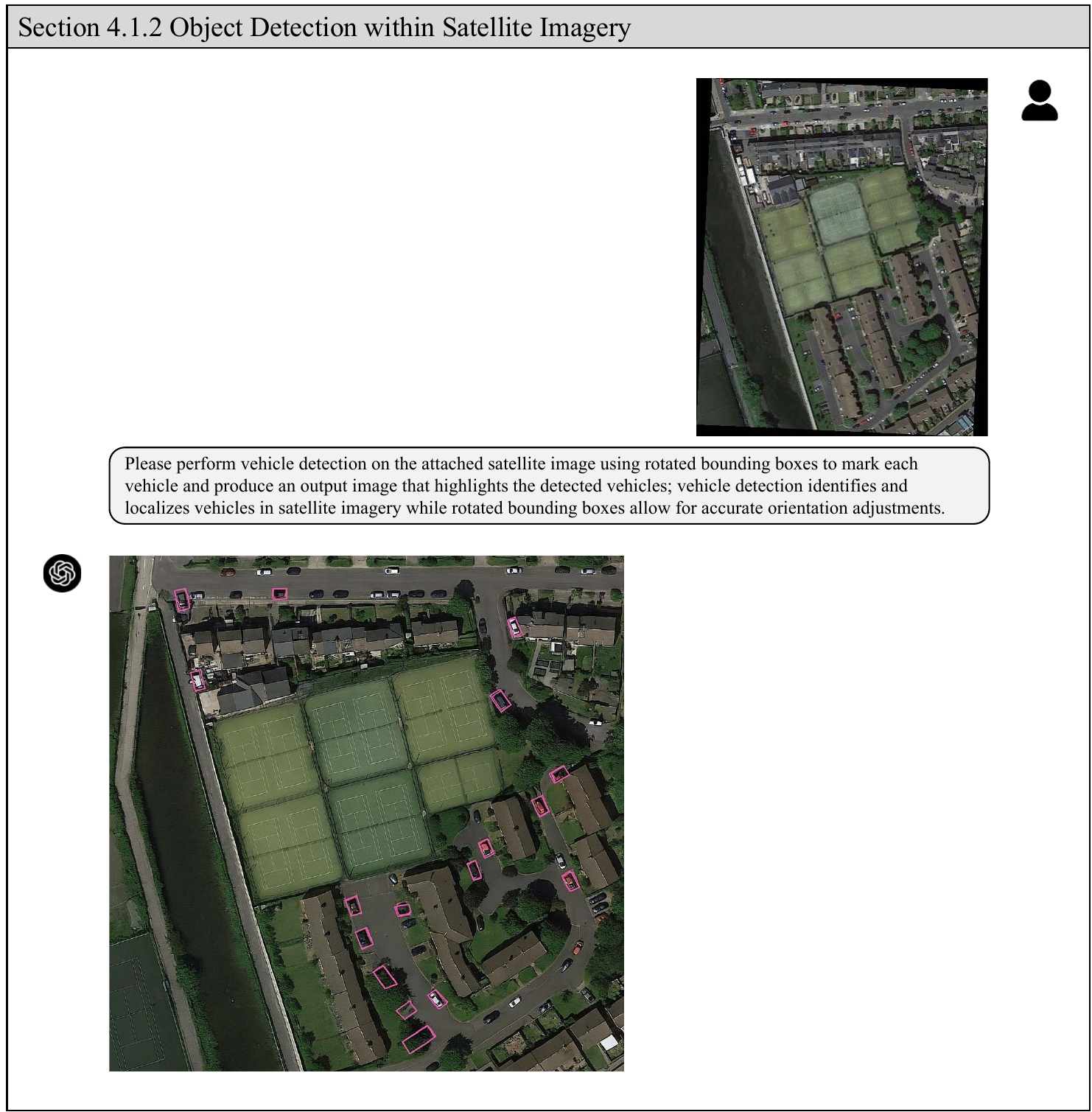}
    \caption[Sec~\ref{sec:satellite}: Object Detection within Satellite Imagery]{Examples of vehicle detection results in satellite imagery generated by \modelname. The model attempts to detect vehicles using rotated bounding boxes. However, a large number of small vehicles are missed, indicating the difficulty of dense small object detection in complex satellite scenes.}
    \label{fig:satellite-inaccurate}
\end{figure}

\begin{figure}[h]
    \centering
    \includegraphics[width=1.0\linewidth]{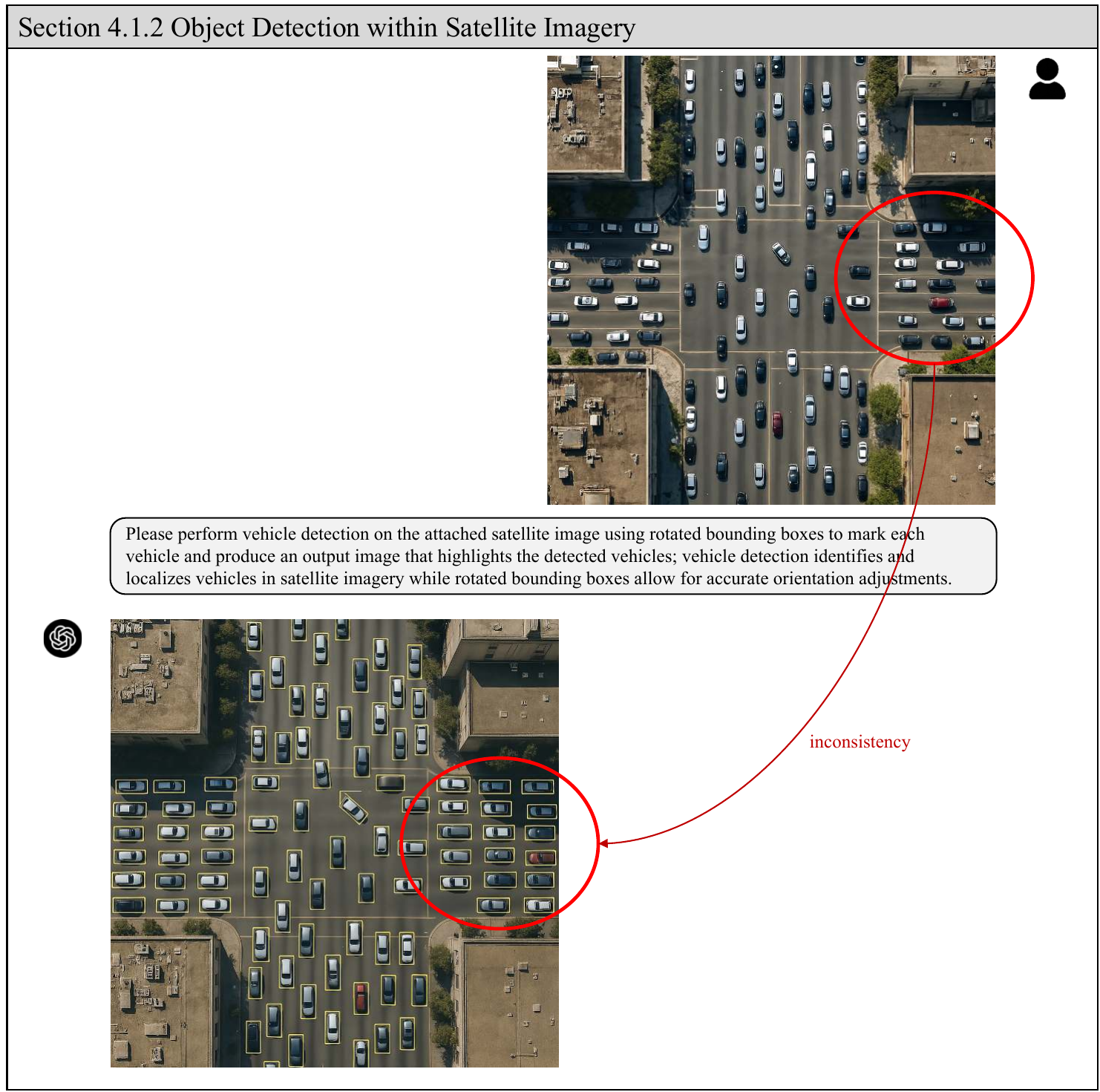}
    \caption[Sec~\ref{sec:satellite}: Object Detection within Satellite Imagery]{Additional examples of vehicle detection results in satellite imagery generated by \modelname. Although the detection accuracy is significantly improved with most vehicles correctly localized, the model tends to modify the original image content, leading to inconsistency with the input image.}
    \label{fig:satellite-change}
\end{figure}

As shown in Fig.\ref{fig:satellite-inaccurate} and Fig.\ref{fig:satellite-change}, we evaluate \modelname's ability to detect vehicles in satellite images using rotated bounding boxes for more accurate localization.
In Fig.~\ref{fig:satellite-inaccurate}, we observe that \modelname suffers from a large number of missed detections in the vehicle detection task. Many small vehicles in the image are not correctly identified, indicating that the model still struggles with dense small object detection in satellite imagery.

Although the detection results in Fig.~\ref{fig:satellite-change} show significantly improved accuracy, with most vehicles being successfully localized, we also observe a critical limitation of the model. Specifically, \modelname tends to modify the input image content during generation. In some regions, the original vehicle distribution has been altered or overwritten, breaking the consistency with the original scene.

These results demonstrate that object detection in satellite imagery remains a highly challenging task for \modelname. The model shows certain potential in generating reasonable detection results, but it still faces serious limitations in maintaining content consistency, dense small object detection, and fine-grained localization accuracy.

\clearpage

\clearpage

\subsubsection{Industrial Visual Inspection}
\label{sec:industry}
Industrial visual inspection is a highly specialized task that requires detecting fine-grained defects in structured and domain-specific environments\cite{thomas1995real,hutten2022vision}, such as manufacturing surfaces or printed circuit boards (PCB). Unlike general object detection tasks, this scenario often requires models to possess domain knowledge about typical defect patterns, material properties, and industrial standards.

\begin{figure}[h]
    \centering
    \includegraphics[width=1.0\linewidth]{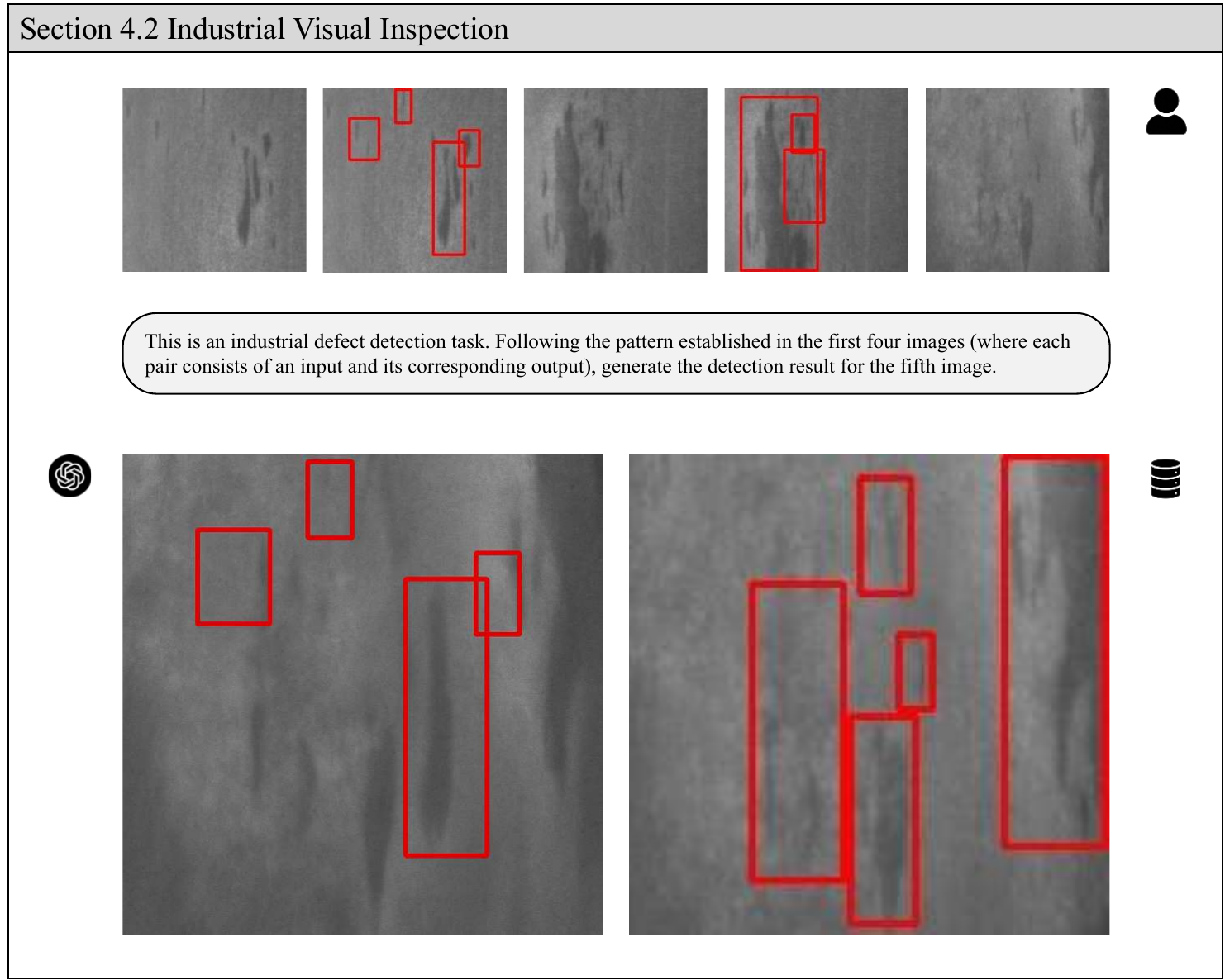}
    \caption[Sec~\ref{sec:industry}: Industrial Defect Detection]{Examples of surface defect detection results generated by \modelname on NEU-DET dataset\cite{song2013noise}. The model attempts to detect defects in industrial textures based on given patterns. However, the detection performance is sub-optimal, with inaccurate bounding box locations and inconsistent defect coverage.}
    \label{fig:industrial1}
\end{figure}

\begin{figure}[h]
    \centering
    \includegraphics[width=1.0\linewidth]{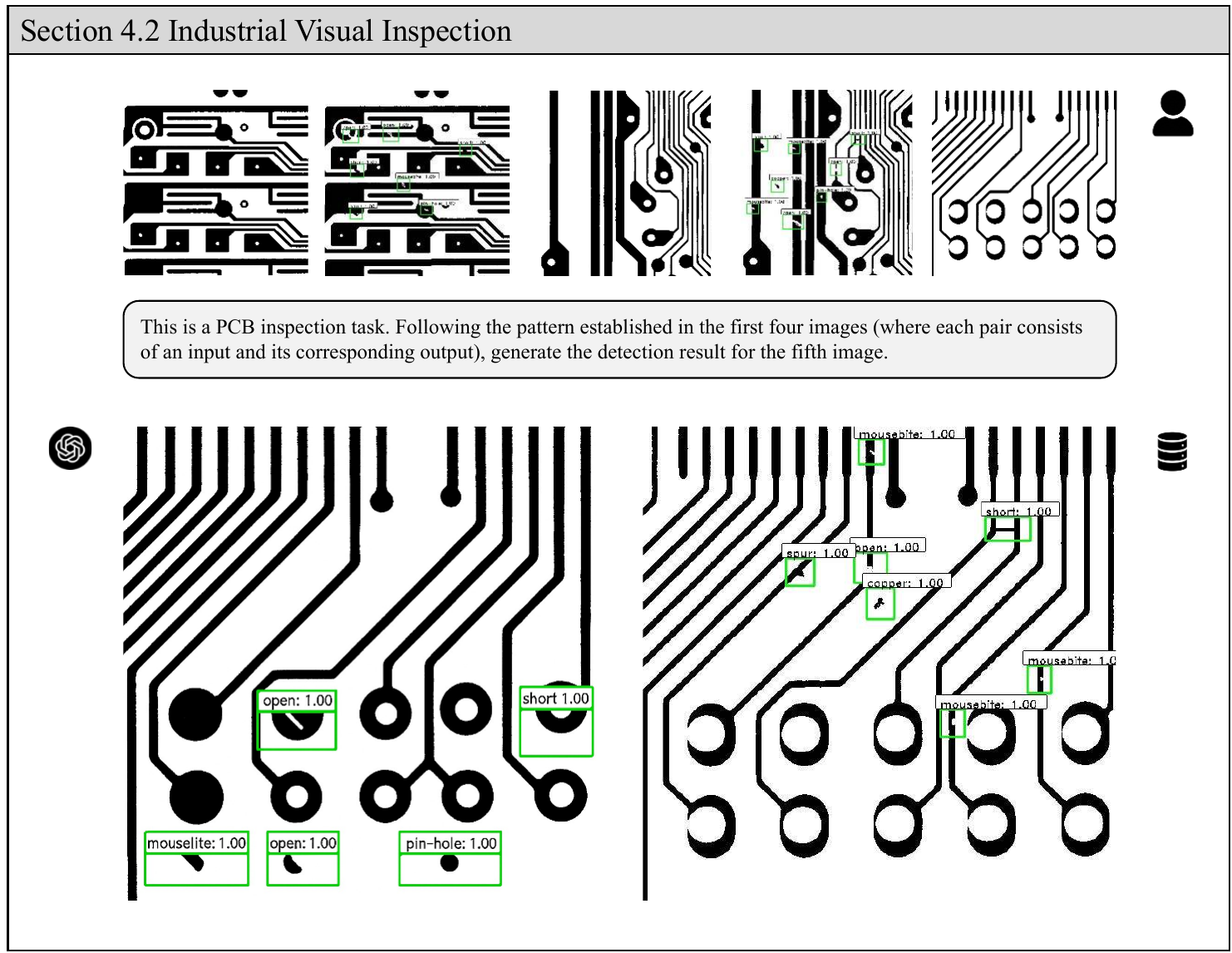}
    \caption[Sec~\ref{sec:industry}: PCB Inspection]{Examples of PCB defect detection results generated by \modelname on DeepDCB dataset\cite{tang2019onlinepcbdefectdetector}. The model detects various defects in PCB layouts, including open, short, mousebite, spur, pin hole and spurious copper. However, the detection accuracy is unsatisfactory, with missed detections and incorrect classifications.}
    \label{fig:industrial2}
\end{figure}

As shown in Fig.\ref{fig:industrial1} and Fig.\ref{fig:industrial2}, we evaluate \modelname in two representative industrial scenarios: surface defect detection\cite{song2013noise} and PCB defect detection\cite{tang2019onlinepcbdefectdetector}.
For surface defect detection, the model attempts to infer defect patterns based on several input-output examples. However, the generated detection results are unsatisfactory. The model fails to correctly localize the defects and produces bounding boxes with inaccurate locations and inconsistent sizes. Moreover, we observe that the model modifies the content of the input image, changing the visual appearance of the original surface, which is unacceptable in industrial inspection scenarios where input integrity must be preserved.
Similarly, in the PCB defect detection task, the model is required to detect predefined categories of defects (e.g., opens, shorts, mousebites). While \modelname can detect part of the defect regions, it suffers from low detection accuracy, frequent false positives, missed detections, and incorrect classifications. In some cases, the model even alters the input PCB pattern, which further impacts its practical usability.

These results indicate that \modelname faces significant challenges in industrial visual inspection tasks. Beyond general visual reasoning, this task requires the model to incorporate domain-specific prior knowledge to recognize subtle defects while strictly preserving the input content. The current generation of models lacks both accurate defect understanding and robust detection ability in such domain-constrained scenarios.

\clearpage

\clearpage

\subsection{Image Segmentation}
\label{sec:seg}

Image segmentation tasks assess a model’s ability to recognize, delineate, and label visual entities at different levels of granularity\cite{zhou2017scene,zhou2019semantic,Cordts2016Cityscapes,ronneberger2015u,cheng2022masked,cheng2021per,chen2014semantic}. In this section, we explore \modelname’s performance on four representative segmentation paradigms: semantic segmentation, instance segmentation, panoptic segmentation, and instruction-based or in-context segmentation. Each task provides a unique challenge in terms of understanding category definitions, spatial boundaries, and output format alignment.

\begin{figure}[h]
    \centering
    \includegraphics[width=1.0\linewidth]{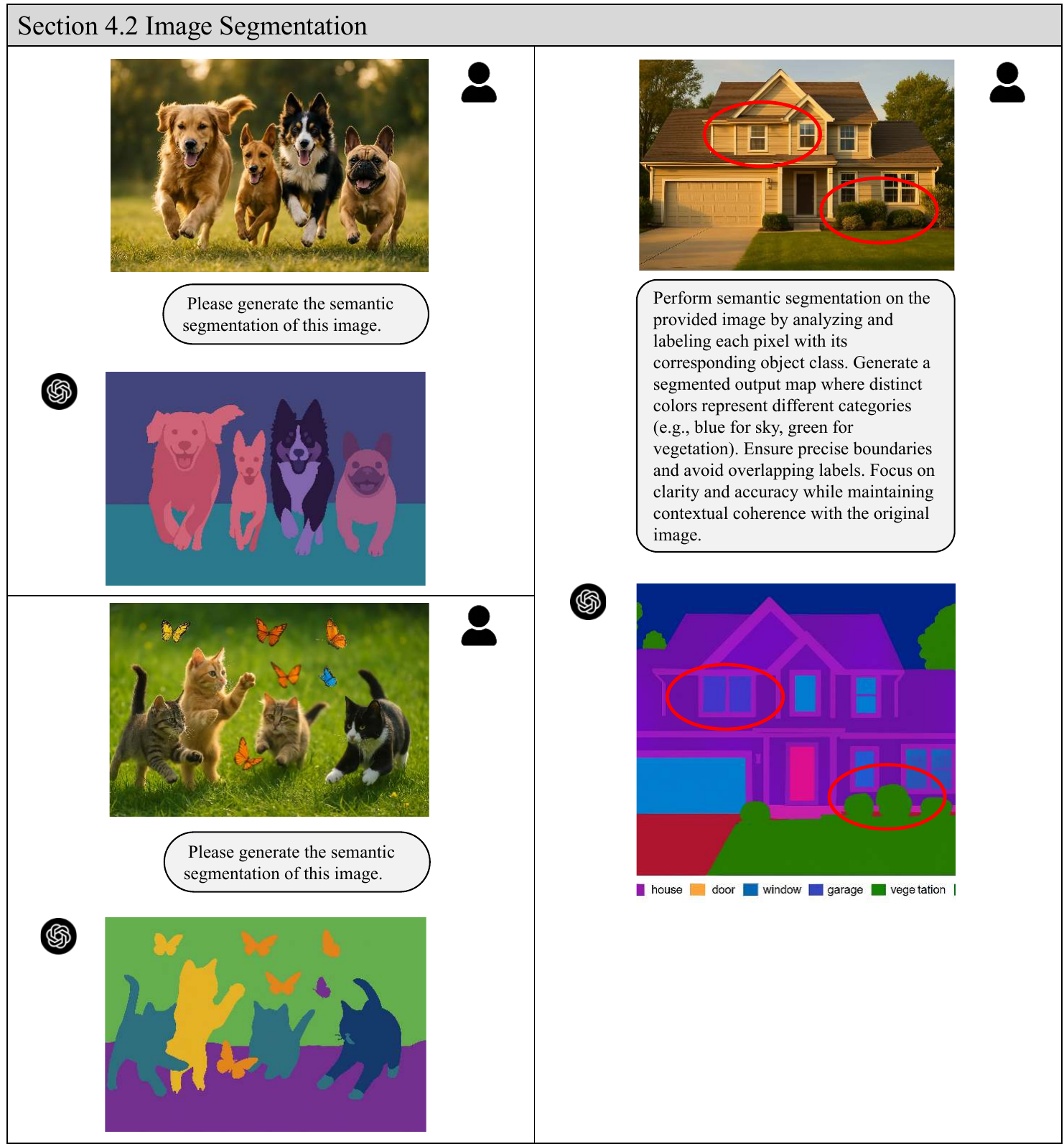}
    \caption[Sec~\ref{sec:seg}: Semantic Segmentation]{Examples of semantic segmentation results generated by \modelname.}
    \label{fig:semseg}
\end{figure}

\textbf{Semantic Segmentation.}  
As shown in Fig.~\ref{fig:semseg}, \modelname is able to perform basic region labeling and distinguish between different object categories. However, the model exhibits significant deviations from standard semantic segmentation conventions. First, the same semantic class is often rendered in inconsistent colors across different regions. Second, instead of grouping pixels solely by object class, the model sometimes performs part-level segmentation (e.g., dividing different window panes or tree parts). Additionally, generated masks are not always spatially accurate—the number of windows and shapes of bushes in the right image differ noticeably from the original input, suggesting semantic misalignment and geometric distortion.

\begin{figure}[h]
    \centering
    \includegraphics[width=1.0\linewidth]{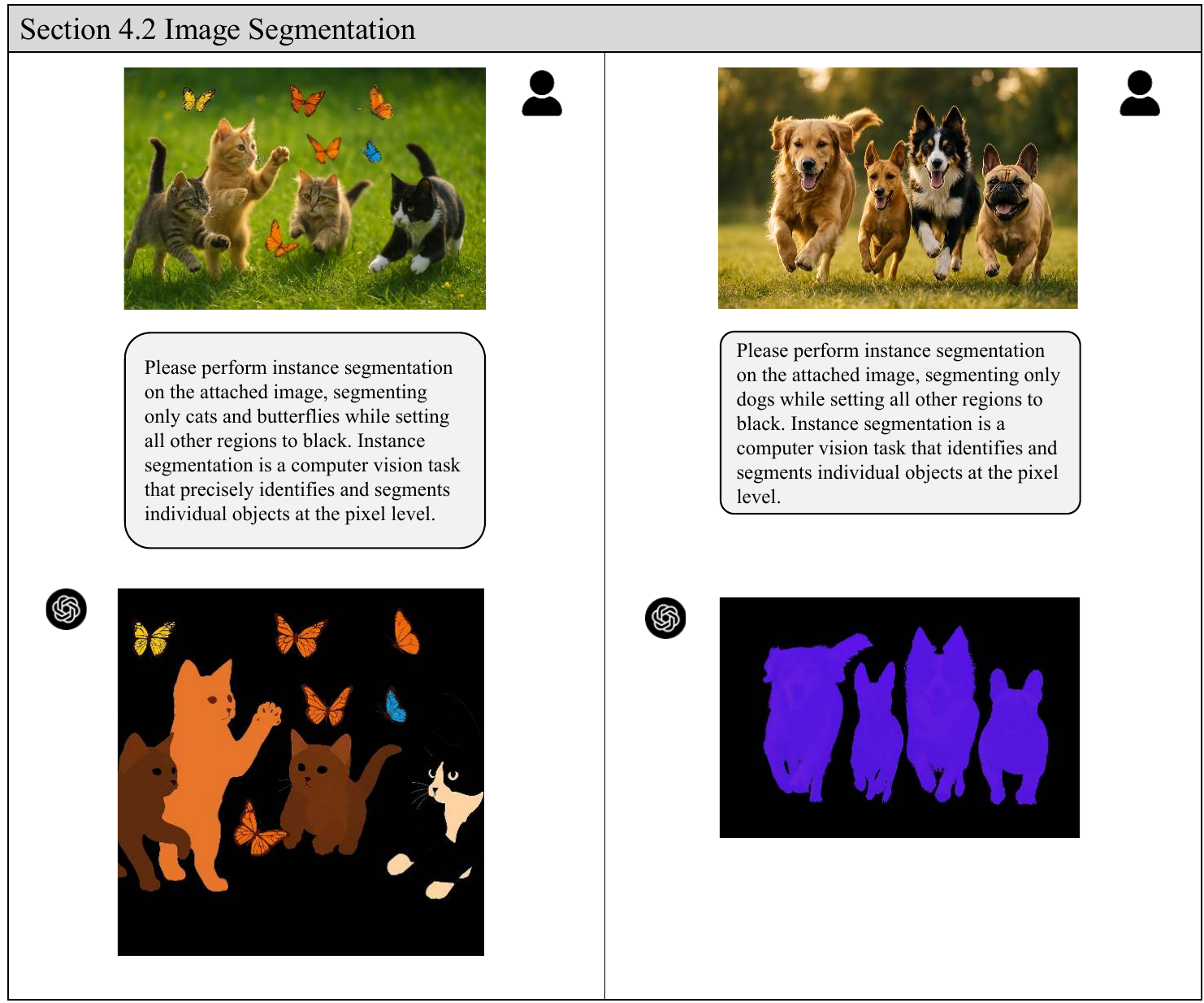}
    \caption[Sec~\ref{sec:seg}: Instance Segmentation]{Examples of instance segmentation results generated by \modelname.}
    \label{fig:insseg}
\end{figure}

\textbf{Instance Segmentation.}  
In instance segmentation (Fig.~\ref{fig:insseg}), the model's output suffers from both visual and functional limitations. In the left example, some masks are rendered in colors too similar to the underlying image content (e.g., a black cat is segmented using a nearly black mask), making it difficult to distinguish object boundaries. In the right example, the model fails to produce instance-level masks and instead defaults to class-level segmentation, essentially reverting to a semantic segmentation output. These results highlight the model’s limited awareness of the distinction between class-level and instance-level segmentation.

\begin{figure}[h]
    \centering
    \includegraphics[width=1.0\linewidth]{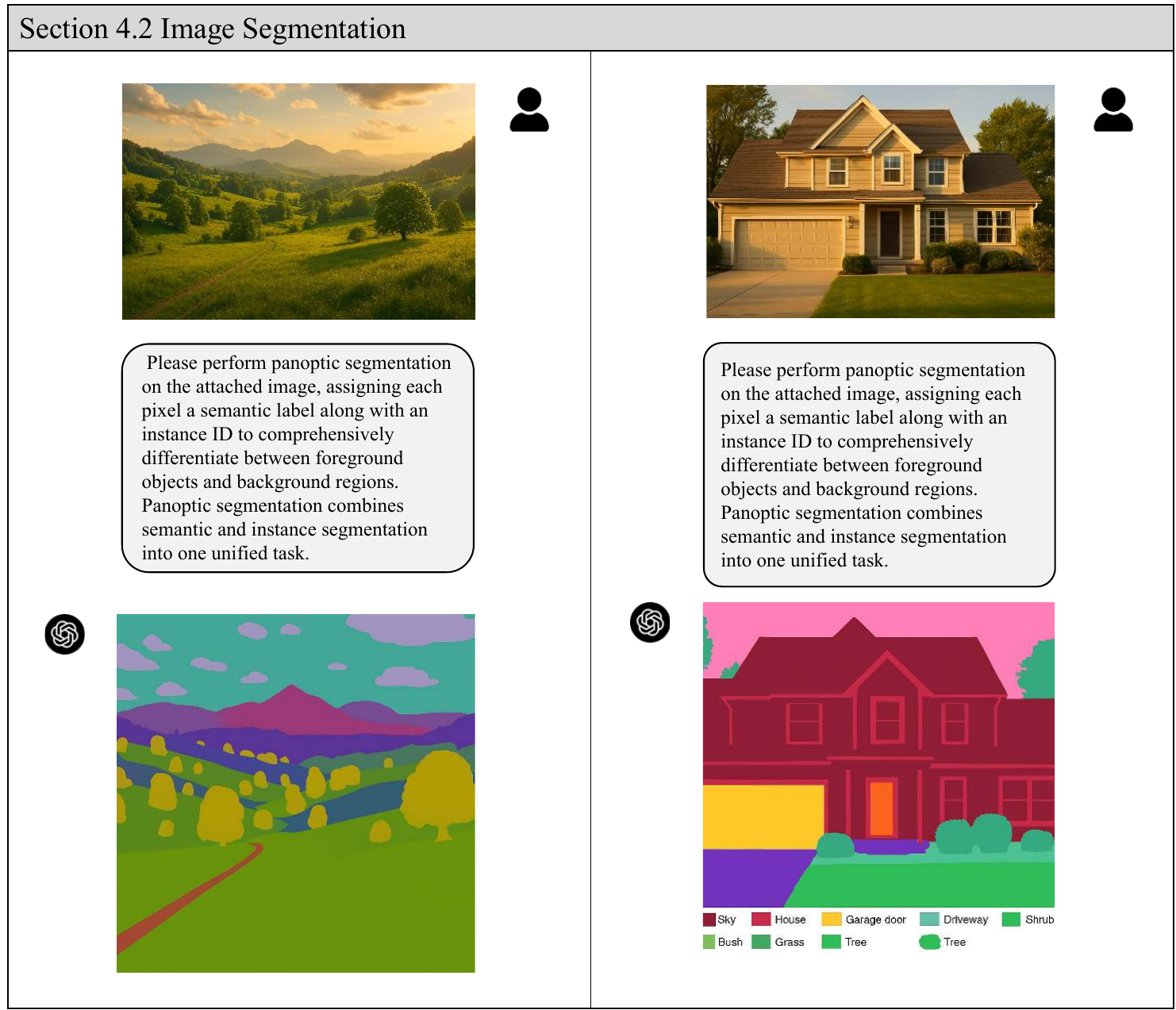}
    \caption[Sec~\ref{sec:seg}: Panoptic Segmentation]{Examples of panoptic segmentation results generated by \modelname.}
    \label{fig:panseg}
\end{figure}

\textbf{Panoptic Segmentation.}  
As shown in Fig.~\ref{fig:panseg}, the model generates panoptic segmentation outputs that are visually coherent and structurally reasonable. It manages to cover both foreground instances and background classes simultaneously. However, category labeling is occasionally incorrect, such as labeling the sky region with unrelated or mismatched class names, revealing a gap in taxonomy grounding and label accuracy.

\begin{figure}[h]
    \centering
    \includegraphics[width=1.0\linewidth]{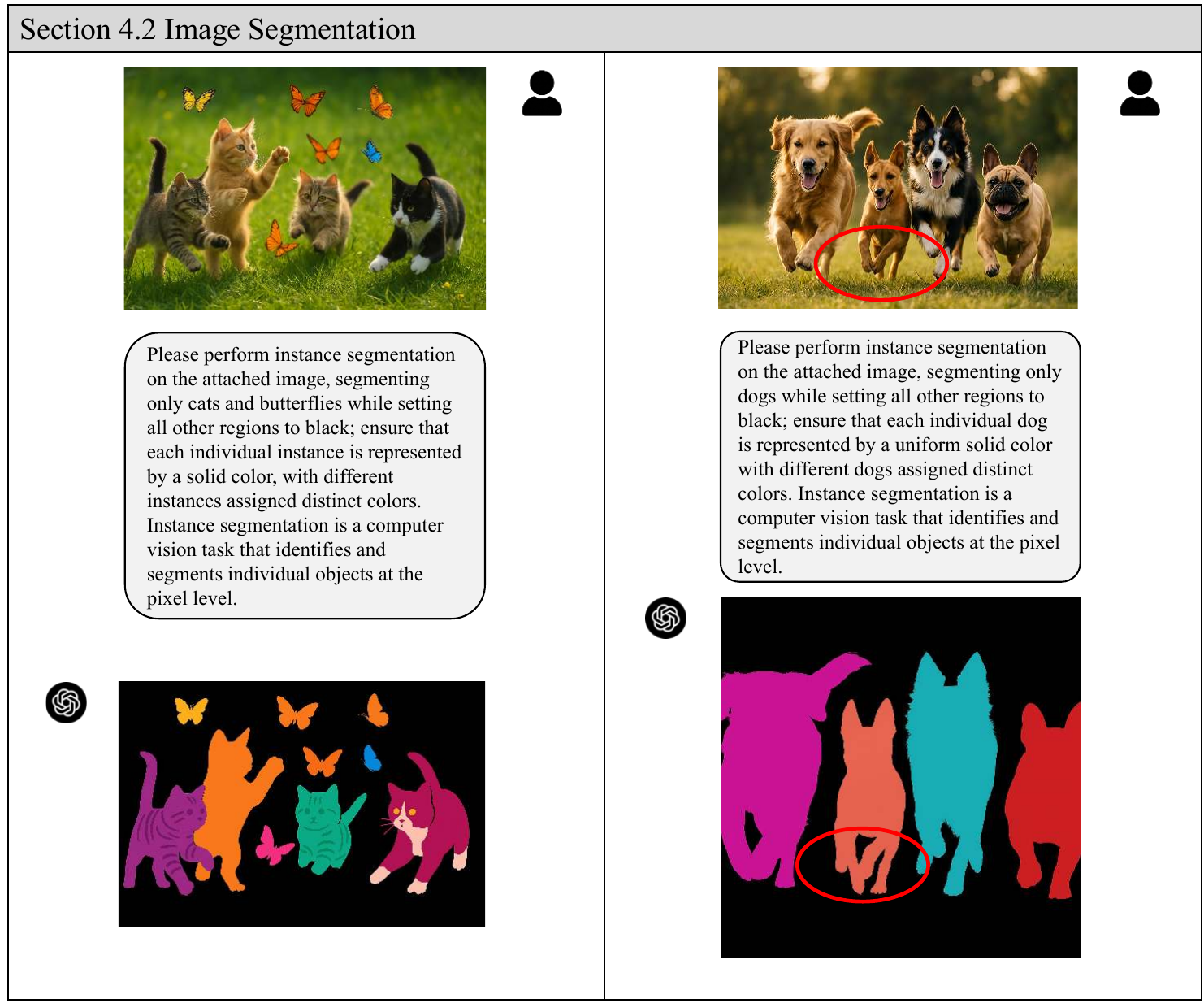}
    \caption[Sec~\ref{sec:seg}: Instance Segmentation with Task Instruction]{Examples of instance segmentation results with task instruction generated by \modelname.}
    \label{fig:insseg-define}
\end{figure}

\begin{figure}[h]
    \centering
    \includegraphics[width=1.0\linewidth]{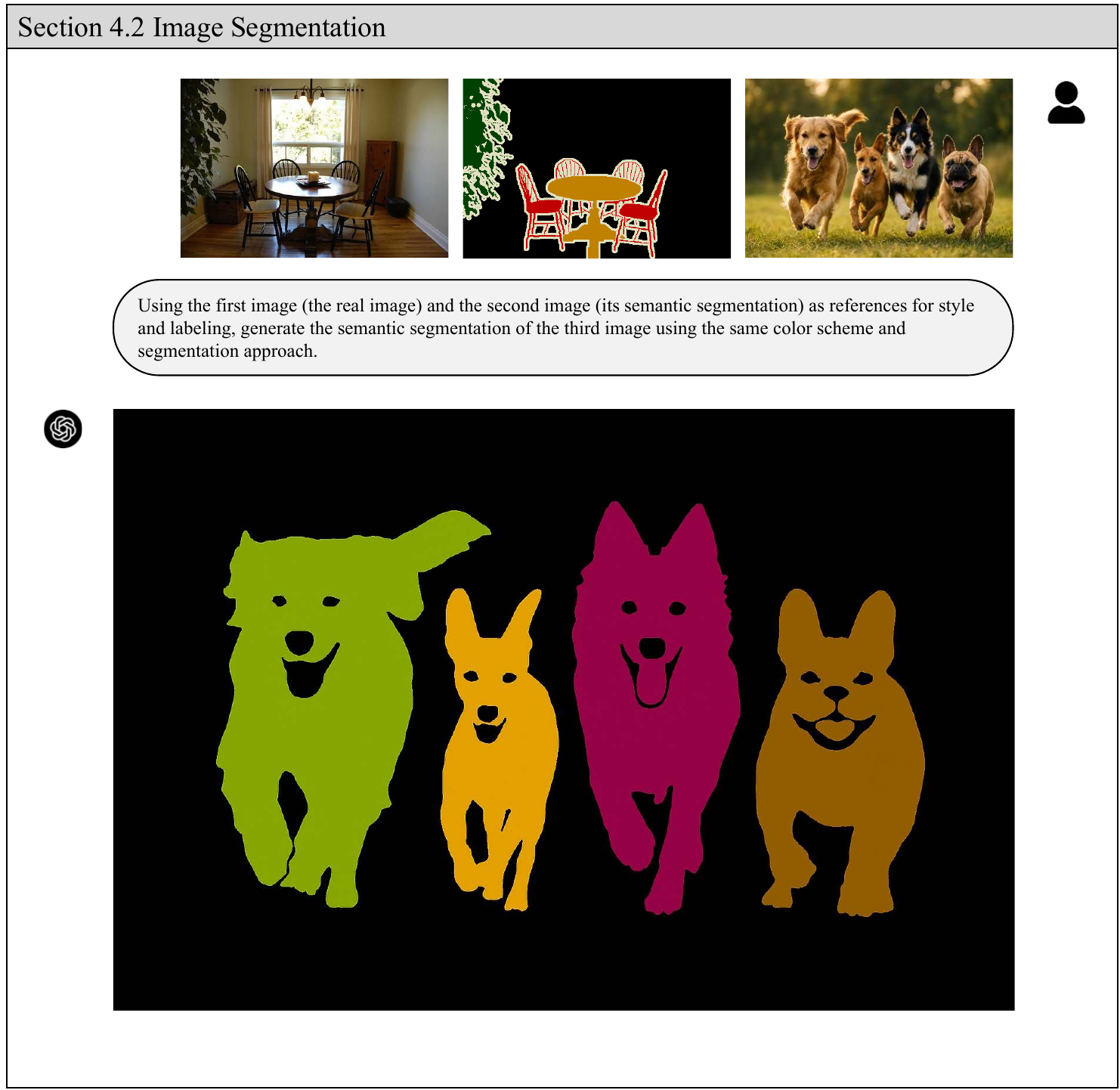}
    \caption[Sec~\ref{sec:seg}: Semantic Segmentation with In-Context Learning]{Examples of semantic segmentation with in-context learning generated by \modelname.}
    \label{fig:incontext1}
\end{figure}

\begin{figure}[h]
    \centering
    \includegraphics[width=1.0\linewidth]{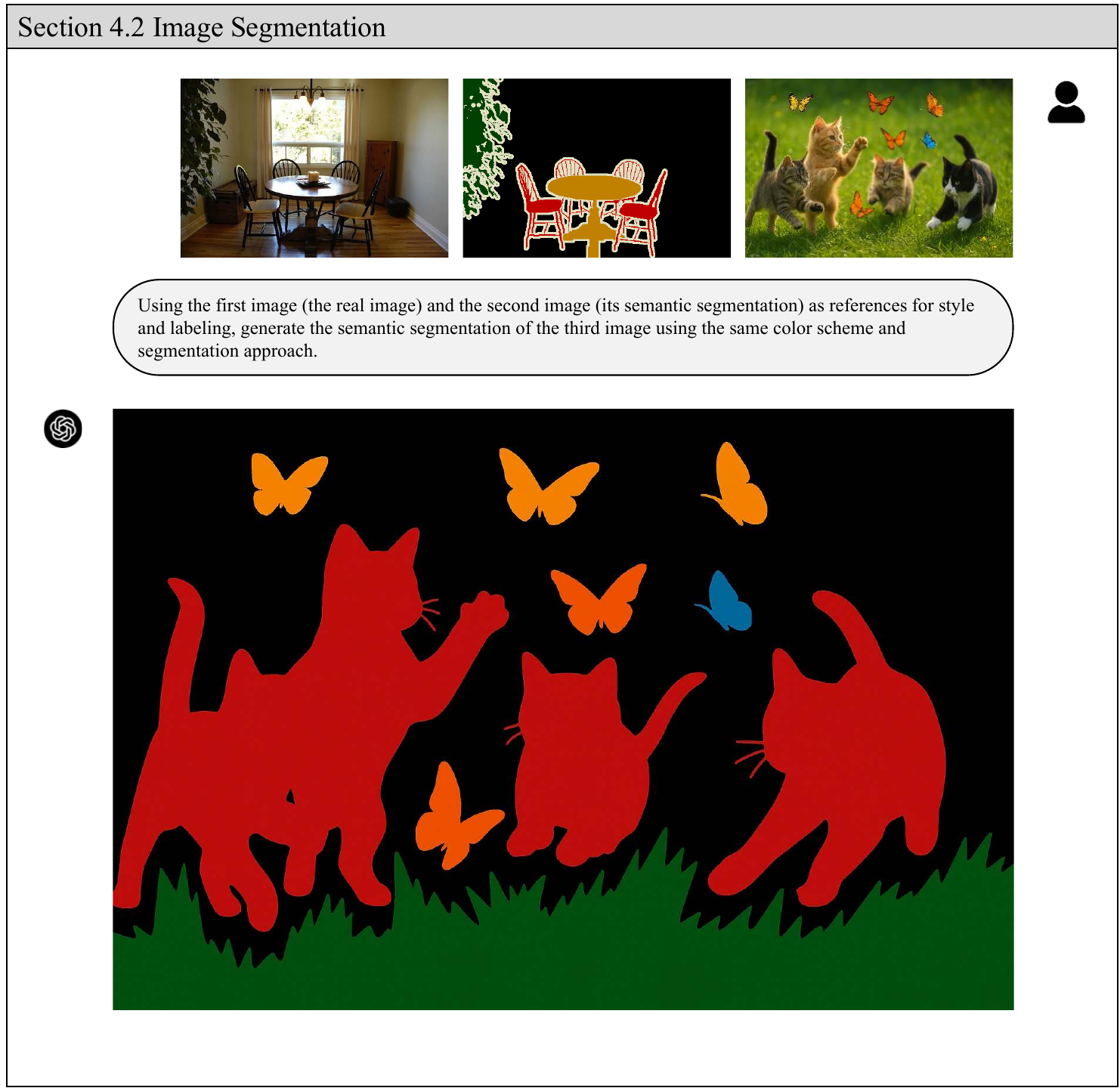}
    \caption[Sec~\ref{sec:seg}: Semantic Segmentation with In-Context Learning]{Examples of semantic segmentation with in-context learning generated by \modelname.}
    \label{fig:incontext2}
\end{figure}

\textbf{Segmentation with Task Instruction and In-context Learning.}  
To address the misalignment between prompts and task-specific outputs observed in the previous settings, we experiment with providing explicit task instructions (Fig.~\ref{fig:insseg-define}) or using in-context learning with visual exemplars (Fig.~\ref{fig:incontext1} and Fig.~\ref{fig:incontext2}). These methods show partial improvement. The segmentation becomes more consistent with the task type, and the visual format aligns better with expected outputs. However, some inconsistencies still remain, especially when input images contain ambiguous or densely packed objects.

Our analysis reveals that \modelname is capable of identifying and segmenting major semantic entities in images. However, it lacks a precise understanding of the differences between segmentation task types and struggles to fully align outputs with specific task requirements. These limitations suggest that while \modelname has strong visual reasoning capabilities, its task grounding and format alignment can benefit from more explicit prompt tuning or modular control mechanisms.

\clearpage

\clearpage

\subsection{Counting}
\label{sec:count}
Counting is a fundamental vision task that requires models to identify and enumerate all instances of specified objects within a scene\cite{onoro2016towards,lin2021object,cholakkal2019object}. It serves as a benchmark for both object detection and fine-grained visual reasoning.

\begin{figure}[h]
    \centering
    \includegraphics[width=1.0\linewidth]{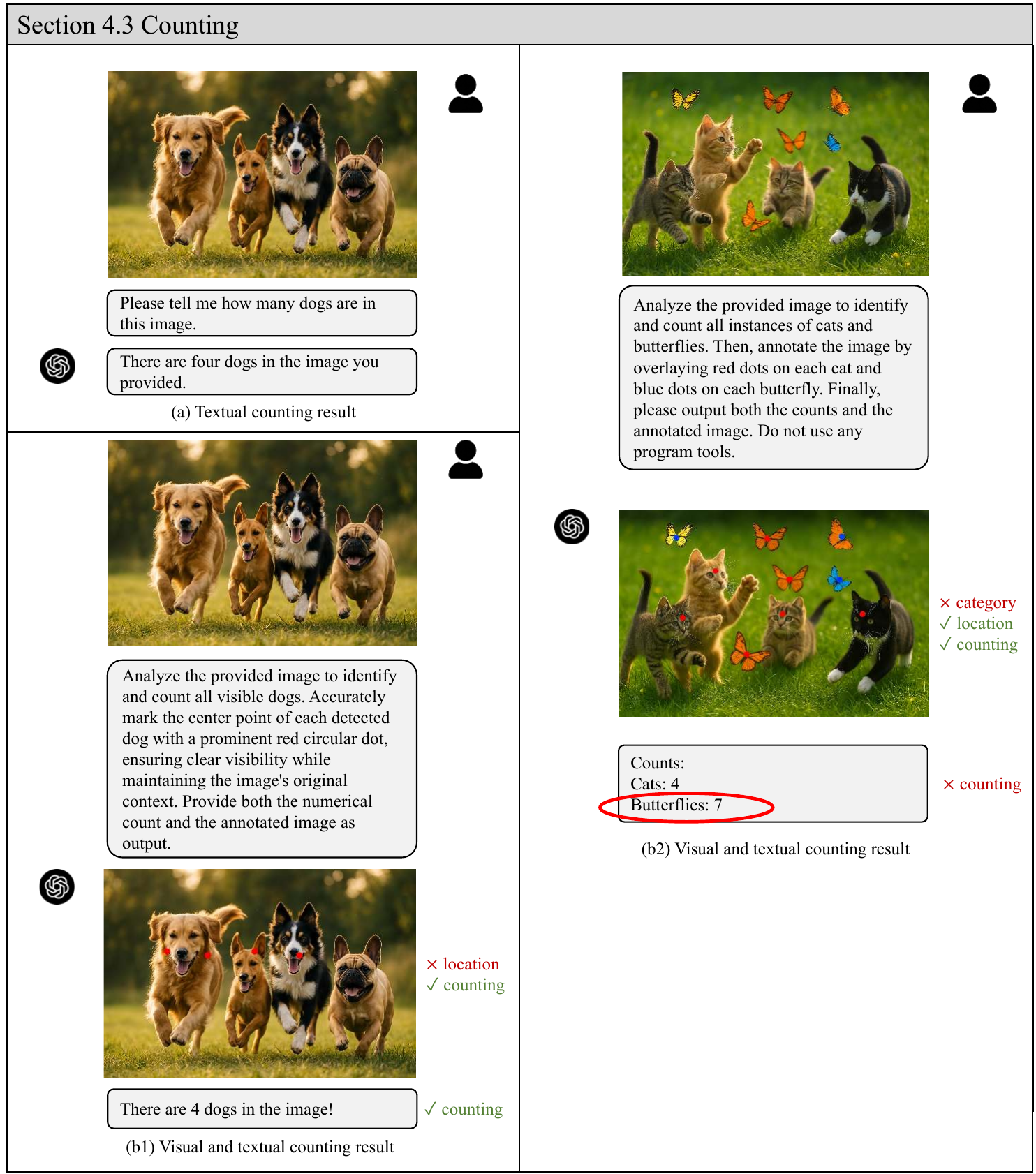}
    \caption[Sec~\ref{sec:count}: Counting]{Examples of object counting results by \modelname.}
    \label{fig:counting}
\end{figure}

Thanks to its strong visual understanding capabilities, \modelname can directly perform counting from natural language instructions alone. As shown in Fig.~\ref{fig:counting} (a), when prompted with “How many dogs are in this image?”, the model correctly outputs the numerical result in text format, without requiring any visual guidance or interactive interface.

To further examine whether visual reasoning could enhance counting accuracy, we explore an additional setup where the model is instructed to annotate the image by placing colored dots on each detected instance before providing the final count (see Fig.~\ref{fig:counting} b1 and b2). In these cases, the model is asked to mark each object (e.g., red dots for dogs, blue dots for butterflies), then return both the annotated image and the object count.
However, our observations show that visual annotation does not lead to a significant improvement in final counting accuracy. Although the model successfully places visual markers on object instances in some cases, errors such as missed counts or incorrect object classification still occur. 

\modelname demonstrates a competent ability to perform direct object counting via textual output, and is capable of localizing objects when prompted. However, integrating visual outputs (e.g., annotated dots) does not necessarily improve performance, suggesting that visual reasoning and numerical generation remain decoupled to some extent in the model’s architecture.

\clearpage
\clearpage

\subsection{Human-centric Understanding}
\label{sec:pose}

\subsubsection{Human Pose Estimation}
\label{sec:hpe}
Human Pose Estimation (HPE) aims to localize keypoints on the human body, such as joints and limbs, and is typically divided into 2D\cite{andriluka20142d,wang2022lite,dang2019deep} and 3D\cite{wang2021deep,yang20183d,zhou2024lifting} variants. This task evaluates a model’s spatial understanding of human anatomy, its ability to detect joint relationships, and alignment with pose estimation protocols.

\begin{figure}[h]
    \centering
    \includegraphics[width=1.0\linewidth]{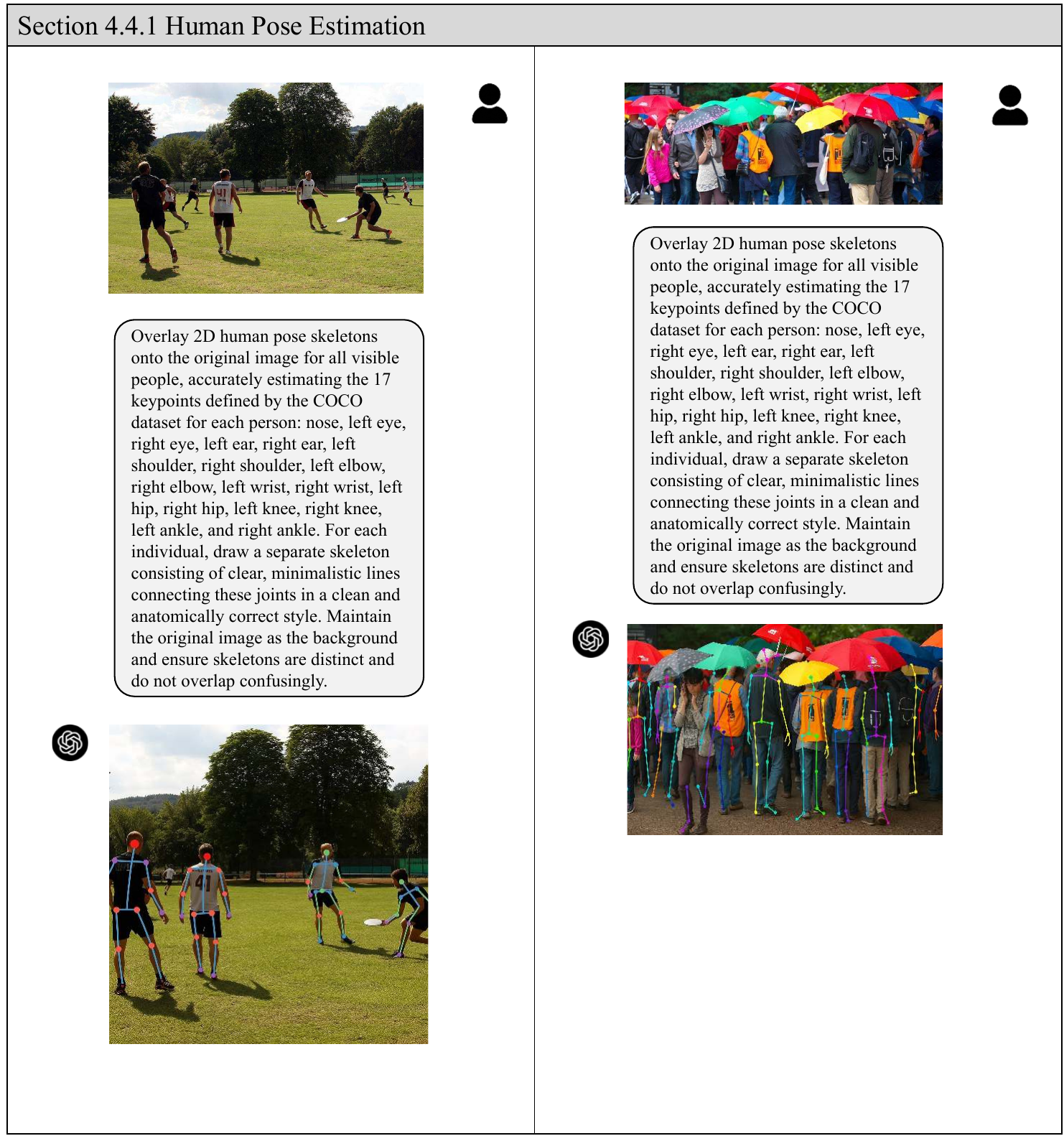}
    \caption[Sec~\ref{sec:hpe}: Human Pose Estimation]{Examples of human pose estimation results by \modelname.}
    \label{fig:human_pose1}
\end{figure}

\begin{figure}[h]
    \centering
    \includegraphics[width=1.0\linewidth]{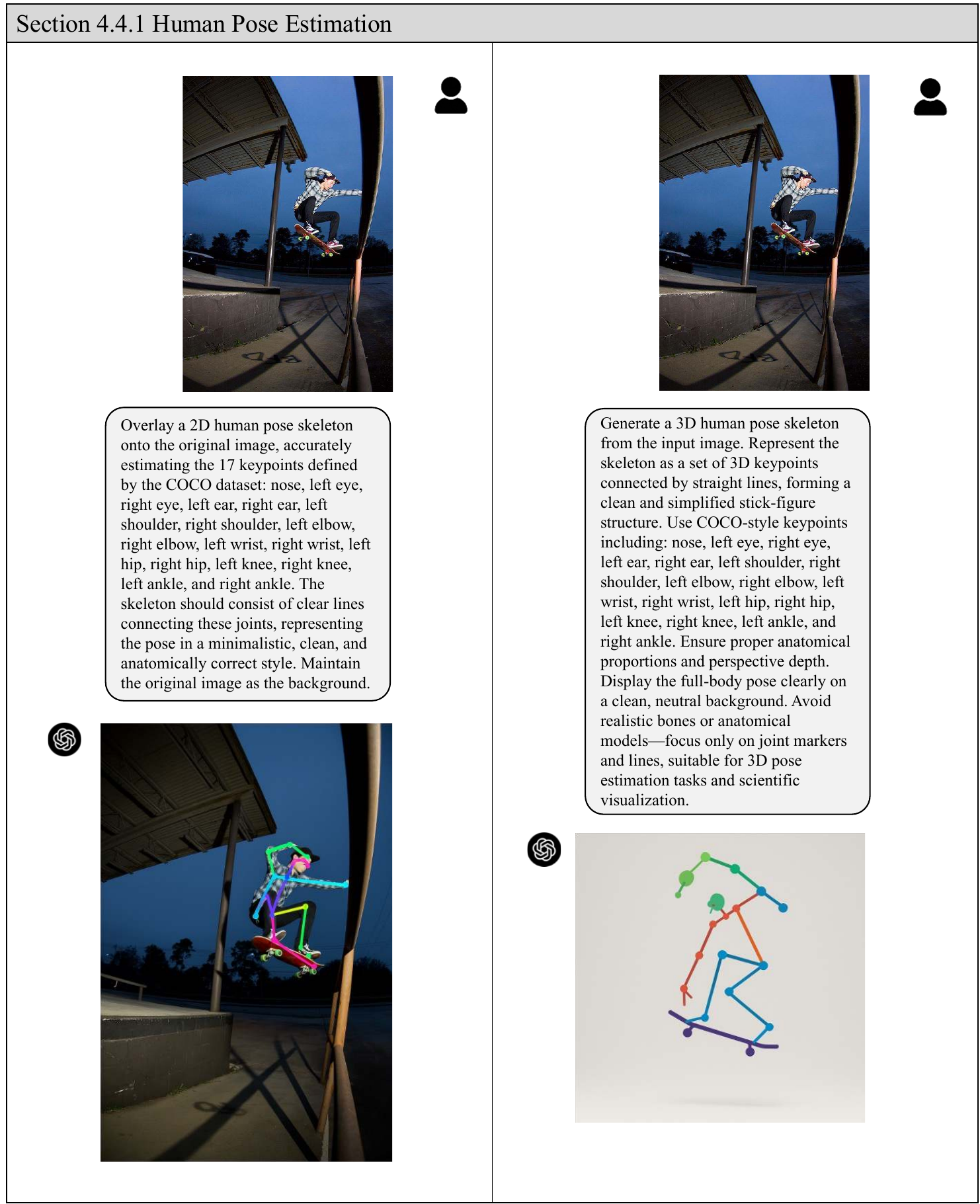}
    \caption[Sec~\ref{sec:hpe}: Human Pose Estimation]{Additional Examples of human pose estimation results by \modelname.}
    \label{fig:human_pose2}
\end{figure}

\begin{figure}[h]
    \centering
    \includegraphics[width=1.0\linewidth]{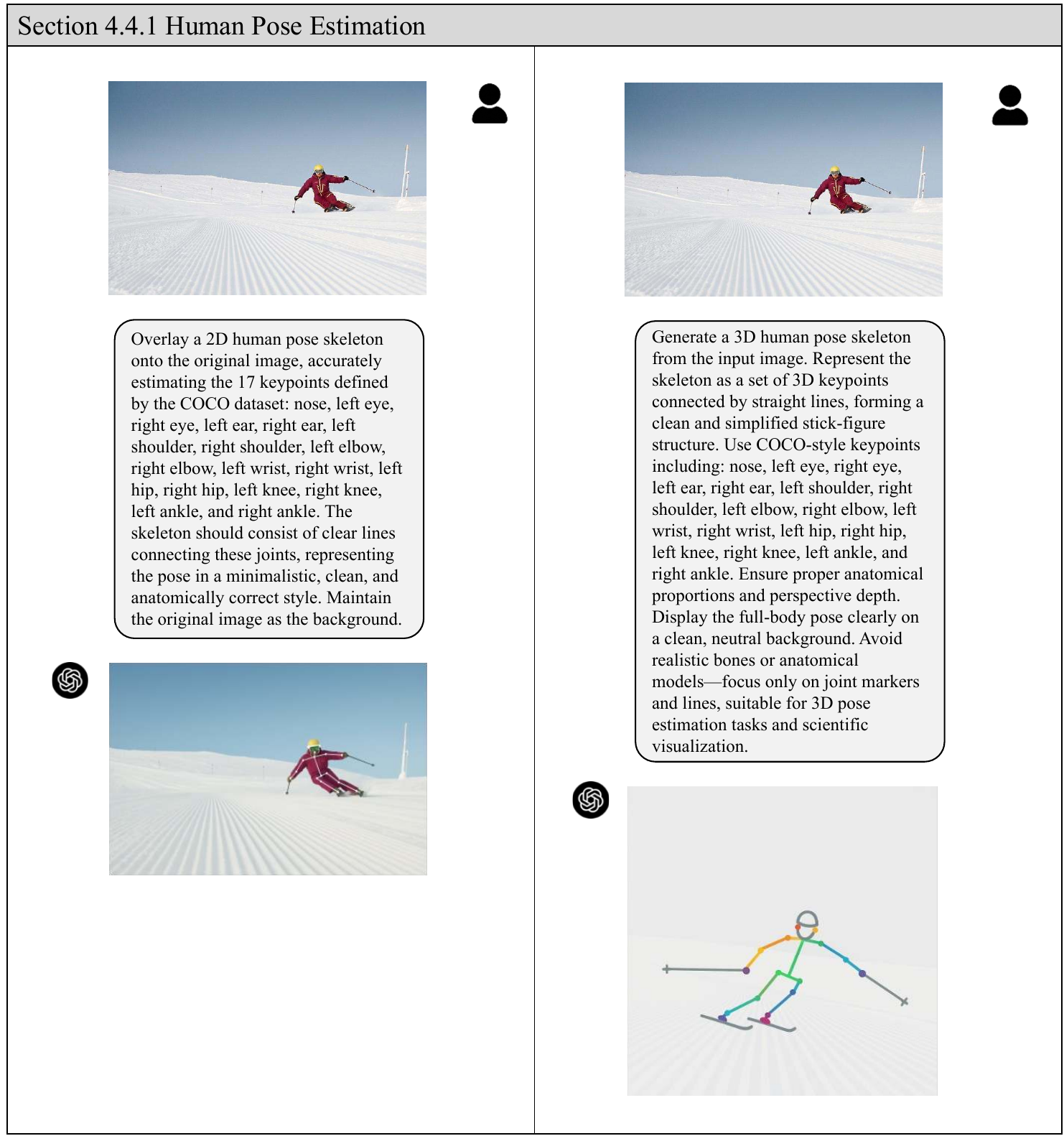}
    \caption[Sec~\ref{sec:hpe}: Human Pose Estimation]{Additional examples of human pose estimation results by \modelname.}
    \label{fig:human_pose3}
\end{figure}

We evaluate \modelname on both 2D and 3D human pose estimation tasks. As illustrated in Fig.~\ref{fig:human_pose1}, Fig.~\ref{fig:human_pose2}, and Fig.~\ref{fig:human_pose3}, \modelname is able to highlight approximate human body structures by generating pose lines or keypoint overlays. However, its performance varies significantly across instances and often fails to align with standard pose estimation requirements.

In the 2D setting, the model sometimes misplaces major joints (e.g., knees or shoulders), produces anatomically implausible connections, or detects an incorrect number of keypoints. In the 3D setting, while the model attempts to simulate spatial depth through shading or perspective cues, the resulting poses lack realistic depth continuity and spatial coherence. Moreover, the outputs are not consistent with any specific annotation format (e.g., COCO 17-keypoint), and the keypoint count often deviates from task specifications.

Overall, \modelname exhibits a rudimentary ability to infer body structure and perform coarse pose visualization. However, it lacks task-specific alignment, standardized output structure, and precise keypoint localization. This suggests that without explicit fine-tuning or structured pose templates, current multimodal models struggle with high-fidelity human pose estimation.

\clearpage

\clearpage

\subsubsection{Human Parsing}
\label{sec:parsing}
Human parsing involves segmenting fine-grained parts of the human body, such as clothing, facial features, limbs, and accessories, into semantically meaningful regions\cite{yang2024deep,yang2022quality,yang2020renovating,ruan2019devil,yang2019parsing}. It is a dense prediction task that requires precise spatial alignment and a detailed understanding of human anatomy and appearance.

\begin{figure}[h]
    \centering
    \includegraphics[width=1.0\linewidth]{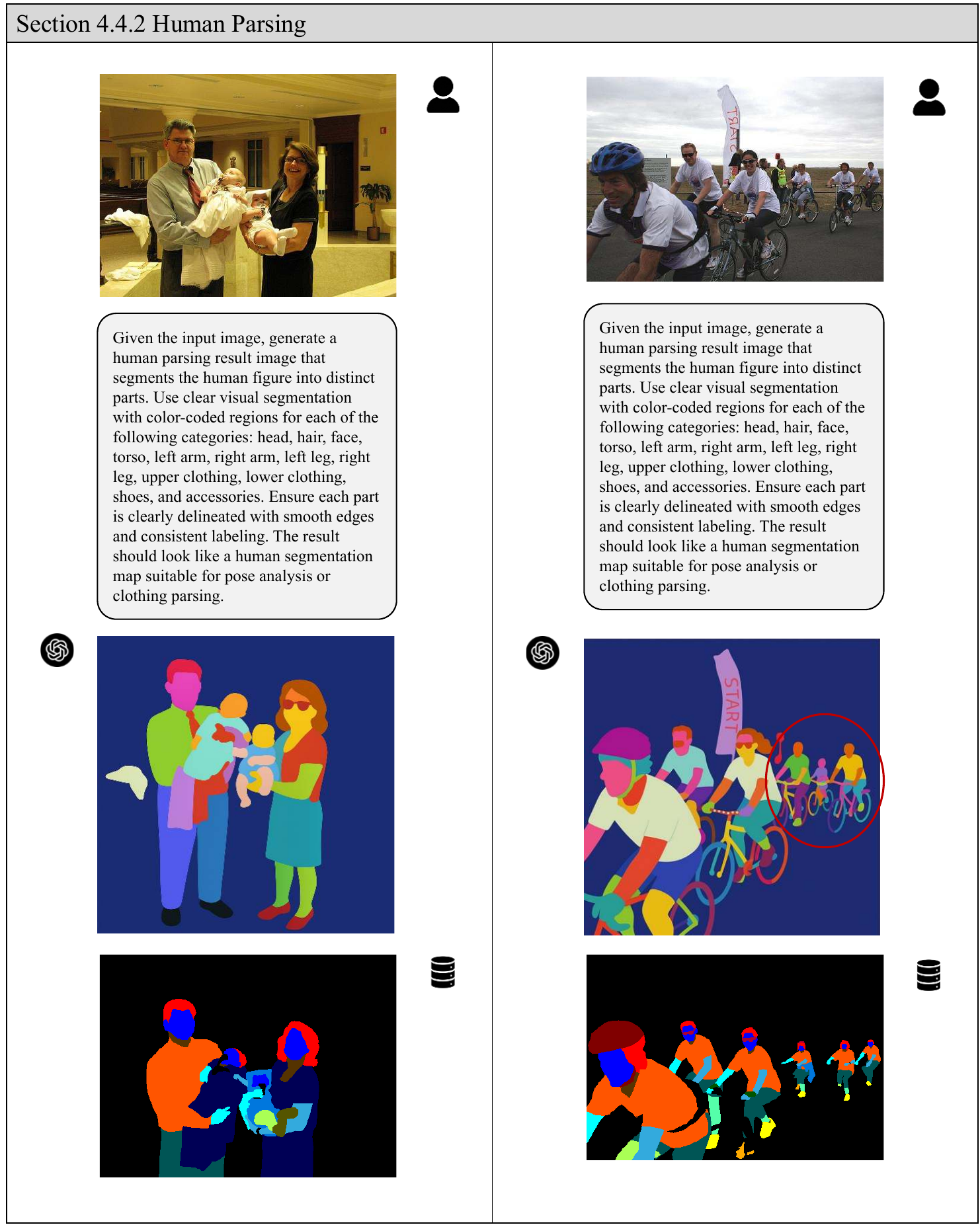}
    \caption[Sec~\ref{sec:parsing}: Human Parsing]{Examples of human parsing generated by \modelname.}
    \label{fig:parsing1}
\end{figure}

\begin{figure}[h]
    \centering
    \includegraphics[width=1.0\linewidth]{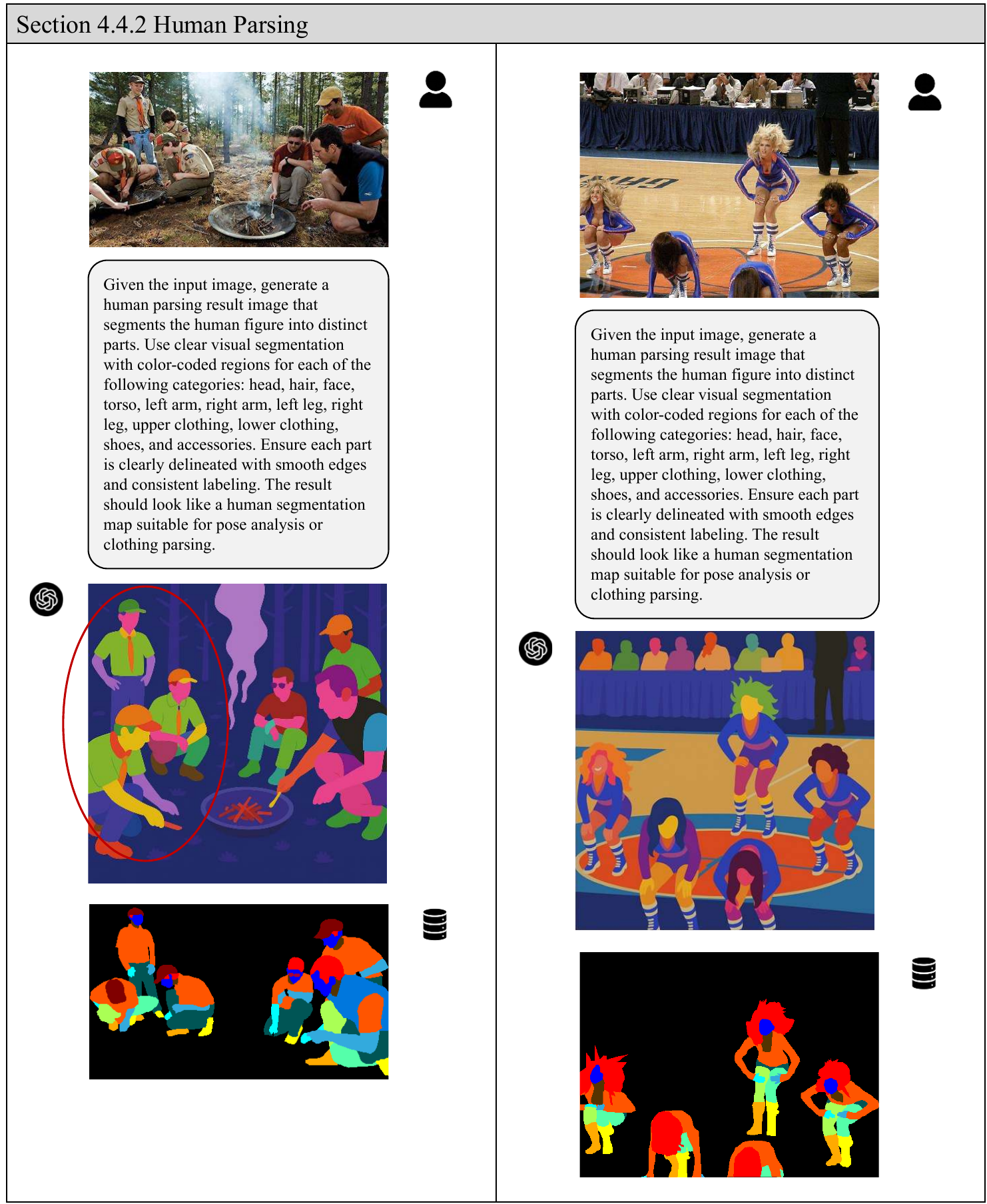}
    \caption[Sec~\ref{sec:parsing}: Human Parsing]{Additional examples of human parsing generated by \modelname.}
    \label{fig:parsing2}
\end{figure}

As shown in Fig.~\ref{fig:parsing1} and Fig.~\ref{fig:parsing2}, \modelname demonstrates the ability to perform semantic segmentation over image regions. However, its results often fail to meet the specific requirements of human parsing. For example, in Fig.~\ref{fig:parsing1} (right), the model incorrectly segments a bicycle as part of the human parsing output and fails to isolate facial accessories like eyeglasses. Additionally, it produces segmentation errors in crowded scenes, as can be seen in right part of Fig.~\ref{fig:parsing1}.

In Fig.~\ref{fig:parsing2}, these issues become more pronounced. The model produces parsing results that appear to be influenced more by the overall image style and composition than by the actual spatial structure of the individuals. For instance, in the leftmost person of the left image, the generated segments do not correspond to meaningful anatomical or clothing regions. This suggests a lack of fine-grained spatial alignment and insufficient grounding in the input image’s local content.

While \modelname can distinguish broad semantic regions and generate plausible masks, it lacks task-specific understanding of human parsing. Its outputs fail to reflect detailed human part structures and often include irrelevant or inconsistent regions. This highlights the need for stronger spatial grounding and part-aware supervision to adapt \modelname to parsing tasks.

\clearpage

\clearpage

\subsubsection{Human Mesh Recovery}
\label{sec:hmr}
Human mesh recovery aims to reconstruct a 3D mesh representation of the human body from a single image\cite{kolotouros2021probabilistic,pavlakos2022human,shen2023learning}, capturing detailed attributes such as pose, shape, and orientation. The ideal output is a mesh overlay with visible contours and vertex structure, following formats like SMPL.

\begin{figure}[h]
    \centering
    \includegraphics[width=1.0\linewidth]{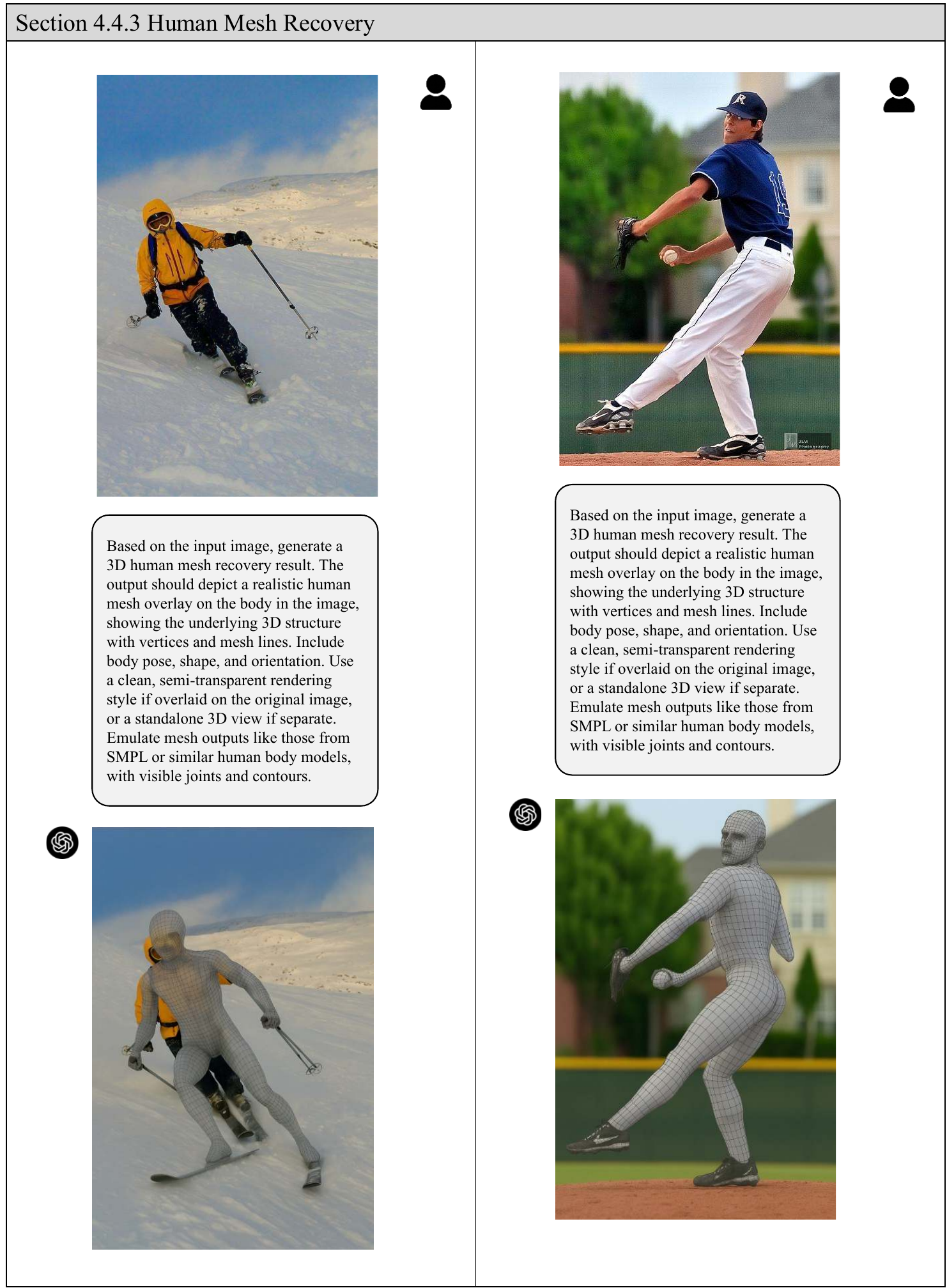}
    \caption[Sec~\ref{sec:hmr}: Human Mesh Recovery]{Examples of human mesh recovery generated by \modelname.}
    \label{fig:hmr}
\end{figure}

As illustrated in Fig.~\ref{fig:hmr}, we use this task to evaluate whether \modelname possesses spatial reasoning and 3D body structure perception capabilities. The results vary across different inputs. In the right example, the generated mesh demonstrates good alignment with the original body pose and proportions, reflecting a well-formed understanding of body orientation and shape. However, in the left example, we observe significant misalignment between the rendered mesh and the original person. The pose and limb configuration of the generated mesh deviate from the input, leading to unrealistic overlays.

These findings suggest that while \modelname shows some ability to infer 3D mesh structure from 2D visual input, its robustness and consistency remain limited. In challenging scenarios, it may struggle with precise geometric alignment and structural integrity.

\clearpage

\clearpage

\clearpage
\subsection{Depth Estimation}
\label{sec:depth}
Depth estimation aims to infer the relative or absolute distance of objects in a scene from a single 2D image\cite{mertan2022single,bhat2021adabins,wofk2019fastdepth}, producing a depth map that captures geometric structure and spatial relationships. This task is fundamental to scene understanding and 3D reasoning.

\begin{figure}[h]
    \centering
    \includegraphics[width=1.0\linewidth]{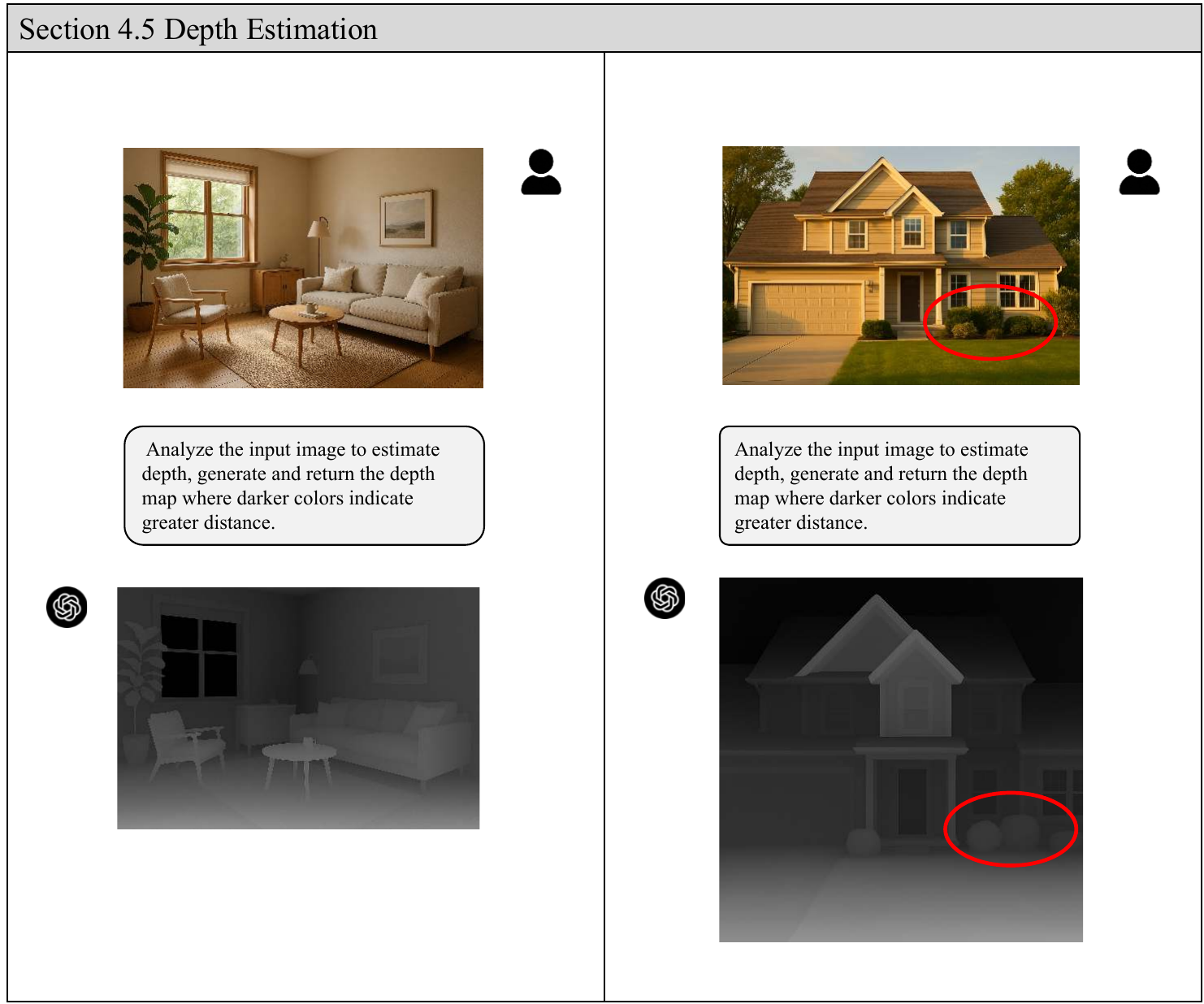}
    \caption[Sec~\ref{sec:depth}: Depth Estimation]{}
    \label{fig:depth1}
\end{figure}

\begin{figure}[h]
    \centering
    \includegraphics[width=1.0\linewidth]{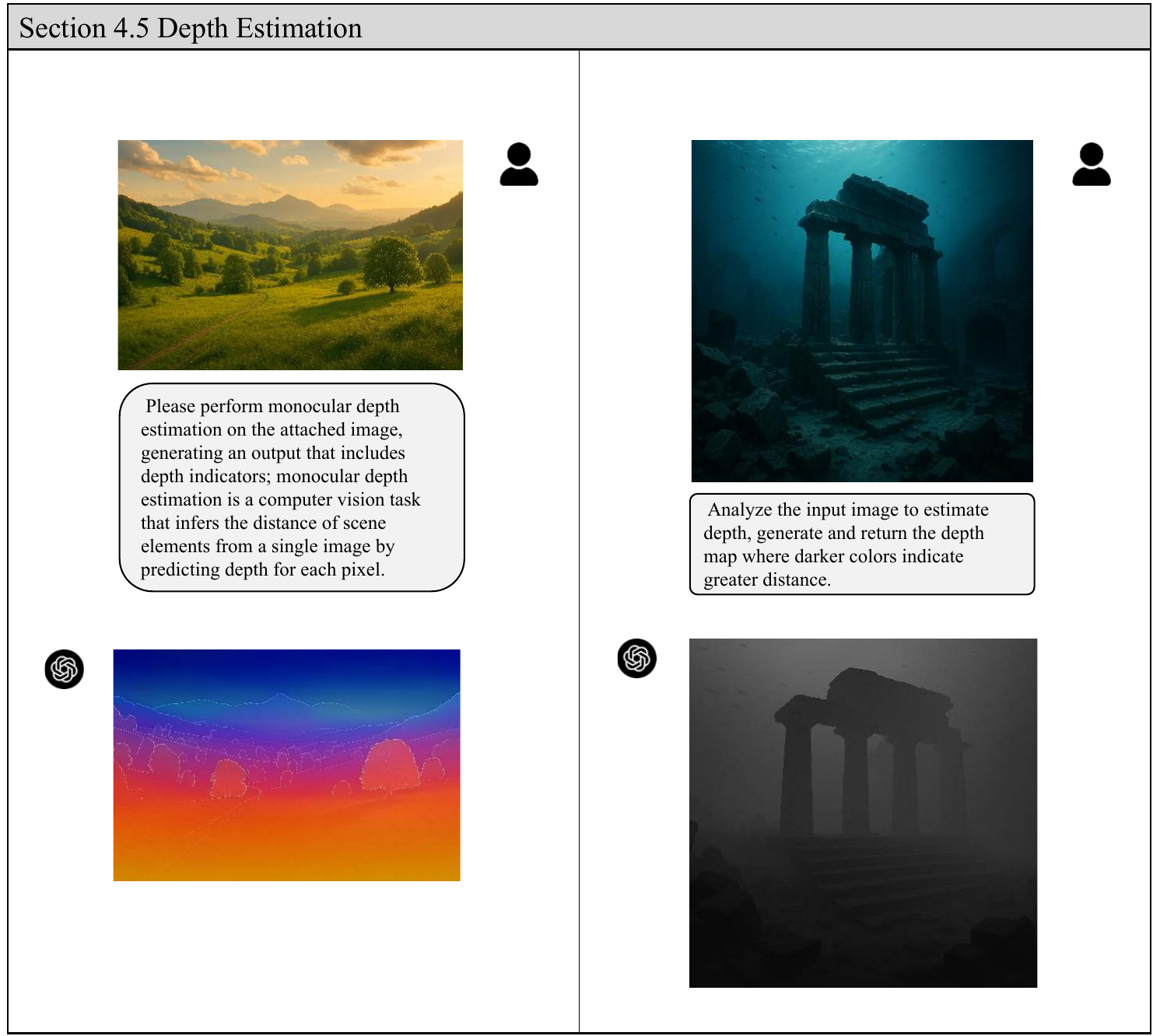}
    \caption[Sec~\ref{sec:depth}: Depth Estimation]{}
    \label{fig:depth2}
\end{figure}

\begin{figure}[h]
    \centering
    \includegraphics[width=1.0\linewidth]{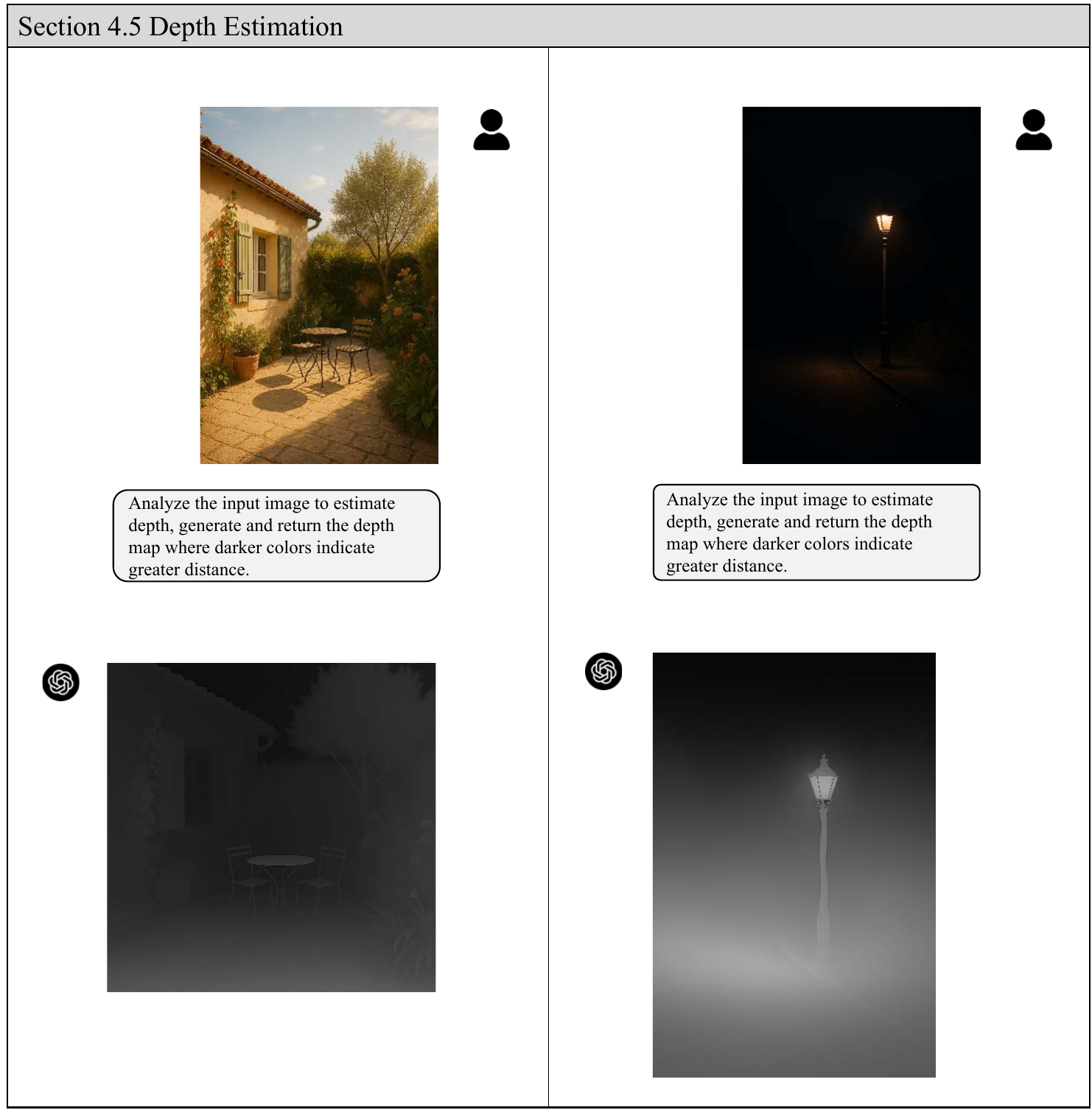}
    \caption[Sec~\ref{sec:depth}: Depth Estimation]{}
    \label{fig:depth3}
\end{figure}

As illustrated in Fig.~\ref{fig:depth1}–\ref{fig:depth3}, \modelname demonstrates a reasonable ability to estimate depth from monocular input. In general, the generated depth maps preserve the overall layout and object boundaries. However, we still observe some inconsistencies in fine details. For example, in Fig.~\ref{fig:depth1}, the bush on the right is inaccurately estimated, showing artifacts and depth discontinuities. 
Moreover, \modelname appears to be sensitive to lighting conditions and global appearance cues. In Fig.~\ref{fig:depth3}, the model misinterprets the depth due to lighting variations and shadows, leading to a less accurate estimation of surface geometry.

While \modelname performs moderately well in capturing general spatial layout, its depth predictions still lack robustness under complex lighting or detailed geometry. Future improvements could focus on enhancing geometry fidelity and reducing sensitivity to photometric factors.

\clearpage

\clearpage

\subsection{Surface Normal Estimation}
\label{sec:normal}

Surface normal estimation aims to infer the per-pixel orientation of surfaces in a 3D scene\cite{qi2018geonet,wang2015designing,bae2024rethinking}, which is essential for understanding object geometry, shape, and spatial layout. In this experiment, we provide a single polarization image as input and ask \modelname to generate a dense surface normal map.

\begin{figure}[h]
    \centering
    \includegraphics[width=1.0\linewidth]{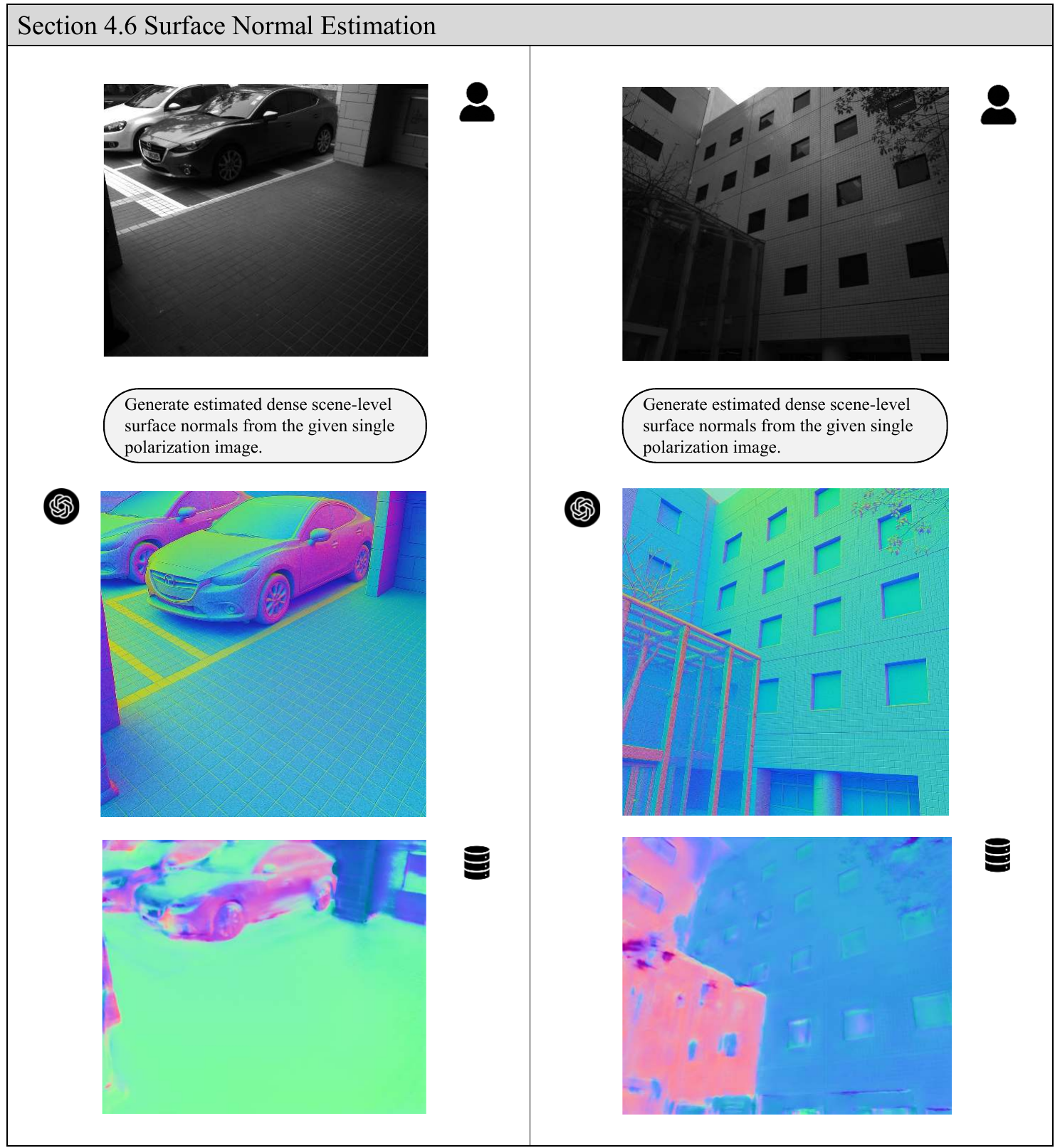}
    \caption[Sec~\ref{sec:normal}: Surface Normal Estimation]{}
    \label{fig:normal}
\end{figure}

As shown in Fig.~\ref{fig:normal}, \modelname is able to produce visually plausible surface normal maps that follow general structural cues. However, there are still noticeable issues in geometry consistency. For example, in the left column, the model incorrectly estimates the surface normal of the ground’s lane markings, despite them being part of the same plane. The vehicle’s angular surface orientation and the wall thickness are also misaligned with the input image.
Similarly, while the right example appears convincing at first glance, it fails to preserve key structural details such as the correct number of windows. This suggests the model struggles with fine-grained spatial geometry and may hallucinate details that do not exist in the input.

Overall, the results indicate that while \modelname can generate surface normal estimations with a reasonable appearance, it lacks precise geometric fidelity and often deviates from the spatial layout of the original image.

\clearpage

\clearpage

\subsection{Optical Flow Estimation}
\label{sec:optical}
Optical flow estimation aims to predict the per-pixel motion between consecutive frames in a video, often visualized as a dense vector field using hue and brightness to encode direction and magnitude\cite{ilg2017flownet,zhai2021optical,ranjan2017optical,hui2018liteflownet}. This task evaluates whether a model can perceive temporal continuity, understand object motion, and differentiate between static background and moving objects.

\begin{figure}[h]
    \centering
    \includegraphics[width=1.0\linewidth]{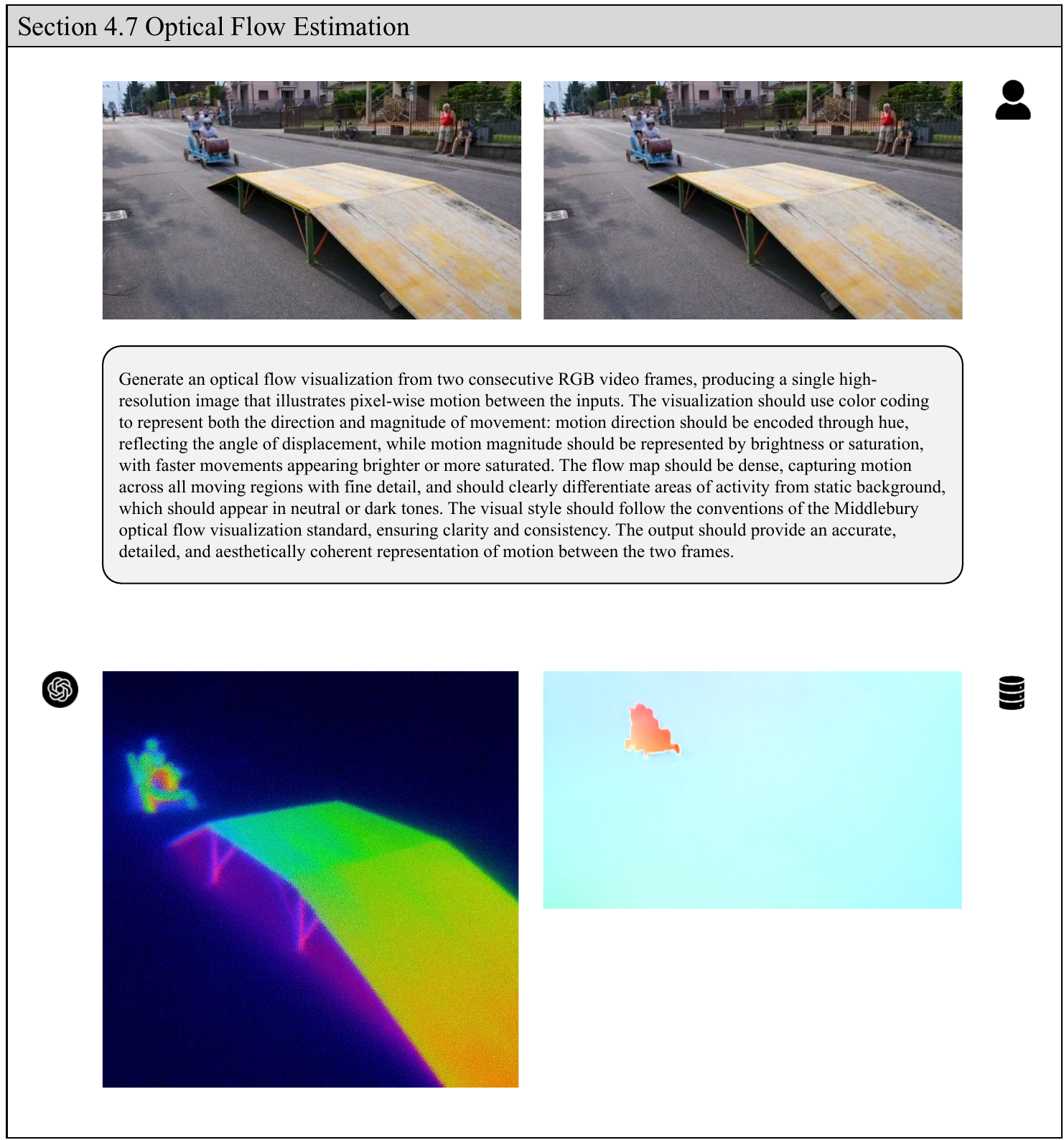}
    \caption[Sec~\ref{sec:optical}: Optical Flow Estimation]{}
    \label{fig:optical}
\end{figure}

\begin{figure}[h]
    \centering
    \includegraphics[width=1.0\linewidth]{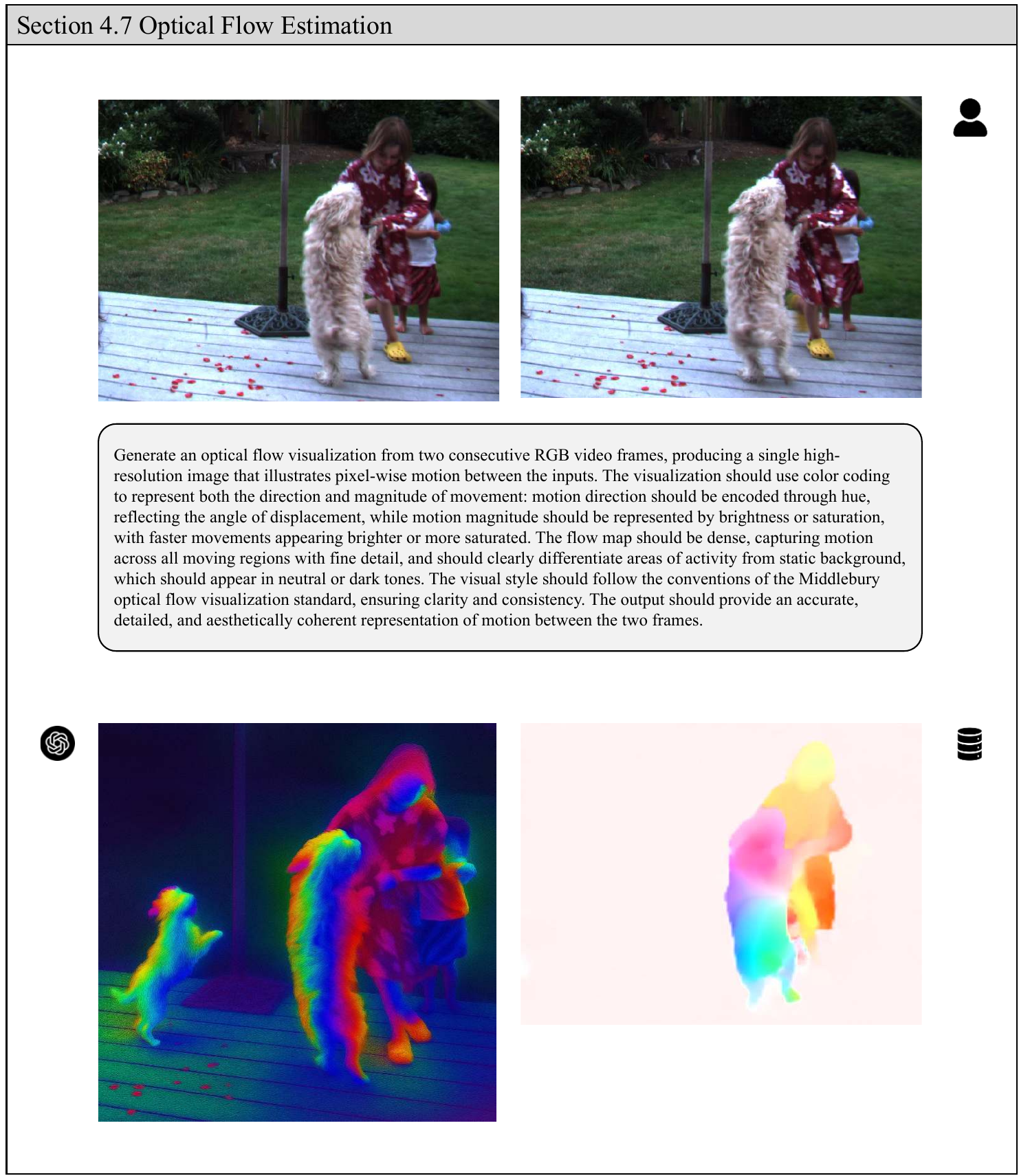}
    \caption[Sec~\ref{sec:optical}: Optical Flow Estimation]{}
    \label{fig:optical2}
\end{figure}

We test \modelname on this task using pairs of consecutive frames and evaluate its ability to produce dense, color-coded flow maps that conform to standard visualization conventions. As shown in Fig.~\ref{fig:optical}, while the model captures the general direction of movement, it introduces hallucinated motion structures—in this case, generating an additional ramp flow not aligned with the ground truth displacement.
In Fig.~\ref{fig:optical2}, the model demonstrates improved perception of human and animal motion. However, it also synthesizes a non-existent dog-like structure in the flow map, revealing a tendency to infer motion beyond the input evidence. This suggests a reliance on visual priors over strict frame-to-frame correlation.

Overall, \modelname shows potential in optical flow estimation but lacks precision in adhering to actual inter-frame motion, often introducing artifacts and inconsistencies in the predicted motion fields.

\clearpage

\clearpage

\subsection{Remote Sensing Change Detection}
\label{sec:rschange}
Remote sensing change detection is a specialized task that requires identifying and localizing changes between two satellite or aerial images captured at different times\cite{asokan2019change,jensen2007remote}. This involves semantic understanding of man-made and natural structures, and distinguishing alterations such as new constructions, deforestation, flooding, or land-use transformation.

\begin{figure}[h]
    \centering
    \includegraphics[width=1.0\linewidth]{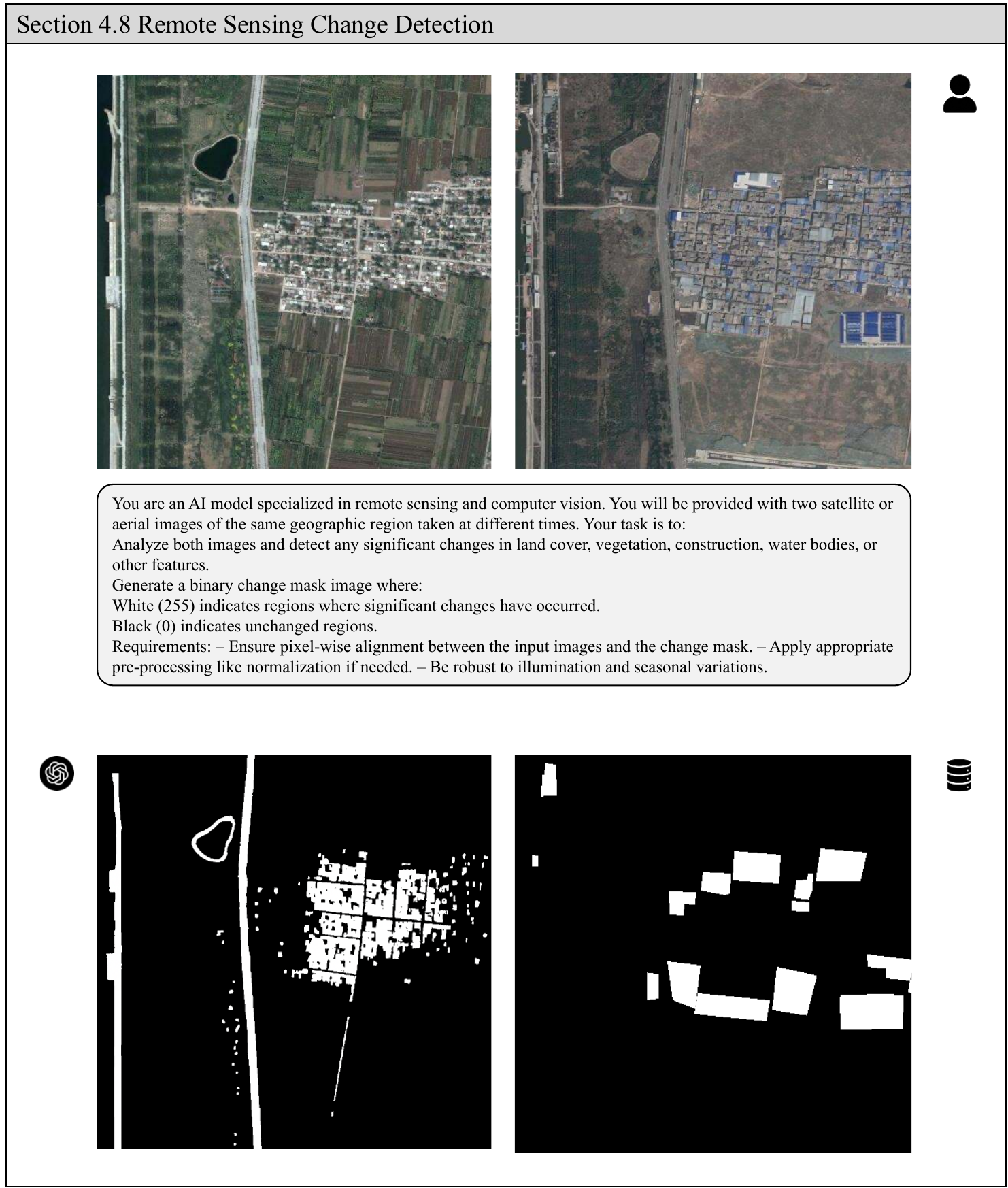}
    \caption[Sec~\ref{sec:rschange}: Remote Sensing Change Detection]{}
    \label{fig:rschange}
\end{figure}

\begin{figure}[h]
    \centering
    \includegraphics[width=1.0\linewidth]{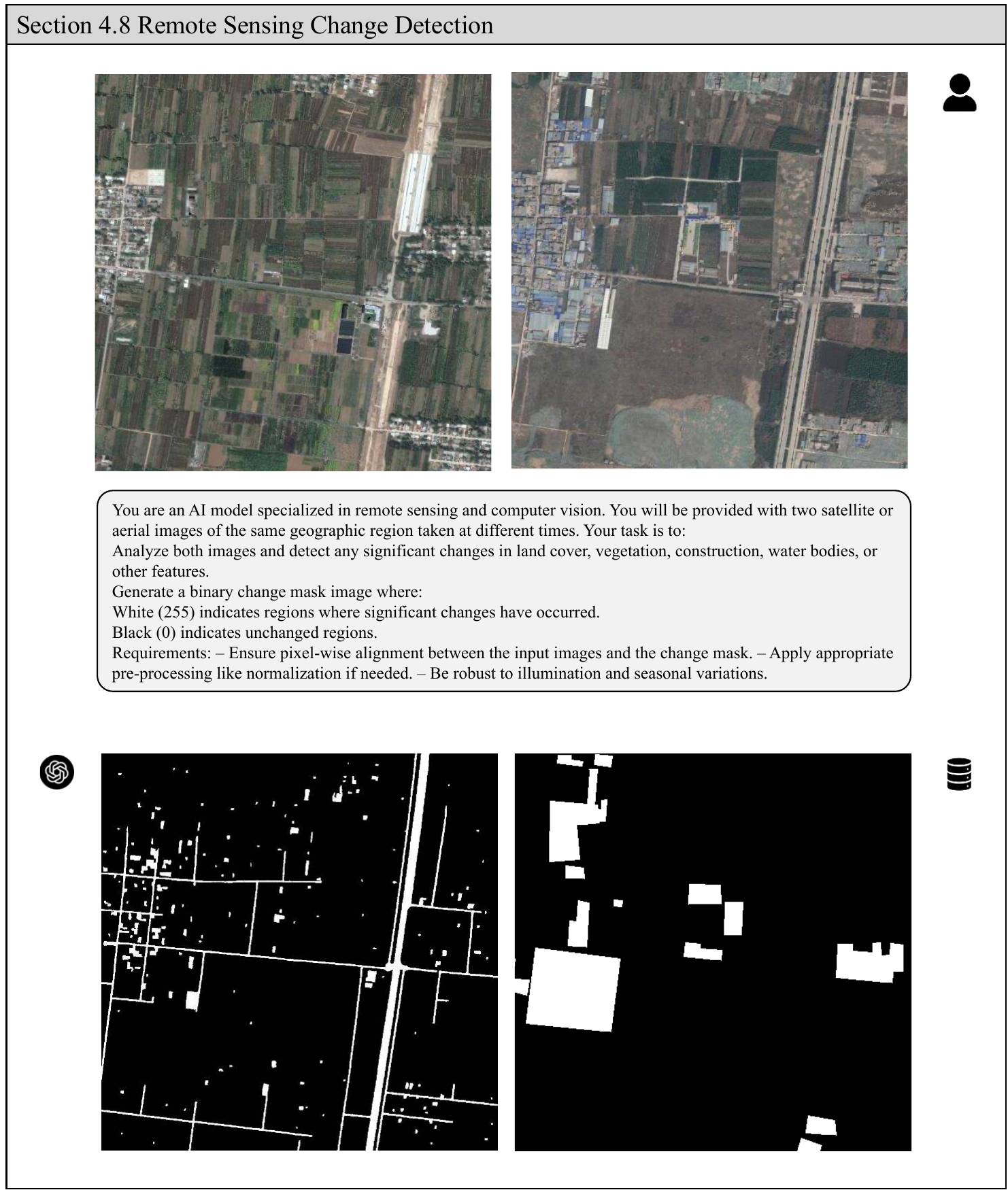}
    \caption[Sec~\ref{sec:rschange}: Remote Sensing Change Detection]{}
    \label{fig:rschange2}
\end{figure}

In our evaluation, we input two temporally distinct remote sensing images and prompt \modelname to generate a change map that highlights the differences. However, as illustrated in Fig.~\ref{fig:rschange} and Fig.~\ref{fig:rschange2}, the model fails to capture the essence of the task. Instead of identifying changes between the two images, the model merely segments out features like roads and buildings from one of the inputs.

This indicates that \modelname has not grasped the concept of temporal comparison or semantic difference analysis required for change detection. The results suggest that either the model lacks task-specific priors or is overly biased towards static semantic segmentation, treating the input more like a single-image understanding task than a comparative one.

Overall, \modelname currently exhibits limited capability in remote sensing change detection and requires further task-aware instruction alignment to generate meaningful temporal comparisons.

\clearpage

\clearpage
\section{Knowledge-based Image Generation}
\label{sec:knowledge}

\subsection{Physics}
\label{sec:physical}
Physics understanding in image generation evaluates the model's ability to reflect fundamental physical laws and commonsense in visual content\cite{meng2024phybench,spitznagel2025physicsgen,liu2024physgen, lin2024phys4dgen, liu2025generative}. This capability is essential for generating plausible and scientifically reasonable scenes, especially in educational or simulation-related applications.
In this section, we follow the taxonomy defined in PhyBench\cite{meng2024phybench}, which categorizes physical knowledge into several sub-fields, including force, optics, thermodynamics, and material properties.

\begin{figure}[h]
    \centering
    \includegraphics[width=1.0\linewidth]{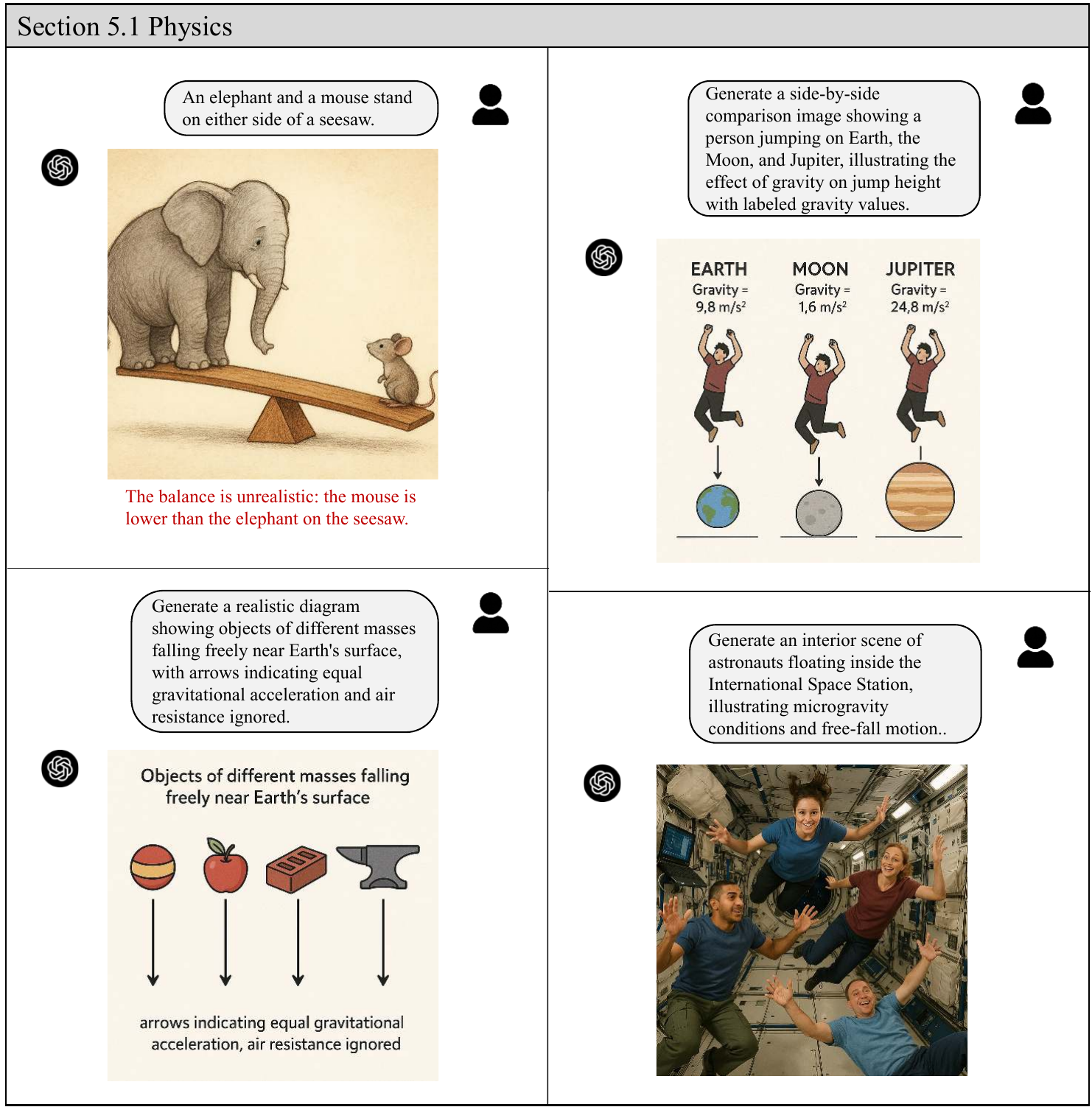}
    \caption[Sec~\ref{sec:physical}: Force]{Examples of force effect results generated by \modelname.}
    \label{fig:force1}
\end{figure}

\begin{figure}[h]
    \centering
    \includegraphics[width=1.0\linewidth]{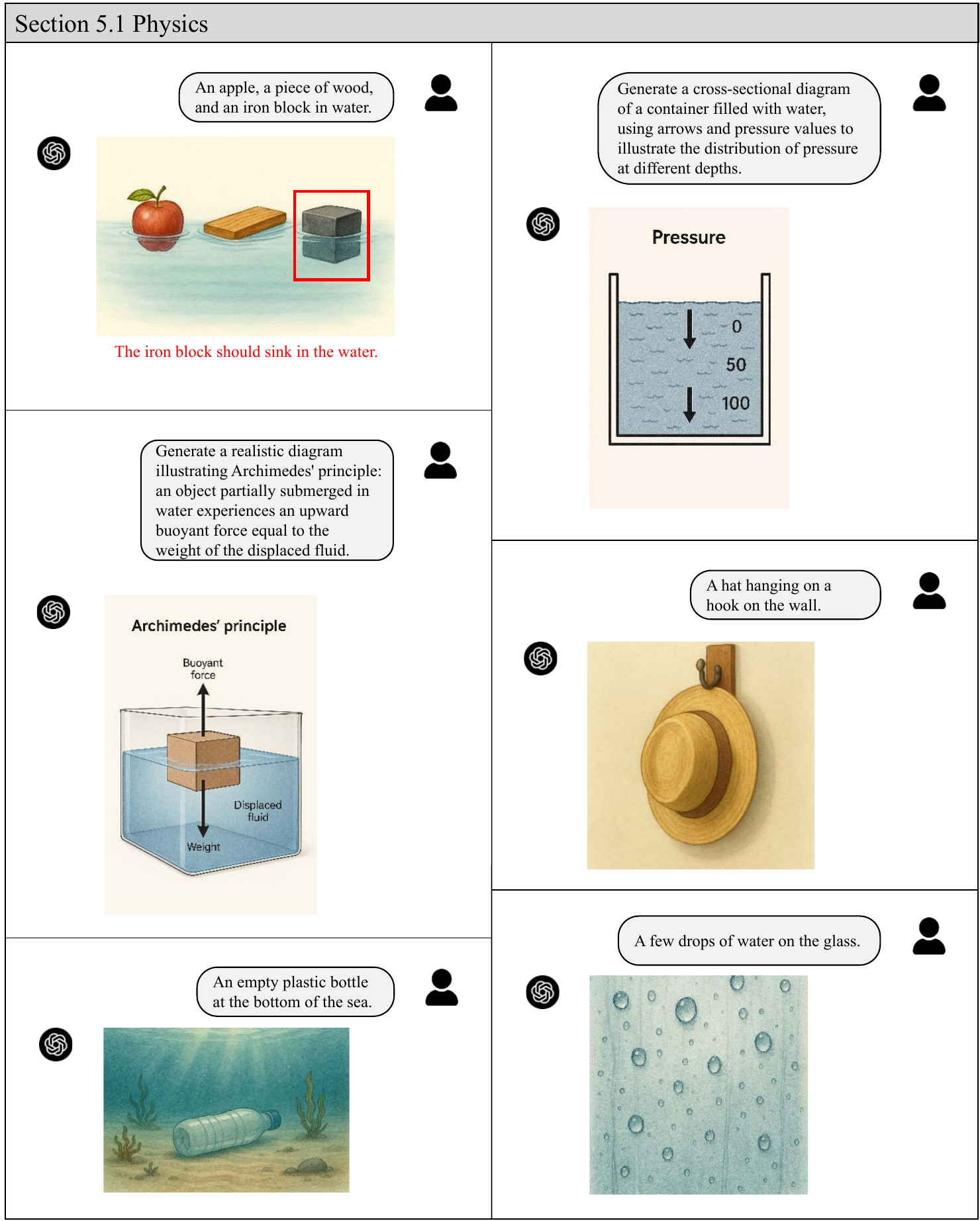}
    \caption[Sec~\ref{sec:physical}: Force]{Examples of force effect results generated by \modelname.}
    \label{fig:force2}
\end{figure}

\begin{figure}[h]
    \centering
    \includegraphics[width=1.0\linewidth]{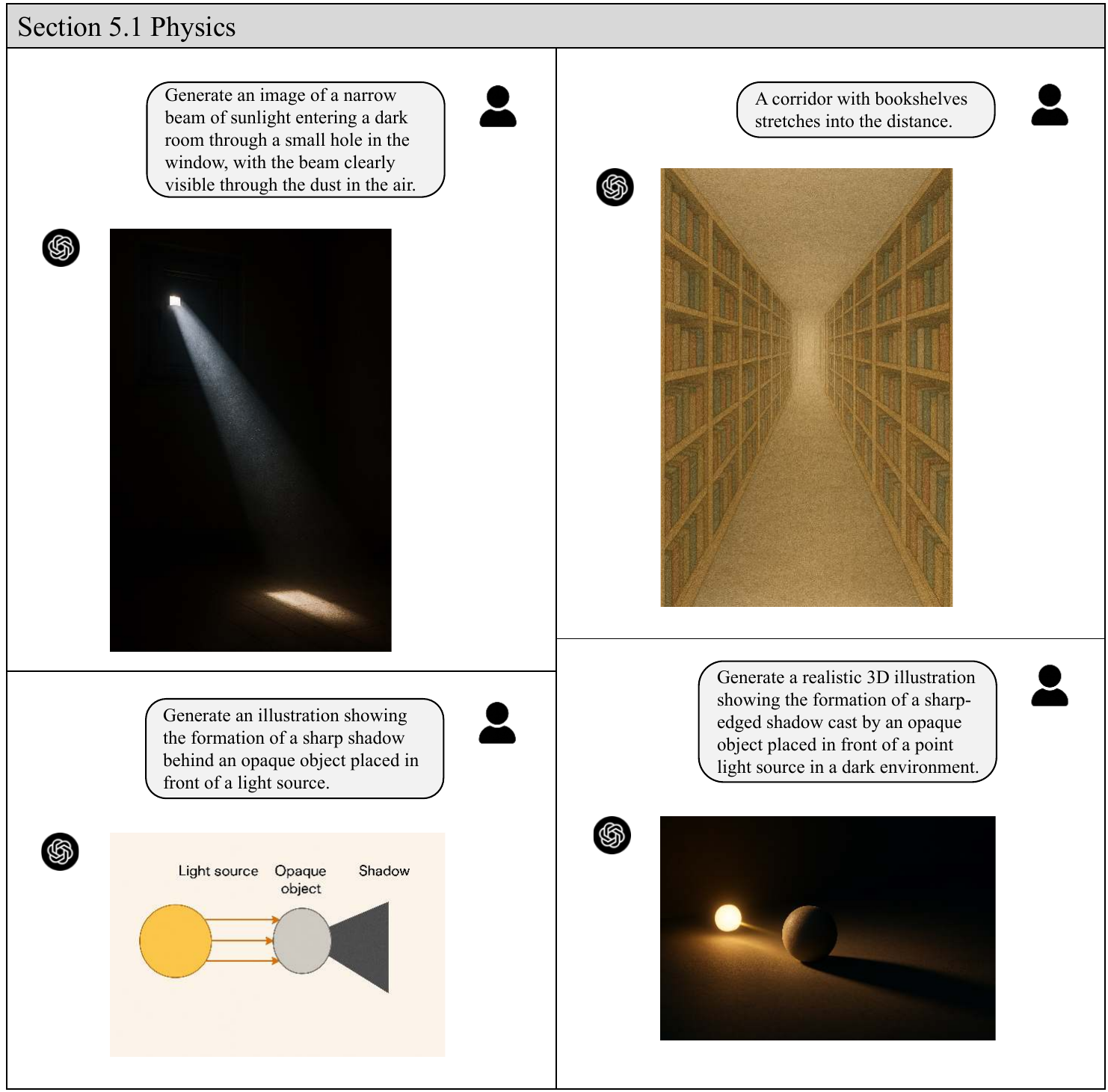}
    \caption[Sec~\ref{sec:physical}: Optics]{Examples of optical results generation by \modelname.}
    \label{fig:optical1}
\end{figure}

\begin{figure}[h]
    \centering
    \includegraphics[width=1.0\linewidth]{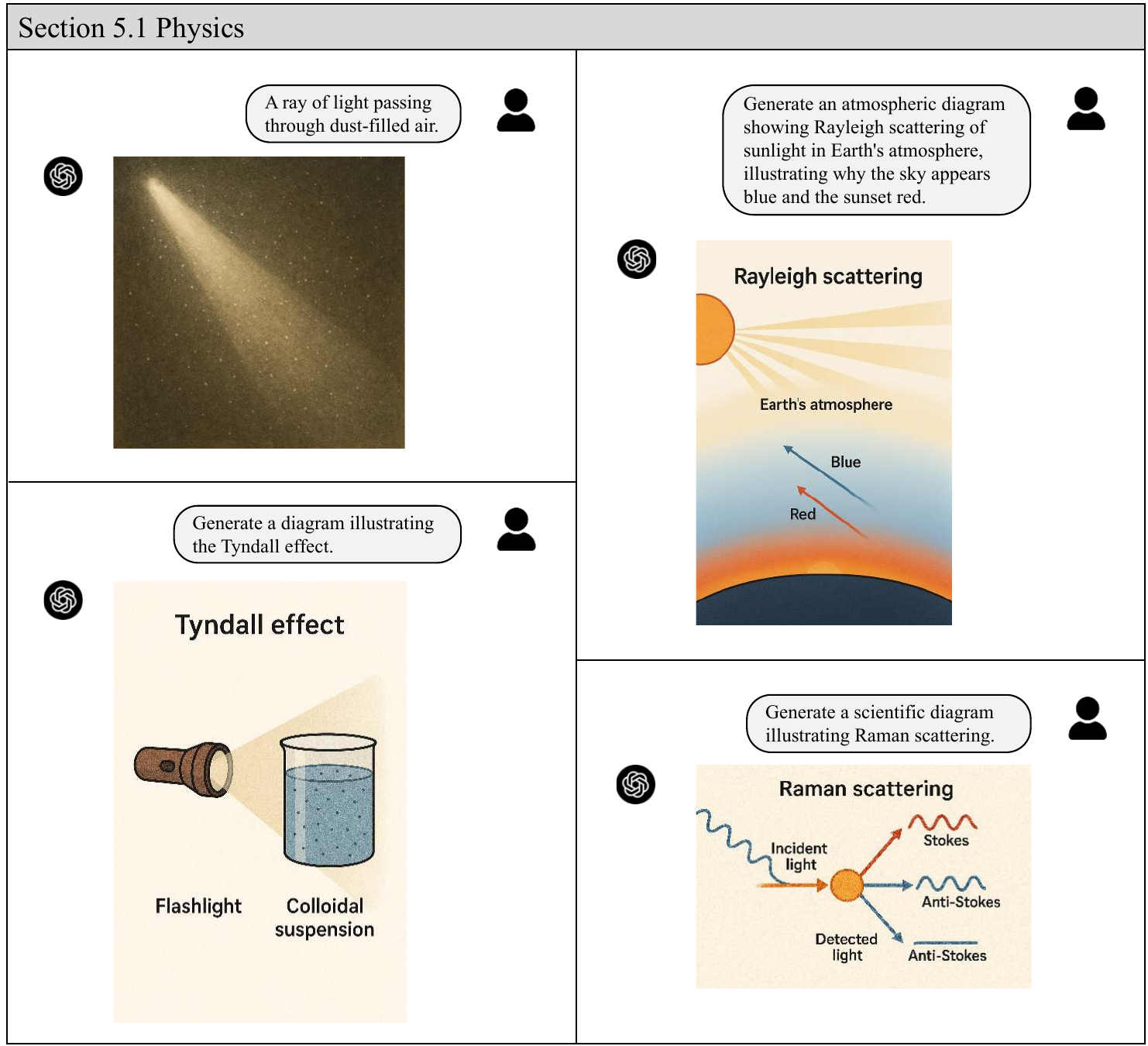}
    \caption[Sec~\ref{sec:physical}: Optics]{Examples of optical results generation by \modelname.}
    \label{fig:optics2}
\end{figure}

\begin{figure}[h]
    \centering
    \includegraphics[width=1.0\linewidth]{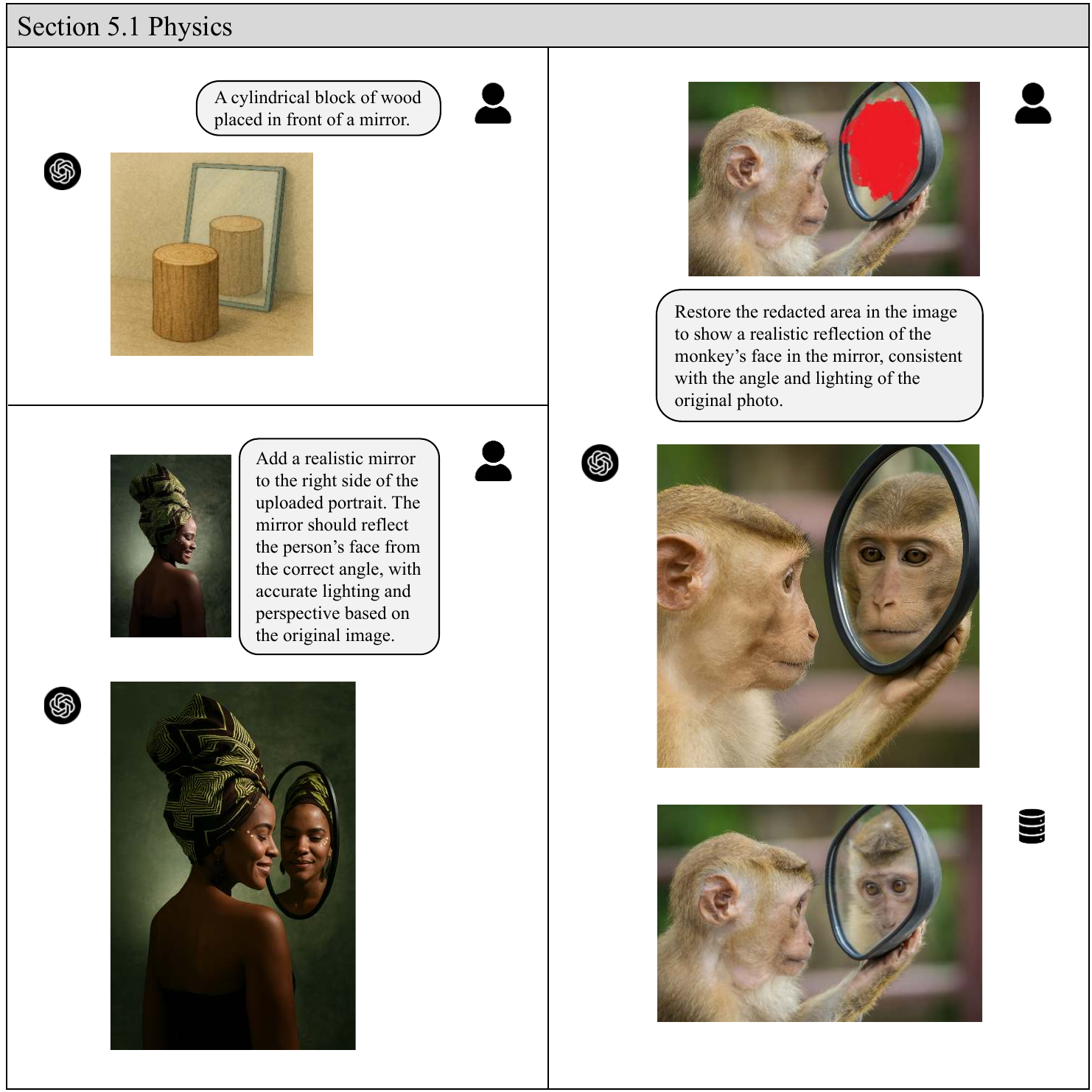}
    \caption[Sec~\ref{sec:physical}: Optics]{Examples of optical results generated by \modelname.}
    \label{fig:optics3}
\end{figure}

\begin{figure}[h]
    \centering
    \includegraphics[width=1.0\linewidth]{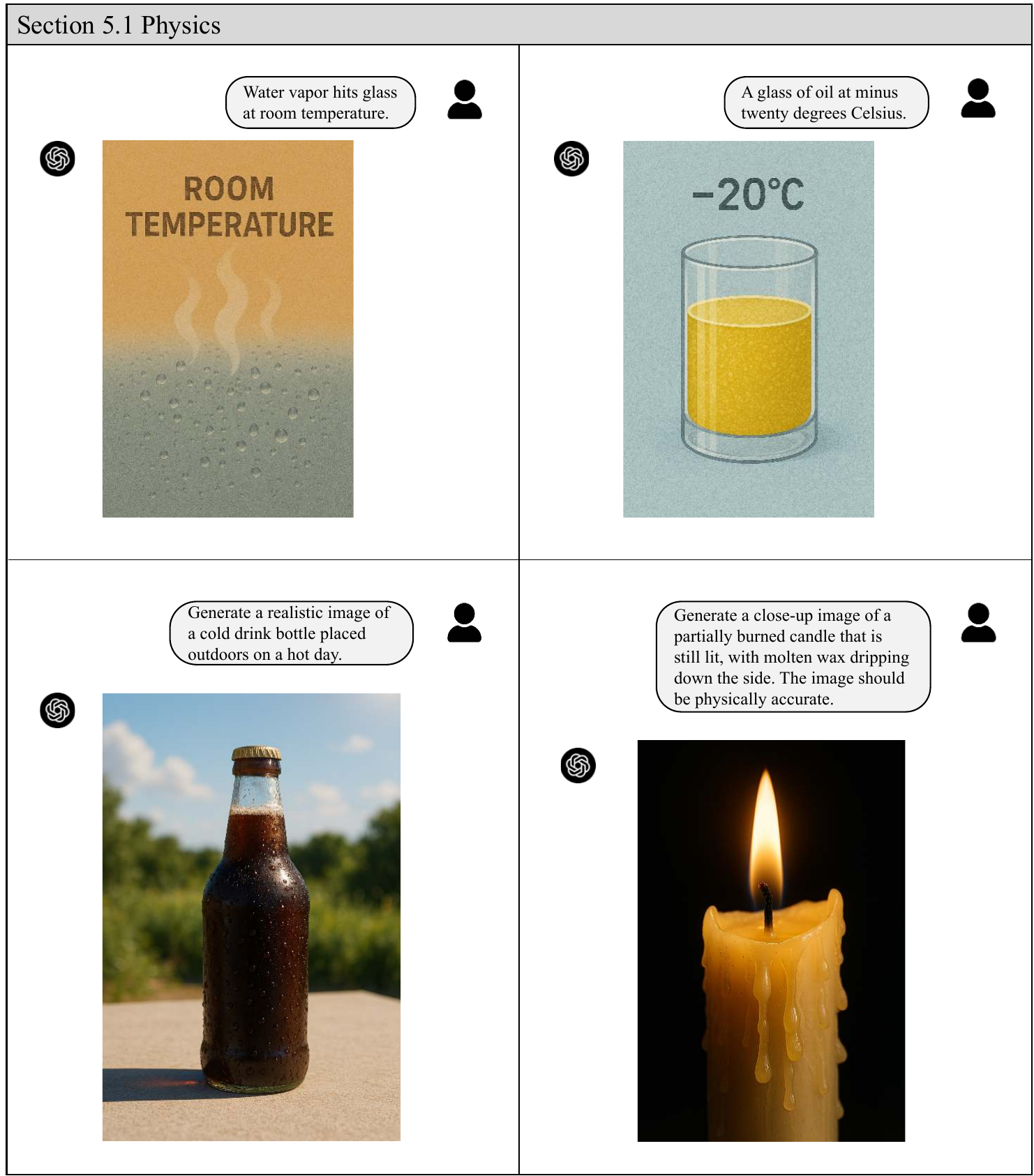}
    \caption[Sec~\ref{sec:physical}: Thermodynamics]{Examples of thermodynamic effect generated by \modelname.}
    \label{fig:heat1}
\end{figure}

\begin{figure}[h]
    \centering
    \includegraphics[width=1.0\linewidth]{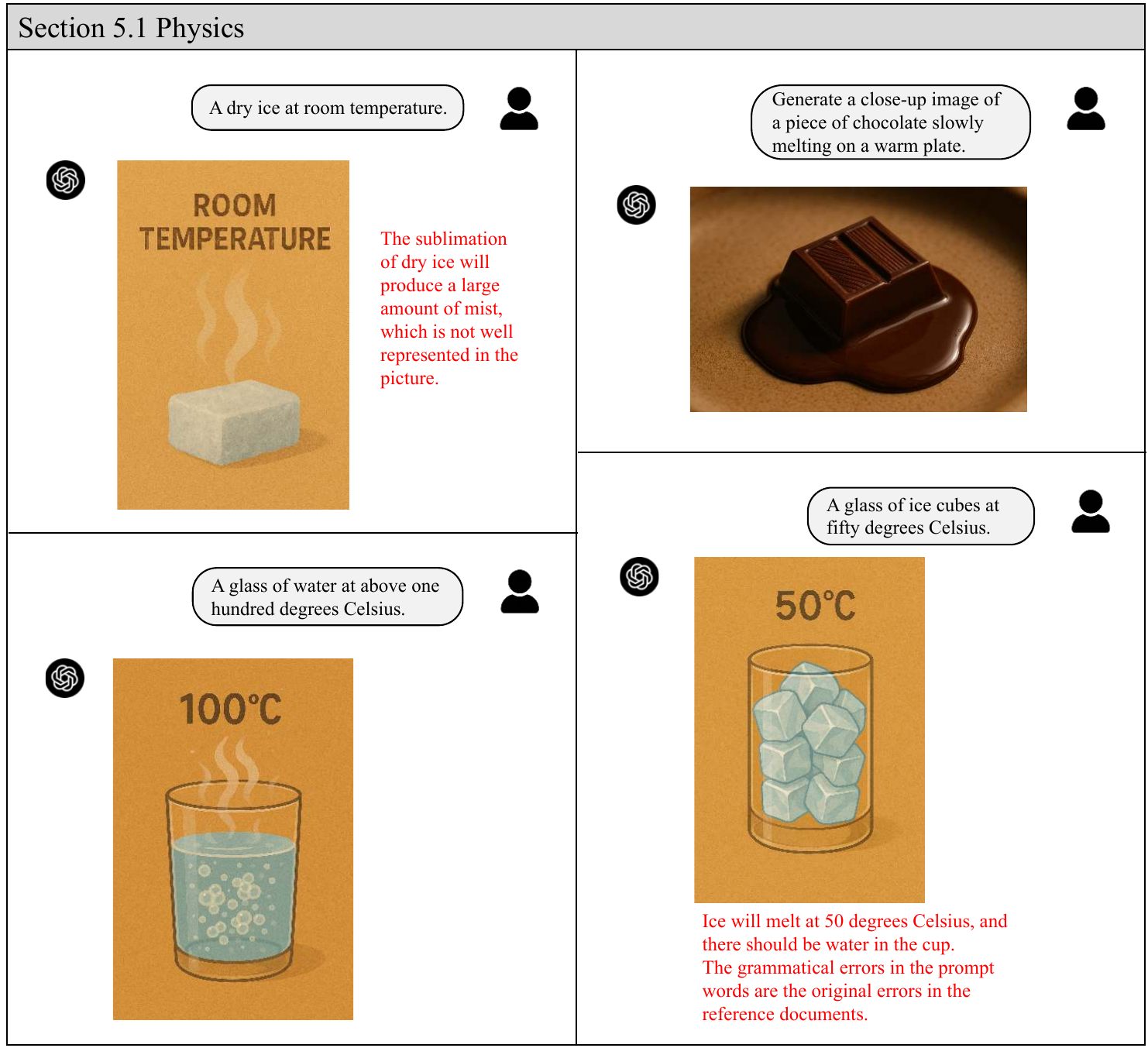}
    \caption[Sec~\ref{sec:physical}: Thermodynamics]{Examples of thermodynamic effect generated by \modelname.}
    \label{fig:heat2}
\end{figure}

\begin{figure}[h]
    \centering
    \includegraphics[width=1.0\linewidth]{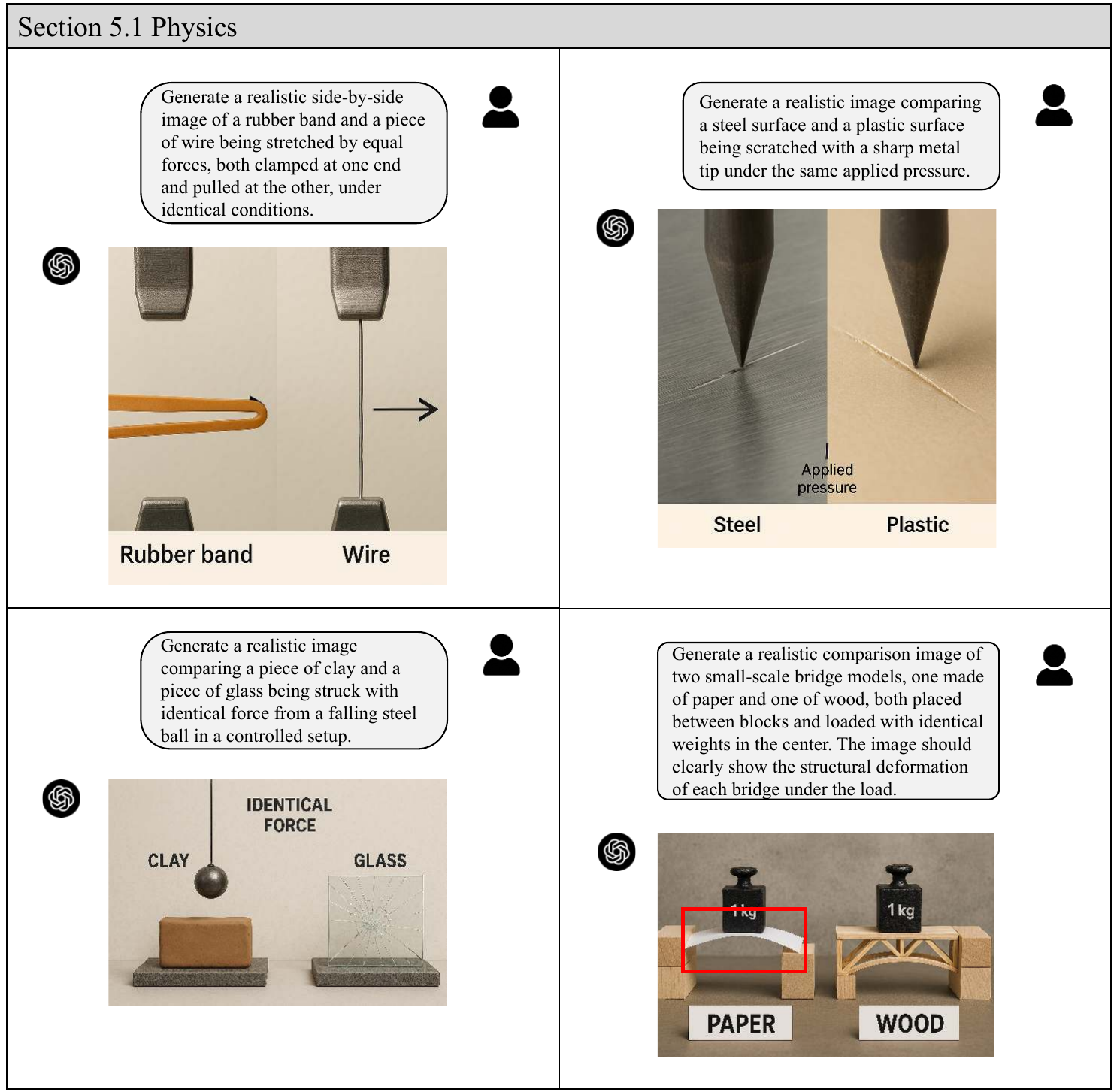}
    \caption[Sec~\ref{sec:physical}: Material]{Examples of material properties generated by \modelname.}
    \label{fig:material1}
\end{figure}

\begin{figure}[h]
    \centering
    \includegraphics[width=1.0\linewidth]{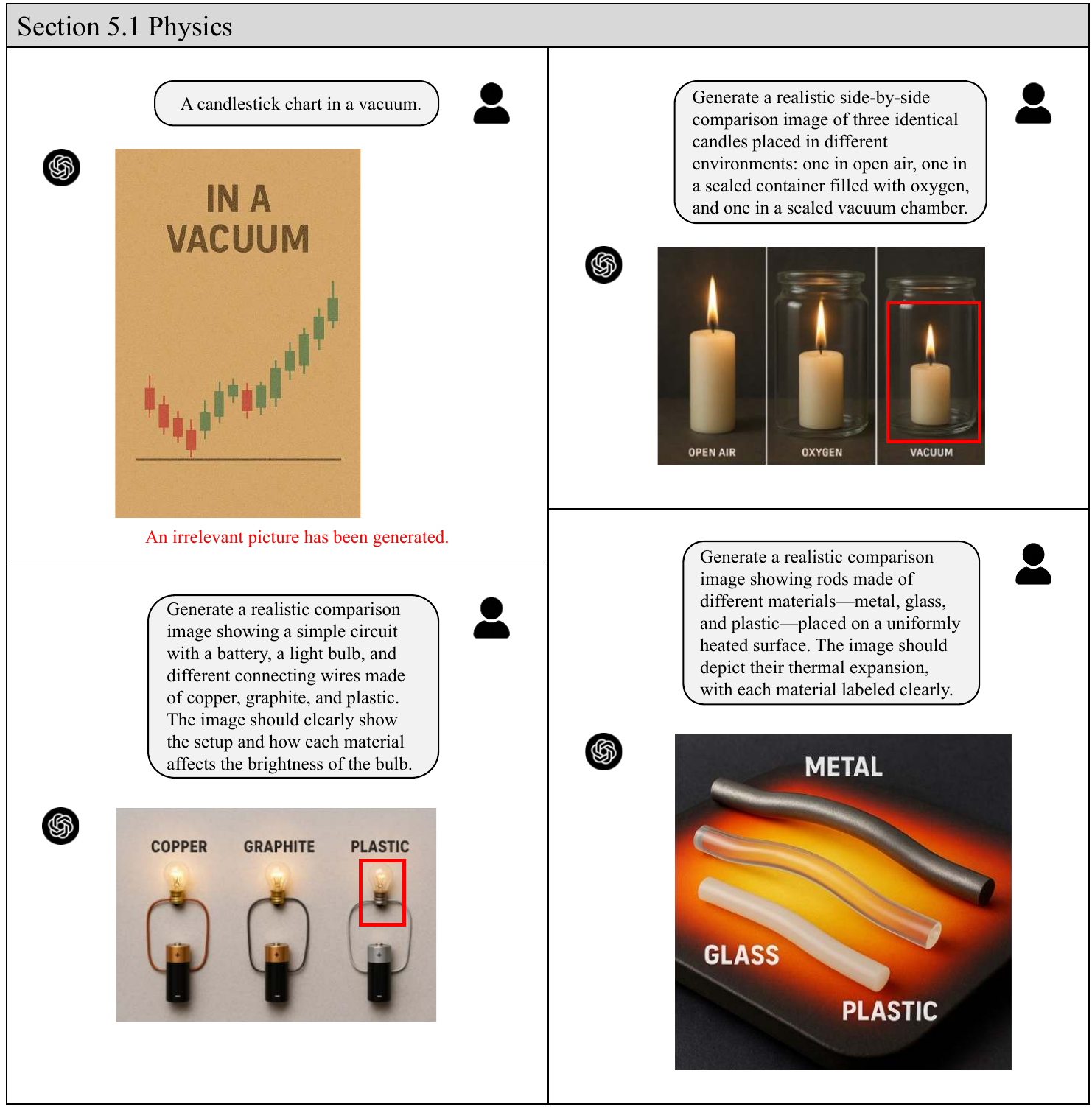}
    \caption[Sec~\ref{sec:physical}: Material]{Examples of material properties generated by \modelname.}
    \label{fig:material2}
\end{figure}

As shown in Fig.~\ref{fig:force1}-\ref{fig:material2}, we evaluate \modelname across these categories. Overall, we find that \modelname demonstrates good performance in most physical commonsense scenarios, generating results that align well with real-world physics. However, we also observe some failure cases. In particular, the model fails to correctly handle light refraction in Fig.~\ref{fig:optics3}, and the thermodynamics-related generation in Fig.~\ref{fig:heat2} exhibits incorrect visual patterns inconsistent with physical laws.

This highlights the potential and limitations of current multi-modal models in understanding and applying physical knowledge through image generation.

\clearpage

\clearpage

\subsection{Chemistry}
\label{sec:chemical}
Chemistry understanding in image generation evaluates the model's ability to correctly reflect chemical knowledge, molecular structures, laboratory scenes, and reaction processes in visual content\cite{liu2024git, qian2023molscribe}.

\begin{figure}[h]
    \centering
    \includegraphics[width=1.0\linewidth]{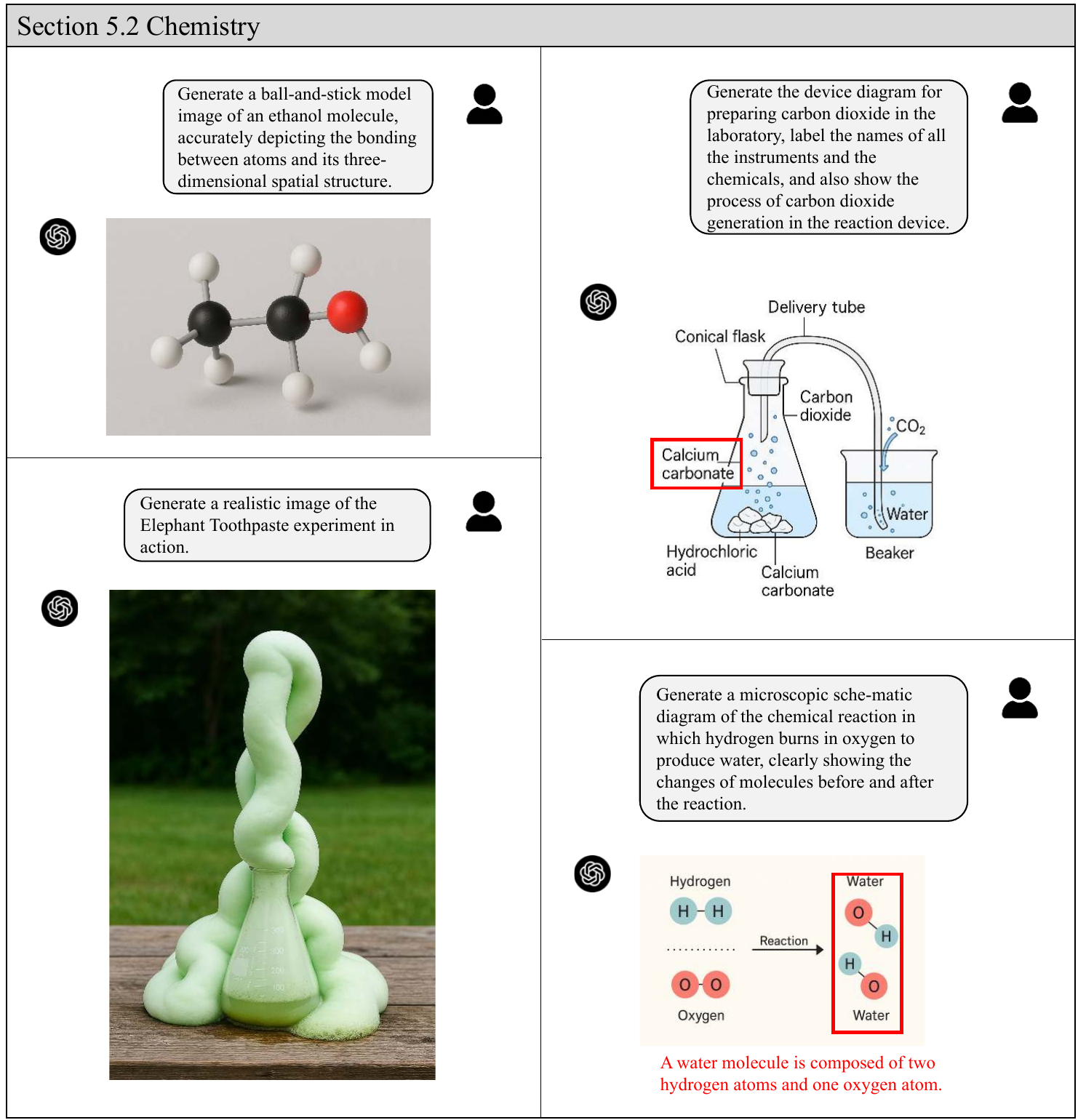}
    \caption[Sec~\ref{sec:chemical}: Chemistry]{Examples of chemistry-related generation results, including molecule models, reaction diagrams, and experiment phenomena.}
    \label{fig:chemistry}
\end{figure}

\begin{figure}[h]
    \centering
    \includegraphics[width=1.0\linewidth]{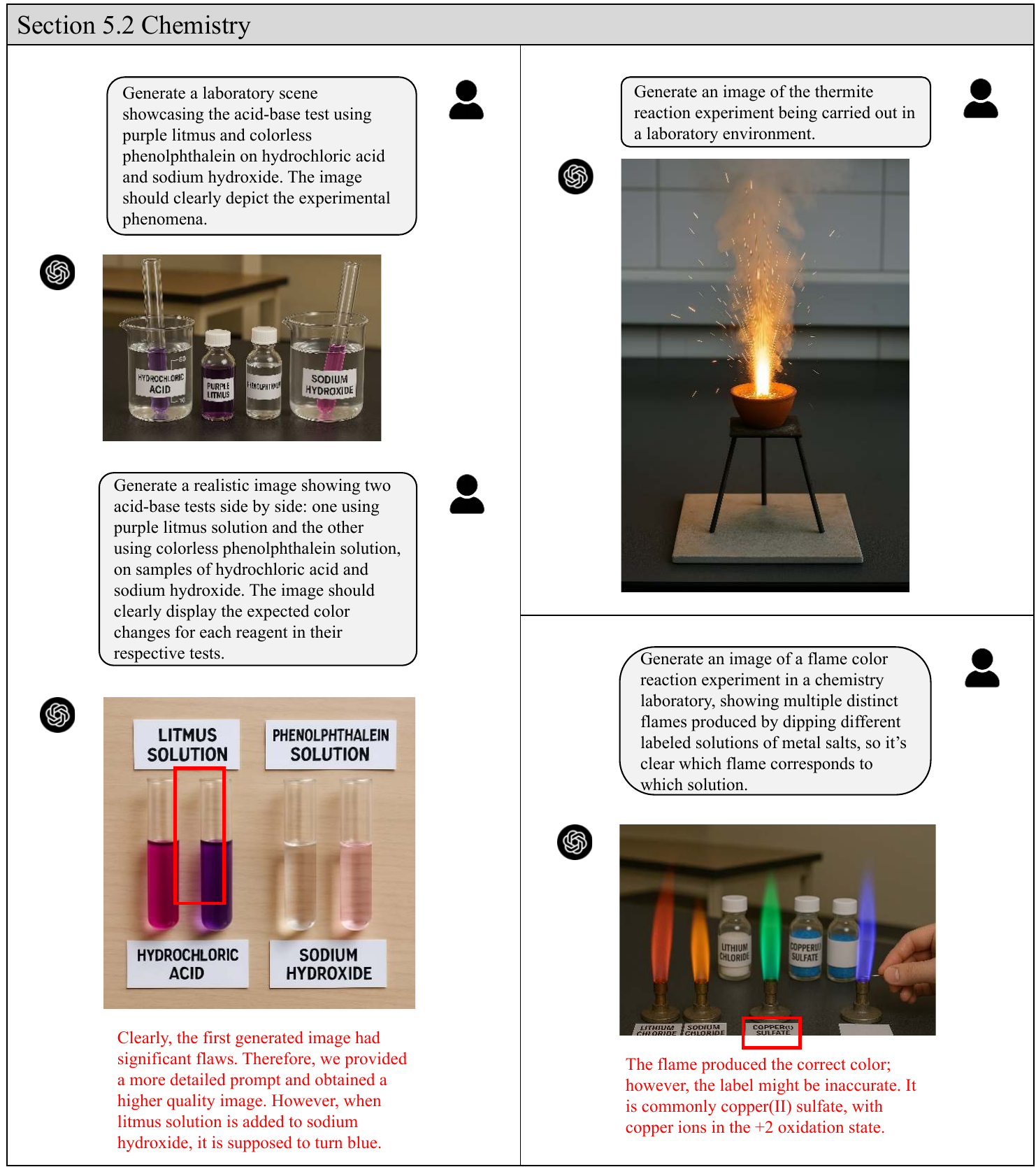}
    \caption[Sec~\ref{sec:chemical}: Chemistry]{Examples of laboratory scene generation for chemistry experiments, with several incorrect labels and inaccurate experiment phenomena.}
    \label{fig:chemistry1}
\end{figure}

\begin{figure}[h]
    \centering
    \includegraphics[width=1.0\linewidth]{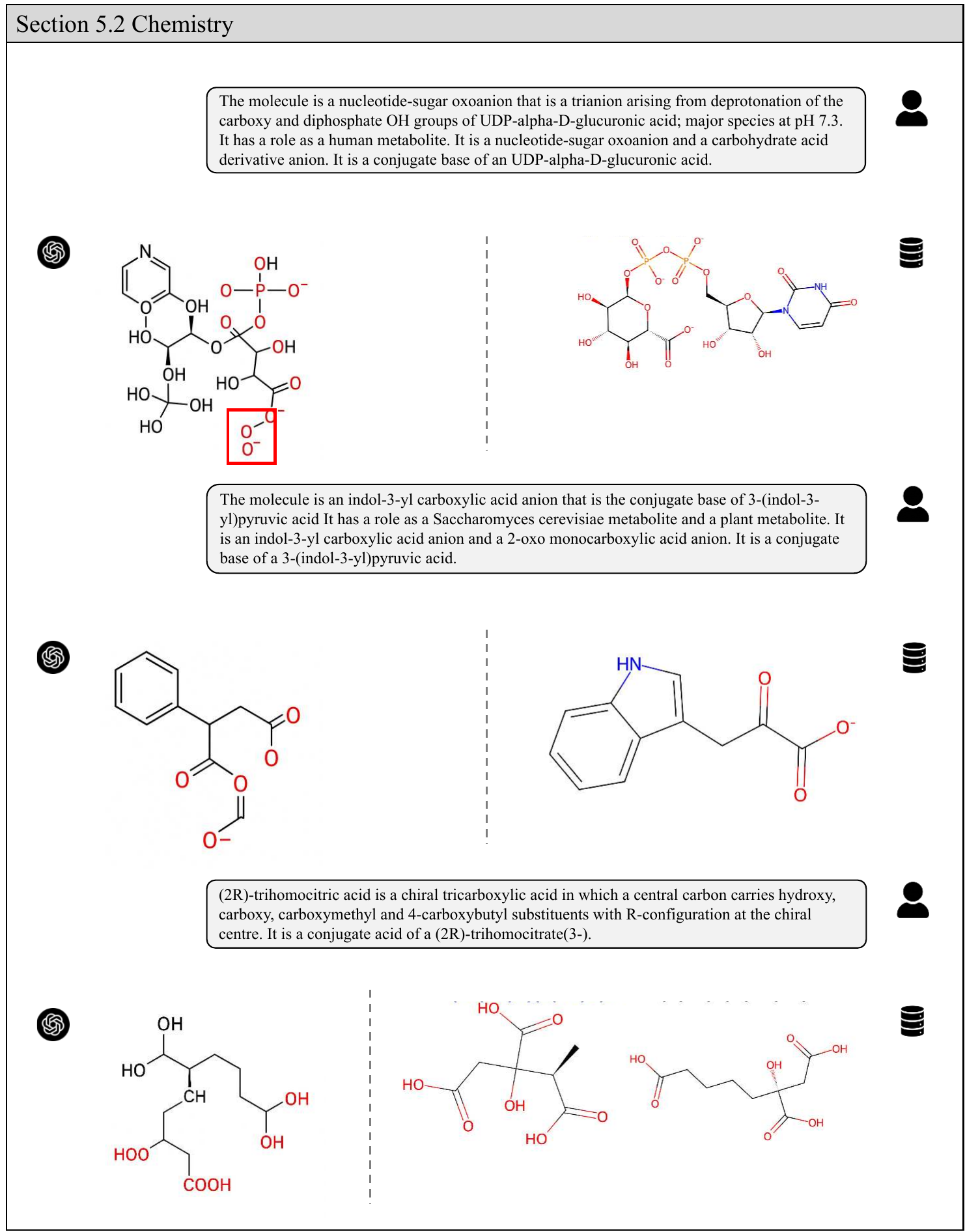}
    \caption[Sec~\ref{sec:chemical}: Chemistry]{Examples of chemical structure generation results, with some structural errors and misconnected molecular diagrams.}
    \label{fig:chemistry2}
\end{figure}

As shown in Fig.~\ref{fig:chemistry} to Fig.~\ref{fig:chemistry2}, we test \modelname across various chemistry-related tasks, including molecular structure generation, laboratory experiment scene generation, and chemical reaction diagram generation.

We find that while \modelname is capable of generating chemistry-related images with reasonable visual style and layout, there are significantly more errors compared to physics generation tasks. These errors mainly include incorrect labeling of chemical substances (e.g., wrong label for copper sulfate in Fig.~\ref{fig:chemistry1}), inaccurate experimental phenomena (e.g., wrong color reaction for litmus test in Fig.~\ref{fig:chemistry1}), and mistakes in structural molecular diagrams (e.g., missing or misconnected atoms in Fig.~\ref{fig:chemistry2}).

These results indicate that \modelname still lacks reliable domain-specific knowledge in the chemistry field, especially for tasks requiring accurate scientific details and strict visual correctness.

\clearpage

\clearpage

\subsection{Biology}
\label{sec:biology}

Biology-related image generation tasks evaluate the model's ability to understand biological knowledge and visually represent biological phenomena, structures, and experimental processes\cite{yilmaz2024cascaded, xing2024can,roy2024gan,elhamod2023discovering}. These tasks cover a wide range of biological concepts, including microbiology, botany, genetics, ecology, and cell biology.

\begin{figure}[h]
    \centering
    \includegraphics[width=1.0\linewidth]{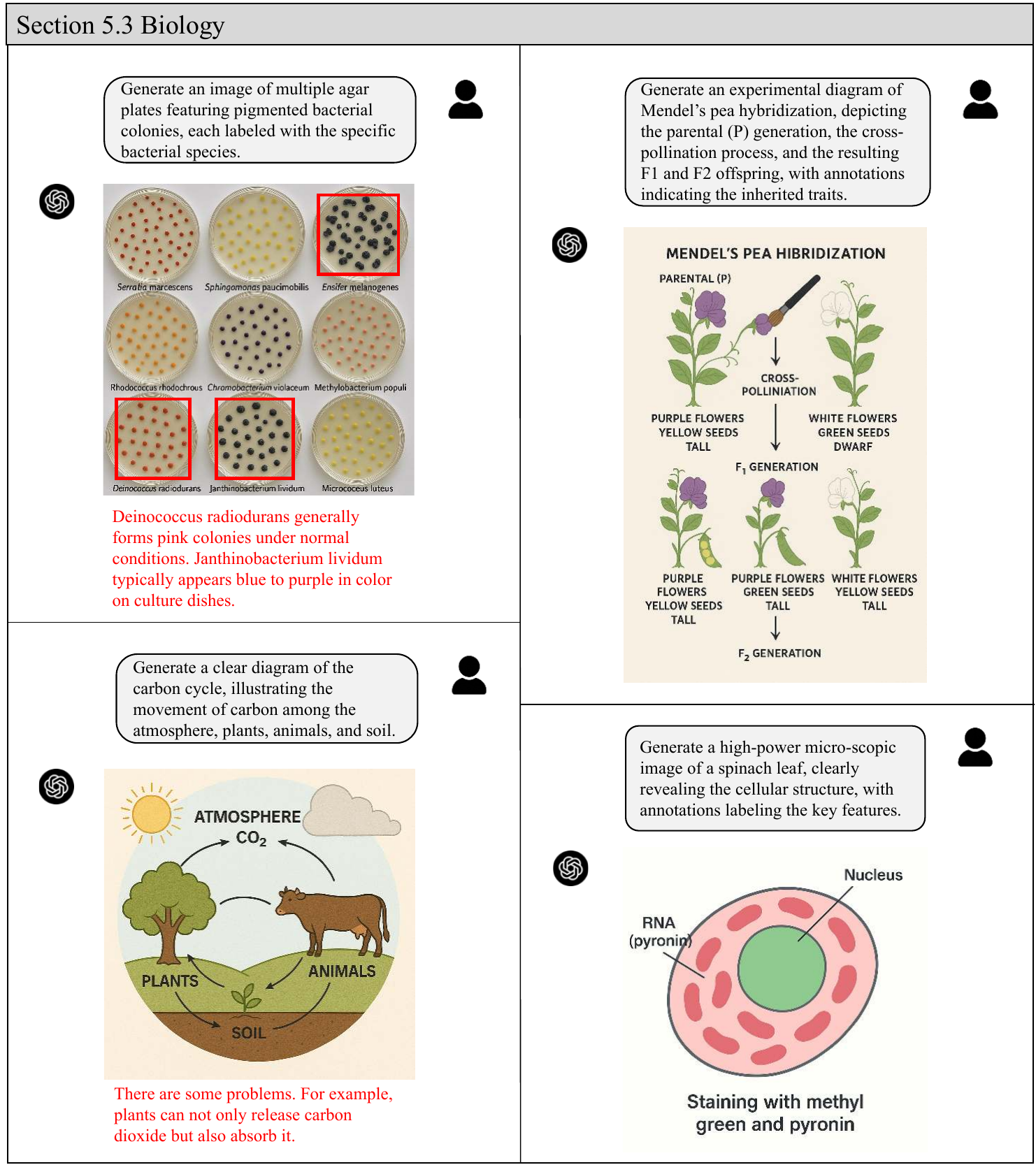}
\caption[Sec~\ref{sec:biology}: Biology]{Examples of biology-related image generation, including bacterial colonies, genetic diagrams, ecological cycles, and cellular structures.}
    \label{fig:biology}
\end{figure}

\begin{figure}[h]
    \centering
    \includegraphics[width=1.0\linewidth]{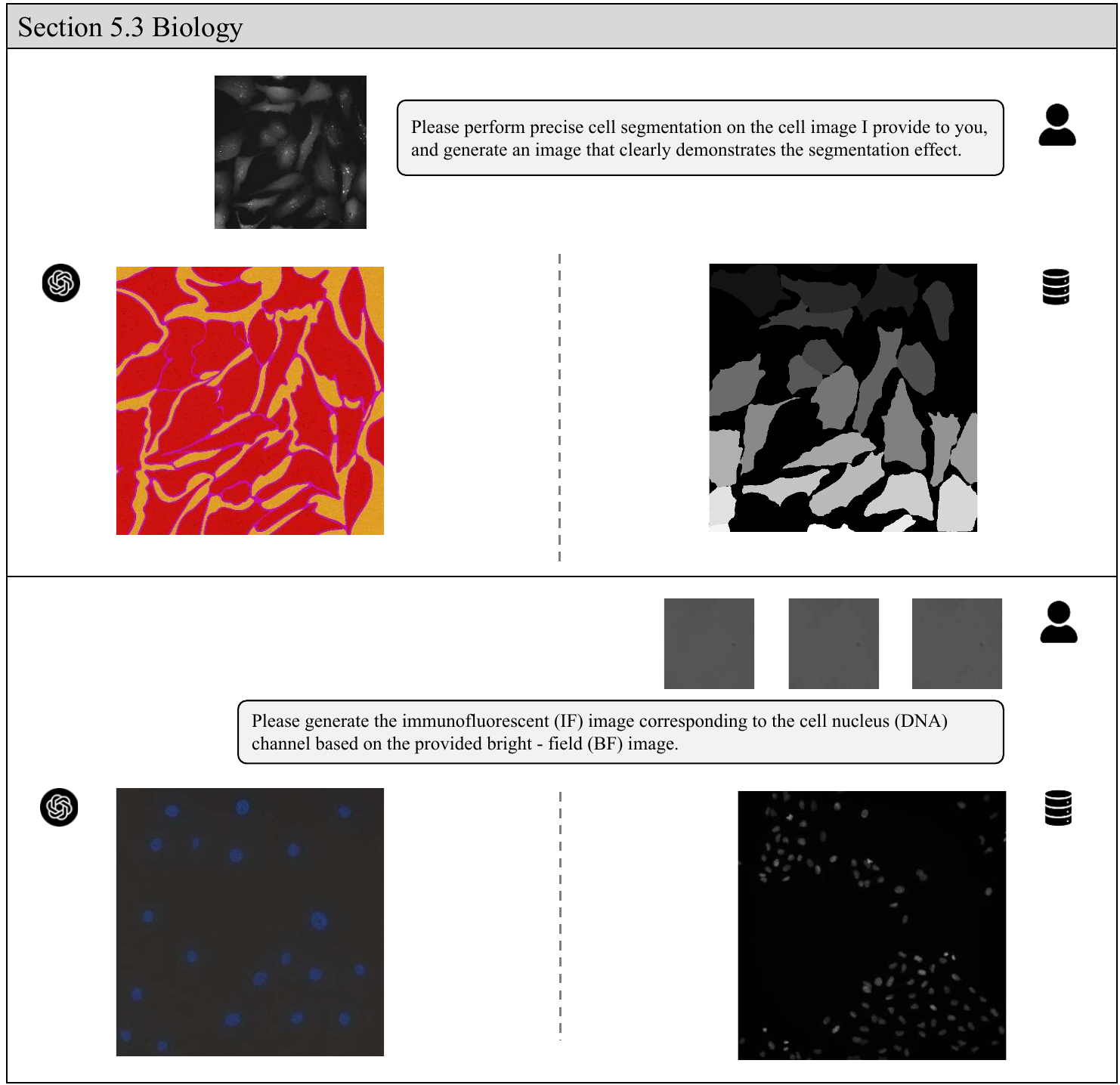}
\caption[Sec~\ref{sec:biology}: Biology]{Examples of biology image analysis tasks, including cell segmentation and immunofluorescence image generation.}

    \label{fig:biology1}
\end{figure}

As shown in Fig.~\ref{fig:biology} and Fig.~\ref{fig:biology1}, we test \modelname in various biology scenarios, such as generating labeled bacterial colony images, genetic experiment diagrams, ecological cycle illustrations, cellular structure generation, and biological image analysis tasks (e.g., cell segmentation, immunofluorescent image synthesis).

We observe that compared to physics generation, biology generation tasks are even more challenging for \modelname. Although the model shows a certain level of understanding in visual style and scene layout, many errors exist in the scientific details. These errors include incorrect colony color representation for specific bacterial species (Fig.~\ref{fig:biology}), misunderstanding of ecological knowledge (e.g., carbon cycle generation in Fig.~\ref{fig:biology}), and mistakes in experimental labeling or biological structures.

In addition, when facing professional biomedical image analysis tasks, such as cell segmentation and fluorescence image synthesis, \modelname still struggles to achieve accurate pixel-level generation, showing a clear performance gap compared to traditional domain-specific models.

These results indicate that while \modelname has basic biological commonsense understanding, there is still a large room for improvement in generating scientifically accurate biology-related images, especially for tasks requiring strict domain knowledge and precision.

\clearpage

\clearpage

\subsection{Mathematics}
\label{sec:math}
Mathematics-related image generation evaluates the model's capability to visualize mathematical concepts, including function plots, geometric diagrams, and mathematical annotations\cite{lee2025text, wang2025magicgeo, huang2024autogeo}. Compared with other domains, mathematical image generation requires precise understanding of mathematical semantics, accurate spatial layout, and strict labeling of axes, points, and annotations.

\begin{figure}[h]
    \centering
    \includegraphics[width=1.0\linewidth]{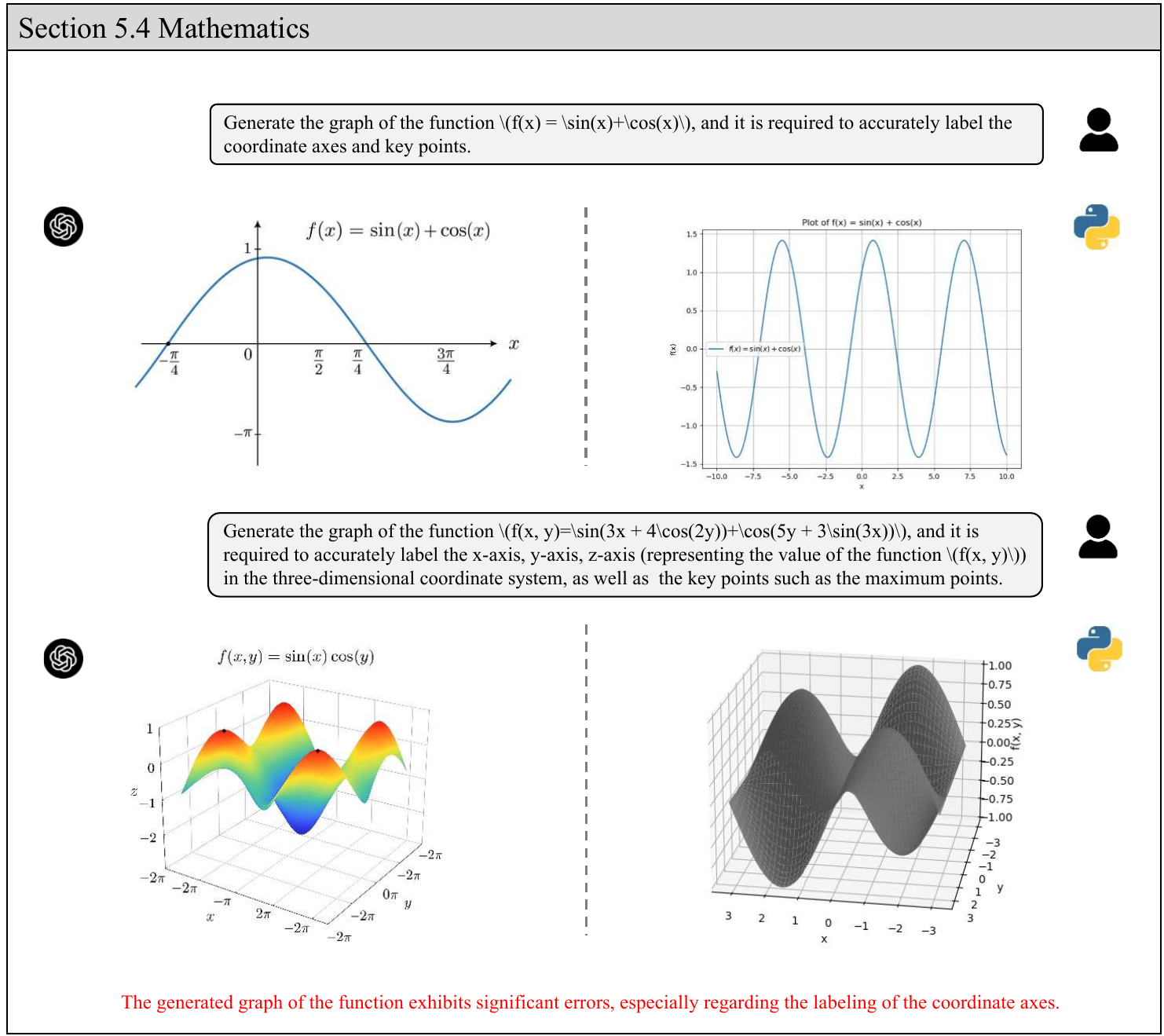}
\caption[Sec~\ref{sec:math}: Mathematics]{Examples of function plotting generation, with common mistakes in axis labeling and key point annotations.}
    \label{fig:math}
\end{figure}

\begin{figure}[h]
    \centering
    \includegraphics[width=1.0\linewidth]{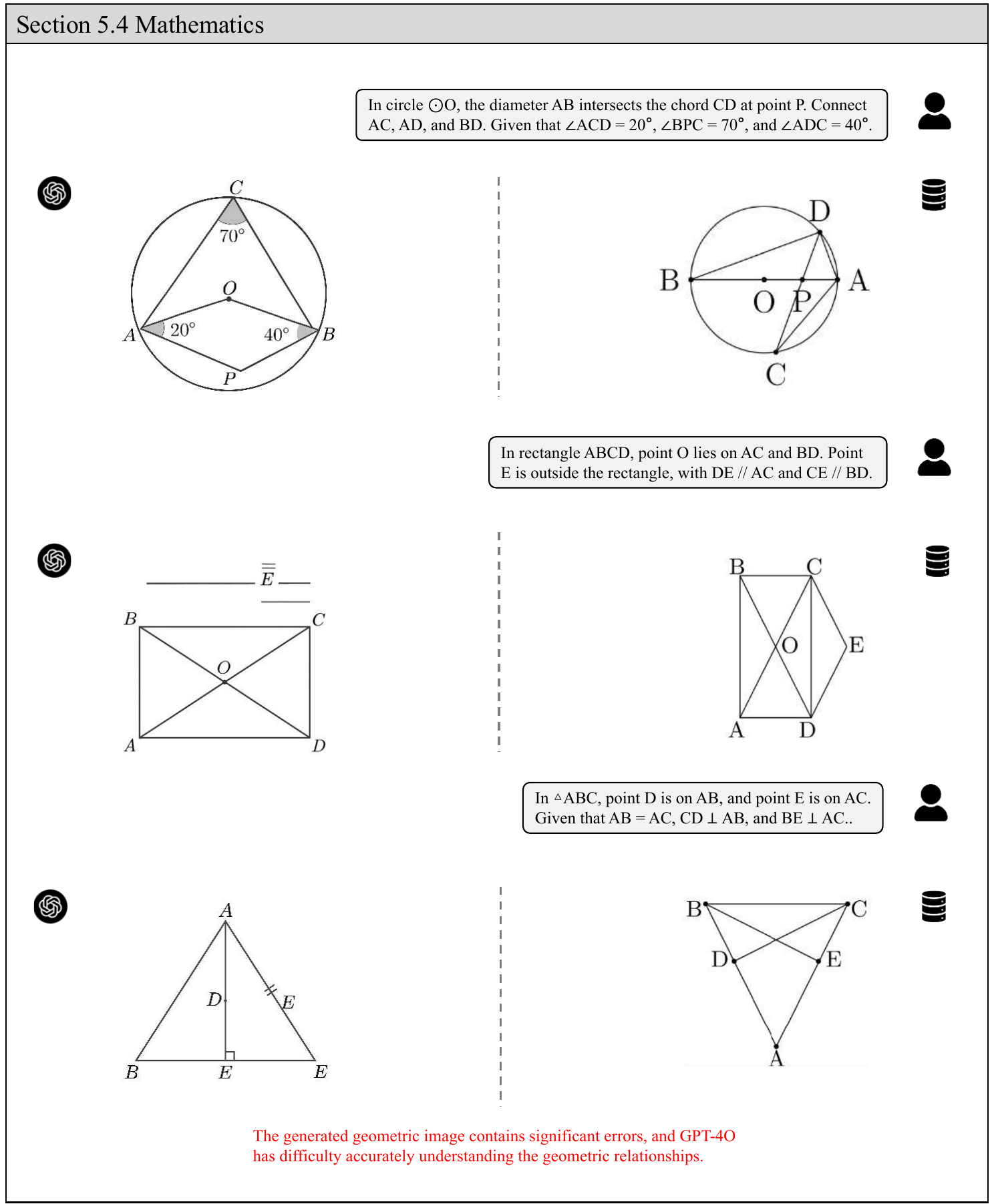}
\caption[Sec~\ref{sec:math}: Mathematics]{Examples of geometry diagram generation, with incorrect geometric relationships and inconsistent structures.}

    \label{fig:math1}
\end{figure}

As shown in Fig.~\ref{fig:math} and Fig.~\ref{fig:math1}, we test \modelname on two typical categories of mathematical generation tasks: function plotting and geometry diagram generation\cite{wang2025magicgeo}.
For function plotting, although \modelname is able to generate visually reasonable curve shapes, the coordinate axes and labels are often incorrect or missing.
For geometry diagram generation, the results are even less satisfactory. The model often fails to correctly understand geometric relationships, resulting in structurally incorrect diagrams, such as point positions inconsistent with the given conditions or lines violating the specified constraints.

These findings reveal that while \modelname shows a certain level of visual generation ability, its understanding of mathematical logic and precise graphical representation is still far from practical applications, especially in educational or scientific scenarios that require high accuracy.

\clearpage

\clearpage

\subsection{Agriculture}
\label{sec:agriculture}

\begin{figure}[h]
    \centering
    \includegraphics[width=1.0\linewidth]{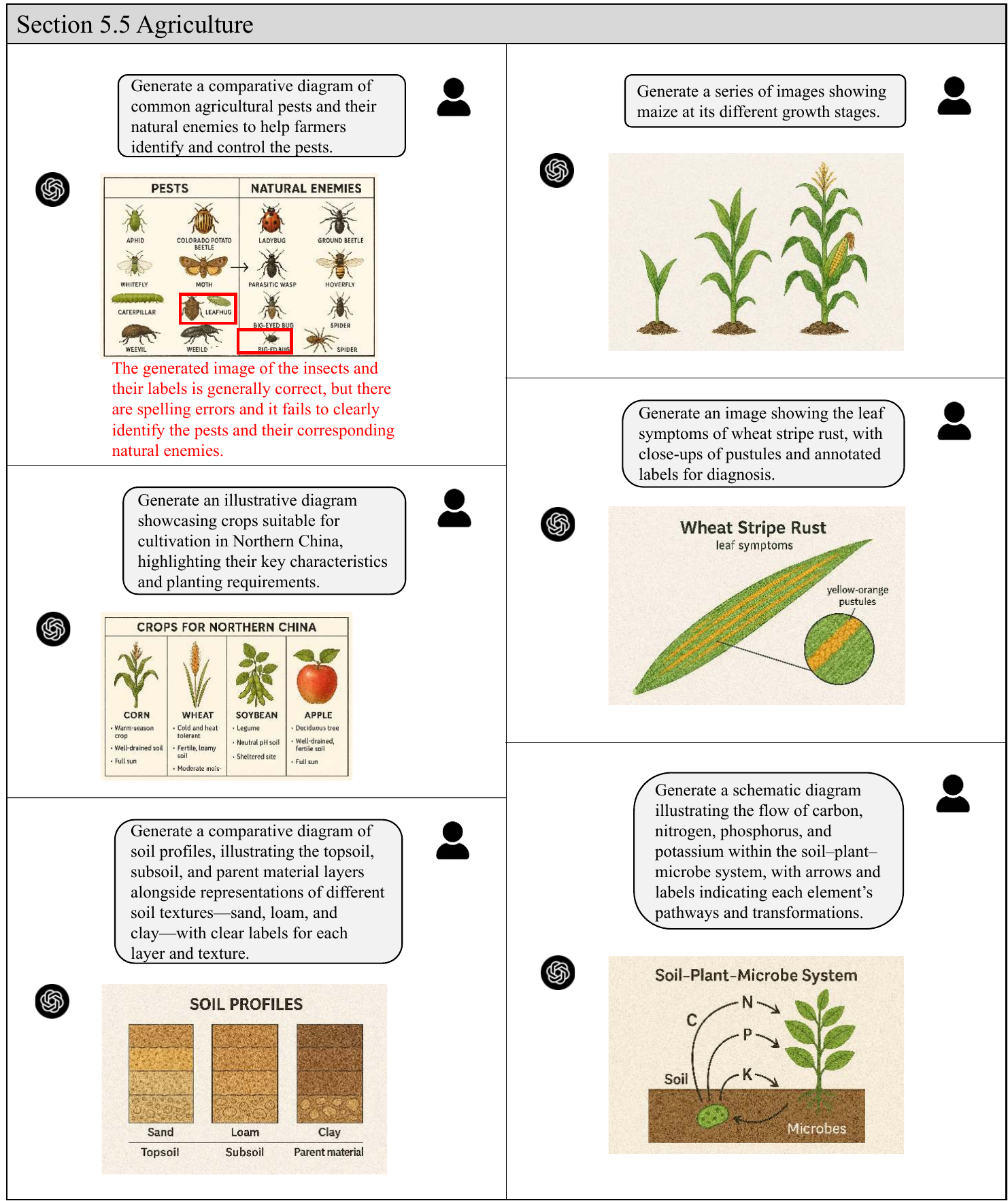}
    \caption[Sec~\ref{sec:agriculture}: Agriculture]{Examples of agriculture knowledge-based generation, including pest identification, crop growth stages, soil profiles, and agricultural cycles.}
    \label{fig:agriculture1}
\end{figure}

Agriculture-related image generation tasks evaluate the model's capability to understand basic agricultural knowledge and generate realistic agricultural scenes\cite{sapkota2023generative, cieslak2024generating, heschl2024synthset, modak2024enhancing, modak2024generative}. This section is divided into two parts: the first part (Fig.~\ref{fig:agriculture1}) focuses on testing the model's basic agricultural knowledge, such as pest identification, crop growth stages, soil profiles, and agricultural cycles. The second part (Fig.~\ref{fig:agriculture1}) explores the model's ability to generate realistic synthetic agricultural images, which are commonly used in agricultural vision tasks and dataset construction.

\begin{figure}[h]
    \centering
    \includegraphics[width=1.0\linewidth]{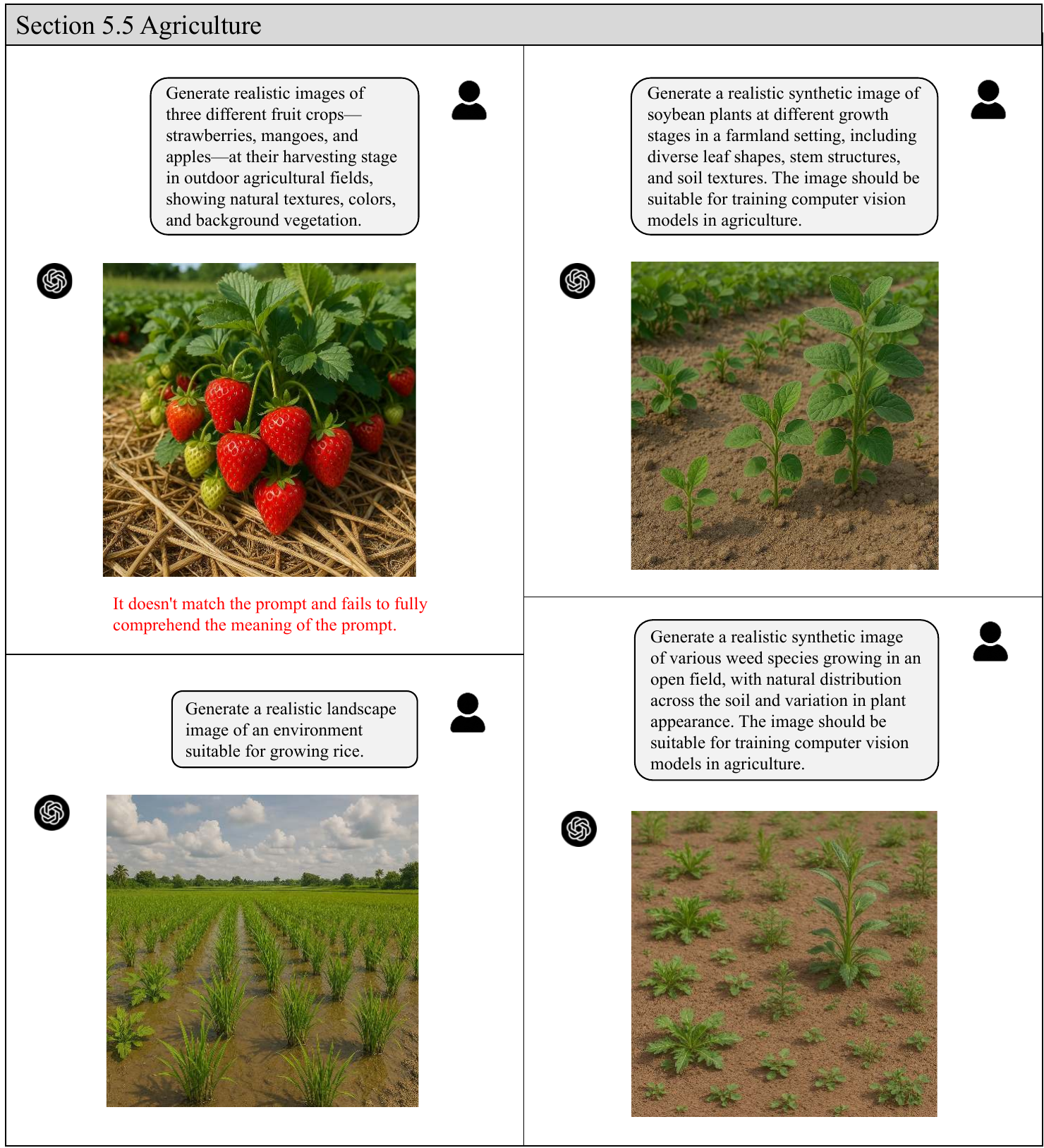}
    \caption[Sec~\ref{sec:agriculture}: Agriculture]{Examples of synthetic agricultural image generation, covering fruit crops, farmland environments, and various plant species in agricultural scenes.}
    \label{fig:agriculture2}
\end{figure}

From Fig.~\ref{fig:agriculture1}, we observe that \modelname shows a reasonable understanding of general agricultural concepts and can generate diagrams with clear visual structures. However, the generated images still suffer from obvious errors, such as incorrect labels (e.g., spelling mistakes in pest names) and inaccurate scientific details (\ie., failing to correctly associate pests).
In Fig.~\ref{fig:agriculture2}, we evaluate the model's capability of generating synthetic agricultural scenes. While the model demonstrates some ability in generating farmland environments, there are noticeable failures in prompt comprehension, such as generating wrong fruit categories or failing to synthesize fine-grained details like crop textures and growth characteristics.

Overall, these results suggest that \modelname possesses basic agricultural knowledge but still lacks domain-specific expertise required for generating high-quality agricultural images, especially when applied to data synthesis tasks for agricultural computer vision.

\clearpage

\clearpage

\section{Commonsense-based Image Generation}
\label{sec:commonsense}

Commonsense-based image generation aims to evaluate the model's ability to understand general knowledge and cultural concepts in the real world. This task is essential for testing the model's capability to generate images that align with human common sense, cultural backgrounds, and general knowledge from different domains.

We design a comprehensive set of categories, including landmark, festival, food, clothing, painting, architecture, literature, logo, and health \& safety. These tasks cover a wide range of cultural and commonsense scenarios, as investigated in previous works~\cite{kannen2024beyond,jeong2025culture,li2024cheffusion,wang2025cookingdiffusion,bayramli2025diffusion,liu2024cultural,lampe2024dicti,srivastava2024wordrobe,dong2024towards,he2024dresscode,singh2024fashionsd,zhang2024texcontrol,qadri2024dialogue}.

Experimental results show that \modelname performs well in most commonsense tasks, especially in landmark, food, clothing, and building generation, demonstrating its impressive cultural awareness and real-world knowledge. However, in the logo generation task, the model exhibits frequent factual errors, generating wrong characters or incorrect brand elements. In addition, we observe several unreasonable generations in the health and safety-related tasks, which deviate from common sense.

These results indicate that while \modelname shows strong commonsense and cultural understanding abilities, there is still room for improvement in precise factual generation and safety-critical scenarios.

\clearpage

\begin{figure}[h]
    \centering
    \includegraphics[width=1.0\linewidth]{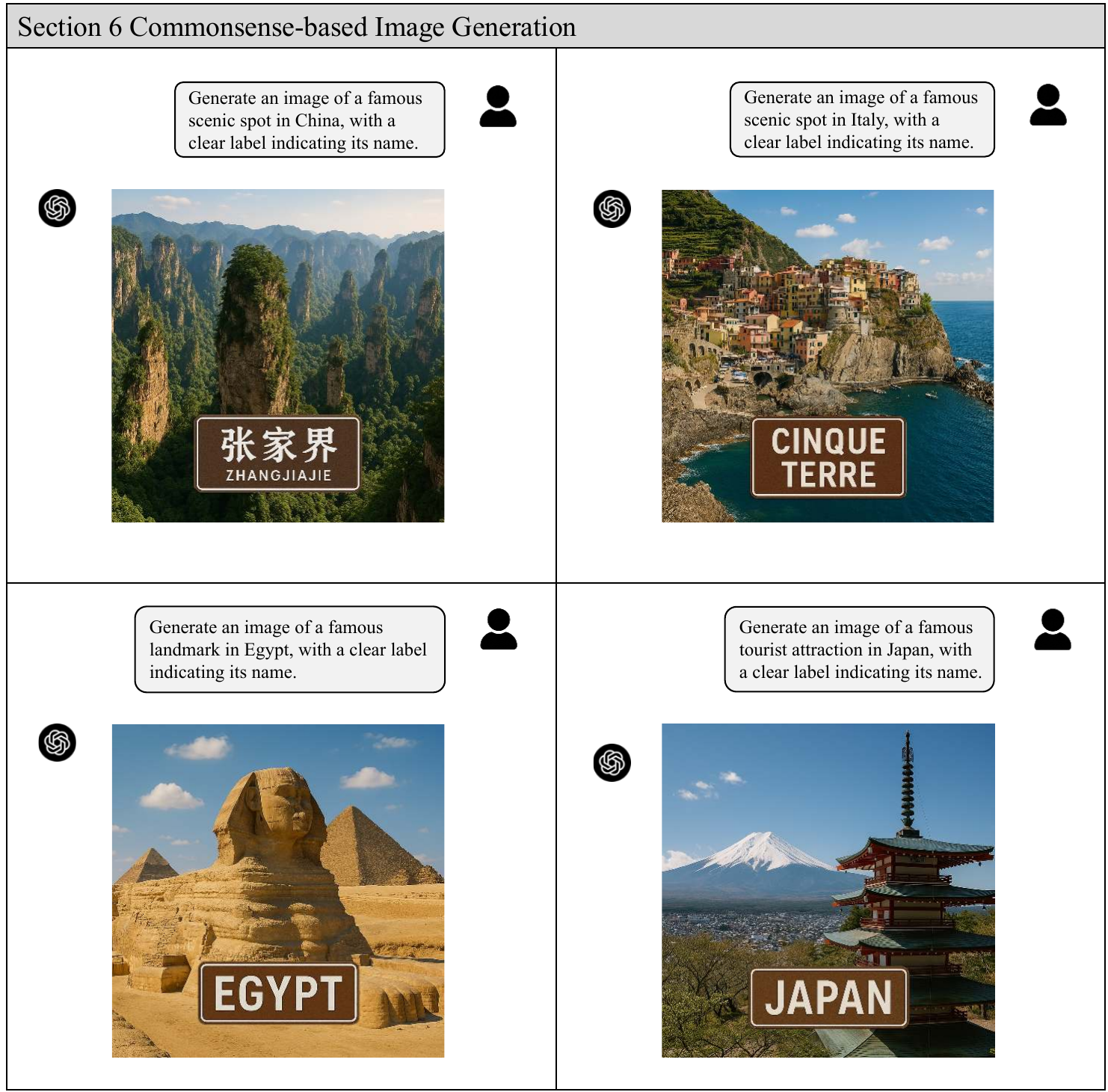}
    \caption[Sec~\ref{sec:commonsense}: Landmark]{Examples of landmark image generation, showcasing different iconic cultural and geographic locations.}
    \label{fig:landmark}
\end{figure}

\begin{figure}[h]
    \centering
    \includegraphics[width=1.0\linewidth]{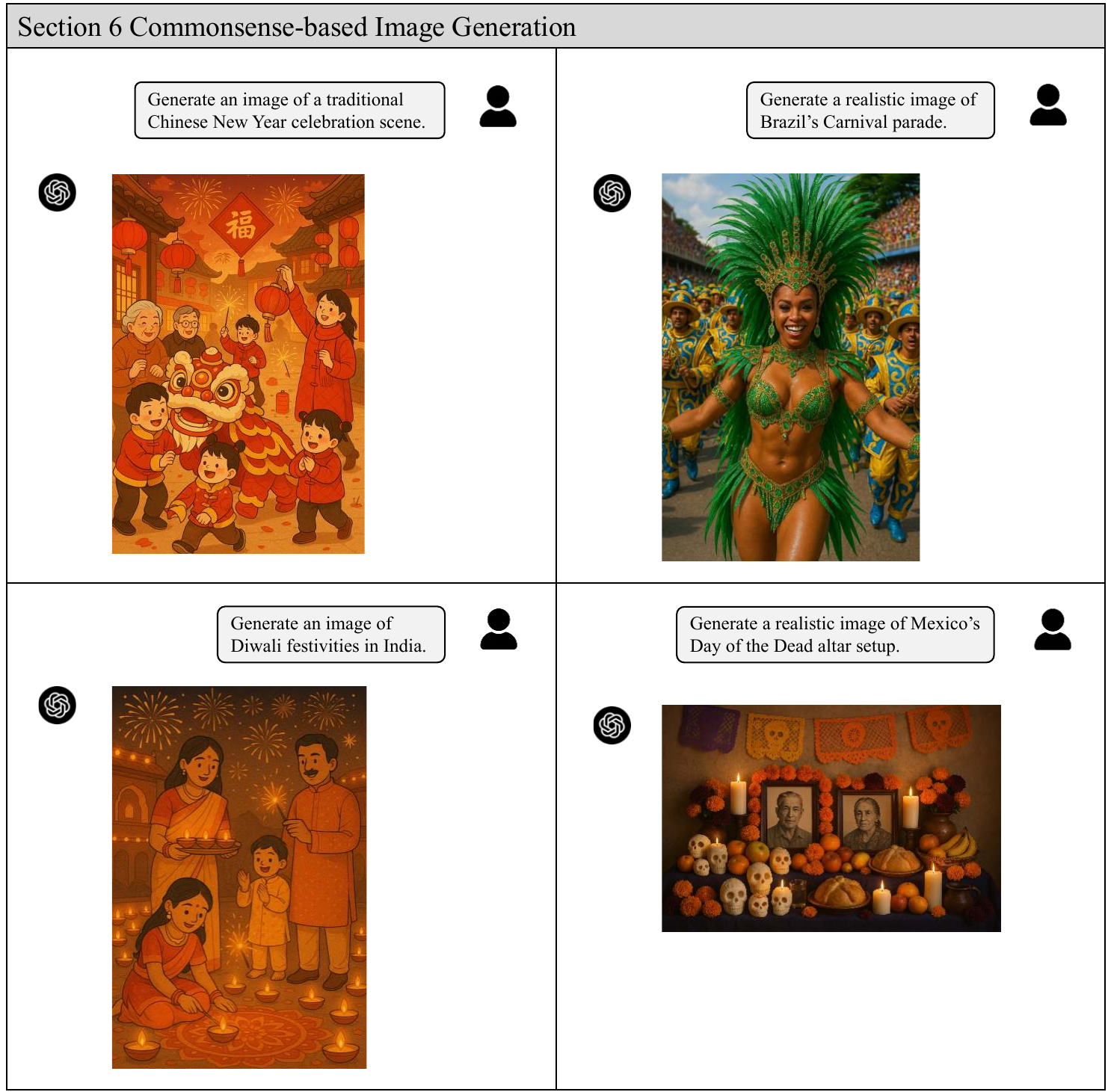}
    \caption[Sec~\ref{sec:commonsense}: Festival]{Examples of festival image generation, showing traditional customs and cultural celebrations.}
    \label{fig:festival}
\end{figure}

\begin{figure}[h]
    \centering
    \includegraphics[width=1.0\linewidth]{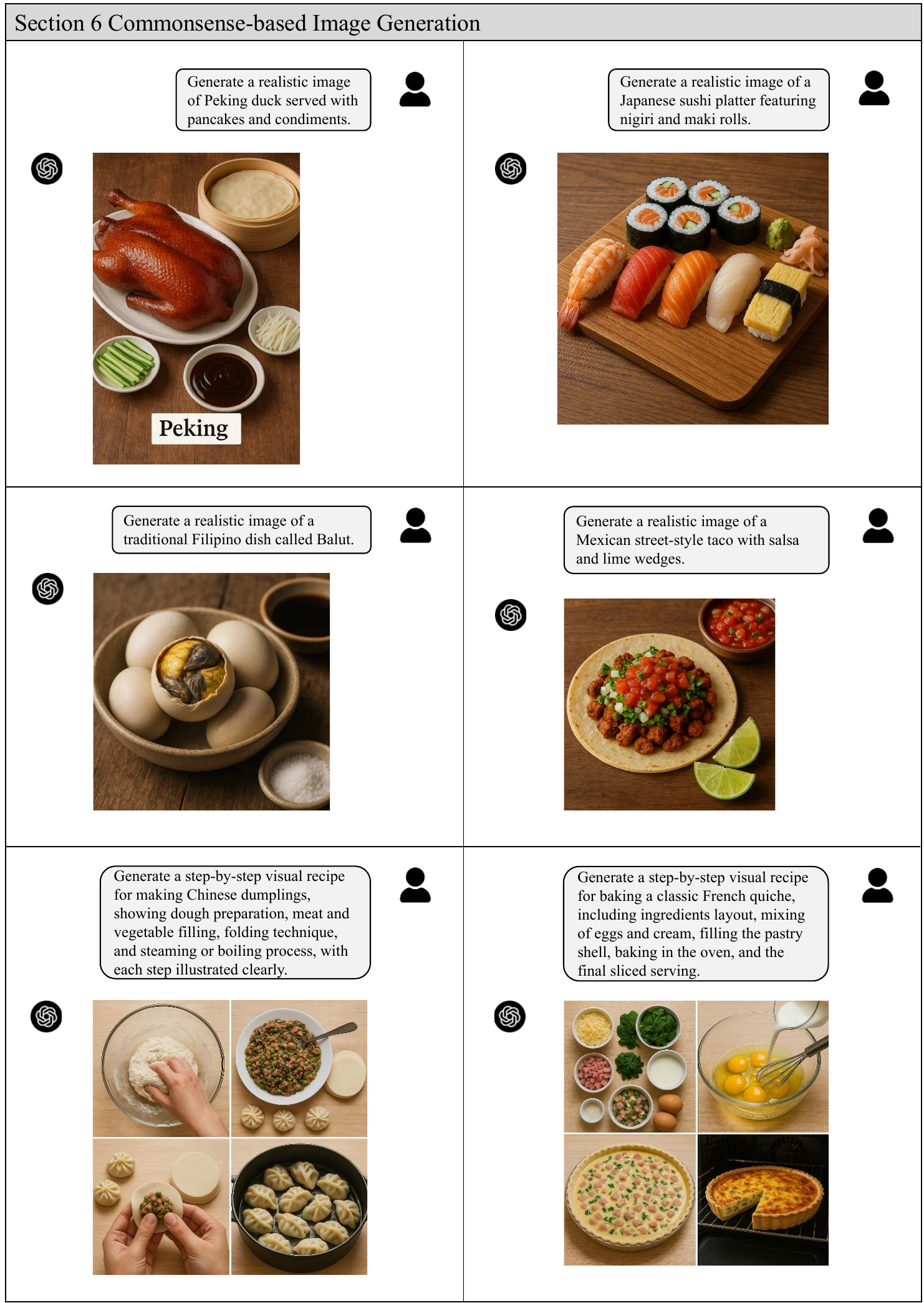}
    \caption[Sec~\ref{sec:commonsense}: Food]{Examples of food generation conditioned on different cultural recipes and cooking styles.}
    \label{fig:food}
\end{figure}

\begin{figure}[h]
    \centering
    \includegraphics[width=1.0\linewidth]{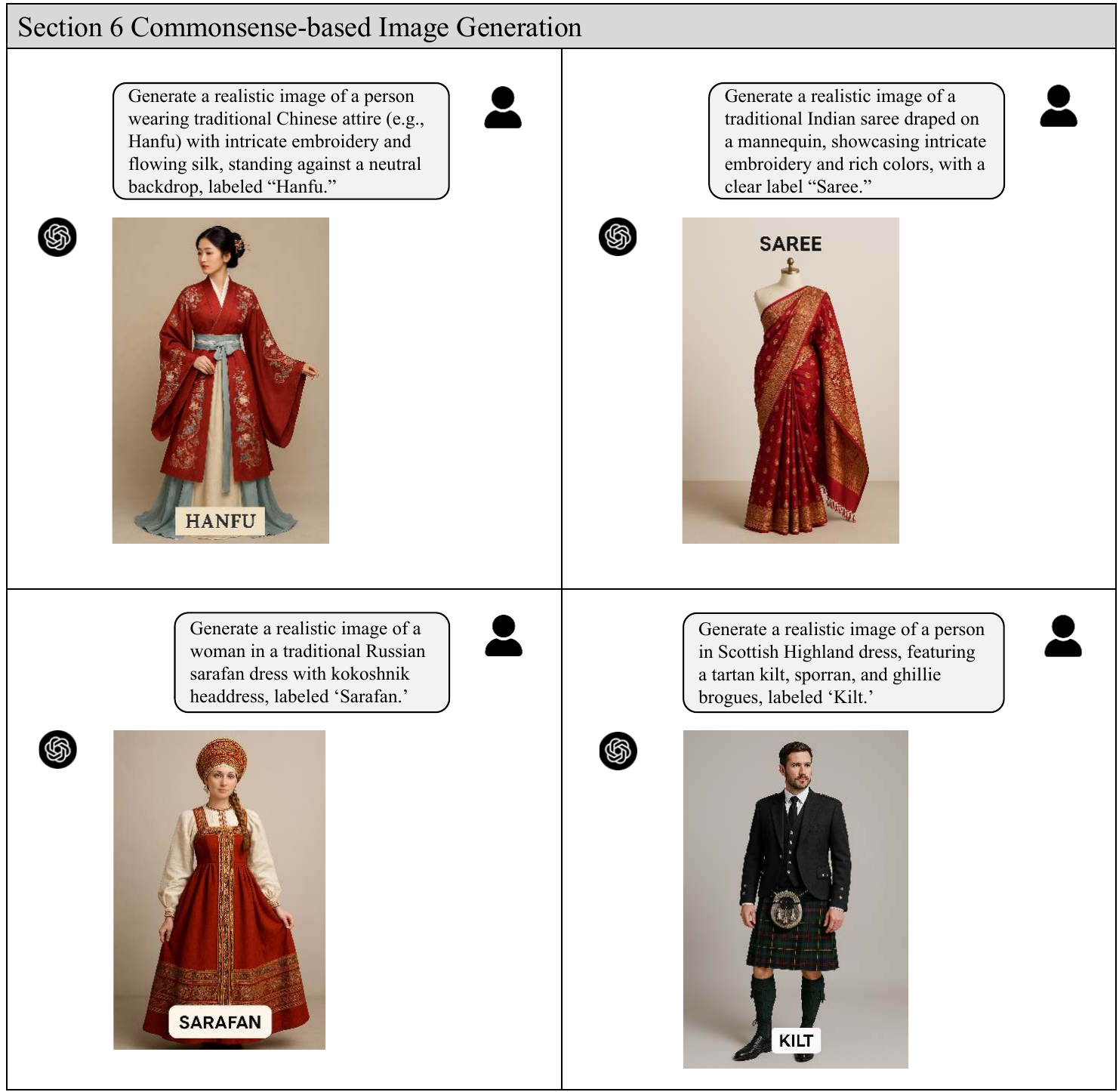}
    \caption[Sec~\ref{sec:commonsense}: Cloth]{Examples of clothing generation guided by regional fashion styles and garment design.}
    \label{fig:cloth}
\end{figure}

\begin{figure}[h]
    \centering
    \includegraphics[width=1.0\linewidth]{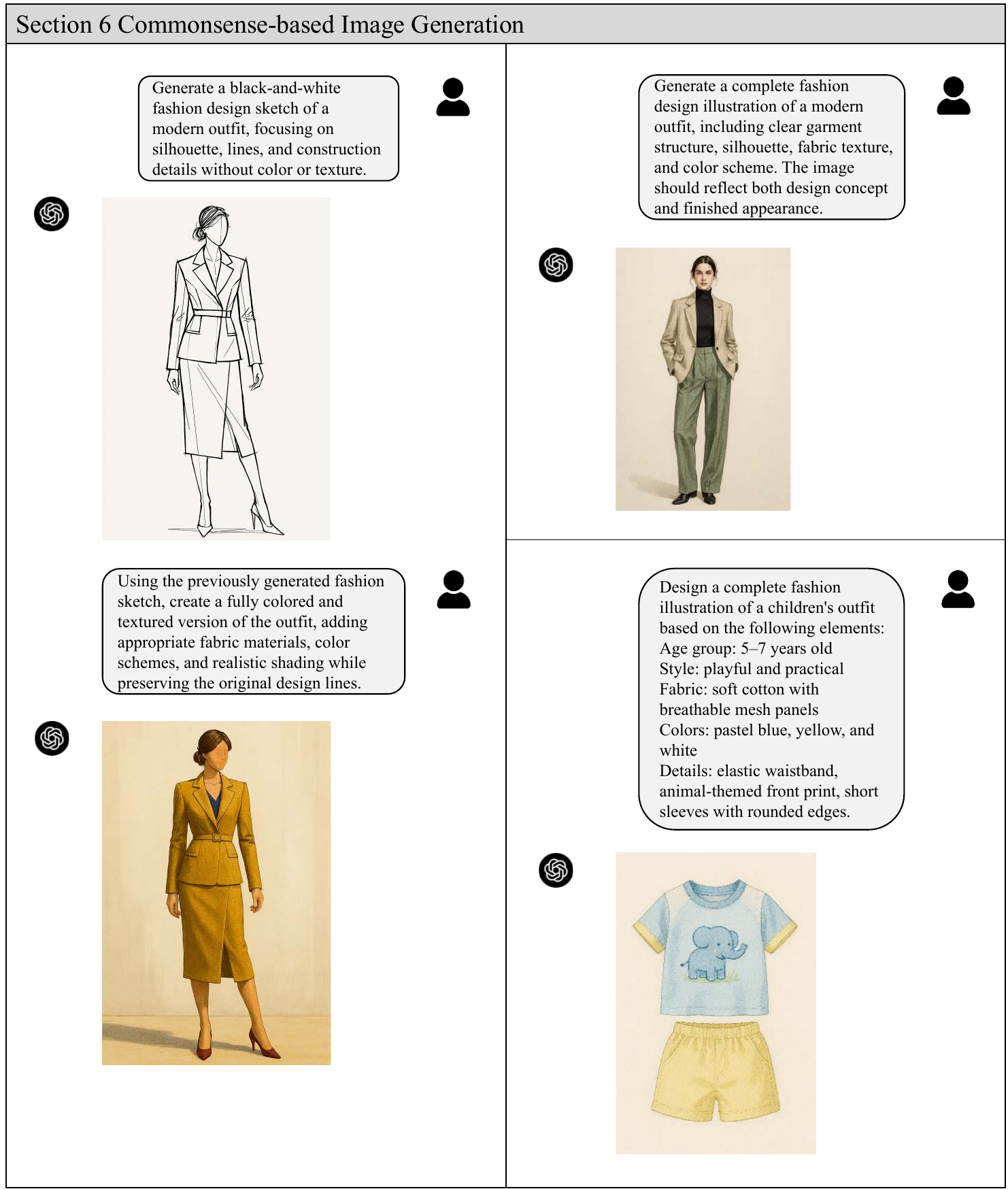}
    \caption[Sec~\ref{sec:commonsense}: Cloth]{More examples of clothing image generation with style control and appearance variation.}
    \label{fig:cloth1}
\end{figure}

\begin{figure}[h]
    \centering
    \includegraphics[width=1.0\linewidth]{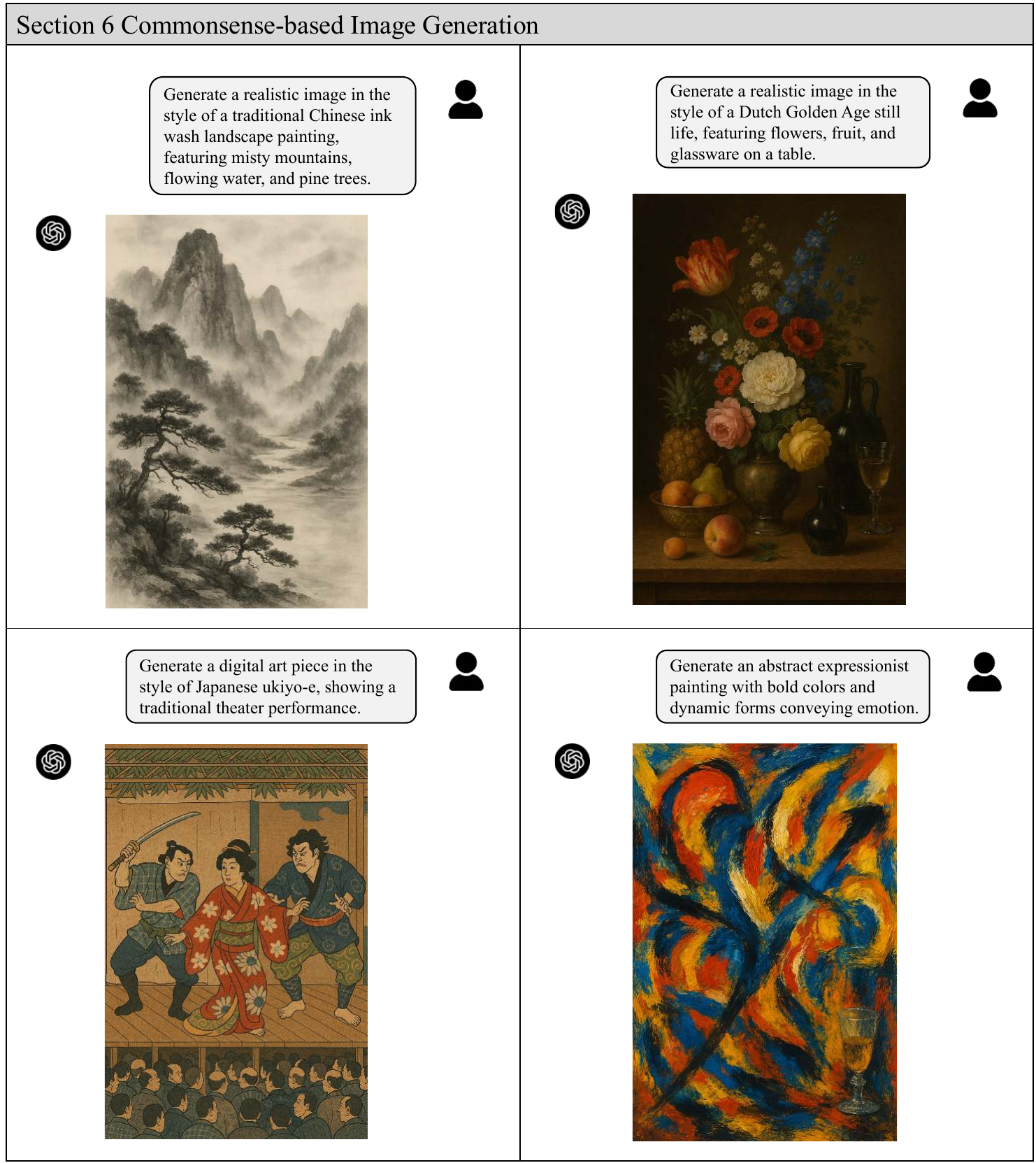}
    \caption[Sec~\ref{sec:commonsense}: Painting]{Examples of painting generation with various art styles and cultural elements.}
    \label{fig:painting}
\end{figure}

\begin{figure}[h]
    \centering
    \includegraphics[width=1.0\linewidth]{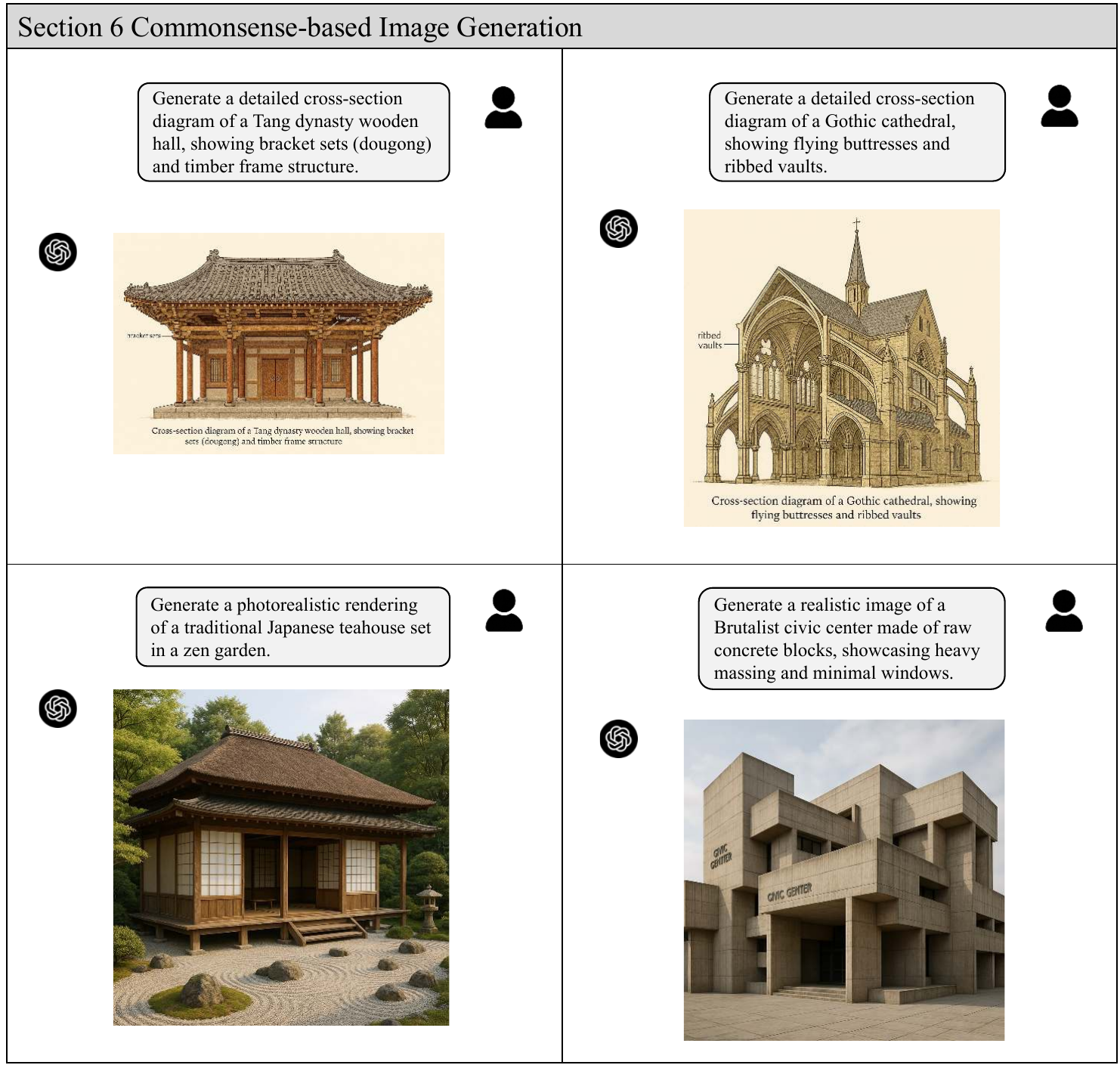}
    \caption[Sec~\ref{sec:commonsense}: Building]{Examples of building generation based on architectural styles and cultural background.}
    \label{fig:building}
\end{figure}

\begin{figure}[h]
    \centering
    \includegraphics[width=1.0\linewidth]{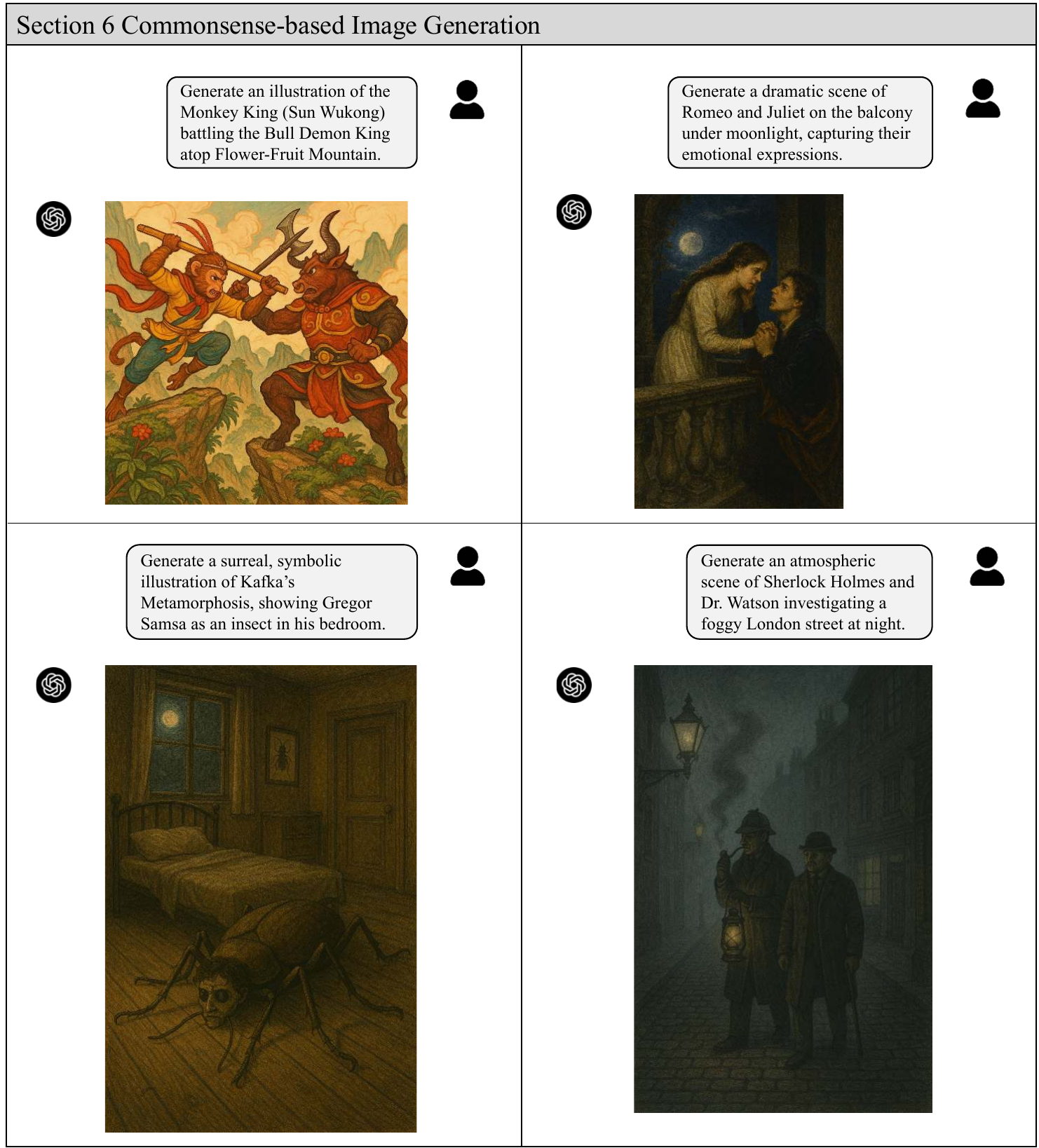}
    \caption[Sec~\ref{sec:commonsense}: Literature]{Examples of literature-related generation, including book covers and literary visual elements.}
    \label{fig:literature}
\end{figure}

\begin{figure}[h]
    \centering
    \includegraphics[width=1.0\linewidth]{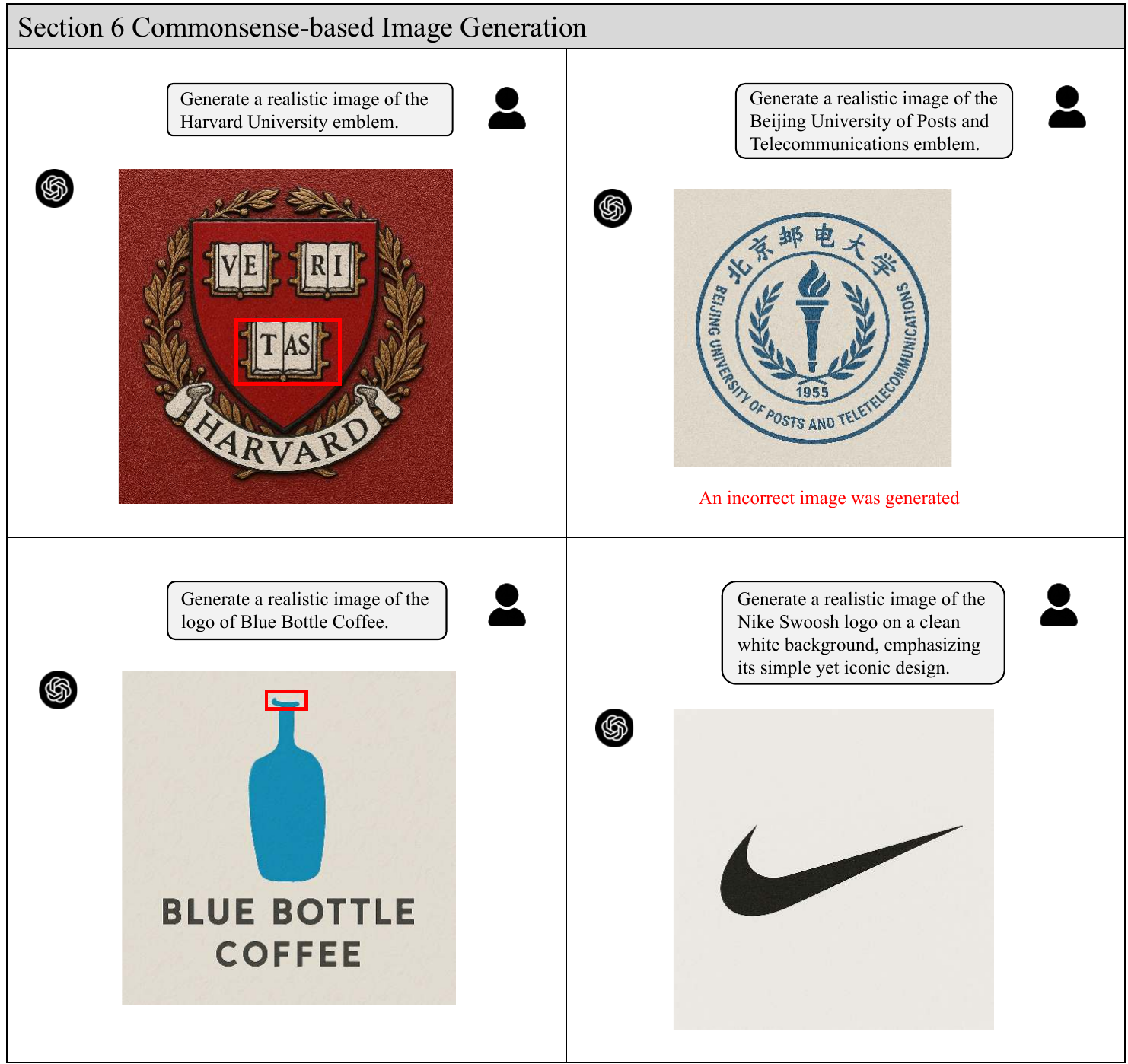}
    \caption[Sec~\ref{sec:commonsense}: LOGO]{Examples of logo generation. \modelname tends to generate factual errors, such as wrong text or symbols.}
    \label{fig:logo}
\end{figure}

\begin{figure}[h]
    \centering
    \includegraphics[width=1.0\linewidth]{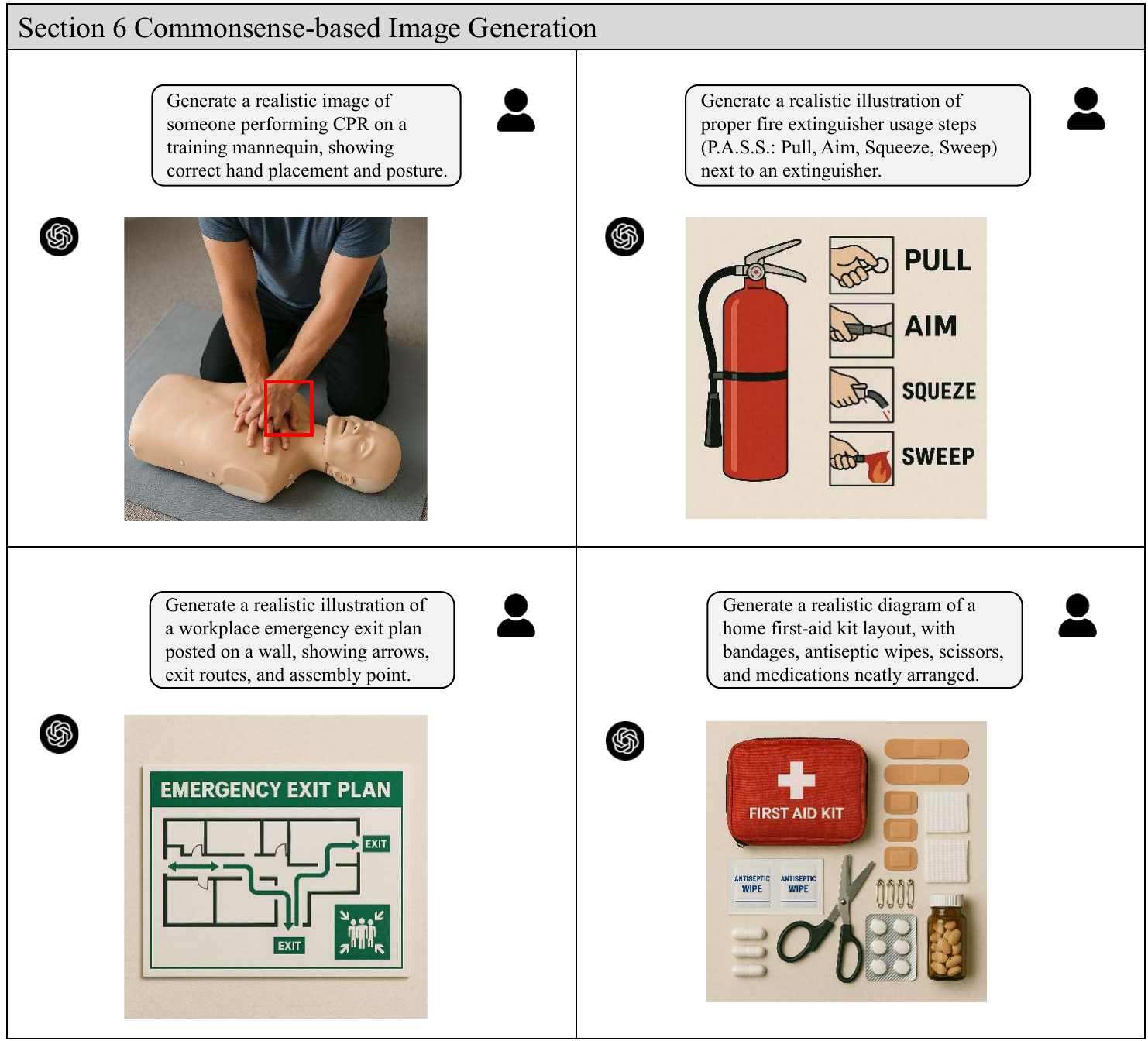}
    \caption[Sec~\ref{sec:commonsense}: Health and Safety]{Examples of health and safety-related generation.}
    \label{fig:Health}
\end{figure}

\clearpage

\section{Spatially-aware Image Generation}

\label{sec:spatial}

\subsection{Multi-view Image Generation}
Our evaluation begins with assessing the model’s target-level reconstruction capability, which involves generating novel-view images conditioned on a single input view of the target.

\textbf{Normal Novel-view Synthesis.}
As shown in Fig.~\ref{fig:7_2_target_0}–\ref{fig:7_2_target_2}, we test \modelname's ability to synthesize novel views for common objects with regular geometric structures. The model successfully produces visually plausible outputs from multiple viewpoints, capturing the overall shape and semantics of the targets. Especially for cartoon-style and simplified furniture objects, the outputs demonstrate strong viewpoint control and global structural consistency. However, we still observe subtle 3D misalignments in details such as window geometry and edge connections, indicating the absence of true volumetric reasoning.

\textbf{Human-centric Novel-view Synthesis.}
We further evaluate \modelname's ability to synthesize new views for human portraits and full-body figures. As illustrated in Fig.~\ref{fig:7_2_human_0}–\ref{fig:7_2_human_2}, the model generates front, side, and back views that maintain identity and pose coherence to a surprising degree. In portrait-level tasks, \modelname preserves key facial features under varying perspectives. For full-body generation, while the general body layout remains intact, issues such as inconsistent limb placement are still visible, reflecting the challenge of maintaining fine-grained spatial correspondence for articulated objects like the human body.

\textbf{Scene-centric Novel-view Synthesis.}
Finally, we evaluate \modelname's capability in synthesizing novel views of complex outdoor scenes (Fig.~\ref{fig:7_2_scene_0}–\ref{fig:7_2_scene_1}). While the model is capable of switching to plausible overhead or oblique viewpoints and preserves large-scale structure and style, its performance degrades significantly under large camera transformations. Inaccurate alignment of roofs, inconsistent vanishing points, and mislocated objects are frequent, suggesting that \modelname lacks sufficient 3D scene understanding to handle complex geometries and deep spatial layouts.

\clearpage
\begin{figure}[h]
    \centering
    \includegraphics[width=1.0\linewidth]{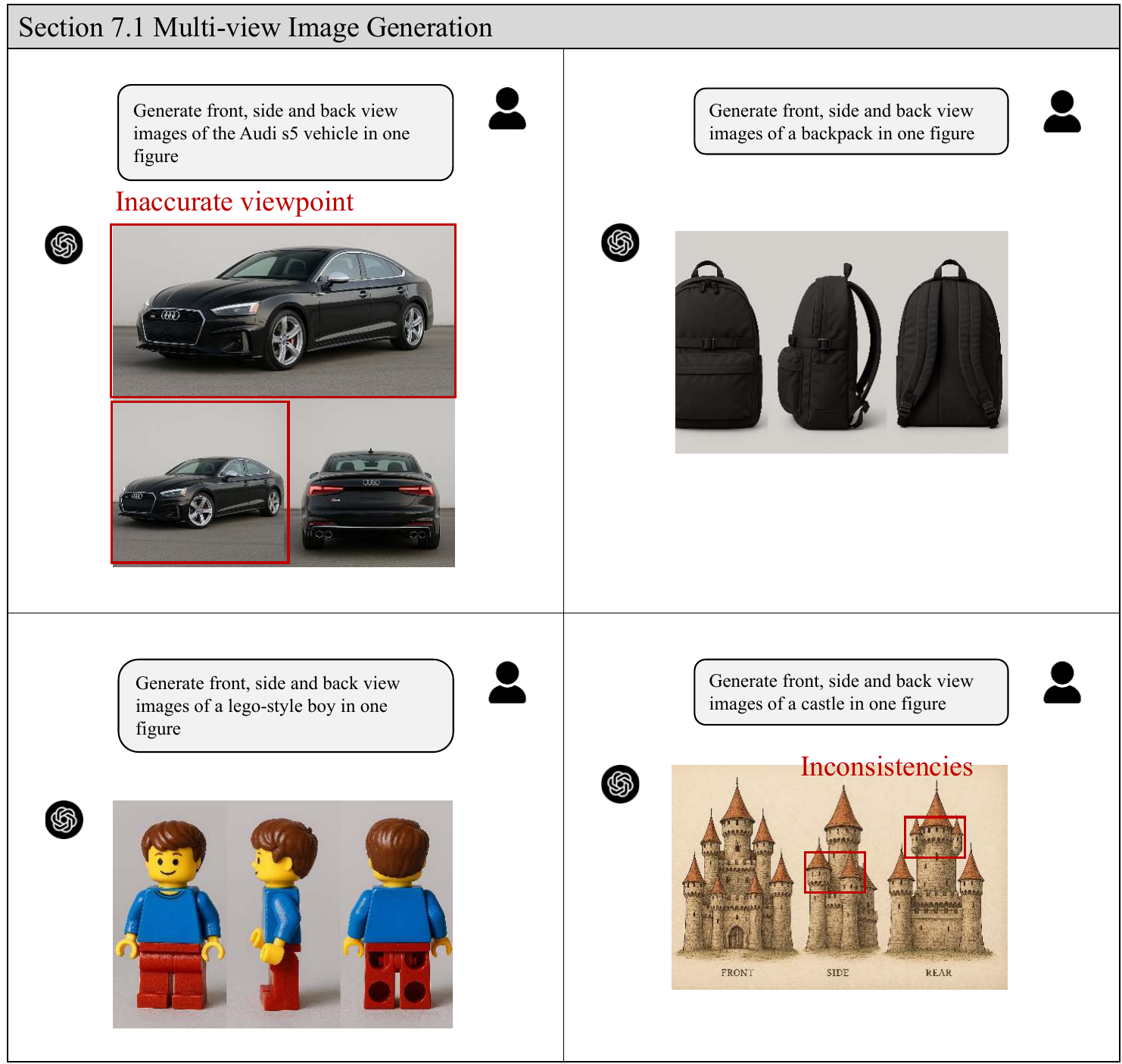}
    \caption[Sec~\ref{sec:spatial}:Multi-view Image Generation]{Examples of multi-view image generation results produced by \modelname. The model is capable of generating coherent front, side, and back views for a variety of objects, such as backpacks and toy figures, maintaining consistent shapes and structures. However, challenges remain in handling complex geometries and rigid objects. For instance, the vehicle example exhibits inaccurate viewpoint alignment (top left), while the castle generation shows structural inconsistencies across views (bottom right). }
    \label{fig:7_1_target_0}
\end{figure}
\clearpage

\begin{figure}[h]
    \centering
    \includegraphics[width=1.0\linewidth]{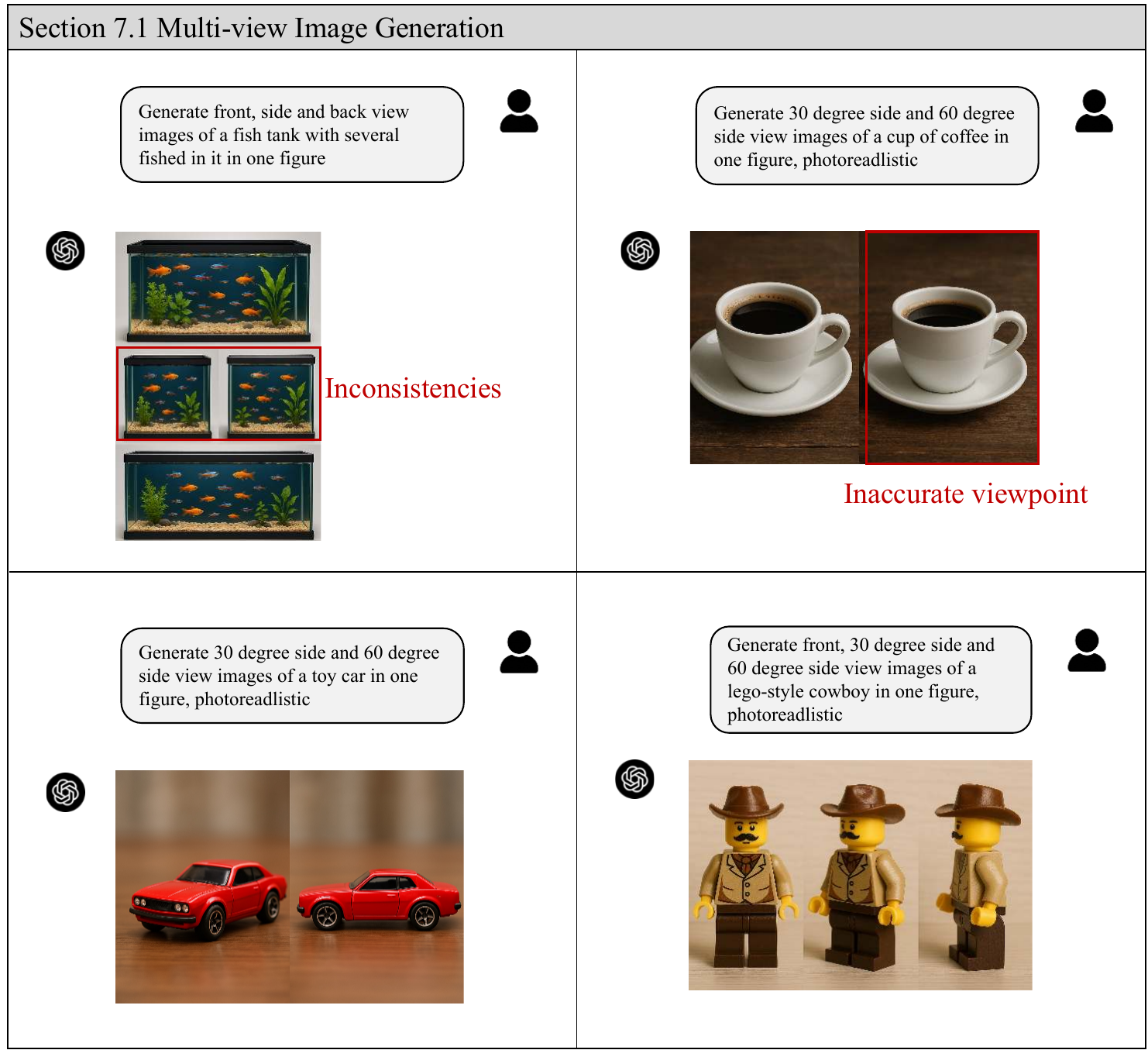}
    \caption[Sec~\ref{sec:spatial}:Multi-view Image Generation]{Examples of angle-specific multi-view image generation results by \modelname. The model successfully renders photorealistic side-view transitions at specified angles (30° and 60°) for structured objects like toy cars and lego-style figures, maintaining high fidelity across views. However, it struggles with consistency in complex or cluttered scenes (e.g., the fish tank) and precise viewpoint transitions for symmetric objects (e.g., the coffee cup), where angular alignment is inaccurate or content appears altered, highlighting limitations under fine-grained view control.}
    \label{fig:7_1_target_1}
\end{figure}
\clearpage

\begin{figure}[h]
    \centering
    \includegraphics[width=1.0\linewidth]{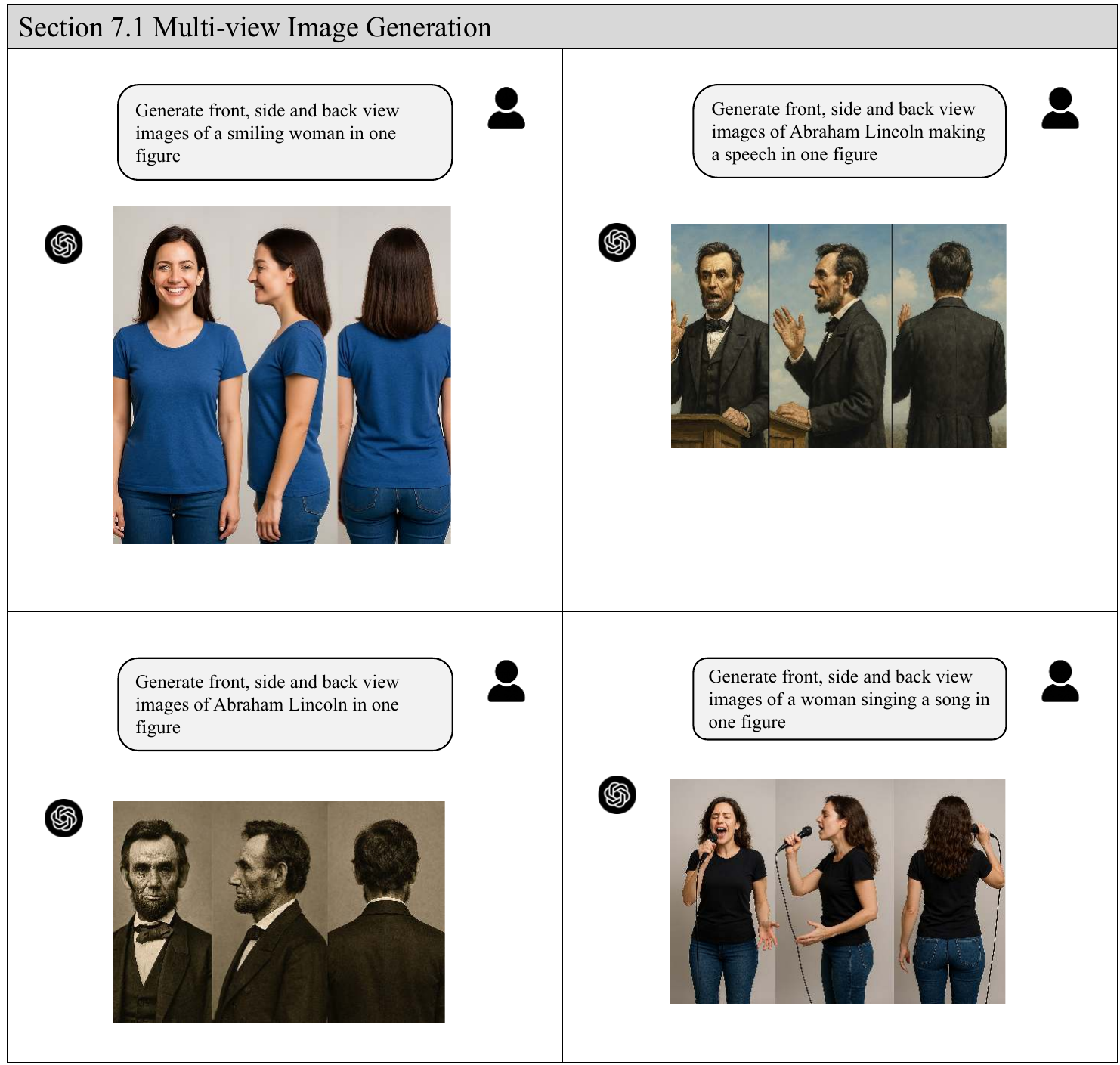}
    \caption[Sec~\ref{sec:spatial}:Multi-view Image Generation]{Examples of multi-view human image generation results by \modelname. The model accurately synthesizes front, side, and back views of human subjects in both static and dynamic scenarios, such as standing poses and singing actions. Viewpoint transitions are smooth and identity preservation is consistent across views, even under varied body orientations and clothing details, demonstrating the model’s strong grasp of human 3D geometry and pose-aware rendering.}
    \label{fig:7_1_human_0}
\end{figure}
\clearpage

\begin{figure}[h]
    \centering
    \includegraphics[width=1.0\linewidth]{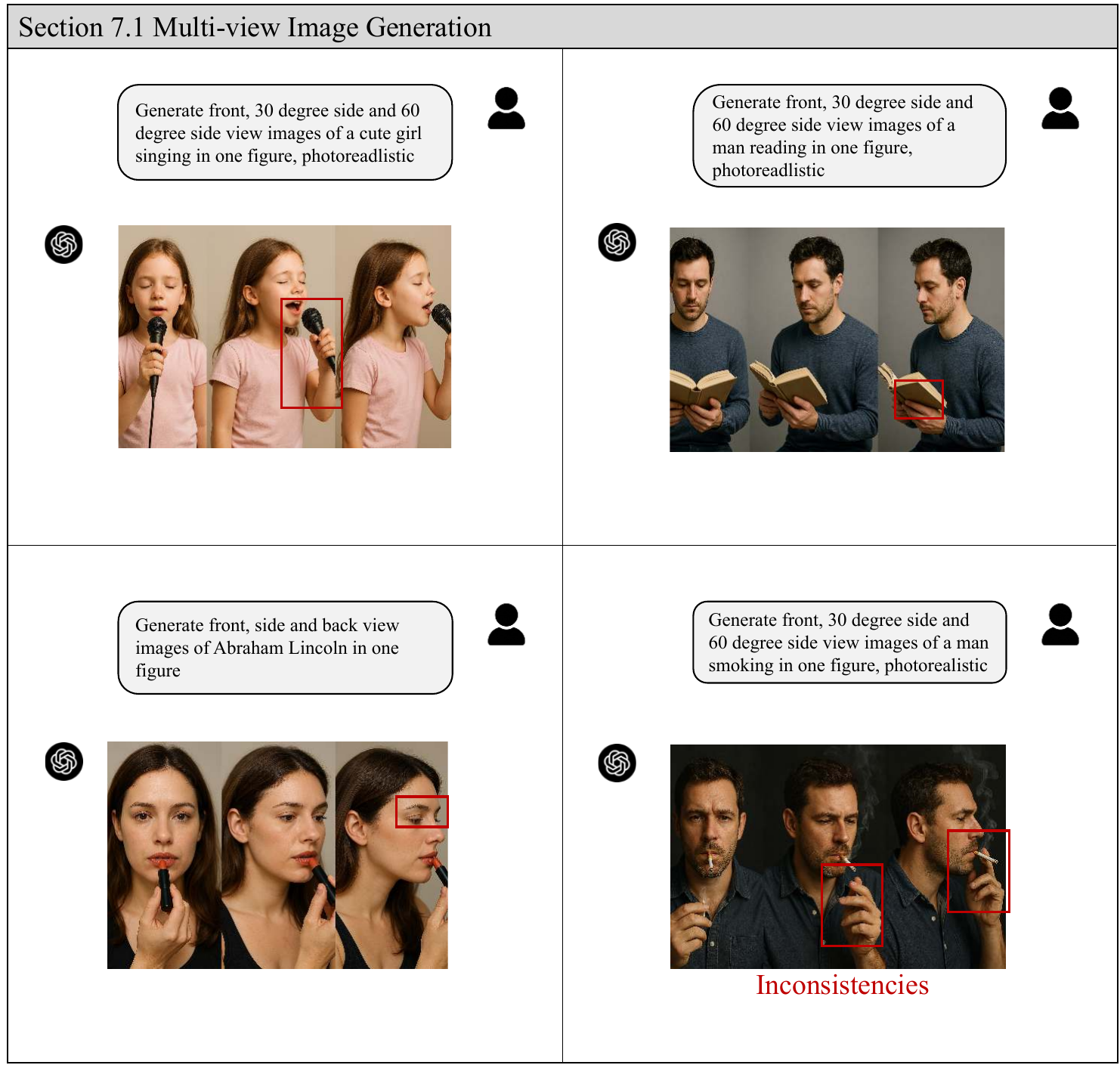}
    \caption[Sec~\ref{sec:spatial}:Multi-view Image Generation]{Examples of photorealistic multi-view human image generation by \modelname. The model generally produces consistent front and angled views of human subjects. However, certain cases still exhibit noticeable inconsistencies, particularly in hand-object interactions and facial details, indicating room for improvement in fine-grained coherence across views.}
    \label{fig:7_1_human_1}
\end{figure}
\clearpage

\begin{figure}[h]
    \centering
    \includegraphics[width=1.0\linewidth]{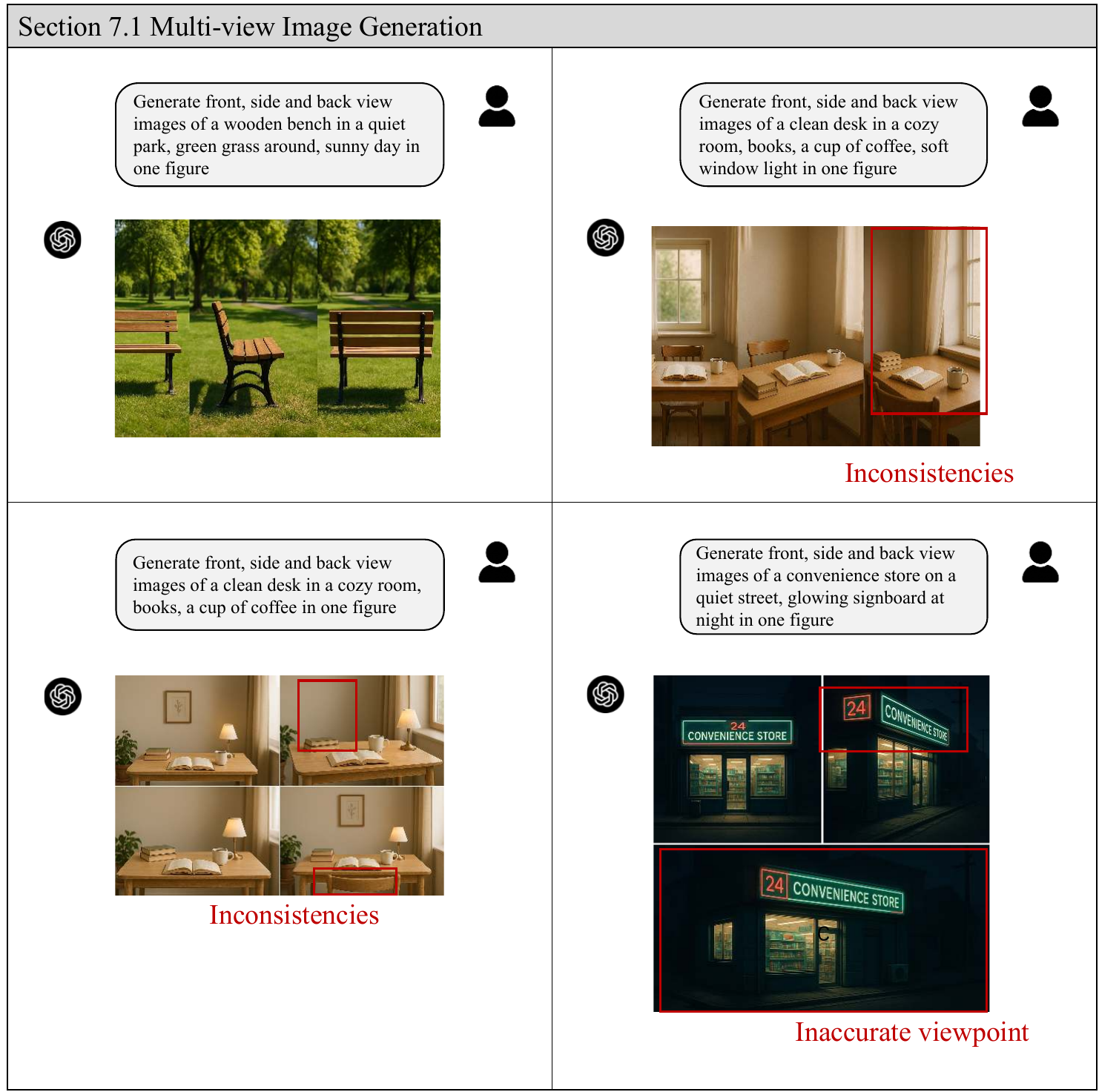}
    \caption[Sec~\ref{sec:spatial}:Multi-view Image Generation]{Examples of multi-view scene generation results by \modelname. The model attempts to produce front, side, and back views of complex indoor and outdoor scenes. However, the outputs frequently suffer from object placement inconsistencies and inaccurate viewpoint transformations, particularly in structured environments like desks and storefronts. These issues reveal \modelname’s current limitations in holistic scene understanding and spatial consistency across views.}
    \label{fig:7_1_scene_0}
\end{figure}
\clearpage

\subsection{Novel-view Synthesis}
The novel-view synthesis task~\cite{watson2022novel, chan2023generative, tseng2023consistent, yu2023long} is a fundamental problem in the field of 3D vision, closely aligned with the goals of 3D reconstruction~\cite{mildenhall2021nerf, kerbl20233d} and image-to-3D~\cite{tatarchenko2019single, liu2023zero, liu2023one, tang2023dreamgaussian, deng20233d, tochilkin2024triposr} generation. In this task, the model is required to generate images of a given object or scene from unseen viewpoints based on a single input image, effectively making it a form of \textbf{ single-image reconstruction}~\cite{wang2024crm, xu2024grm}. This setup allows us to assess \modelname’s spatial awareness capability from a reconstruction perspective. Similar to the previous section, we design a set of logically structured tasks with varying levels of difficulty to systematically evaluate the model’s performance.

\subsubsection{Traditional View Synthesis }
Our evaluation begins with assessing the model’s target-level reconstruction capability, which involves generating novel-view images conditioned on a single input view of the target.

\begin{figure}[h]
    \centering
    \includegraphics[width=1.0\linewidth]{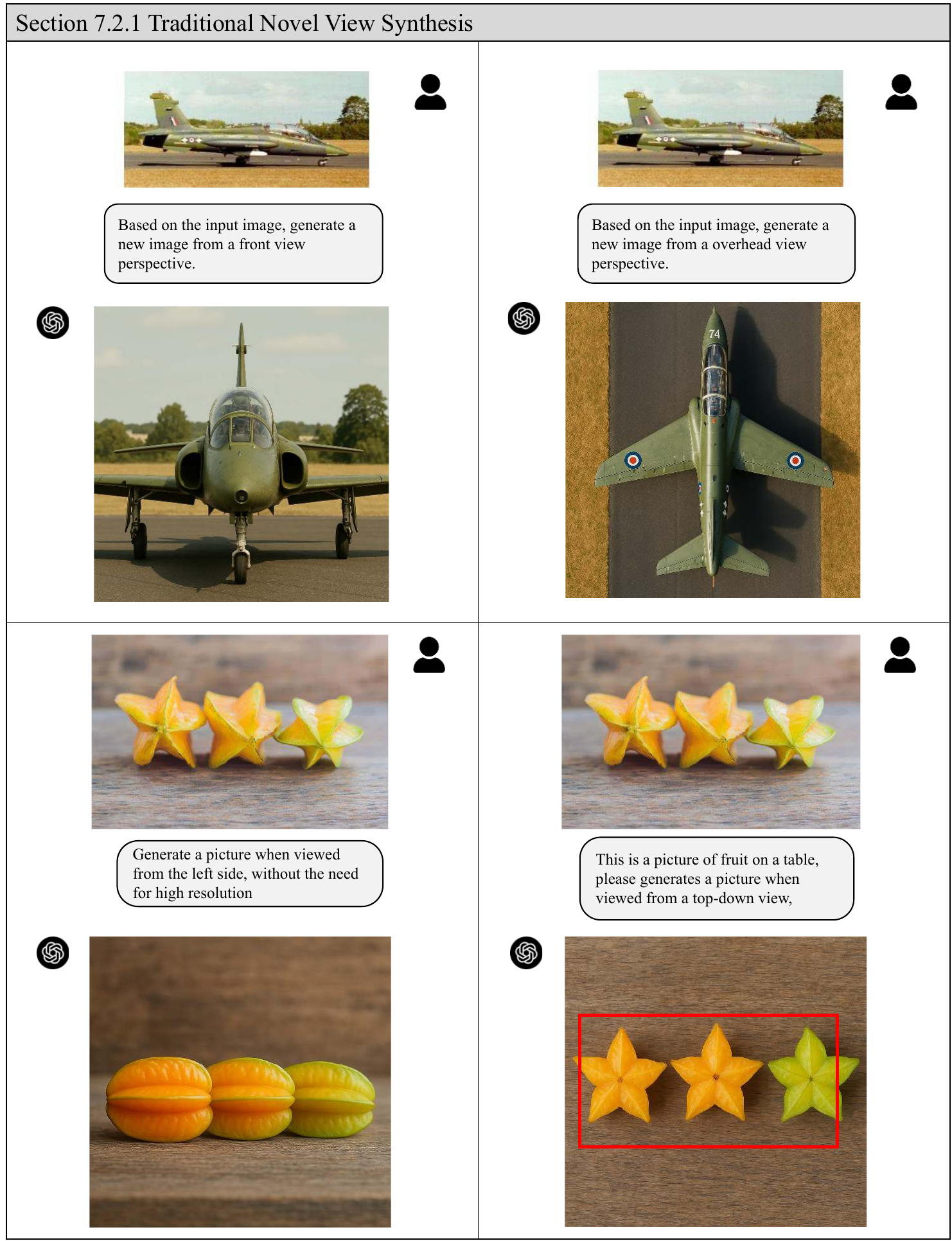}
    \caption[Sec~\ref{sec:spatial}:Novel View Synthesis]{Examples of novel-view synthesis on normal targets by \modelname. The model generates visually plausible images from new viewpoints given a single input view, showing reasonable spatial understanding for both objects and scenes. However, subtle 3D inconsistencies remain—such as misalignment in geometry or structure—indicating that perfect reconstruction fidelity has not yet been achieved.}
    \label{fig:7_2_target_0}
\end{figure}

\begin{figure}[h]
    \centering
    \includegraphics[width=1.0\linewidth]{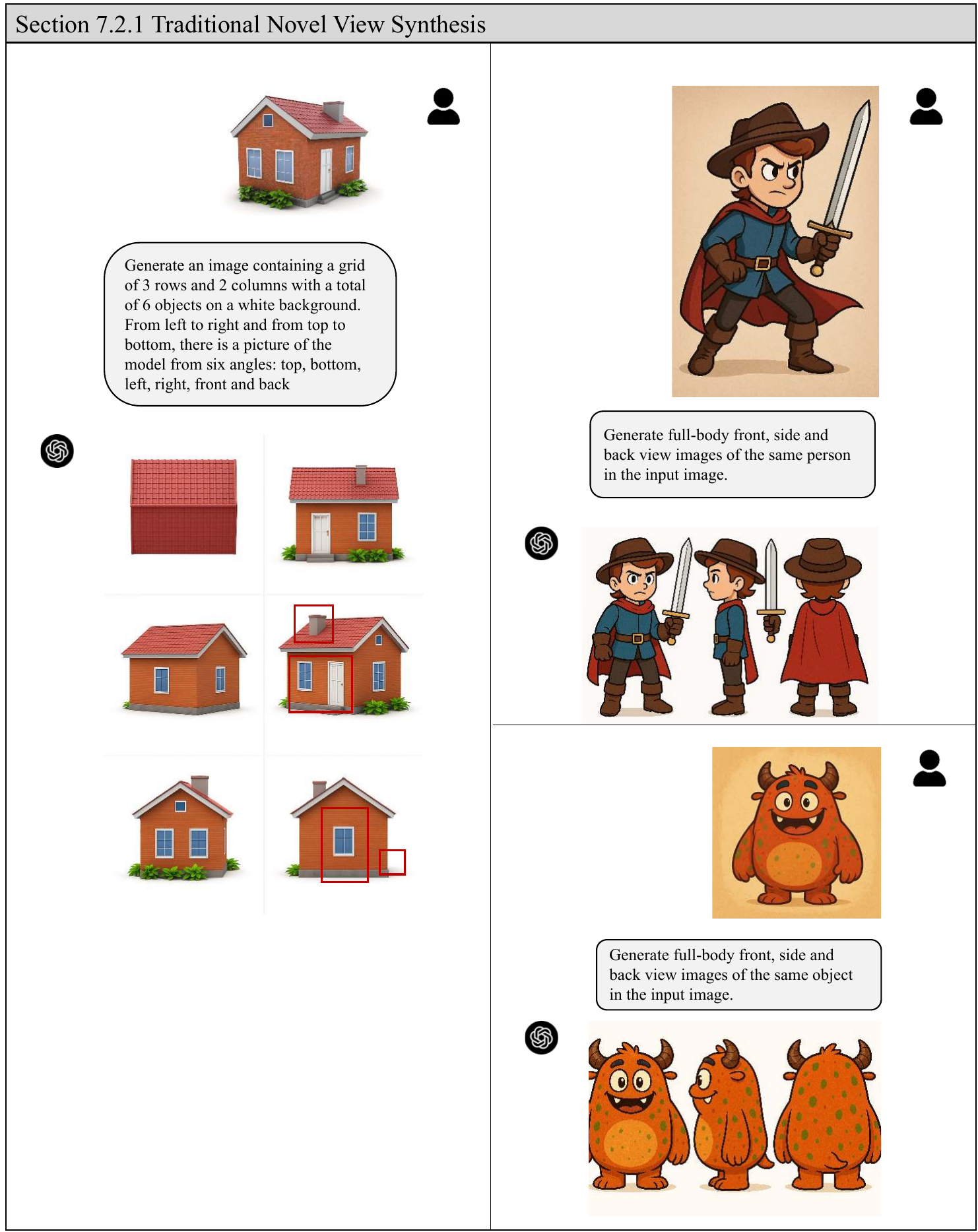}
    \caption[Sec~\ref{sec:spatial}:Novel View Synthesis]{Examples of novel-view synthesis on cartoon-style normal targets by \modelname. The model generates coherent and visually consistent images from multiple new viewpoints, capturing the overall structure and appearance of stylized characters and objects. Despite minor geometric inaccuracies in complex shapes (e.g., roof or window details), the outputs demonstrate strong spatial reasoning and effective viewpoint control in simplified, non-photorealistic domains.}
    \label{fig:7_2_target_1}
\end{figure}

\begin{figure}[h]
    \centering
    \includegraphics[width=1.0\linewidth]{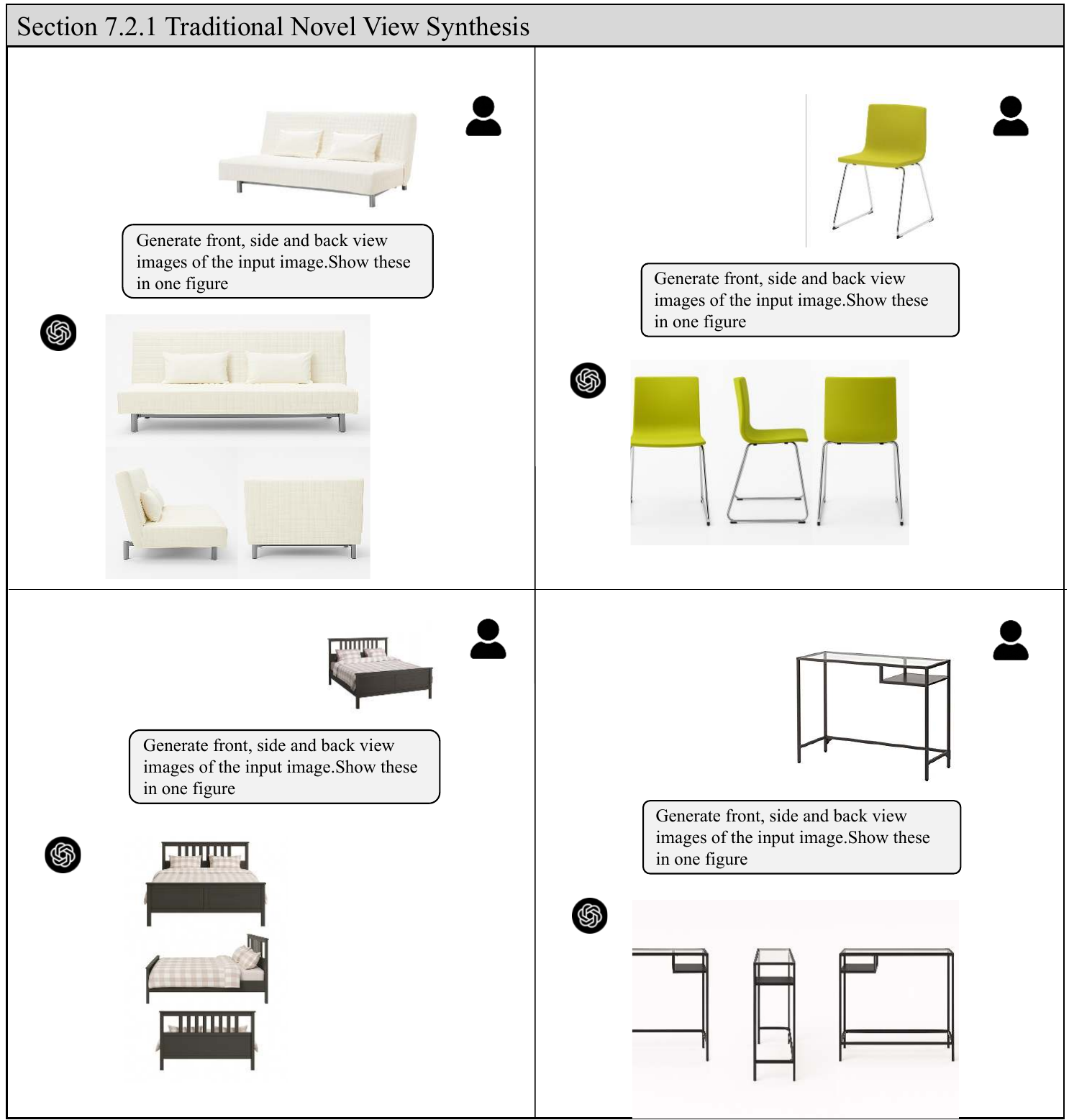}
    \caption[Sec~\ref{sec:spatial}:Novel View Synthesis]{Examples of novel-view synthesis on structurally simple normal targets by \modelname. Given a single input image, the model accurately reconstructs front, side, and back views of furniture items with clean lines and regular shapes. The outputs exhibit strong geometric consistency and precise viewpoint transitions, demonstrating \modelname’s capability to handle objects with simple, well-defined structures in controlled settings.}
    \label{fig:7_2_target_2}
\end{figure}

\textbf{Normal Novel-view Synthesis.}
As shown in Fig~\ref{fig:7_2_target_0},~\ref{fig:7_2_target_1},~\ref{fig:7_2_target_2}, \modelname demonstrates reasonable viewpoint control in novel view synthesis of common and simple targets. However, the generated results still lack perfect 3D consistency.

\begin{figure}[h]
    \centering
    \includegraphics[width=1.0\linewidth]{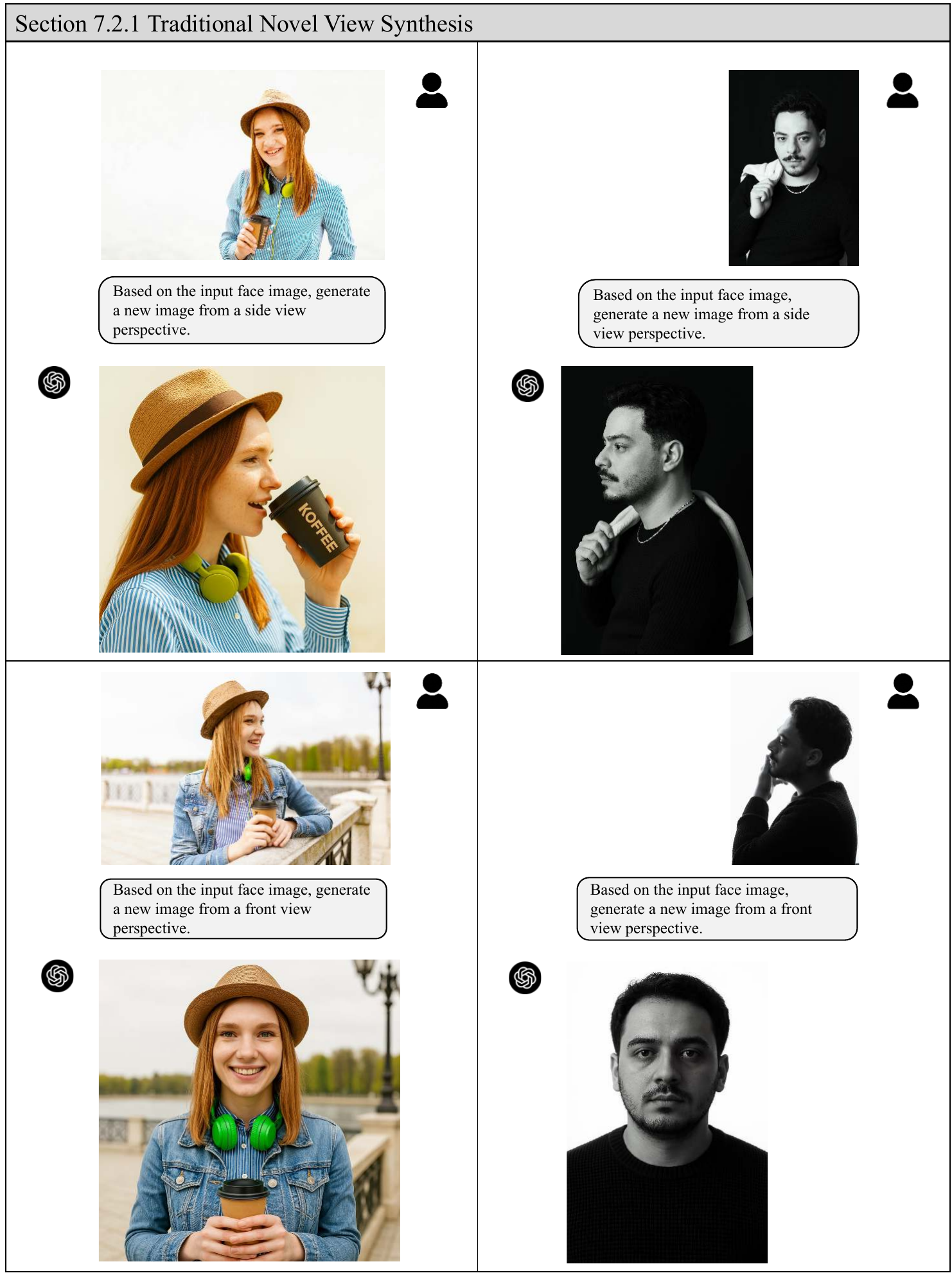}
    \caption[Sec~\ref{sec:spatial}:Novel View Synthesis]{Examples of novel-view synthesis on face portrait (ID-style) images by \modelname. The model generates realistic and identity-consistent views from both front and side perspectives based on a single input image. Despite occasional artifacts in facial features or accessories, the overall viewpoint transformations demonstrate \modelname’s strong 3D understanding of human faces.}
    \label{fig:7_2_human_0}
\end{figure}

\begin{figure}[h]
    \centering
    \includegraphics[width=1.0\linewidth]{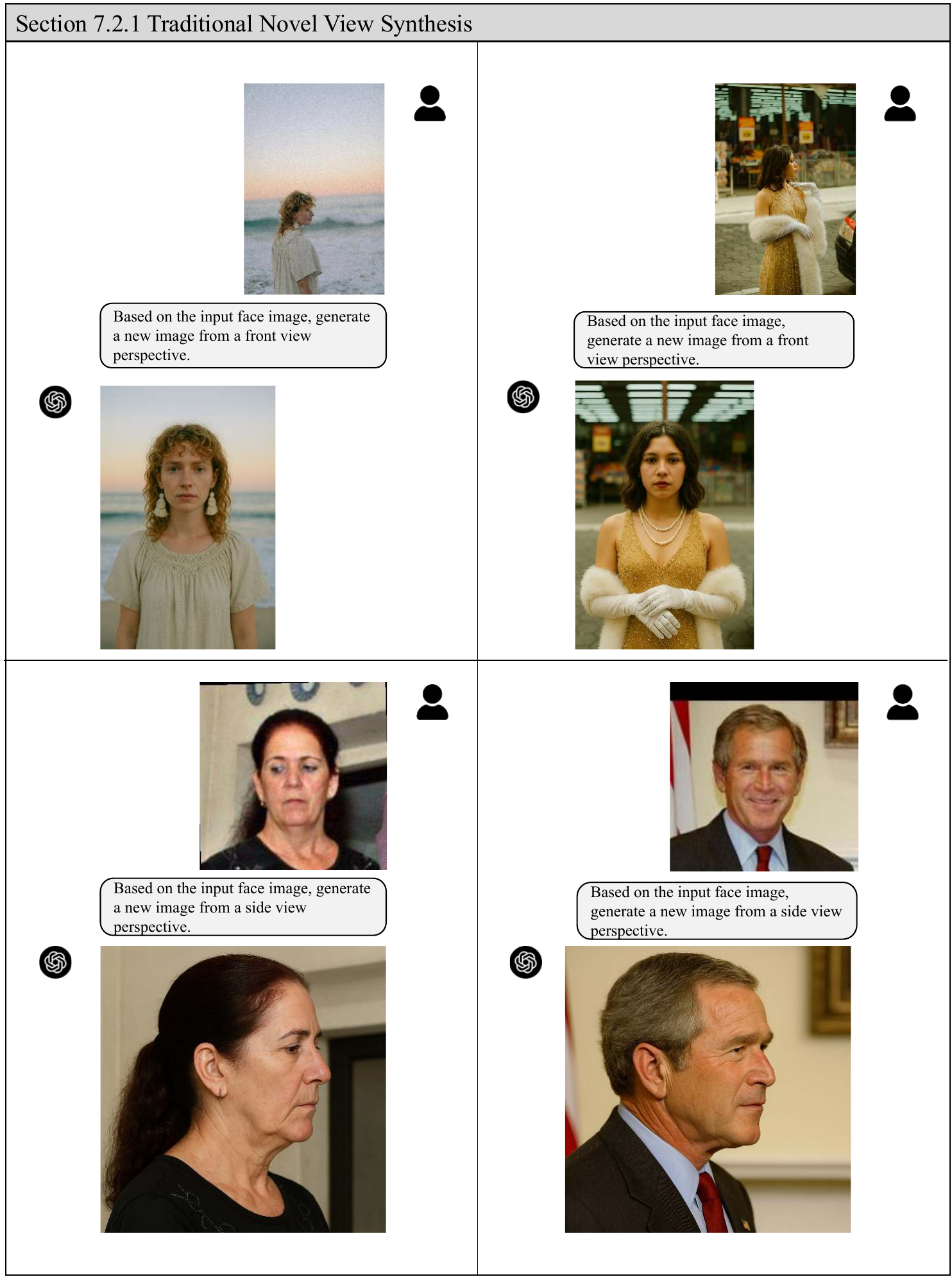}
    \caption[Sec~\ref{sec:spatial}:Novel View Synthesis]{Additional examples of novel-view synthesis on face portrait (ID-style) images by \modelname. The model produces front and side views that are generally consistent with the input identity, showing a solid grasp of 3D facial structure with only minor variations.}
    \label{fig:7_2_human_1}
\end{figure}

\begin{figure}[h]
    \centering
    \includegraphics[width=1.0\linewidth]{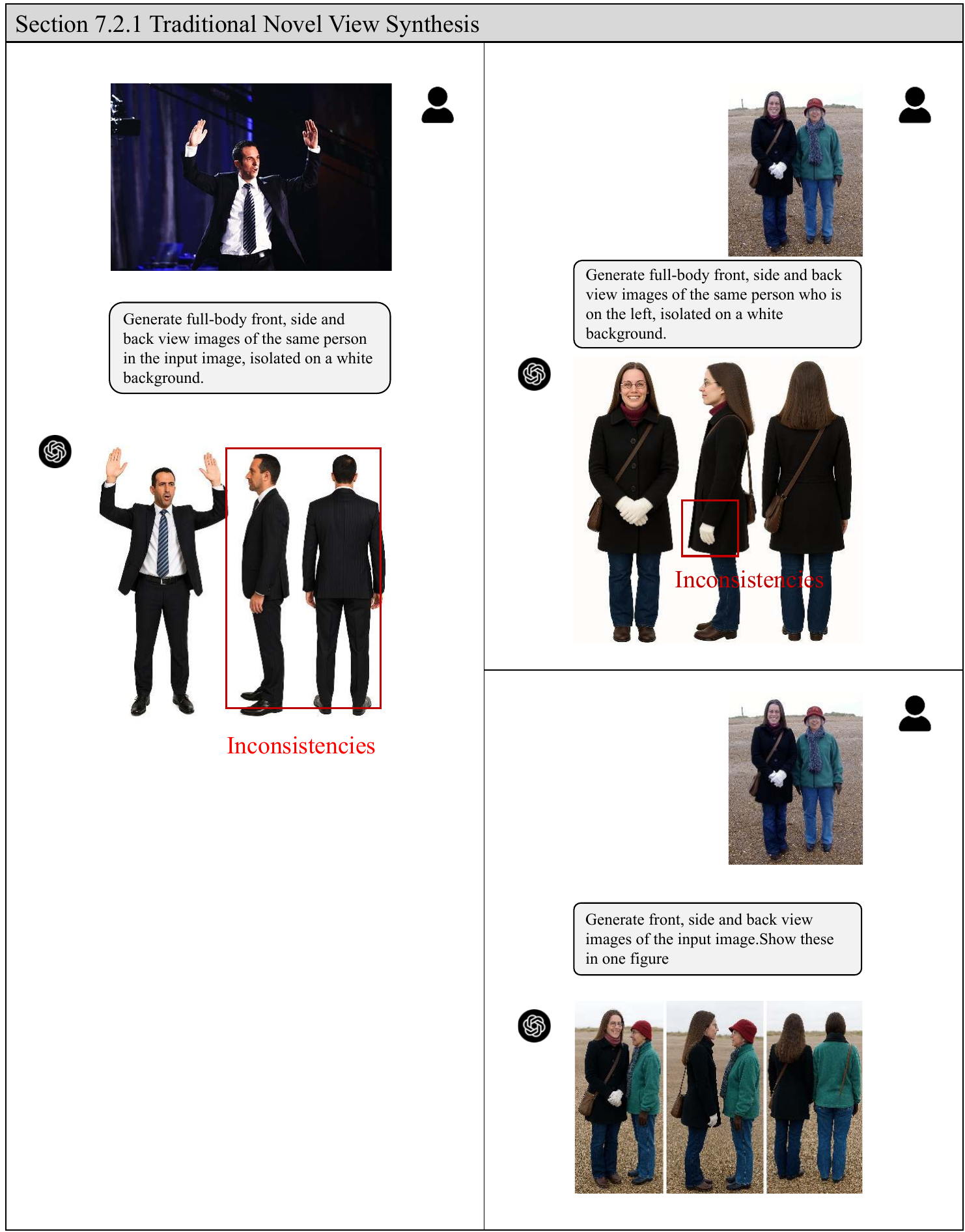}
    \caption[Sec~\ref{sec:spatial}:Novel View Synthesis]{Examples of novel-view synthesis on full-body human images by \modelname. The model is able to generate front, side, and back views that broadly preserve body posture and clothing details. However, inconsistencies in limb position, accessories, and clothing geometry are still common, highlighting the challenge of achieving precise spatial alignment in full-body reconstructions.}
    \label{fig:7_2_human_2}
\end{figure}

\textbf{Human-centric Novel-view Synthesis. }
Next, we evaluate \modelname’s ability to generate novel views of humans, including both full-body figures and facial portraits. Surprisingly, despite some remaining inconsistencies, \modelname demonstrates an impressive level of 3D understanding when it comes to human subjects, as shown in Fig~\ref{fig:7_2_human_0},~\ref{fig:7_2_human_1},~\ref{fig:7_2_human_2}.

\begin{figure}[h]
    \centering
    \includegraphics[width=1.0\linewidth]{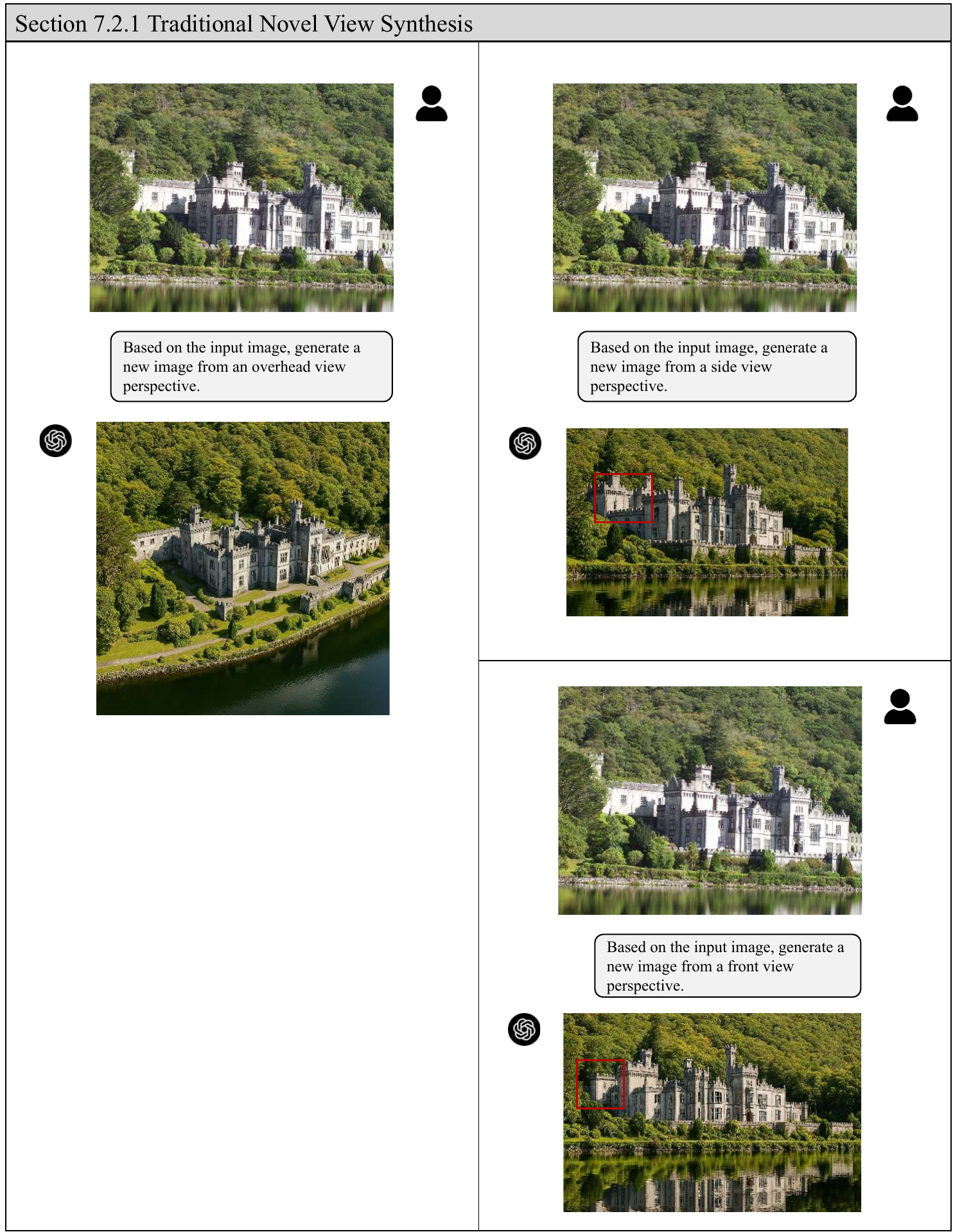}
    \caption[Sec~\ref{sec:spatial}:Novel View Synthesis]{Examples of novel-view synthesis on complex outdoor scenes by \modelname. The model produces plausible overhead, side, and front views from a single input image, preserving the overall structure of the scene. However, structural inaccuracies emerge under large viewpoint shifts, revealing the model’s limited ability to maintain full 3D consistency in complex environments.}
    \label{fig:7_2_scene_0}
\end{figure}

\begin{figure}[h]
    \centering
    \includegraphics[width=1.0\linewidth]{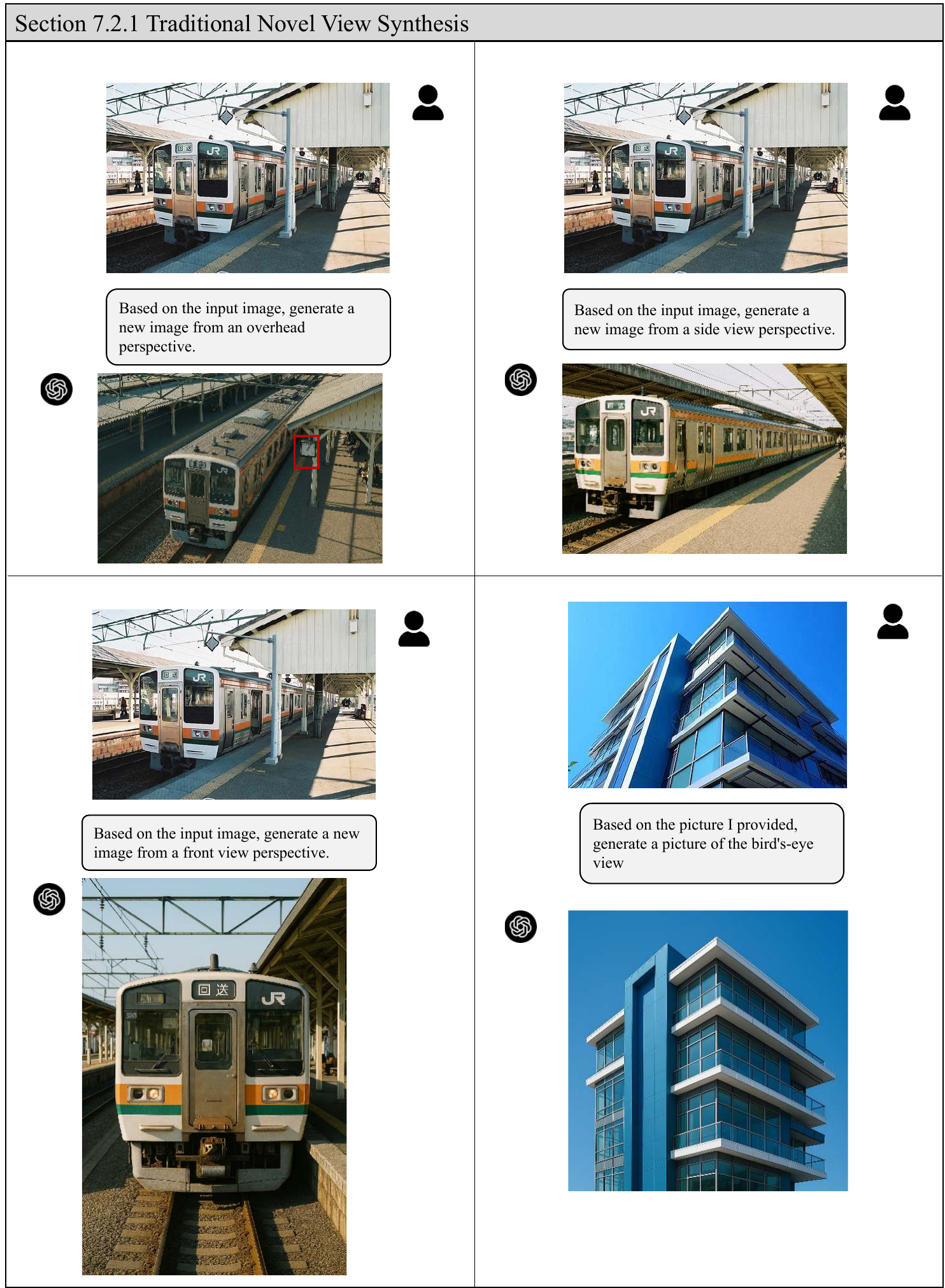}
    \caption[Sec~\ref{sec:spatial}:Novel View Synthesis]{More examples of novel-view synthesis on complex scenes by \modelname. The model is able to preserve overall structure and visual fidelity across different viewpoints. However, the generated outputs often suffer from inaccurate viewpoint alignment, suggesting that \modelname still faces challenges in precisely controlling camera angles in complex real-world scenes.}
    \label{fig:7_2_scene_1}
\end{figure}

\textbf{Scene-centric Novel-view Synthesis. }
We assess \modelname’s novel-view synthesis capabilities on complex outdoor scenes, as illustrated in Fig~\ref{fig:7_2_scene_0},~\ref{fig:7_2_scene_1}. The results show that, while the model maintains a relatively strong degree of 3D consistency, \modelname fails to handle significant viewpoint shifts effectively, highlighting its limited capacity for 3D understanding in complex scene settings.

\clearpage
\clearpage

\clearpage
\clearpage
\clearpage

\clearpage

\clearpage

\clearpage

\clearpage

\subsubsection{View Synthesis under Embodied Scene}
Viewpoint transformation in embodied or indoor environments is particularly critical for embodied intelligence research. Therefore, we specifically evaluate \modelname’s spatial understanding based on indoor scene images and layouts. Through our test, \modelname demonstrates strong viewpoint control but relatively weak consistency across views.

\begin{figure}[h]
    \centering
    \includegraphics[width=1.0\linewidth]{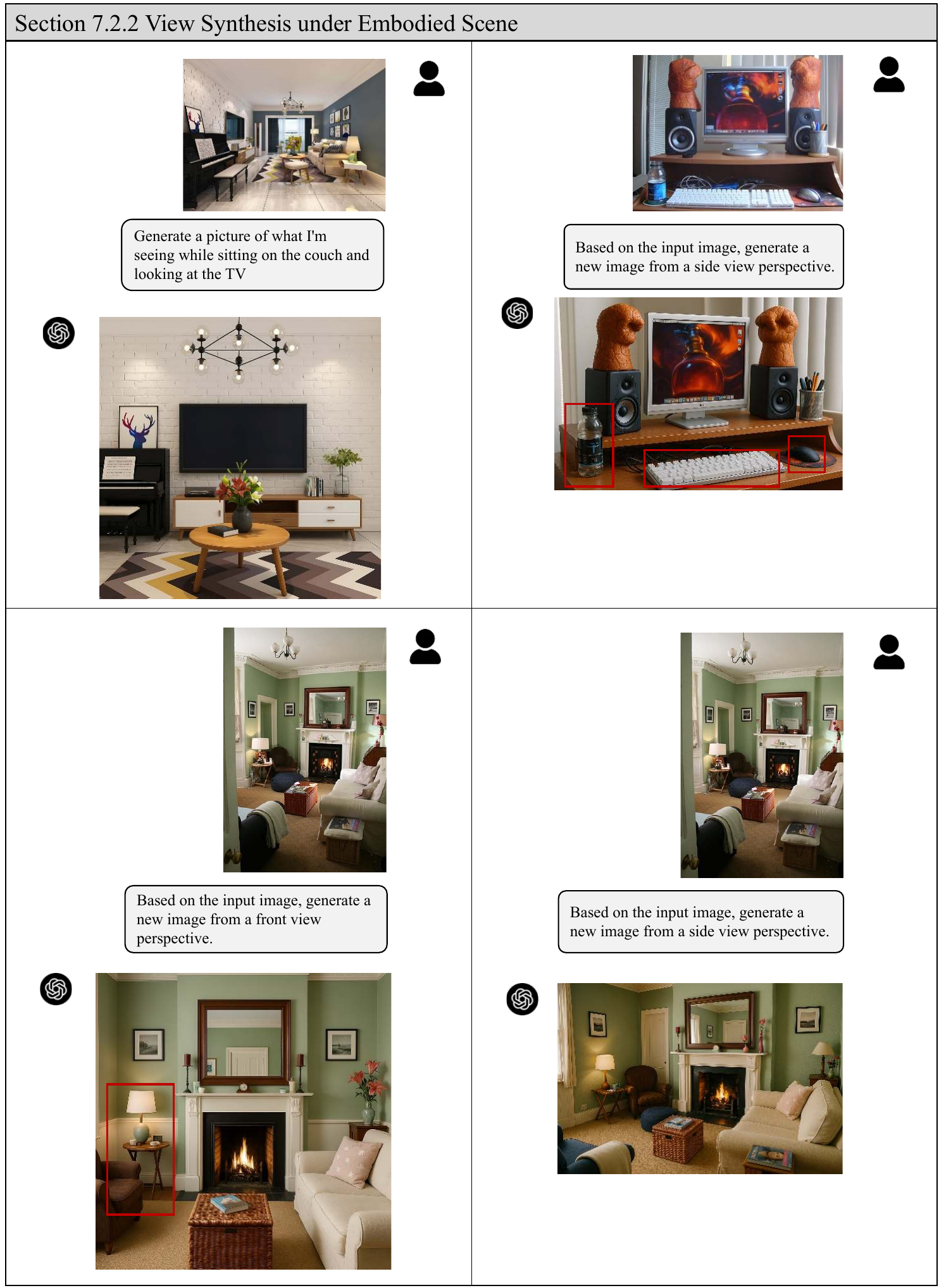}
    \caption[Sec~\ref{sec:spatial}:Novel View Synthesis]{Examples of novel-view synthesis on indoor (embodied) scenes by \modelname. The model shows strong viewpoint control and can generate plausible views from new perspectives based on a single input. However, inconsistencies in object placement and appearance—such as furniture, lighting, and layout—are still noticeable, indicating limited spatial consistency in complex embodied environments.}
    \label{fig:7_2_embodied_0}
\end{figure}

\textbf{Image-based View Synthesis.}
As shown in Fig.~\ref{fig:7_2_embodied_0}, we evaluate \modelname on a set of tasks where the model is asked to generate images from new viewpoints based on a single input photograph of an indoor scene. The model shows promising performance in generating visually plausible results from new perspectives, demonstrating a reasonable grasp of camera positioning and viewpoint shift. However, the generated outputs still exhibit noticeable inconsistencies in spatial alignment. In particular, some objects undergo geometric distortion, positional shifts, or changes in appearance. For instance, the same lamp or keyboard may appear altered or misplaced between the original and generated view, indicating that while \modelname is capable of inferring approximate scene layouts, it lacks fine-grained 3D scene understanding and consistency-preserving mechanisms.

\begin{figure}[h]
    \centering
    \includegraphics[width=1.0\linewidth]{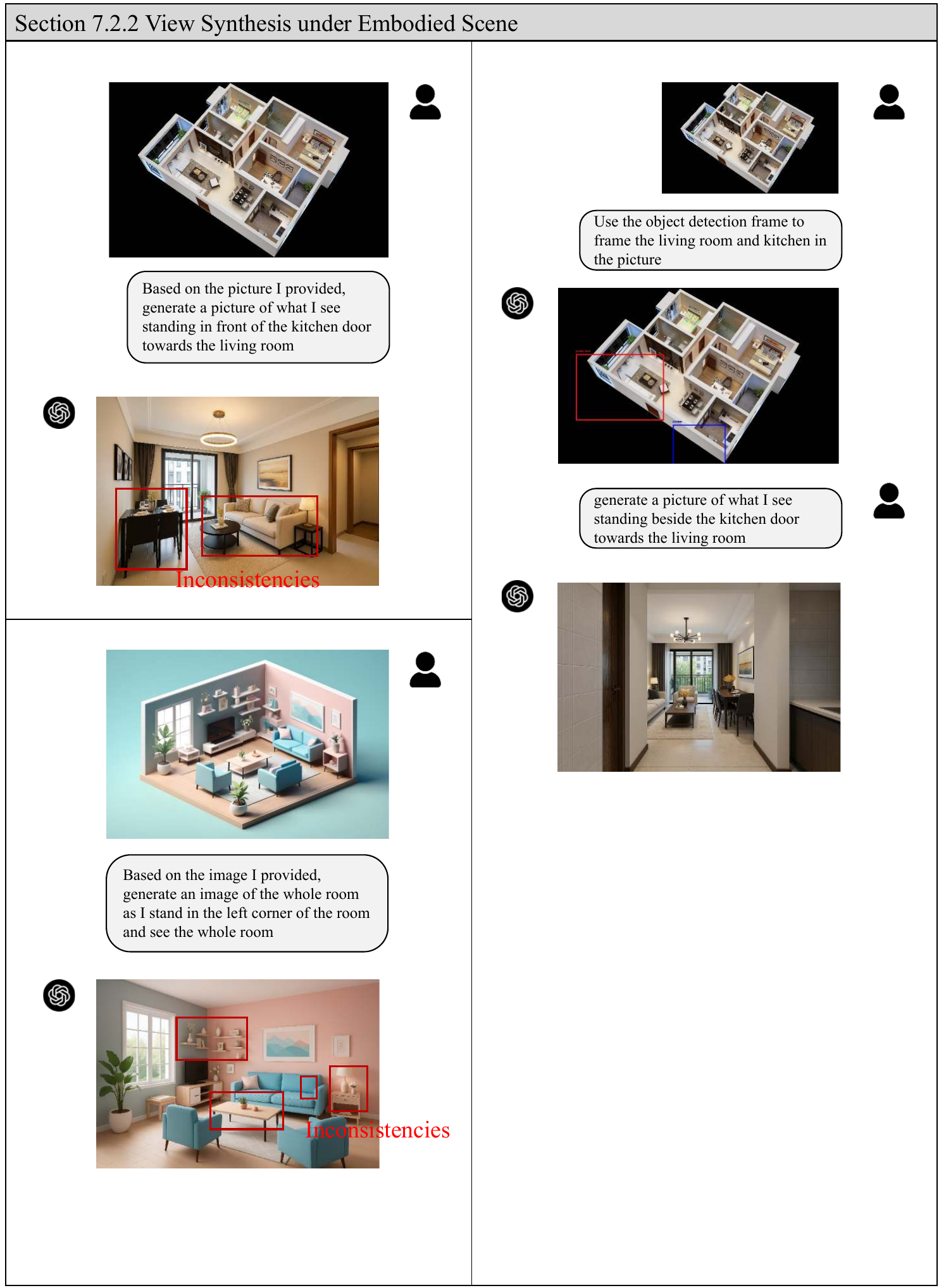}
    \caption[Sec~\ref{sec:spatial}:Novel View Synthesis]{Examples of novel-view synthesis based on layout images by \modelname. The model can generate plausible first-person perspectives from specific locations in the scene, conditioned on 3D floor plans or stylized layouts. While the overall spatial arrangement is preserved, the generated views often contain object-level inaccuracies, indicating challenges in fine-grained grounding and geometric precision from abstract layout inputs.}
    \label{fig:7_2_embodied_1}
\end{figure}

\textbf{Layout-based View Synthesis.}
As shown in Fig.~\ref{fig:7_2_embodied_1}, we further evaluate \modelname using top-down scene layouts or stylized 3D room schematics as input conditions. The goal is to synthesize a plausible first-person perspective from a described location (e.g., “standing at the kitchen door looking at the living room”). \modelname successfully generates semantically reasonable images and often captures the high-level spatial arrangement outlined in the layout. Nonetheless, there are notable issues at the object level, including furniture discrepancies, missing elements, and incorrect room proportions. These problems highlight the difficulty in faithfully grounding generation based on abstract scene layouts and converting spatial instructions into accurate visual reconstructions.

Overall, these experiments demonstrate that while \modelname possesses a strong ability for global viewpoint control, its internal scene representations remain limited. The model often fails to maintain object permanence and spatial consistency across views, revealing a lack of true 3D reasoning capabilities. Improving these aspects is crucial for tasks such as embodied AI, navigation, and robotic perception, where viewpoint fidelity and spatial grounding are essential.

\clearpage
\clearpage

\clearpage

\subsection{Spatial Reasoning}
The spatial reasoning task~\cite{chen2024spatialvlm, cheng2024spatialrgpt, wang2024picture} evaluates a model’s ability to interpret and visually realize spatial relationships described in a textual prompt. In this task, the model is asked to generate images that accurately reflect specific object placements or relative spatial configurations—for example, “a person standing behind a tree” or “a cat sitting between two chairs.” Successfully completing this task requires not only semantic understanding of the involved entities, but also the ability to reason about depth, occlusion, and relative positioning in a physically coherent manner. As such, this task serves as a critical benchmark for assessing \modelname’s spatial awareness, particularly in terms of scene composition and depth-based reasoning.

\begin{figure}[h]
    \centering
    \includegraphics[width=1.0\linewidth]{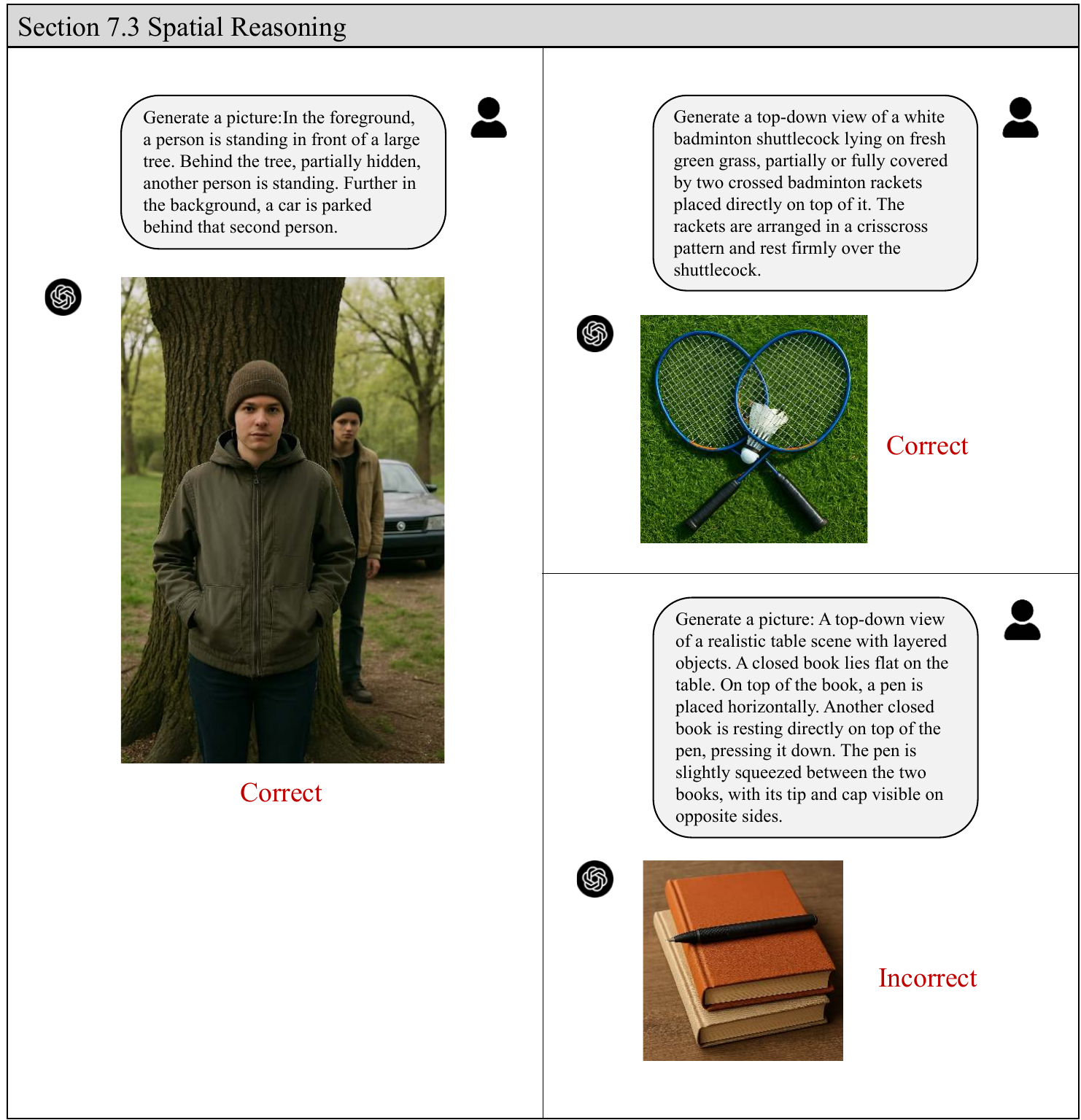}
    \caption[Sec~\ref{sec:spatial}:Spatical Reasoning]{Examples of spatial reasoning through textual prompts by \modelname. The model demonstrates a strong grasp of relative spatial relationships and occlusion in certain scenarios—such as depth-based person placement and object layering (top). However, in more complex configurations involving subtle interactions (e.g., multiple levels of layering and pressure), failures occur, suggesting limitations in fine-grained spatial composition and physical plausibility.}
    \label{fig:7_3_prompt_0}
\end{figure}

\textbf{Prompt-based Reasoning. }
As shown in Fig.~\ref{fig:7_3_prompt_0}, we evaluate the model’s ability to control spatial relationships through textual prompts. \modelname demonstrates a certain level of spatial understanding, particularly in interpreting depth-related relationships.

\begin{figure}[h]
    \centering
    \includegraphics[width=1.0\linewidth]{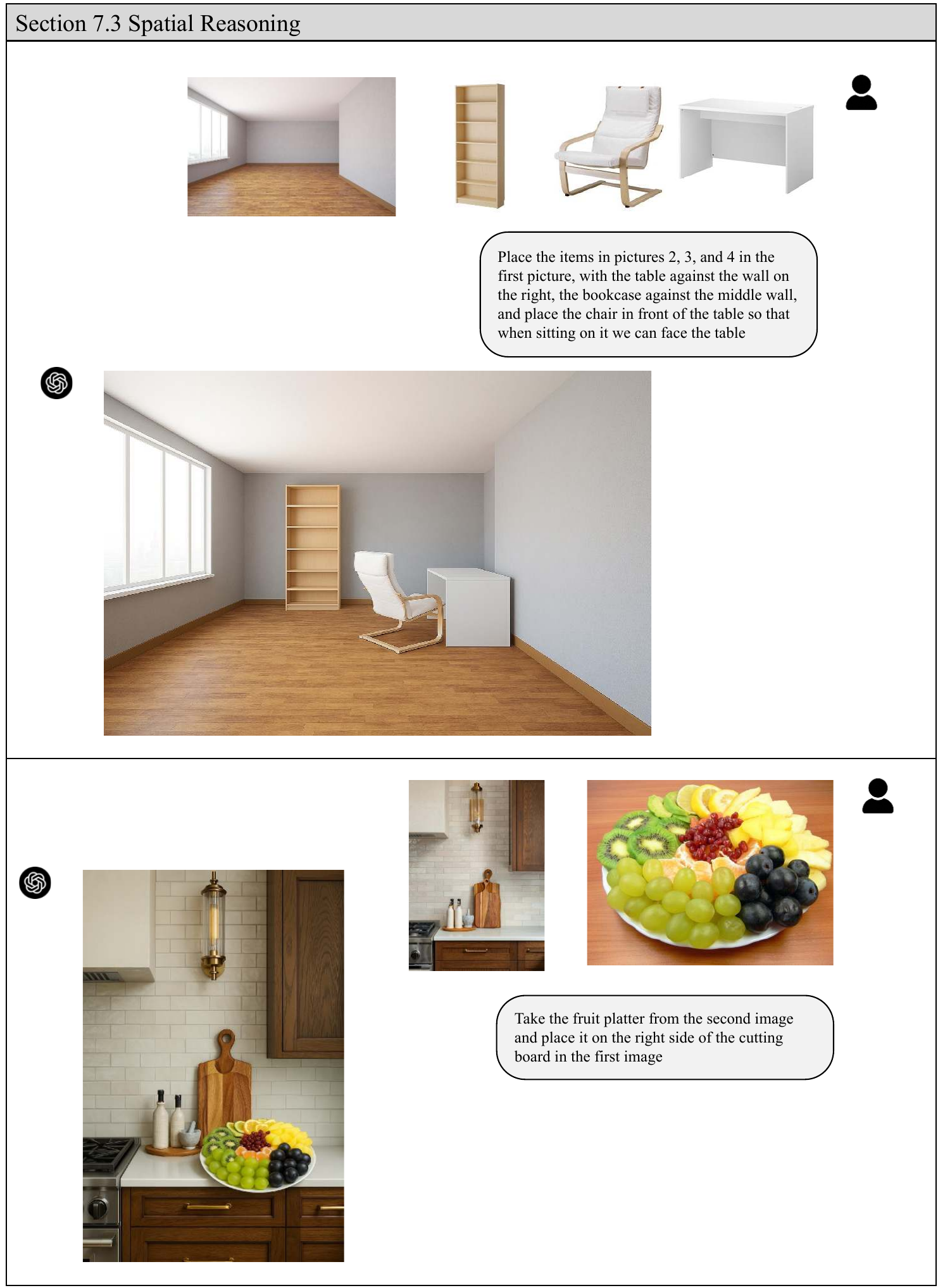}
    \caption[Sec~\ref{sec:spatial}:Spatical Reasoning]{Examples of image-based spatial reasoning by \modelname. The model successfully integrates multiple objects into a coherent scene according to visual and textual instructions, demonstrating an ability to handle basic spatial placement and alignment. While overall composition is reasonable, some positional inaccuracies remain, indicating room for improvement in precise spatial grounding and instruction following.}
    \label{fig:7_3_image_0}
\end{figure}

\begin{figure}[h]
    \centering
    \includegraphics[width=1.0\linewidth]{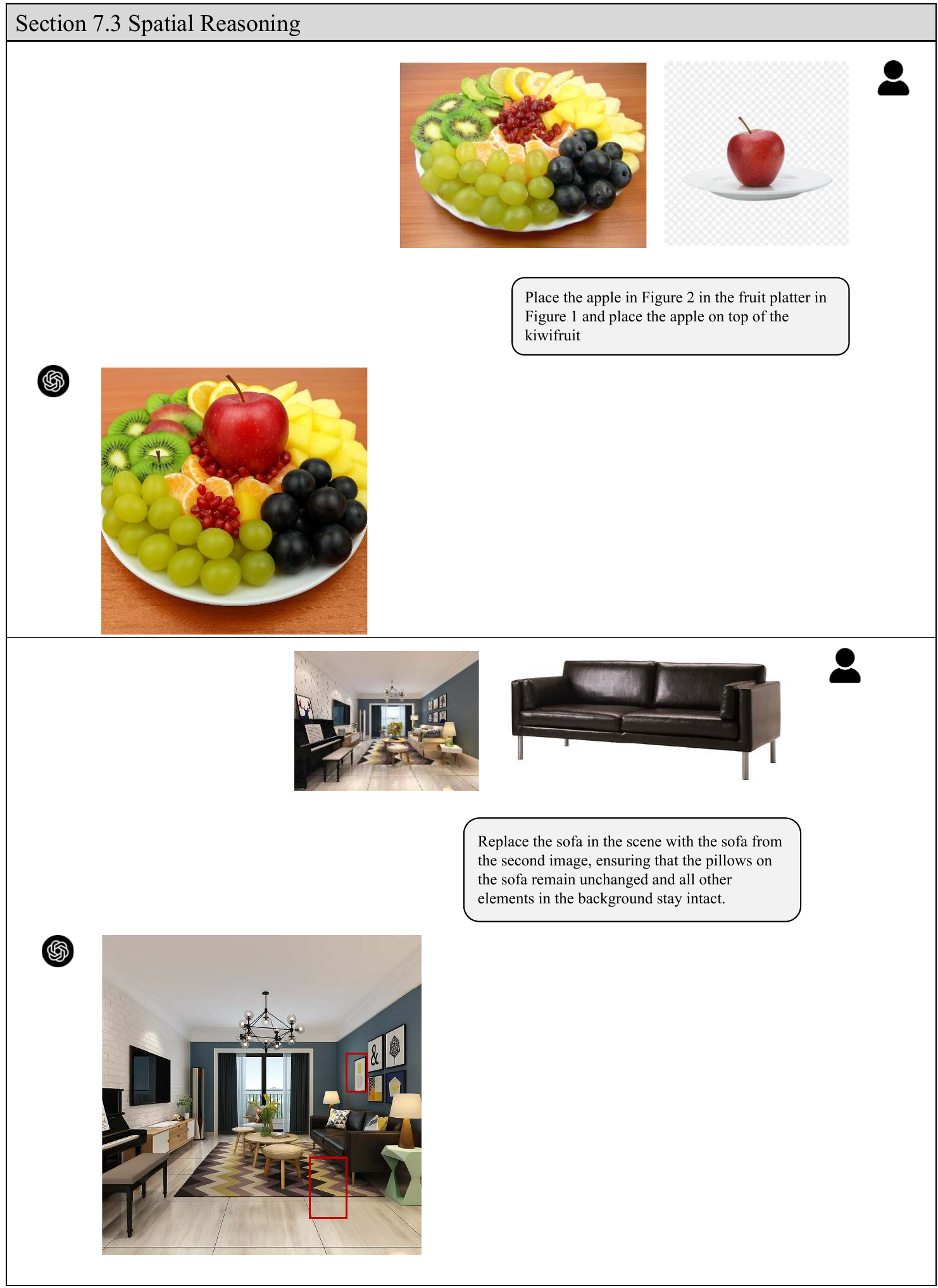}
    \caption[Sec~\ref{sec:spatial}:Spatical Reasoning]{More examples of image-based spatial reasoning by \modelname. The model is able to follow compositional instructions such as object placement and substitution with reasonable success. While the apple is correctly added with appropriate layering, the sofa replacement introduces minor inconsistencies in lighting and background alignment, indicating limitations in fine-grained control and global scene consistency.}
    \label{fig:7_3_image_1}
\end{figure}

\begin{figure}[h]
    \centering
    \includegraphics[width=1.0\linewidth]{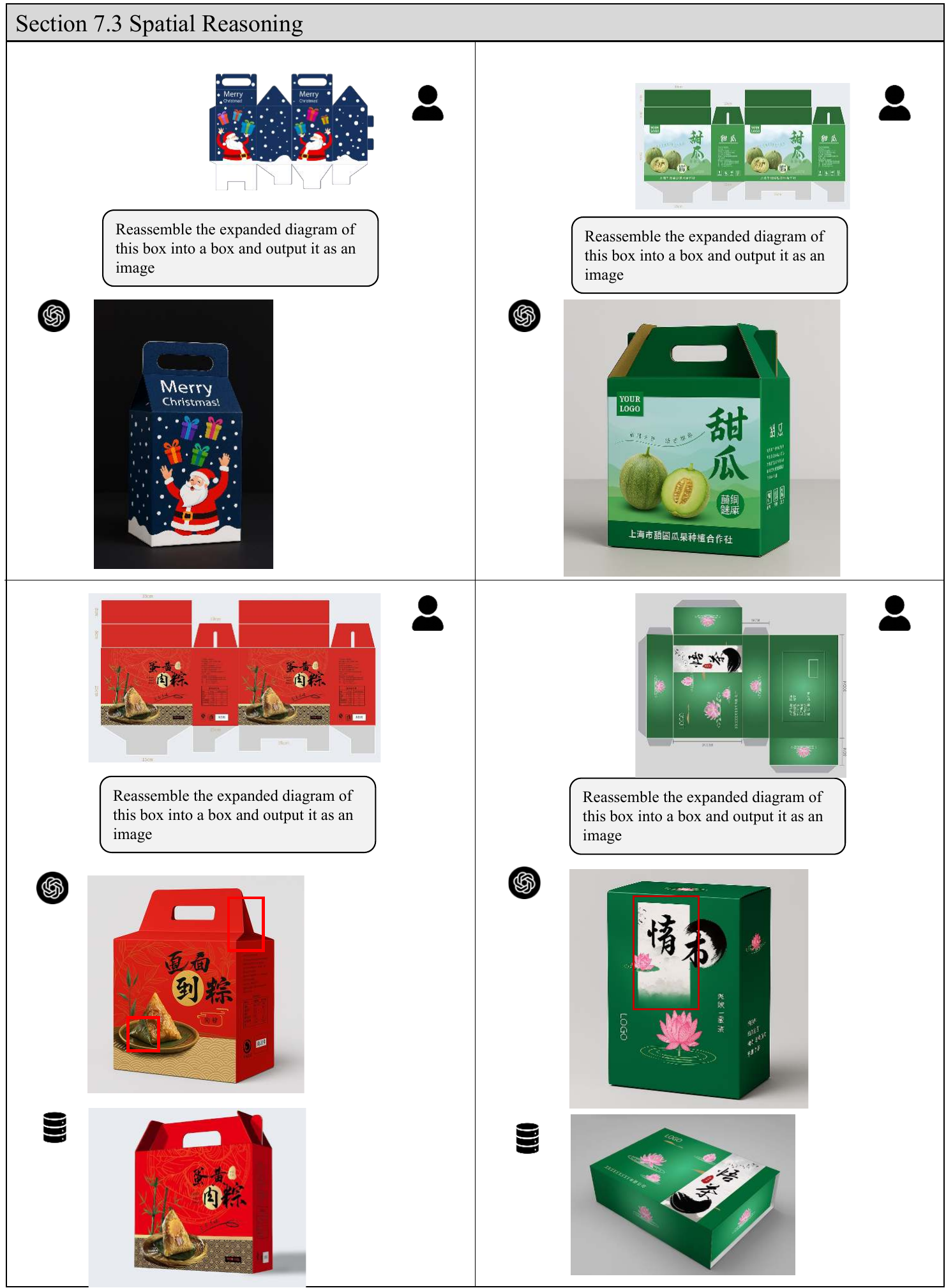}
    \caption[Sec~\ref{sec:spatial}:Spatical Reasoning]{Examples of spatial reasoning through 2D-to-3D reassembly by \modelname. Given flat box layouts, the model is able to generate plausible 3D box renderings, capturing overall structure, folding logic, and surface continuity. While most reassembled results are visually coherent, minor inconsistencies in texture alignment and panel orientation still occur, reflecting the challenge of precise spatial reconstruction from UV-like input maps.}
    \label{fig:7_3_image_2}
\end{figure}

\begin{figure}[h]
    \centering
    \includegraphics[width=1.0\linewidth]{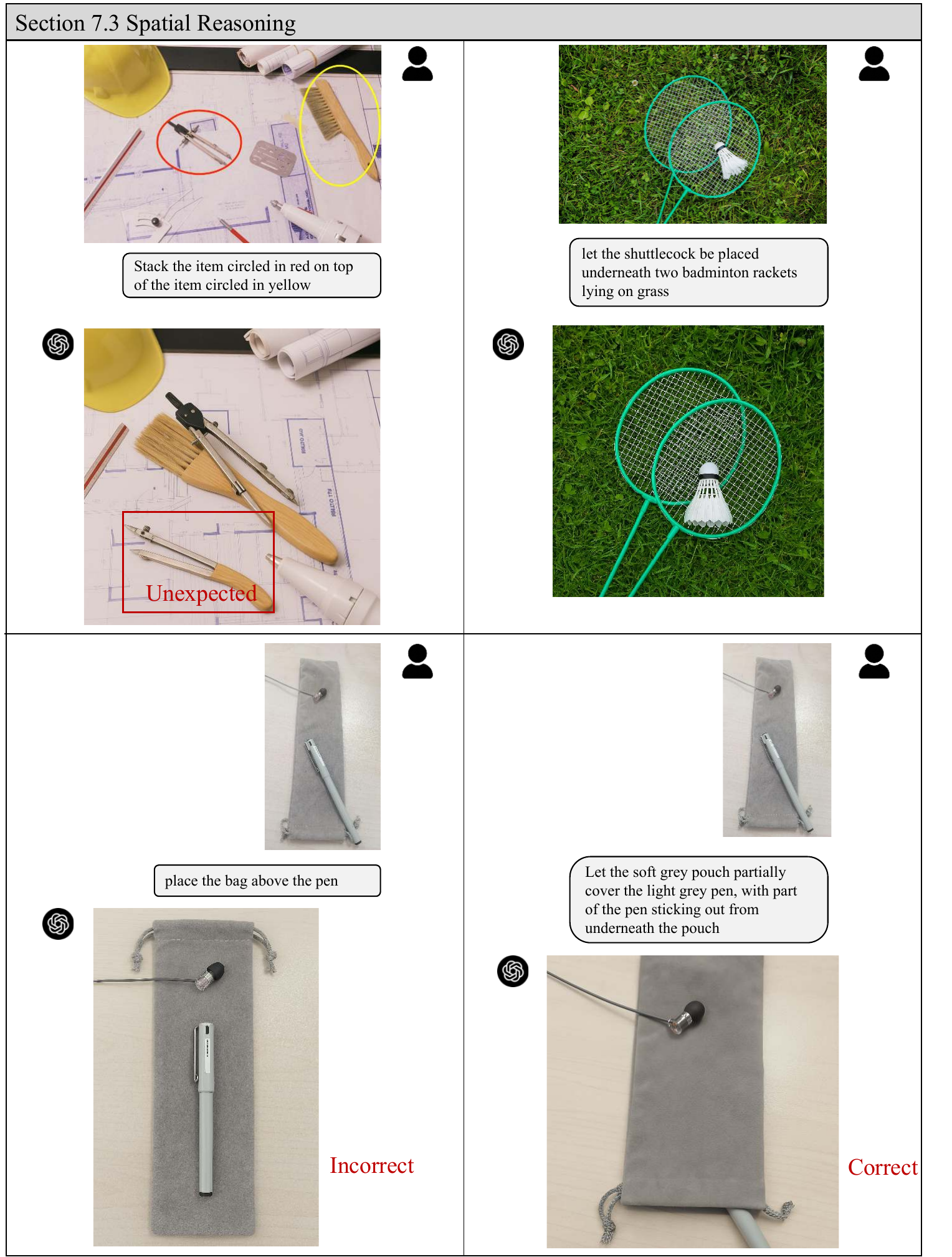}
    \caption[Sec~\ref{sec:spatial}:Spatical Reasoning]{Examples of spatial reasoning through localized object editing by \modelname. The model shows partial success in interpreting fine-grained spatial instructions involving occlusion, stacking, and partial covering. While some generations correctly reflect the intended relationships (e.g., partial occlusion), others misinterpret layering or spatial order, indicating ongoing challenges in precise object-level spatial manipulation.}
    \label{fig:7_3_image_3}
\end{figure}

\textbf{Image-based Reasoning. }
As shown in Fig.~\ref{fig:7_3_image_0},~\ref{fig:7_3_image_1},~\ref{fig:7_3_image_2},~\ref{fig:7_3_image_3}, we test \modelname’s ability to arrange multiple objects in space based on a given prompt. The model demonstrates promising capabilities in composing multiple targets within a scene, but still exhibits inconsistencies with the specified spatial instructions.

\clearpage

\clearpage

\clearpage

\clearpage

\clearpage

%\subsubsection{Challenging Spatial Reasoning Generation}
%To probe the boundaries of \modelname’s spatial reasoning ability, we constructed a series of particularly challenging generation tasks that demand accurate interpretation of complex spatial instructions. As illustrated in Fig~\ref{}, \modelname was unable to successfully complete these tasks.
%\label{sec:multiview}

\section{Temporally-aware Image Generation}
\label{sec:temporal}

Temporally-aware image generation~\cite{li2024image, lyu2024image} is a critical task for assessing a model’s ability to maintain coherence over time and understand dynamic visual processes. In this task, the model is required to generate a sequence of images that depict consistent temporal evolution—such as motion, transformation, or progression—based on a given series of frame images. This setup enables the evaluation of \modelname’s understanding of causality, temporal continuity, and object permanence, all of which are essential for applications involving video generation, animation, and real-world scene understanding.
We evaluate \modelname's capability across three representative tasks: future frame prediction, intermediate frame prediction, and past frame prediction. Each task provides distinct challenges in terms of temporal reasoning and spatial consistency.

\begin{figure}[h]
    \centering
    \includegraphics[width=1.0\linewidth]{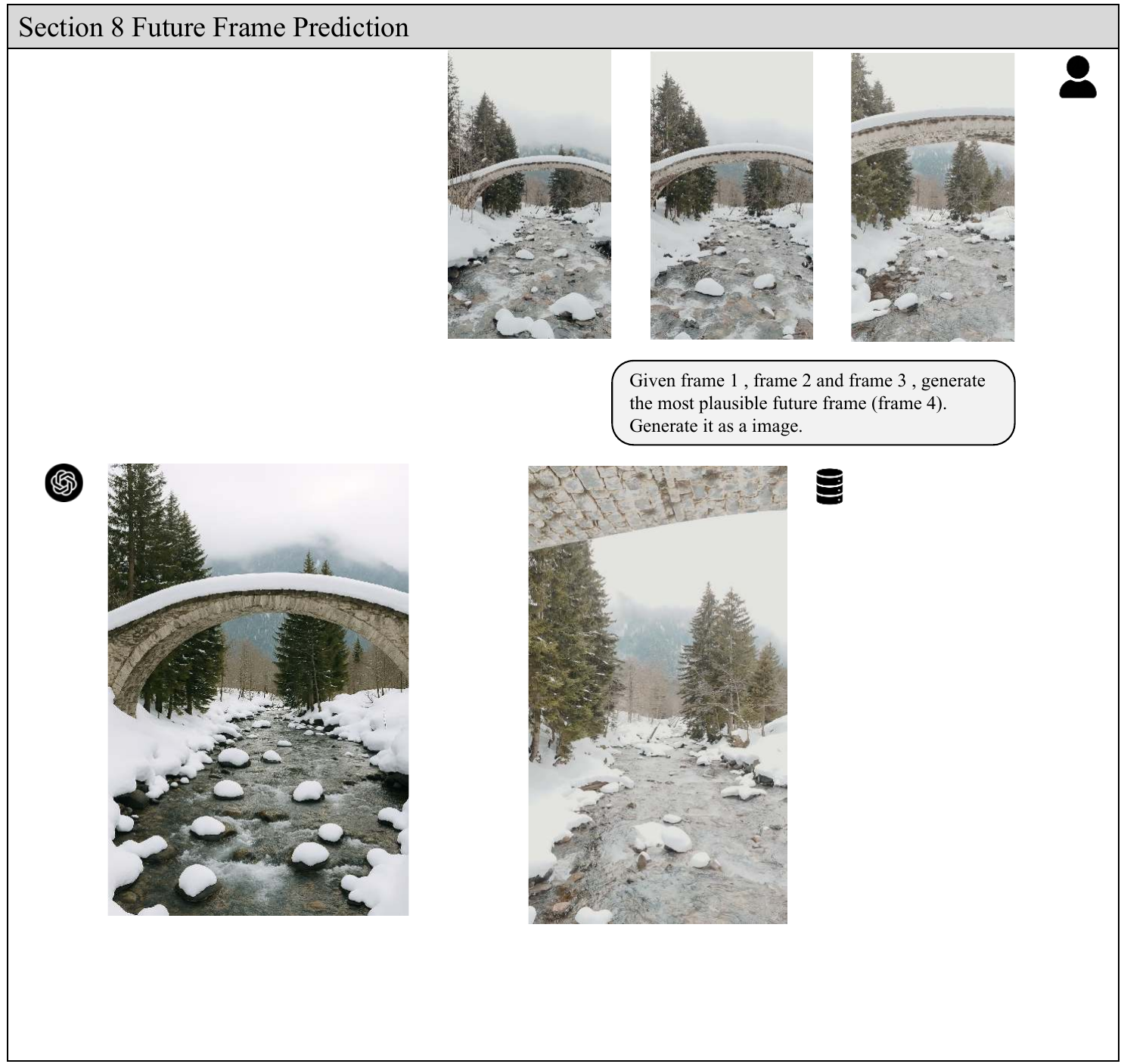}
    \caption[Sec~\ref{sec:temporal}: Future Frame Prediction]{Example of future frame prediction by \modelname. Given three input frames showing a gradual camera movement under a bridge, the model attempts to generate the next plausible frame. While the general scene layout is preserved, the predicted frame fails to accurately continue the motion trajectory of the bridge, indicating limited temporal understanding of the scene’s main structure. }
    \label{fig:8_1_0}
\end{figure}

\begin{figure}[h]
    \centering
    \includegraphics[width=1.0\linewidth]{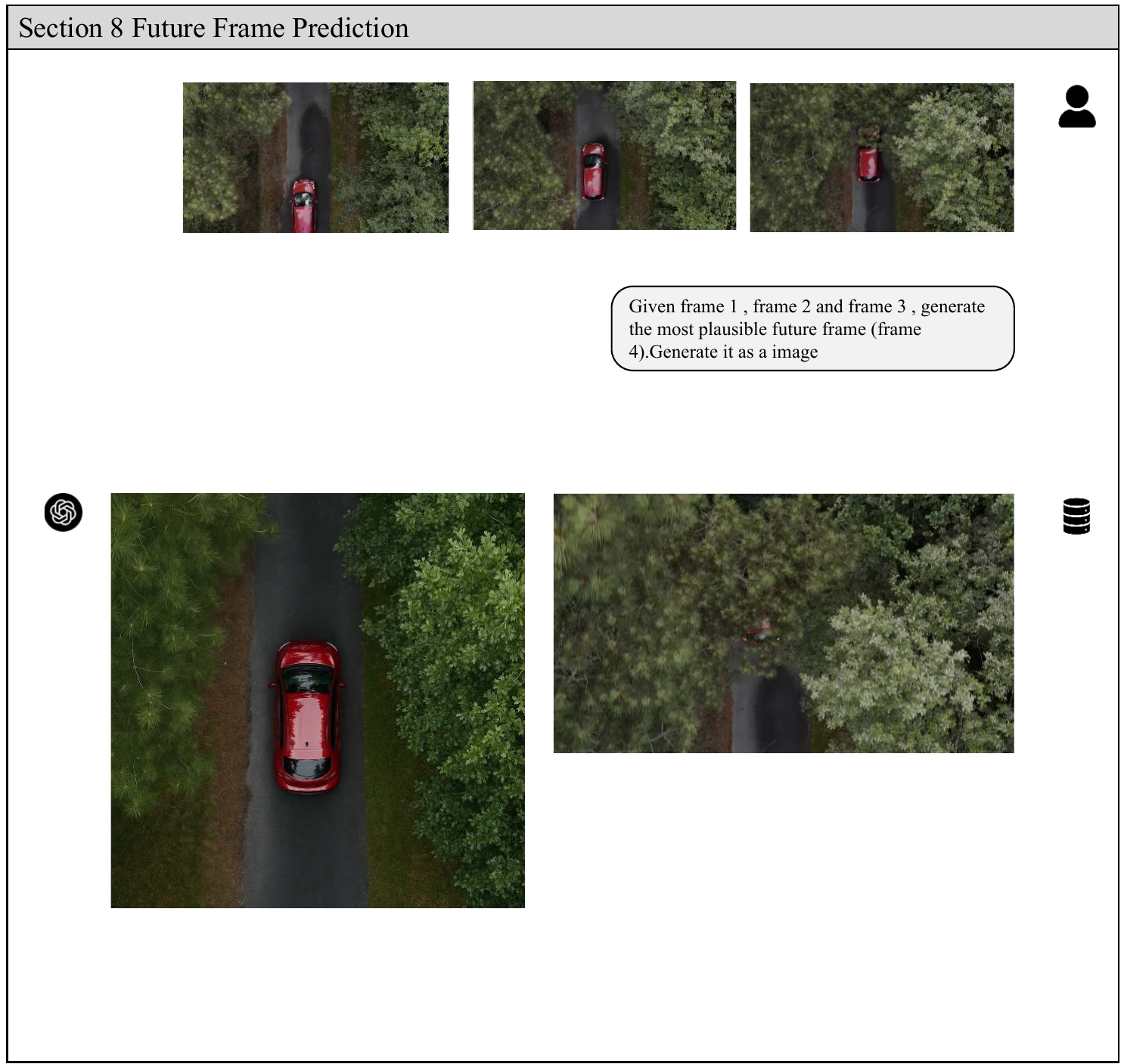}
    \caption[Sec~\ref{sec:temporal}: Future Frame Prediction]{Example of future frame prediction by \modelname. Although the model produces a visually plausible image, the predicted frame does not reflect the expected temporal progression of the moving vehicle. Instead, it appears to focus solely on reconstructing a static scene, lacking awareness of motion cues present in the input sequence.}
    \label{fig:8_1_1}
\end{figure}

\begin{figure}[h]
    \centering
    \includegraphics[width=1.0\linewidth]{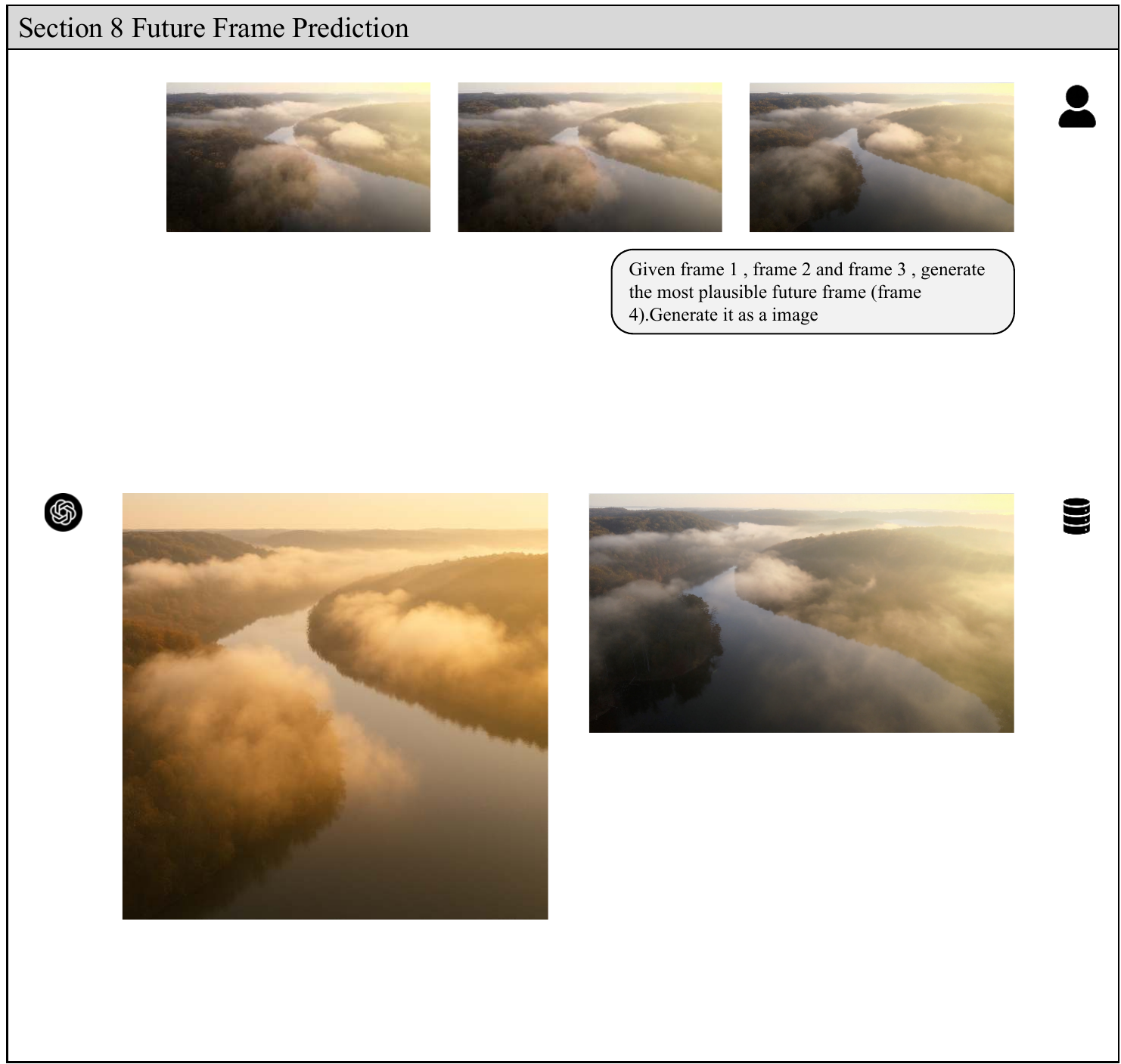}
    \caption[Sec~\ref{sec:temporal}: Future Frame Prediction]{Example of future frame prediction by \modelname. While the generated image maintains overall scene appearance and lighting, it fails to accurately capture the motion trajectory of the drifting clouds observed in the input frames. This suggests limited capability in modeling dynamic elements within temporally evolving natural scenes.}
    \label{fig:8_1_2}
\end{figure}

\begin{figure}[h]
    \centering
    \includegraphics[width=1.0\linewidth]{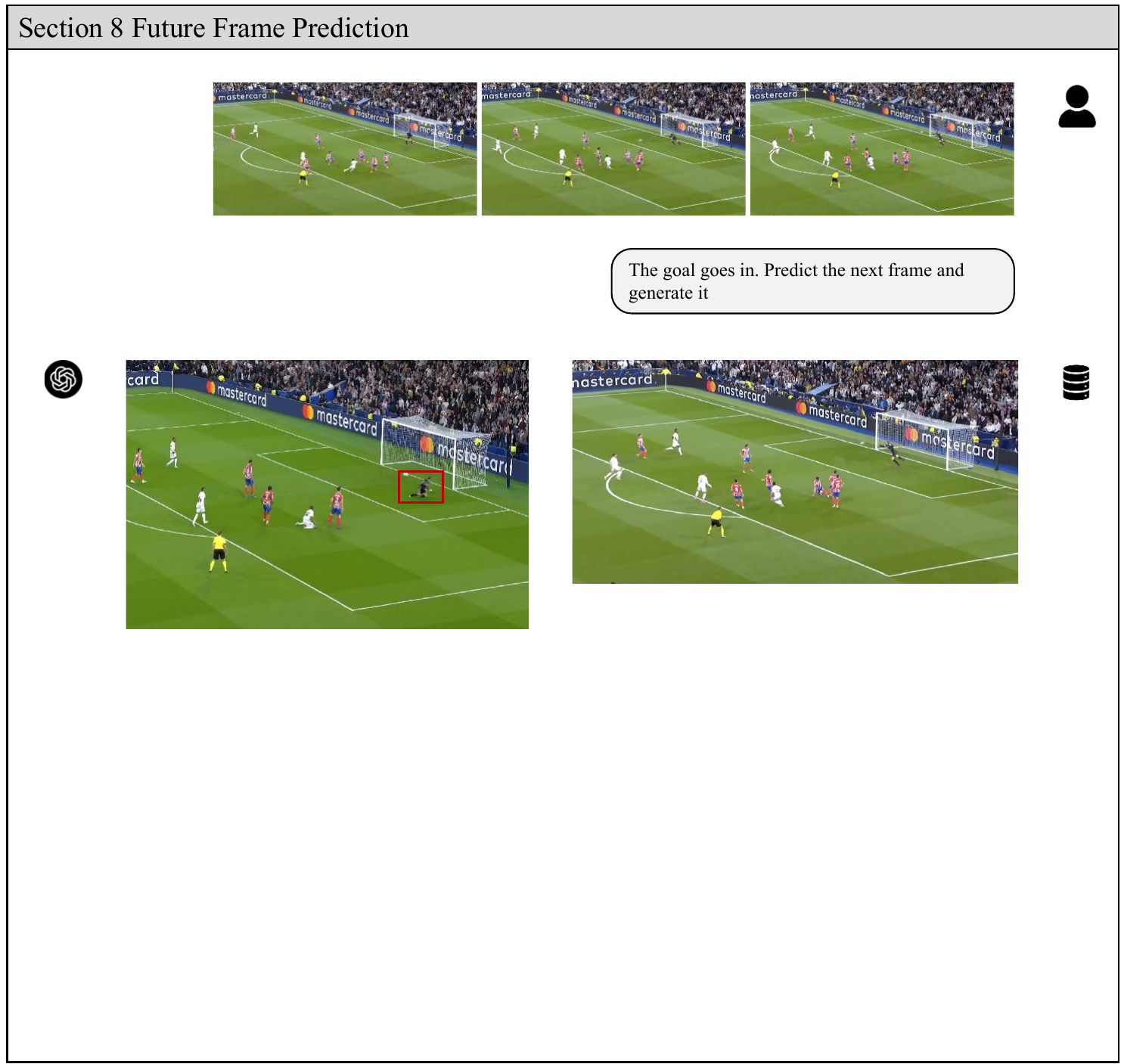}
    \caption[Sec~\ref{sec:temporal}: Future Frame Prediction]{Example of future frame prediction by \modelname. Although the overall scene and context of a goal being scored are maintained, the predicted frame fails to capture the dynamic motion of the goalkeeper accurately.  }
    \label{fig:8_1_3}
\end{figure}

\begin{figure}[h]
    \centering
    \includegraphics[width=1.0\linewidth]{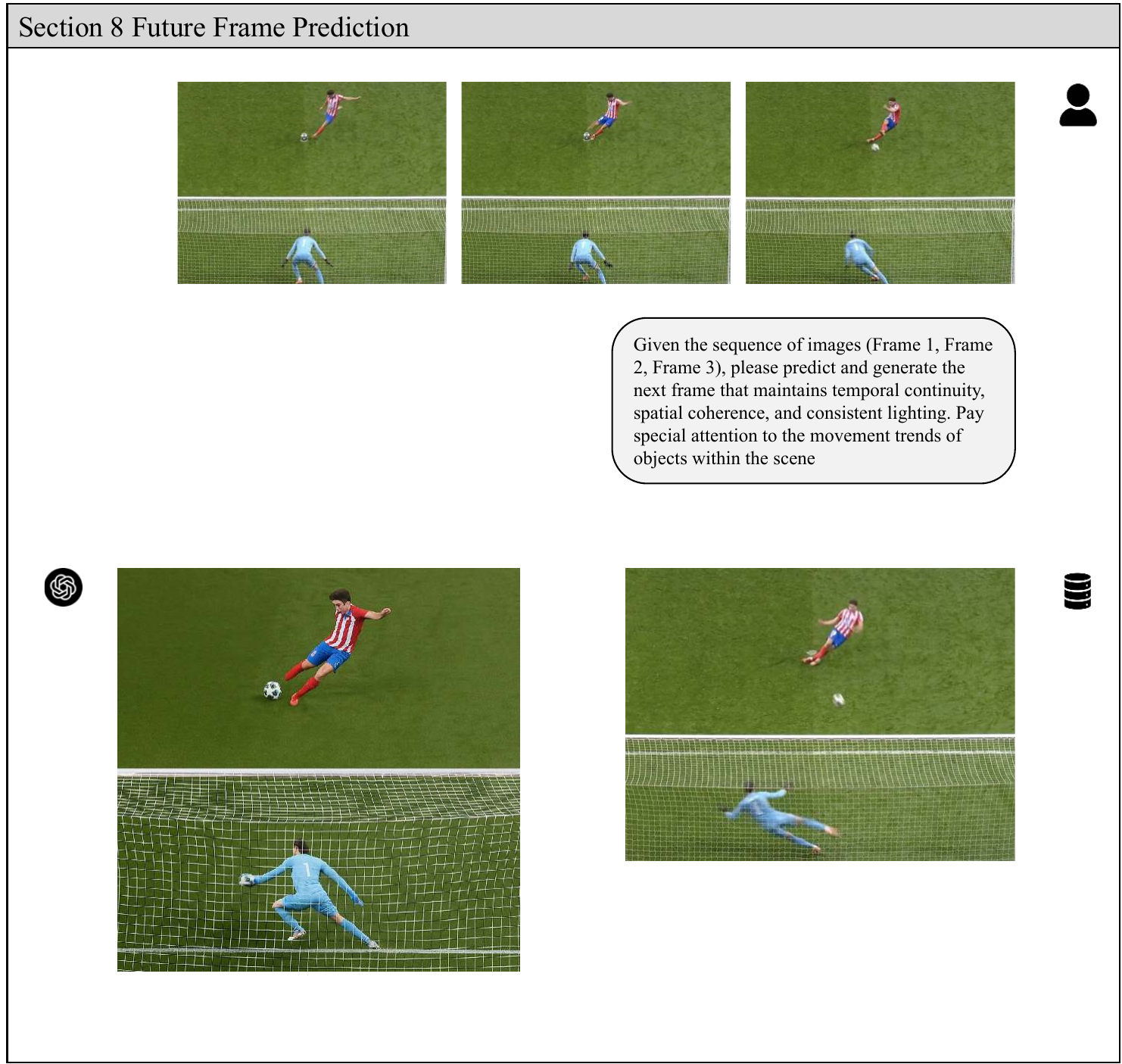}
    \caption[Sec~\ref{sec:temporal}: Future Frame Prediction]{Example of future frame prediction by \modelname. The generated frame maintains strong spatial coherence and visual quality, accurately reconstructing the scene layout. However, it fails to capture the correct motion trajectory of key elements—such as the player and the ball—indicating limited understanding of action dynamics despite solid scene rendering.}
    \label{fig:8_1_4}
\end{figure}

\textbf{Future Frame Prediction.}
This task involves generating the next plausible frame in a visual sequence, given the initial few frames. The model is expected to predict a visually consistent and temporally reasonable continuation of scene dynamics, object motion, and structural transitions.
As shown in Fig.~\ref{fig:8_1_0} to Fig.~\ref{fig:8_1_4}, \modelname demonstrates the ability to maintain global scene layouts and preserve image fidelity. However, it consistently struggles to model precise object motion and directionality. For instance, in Fig.~\ref{fig:8_1_0}, although the spatial structure under the bridge is roughly preserved, the bridge’s predicted continuation fails to match the previous trajectory. In Fig.~\ref{fig:8_1_1}, the model generates a plausible car scene, but completely ignores the expected motion of the vehicle, indicating weak motion extrapolation. Similarly, in Fig.~\ref{fig:8_1_2}, the drifting clouds fail to align with the observed trajectory, suggesting difficulties in modeling natural dynamics. While Fig.~\ref{fig:8_1_3} and Fig.~\ref{fig:8_1_4} show correct semantic context (e.g., scoring a goal), the details of the ball and goalkeeper's motion are inaccurate. These examples highlight that \modelname primarily prioritizes static scene generation over temporal causality.

\begin{figure}[h]
    \centering
    \includegraphics[width=1.0\linewidth]{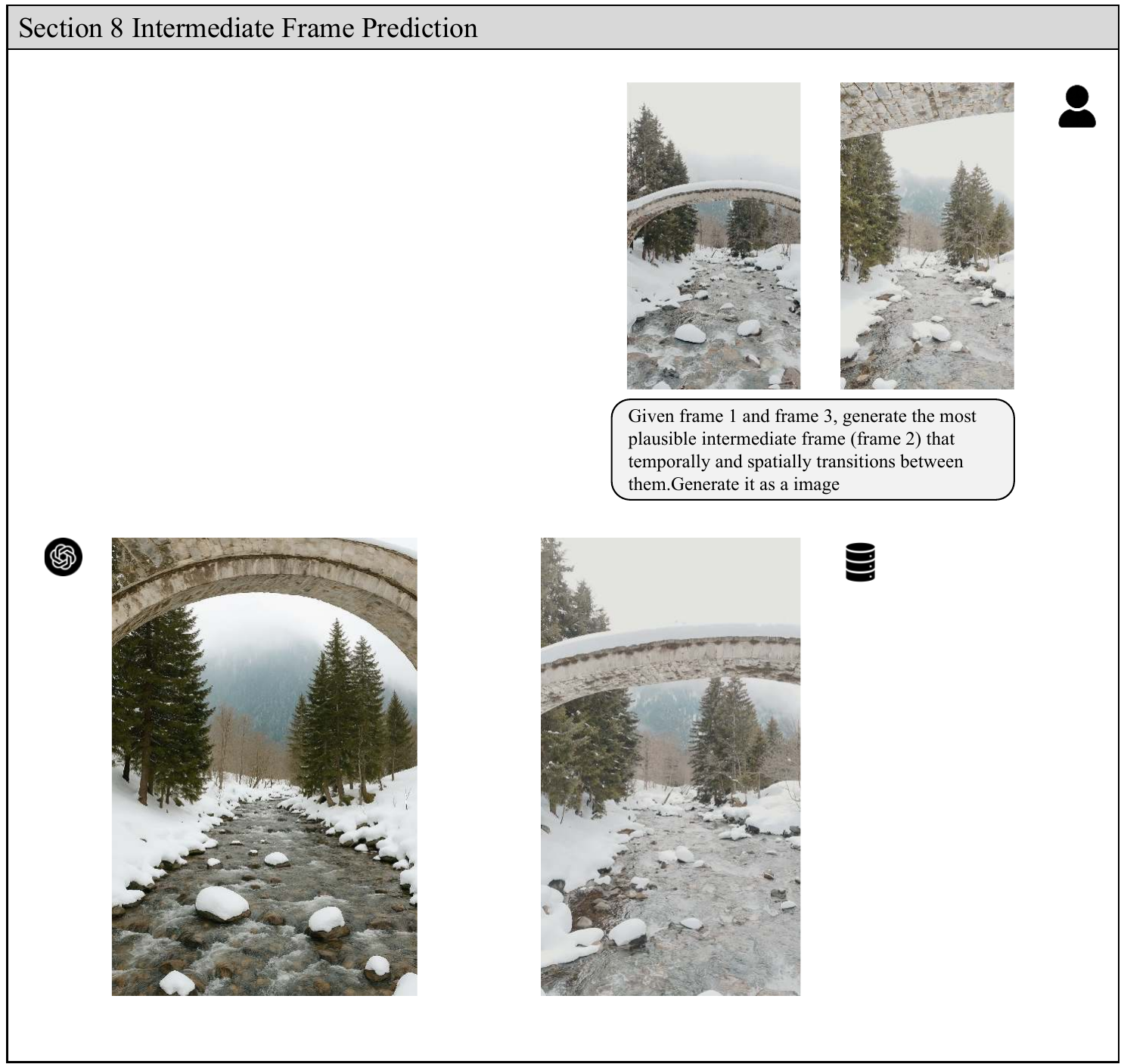}
    \caption[Sec~\ref{sec:temporal}: Intermediate Frame Prediction]{Example of intermediate frame prediction by \modelname. Given the first and third frames, the model successfully generates a visually coherent intermediate frame that preserves spatial layout and motion continuity. Aside from slight color tone discrepancies, the reconstruction and temporal prediction are handled effectively.}
    \label{fig:8_2_0}
\end{figure}

\begin{figure}[h]
    \centering
    \includegraphics[width=1.0\linewidth]{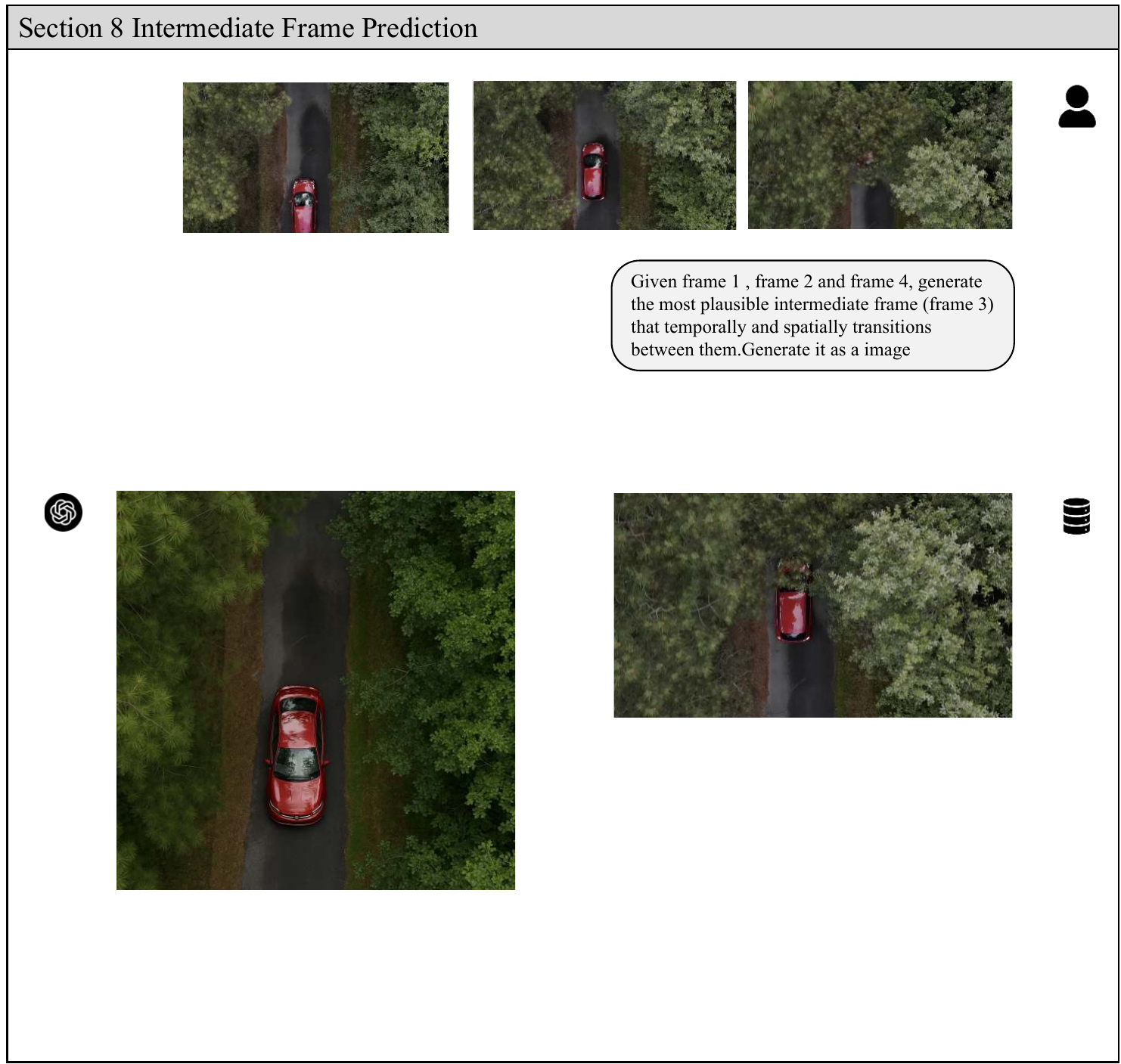}
    \caption[Sec~\ref{sec:temporal}: Intermediate Frame Prediction]{Example of intermediate frame prediction by \modelname. The generated frame fails to align with the spatial and temporal patterns observed in the input sequence. In particular, the car’s orientation is incorrectly reconstructed, and the overall scene lacks visual coherence, indicating poor understanding of motion continuity and spatial consistency. }
    \label{fig:8_2_1}
\end{figure}

\begin{figure}[h]
    \centering
    \includegraphics[width=1.0\linewidth]{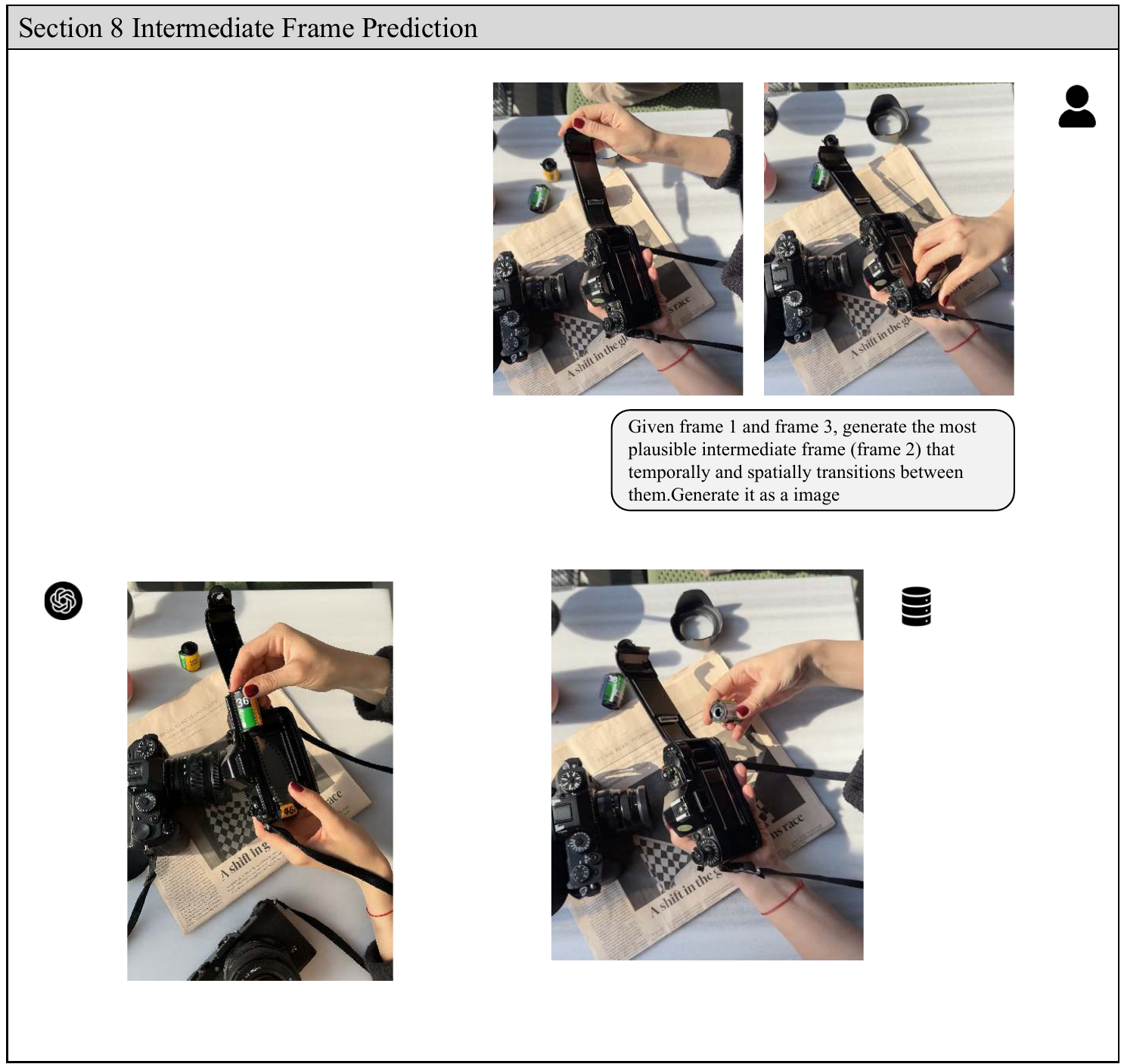}
    \caption[Sec~\ref{sec:temporal}: Intermediate Frame Prediction]{Example of intermediate frame prediction by \modelname. The generated frame demonstrates accurate spatial understanding and smooth temporal transition, effectively bridging the motion and object positions between the two input frames. Both the hand pose and film placement are well reconstructed, showcasing strong reasoning and visual consistency.}
    \label{fig:8_2_2}
\end{figure}

\begin{figure}[h]
    \centering
    \includegraphics[width=1.0\linewidth]{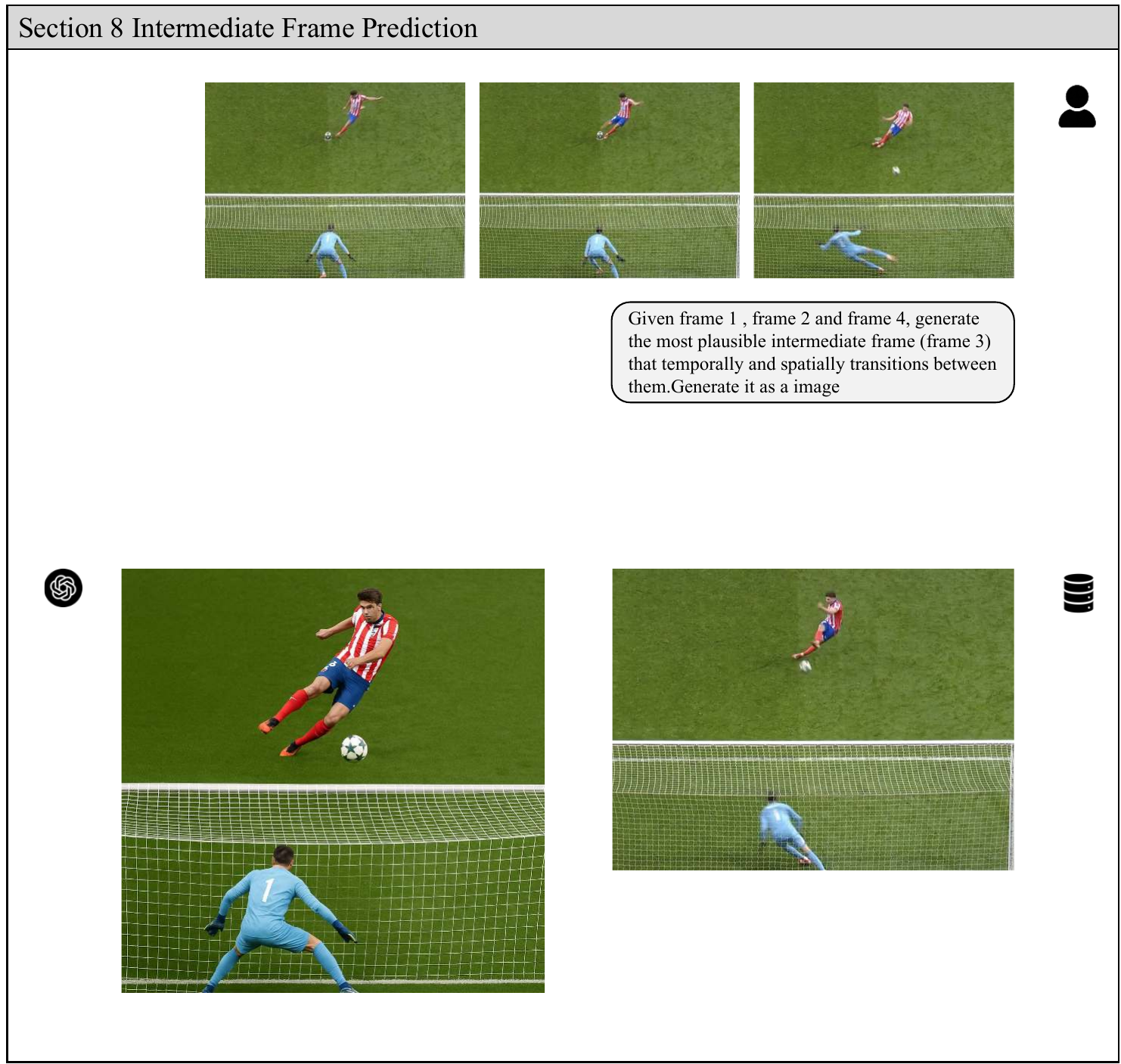}
    \caption[Sec~\ref{sec:temporal}: Intermediate Frame Prediction]{Example of intermediate frame prediction by \modelname. The generated frame fails to preserve the spatial and temporal coherence of the input sequence. The positions and interactions between the player, ball, and goalkeeper are inconsistent with the surrounding frames, indicating a breakdown in understanding the underlying motion dynamics and object relationships.}
    \label{fig:8_2_3}
\end{figure}

\begin{figure}[h]
    \centering
    \includegraphics[width=1.0\linewidth]{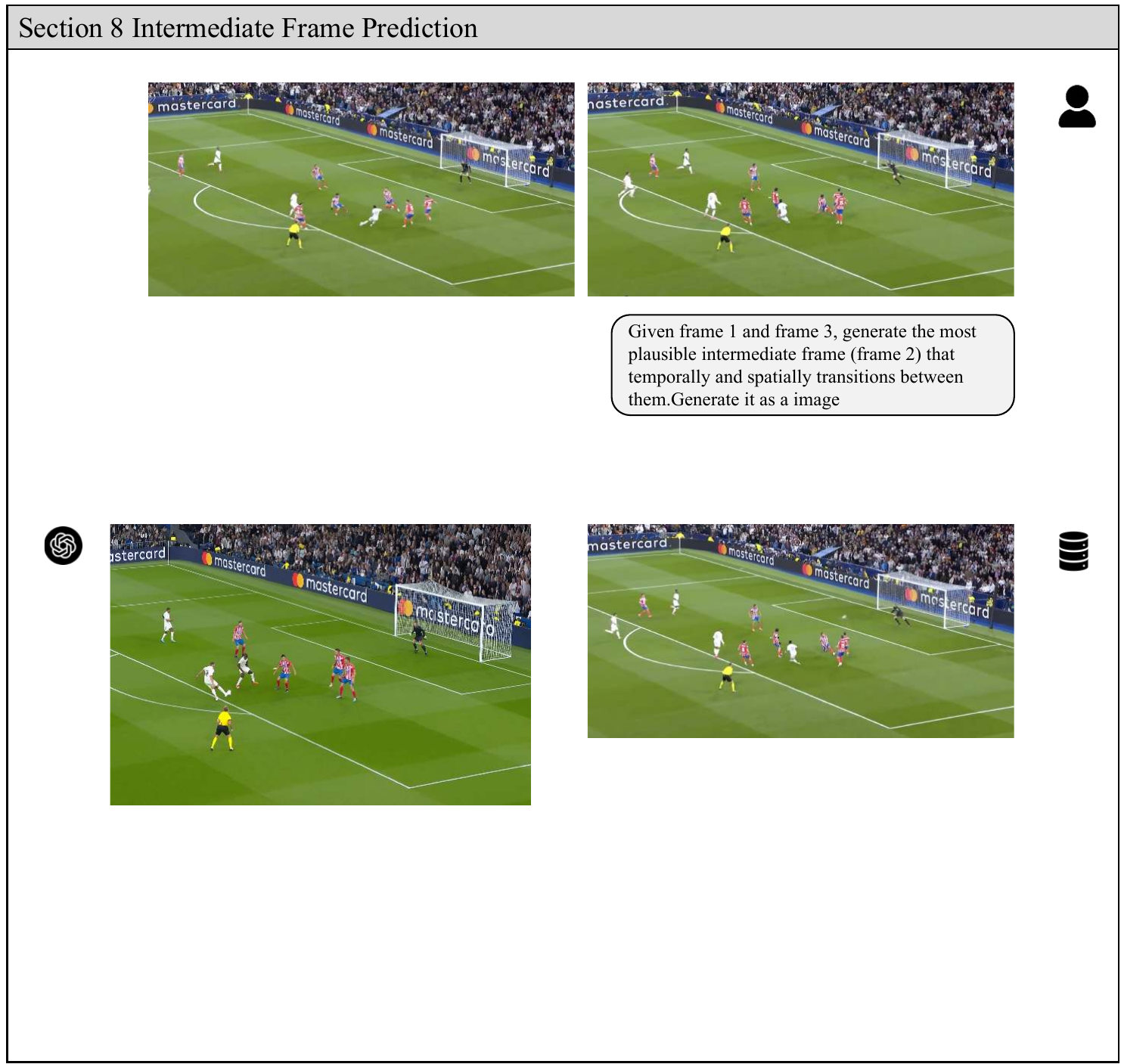}
    \caption[Sec~\ref{sec:temporal}: Intermediate Frame Prediction]{Example of intermediate frame prediction by \modelname. The generated frame exhibits major inconsistencies in the spatial arrangement of players and the shooting action. The kicker’s pose and position, as well as the placement of surrounding players, deviate significantly from the temporal flow established in the input frames, revealing poor motion understanding and scene continuity.}
    \label{fig:8_2_4}
\end{figure}

\textbf{Intermediate Frame Prediction.}
This task requires the model to generate a plausible intermediate frame, given both the starting and ending frames. The model must infer continuous motion trajectories and interpolate realistic transitions between visual states.
As shown in Fig.~\ref{fig:8_2_0} to Fig.~\ref{fig:8_2_4}, the results are mixed. In Fig.~\ref{fig:8_2_0}, the model produces a semantically coherent intermediate frame with reasonable alignment and motion. However, in Fig.~\ref{fig:8_2_1}, the car’s orientation is reversed, indicating poor spatial continuity. While Fig.~\ref{fig:8_2_2} shows strong temporal alignment between hand and film roll, Fig.~\ref{fig:8_2_3} and Fig.~\ref{fig:8_2_4} reveal serious failures in understanding player position, motion trajectory, and semantic relationships in soccer scenes. These examples demonstrate the challenge of fine-grained motion interpolation, where \modelname often falls short in preserving both local interactions and global structure.

\begin{figure}[h]
    \centering
    \includegraphics[width=1.0\linewidth]{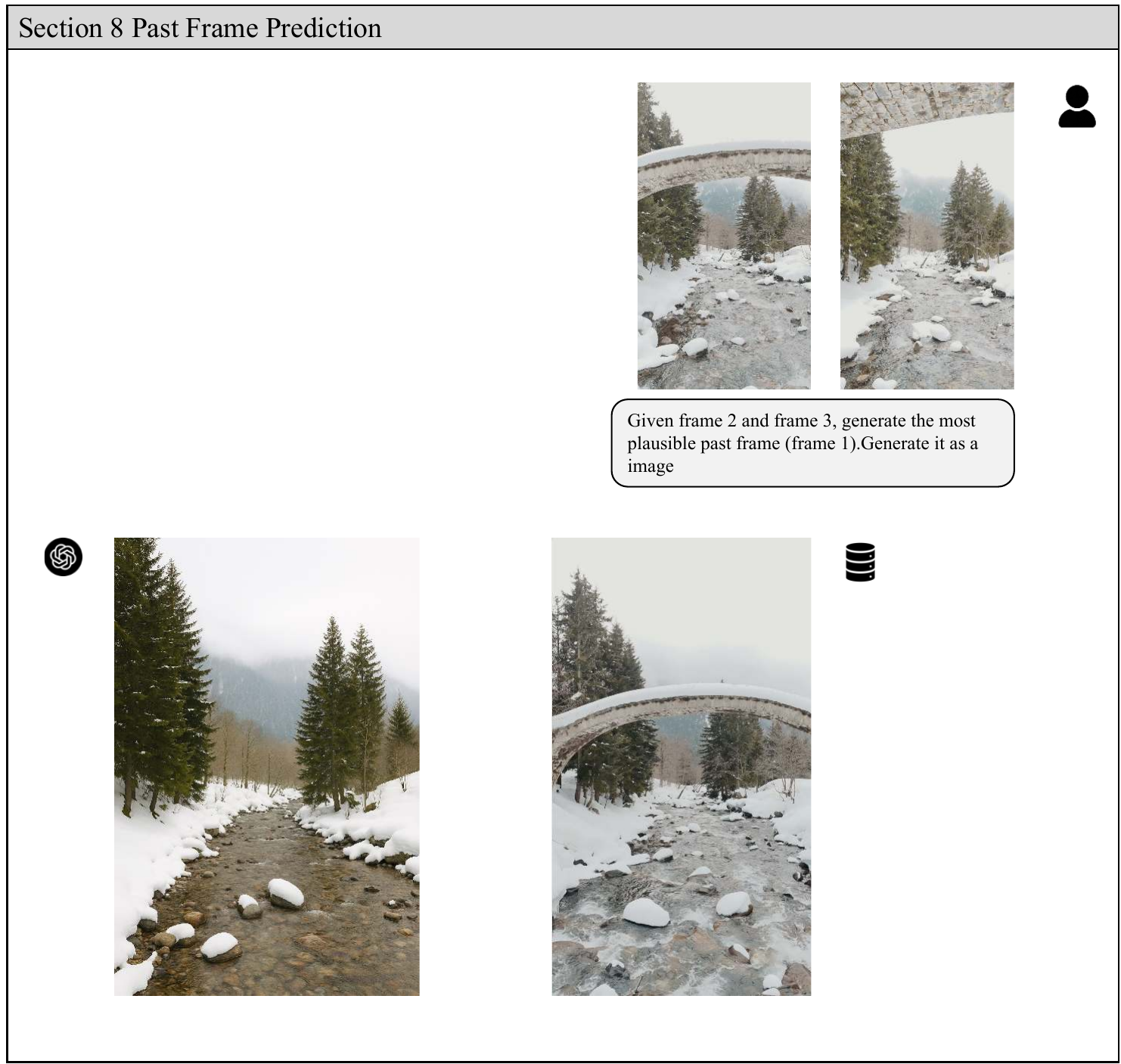}
    \caption[Sec~\ref{sec:temporal}: Past Frame Prediction]{Example of past frame prediction by \modelname. Although the model attempts to infer a plausible earlier frame from the given sequence, it fails to preserve key structural elements—most notably the bridge, which is entirely missing. This highlights significant limitations in scene reconstruction under backward temporal reasoning.}
    \label{fig:8_3_0}
\end{figure}

\begin{figure}[h]
    \centering
    \includegraphics[width=1.0\linewidth]{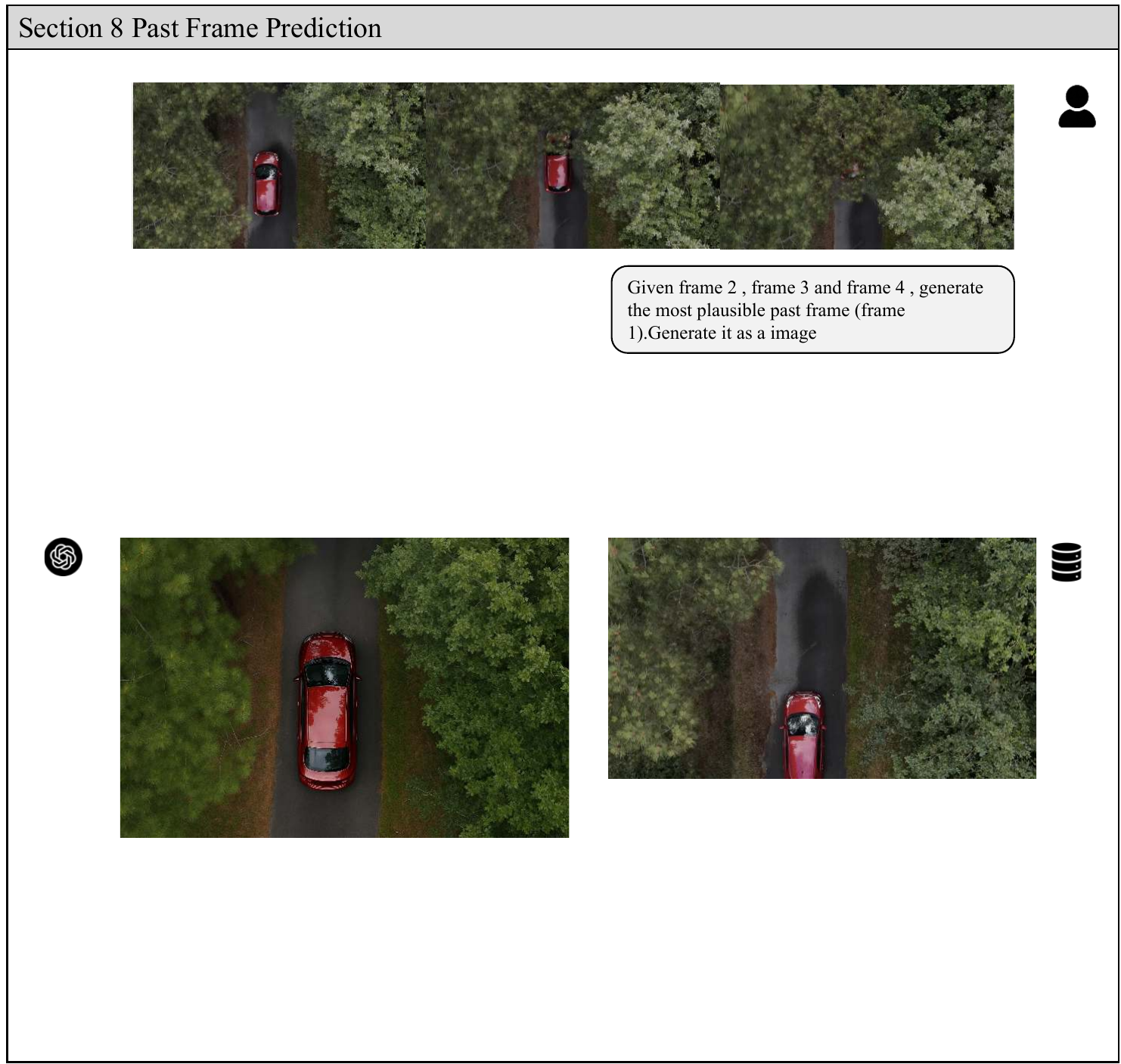}
    \caption[Sec~\ref{sec:temporal}: Past Frame Prediction]{Example of past frame prediction by \modelname. The generated image appears to prioritize scene reconstruction, producing a visually coherent frame. However, it fails to reflect the expected temporal regression of the car’s position, indicating that the model neglects backward motion reasoning in favor of static visual fidelity. }
    \label{fig:8_3_1}
\end{figure}

\begin{figure}[h]
    \centering
    \includegraphics[width=1.0\linewidth]{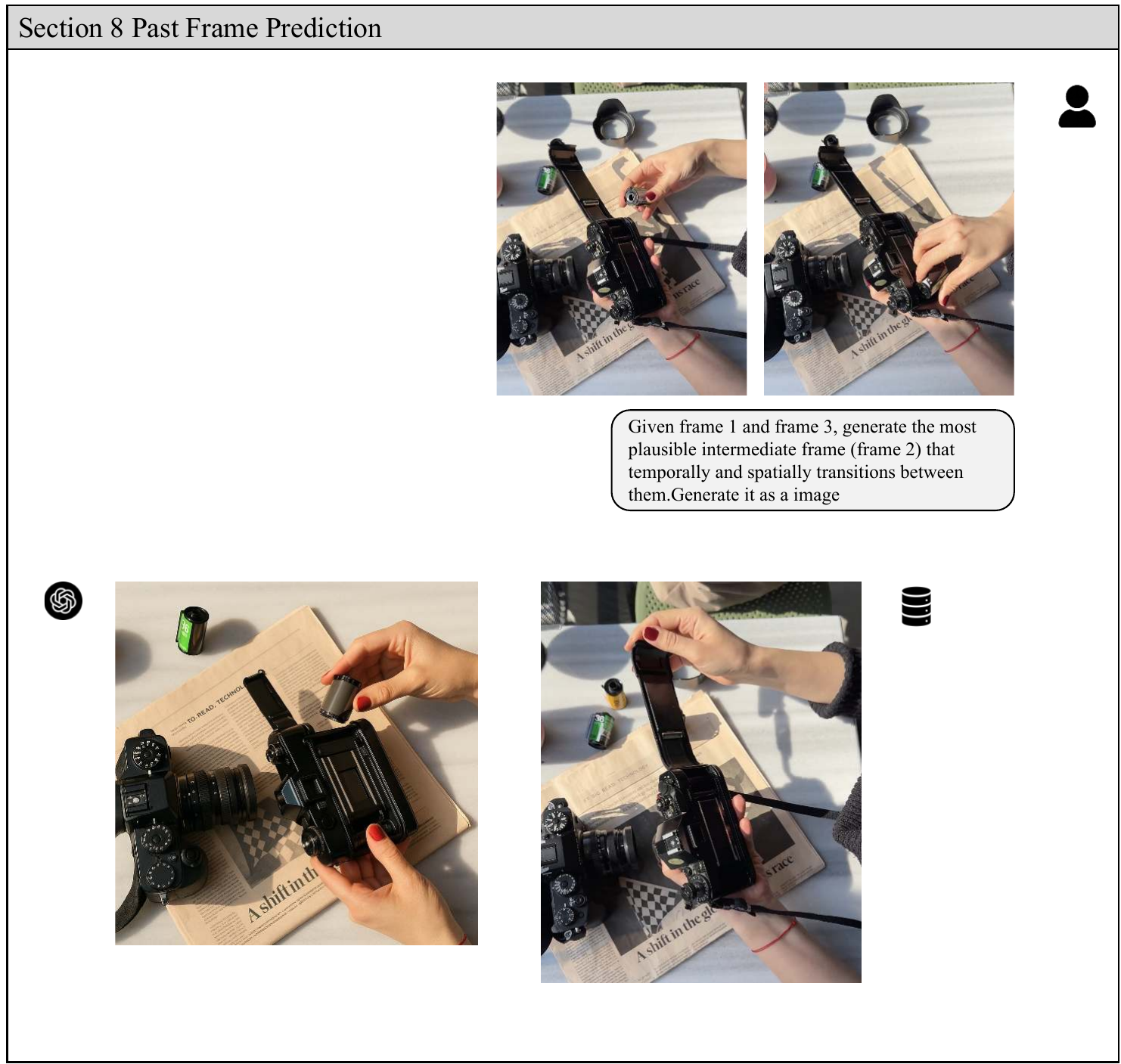}
    \caption[Sec~\ref{sec:temporal}: Past Frame Prediction]{Example of past frame prediction by \modelname. Although the generated image visually resembles the surrounding frames, it fails to capture the correct temporal progression of the action—specifically, the interaction between the hand and the film roll. This suggests limited capability in reasoning about the sequence of physical actions over time.}
    \label{fig:8_3_2}
\end{figure}

\begin{figure}[h]
    \centering
    \includegraphics[width=1.0\linewidth]{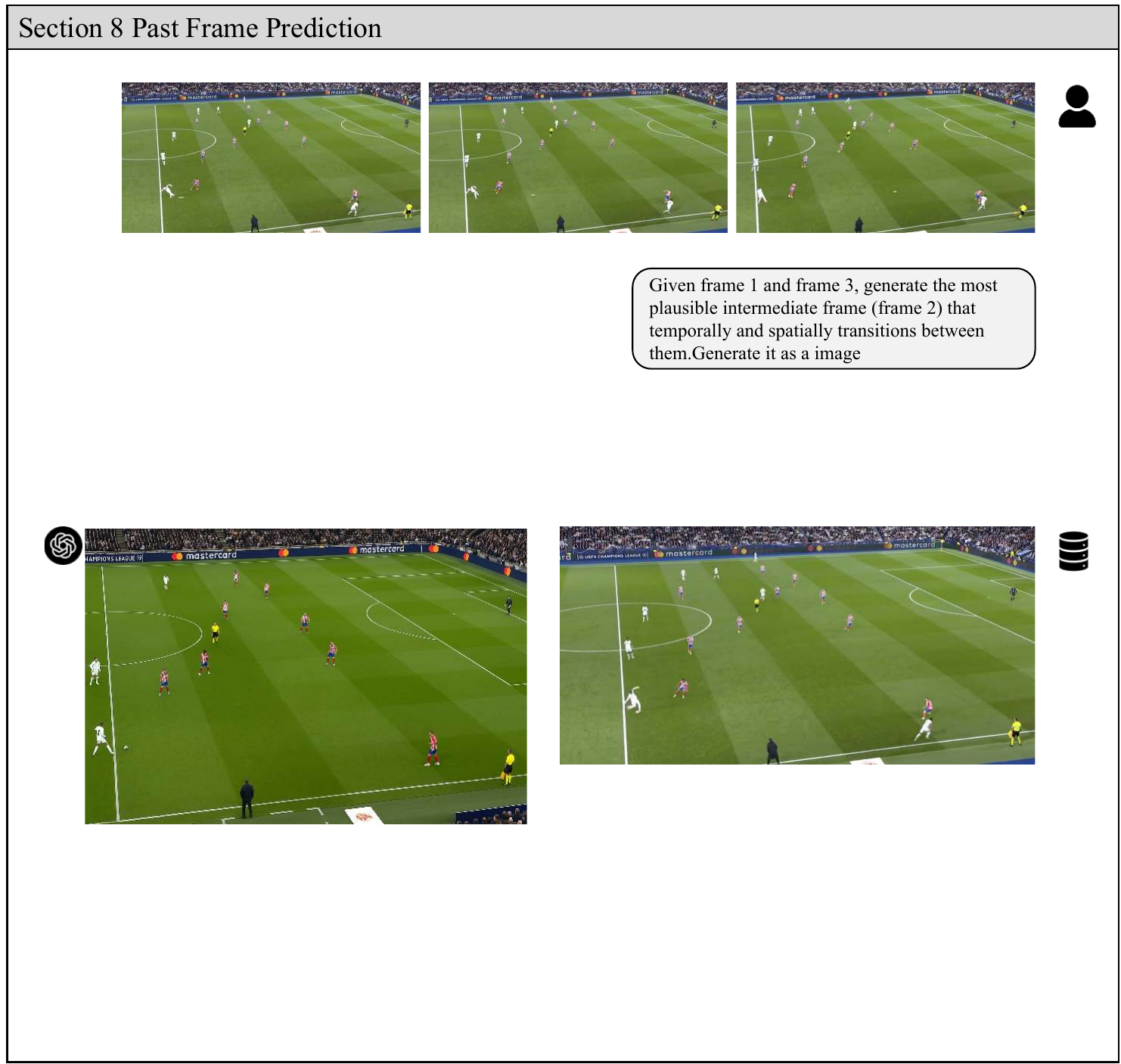}
    \caption[Sec~\ref{sec:temporal}: Past Frame Prediction]{Example of past frame prediction by \modelname. The model demonstrates a reasonable understanding of the overall temporal progression in the scene. However, the interaction between the goalkeeper and the goal structure is physically inconsistent, revealing challenges in accurately modeling fine-grained spatial relationships.}
    \label{fig:8_3_3}
\end{figure}

\begin{figure}[h]
    \centering
    \includegraphics[width=1.0\linewidth]{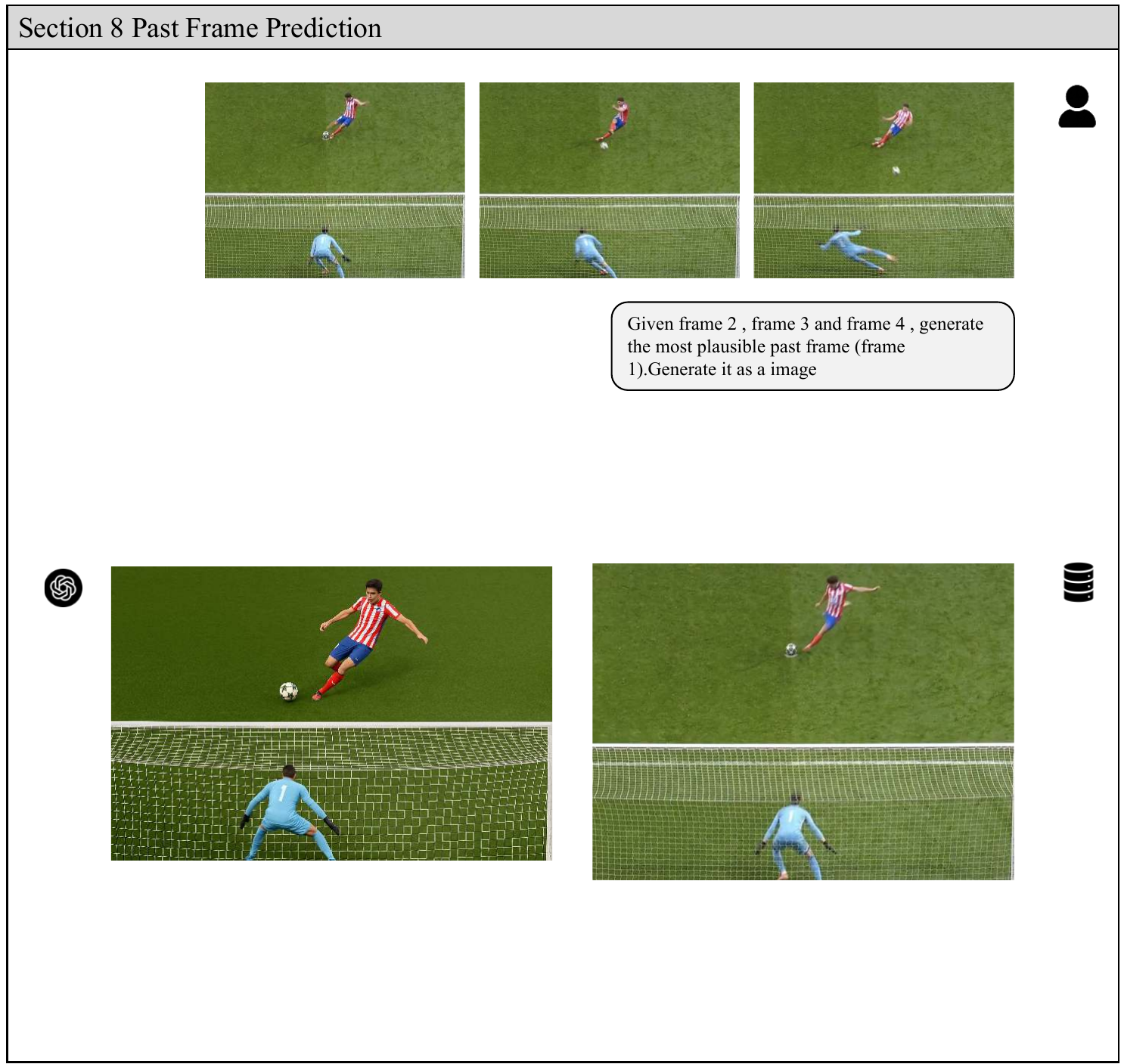}
    \caption[Sec~\ref{sec:temporal}: Past Frame Prediction]{Example of past frame prediction by \modelname. The model successfully captures the temporal dynamics of the ball-kicking action, demonstrating decent motion reasoning. However, the reconstruction of the surrounding players is inaccurate, with noticeable inconsistencies in their positions and postures, revealing challenges in maintaining global scene consistency.}
    \label{fig:8_3_4}
\end{figure}

\textbf{Past Frame Prediction.}
This task evaluates the model’s ability to hallucinate plausible earlier frames given subsequent ones. It requires temporal inversion, causal reasoning, and backward consistency of object positions and motion.
As shown in Fig.~\ref{fig:8_3_0} to Fig.~\ref{fig:8_3_4}, the model tends to focus on generating semantically valid but temporally inconsistent scenes. In Fig.~\ref{fig:8_3_0}, the predicted past frame omits critical structural elements like the bridge, revealing flaws in memory-based reconstruction. In Fig.~\ref{fig:8_3_1}, the car's position does not reflect reverse movement, and in Fig.~\ref{fig:8_3_2}, the hand-action interaction is misaligned with the temporal flow. While Fig.~\ref{fig:8_3_3} shows some temporal consistency, interactions remain physically implausible. Fig.~\ref{fig:8_3_4} captures the kicking motion effectively, but inconsistencies in surrounding players highlight the model's limited understanding of spatial continuity in dynamic group interactions.

Across all tasks, \modelname demonstrates a general capability to reconstruct coherent individual frames with high visual fidelity. However, its understanding of temporal consistency, motion dynamics, and causality remains limited. The model often focuses on plausible scene completion rather than faithfully continuing motion or reasoning over time. Failures are especially apparent in sports scenes and scenes involving complex interactions or occlusions, where precise alignment across frames is critical.

\clearpage

\clearpage

\clearpage

\clearpage

\clearpage

\clearpage

\clearpage

\clearpage

\clearpage

\clearpage

\clearpage

\clearpage

\clearpage

\clearpage

\clearpage
\clearpage

\section{Limitations}
\label{sec:sc}

\subsection{\modelname is Not Yet a World Model}
\label{sec:world_model_limitation}

The ability to imitate and reason about the real world is a foundational requirement for a model to be considered a \textit{world model}. In principle, a world model should be able to understand physical structure, causality, common sense, and the temporal evolution of events, allowing it to reason about how the world behaves under various conditions.

However, our experiments reveal that \modelname falls short of reliably generating knowledge-grounded visual content across multiple scientific domains. In physics-related generation (Sec.~\ref{sec:physical}), while the model succeeds in many commonsense cases, it fails to capture key physical principles in scenarios involving light refraction and thermodynamics. In chemistry tasks (Sec.~\ref{sec:chemical}), the model often produces scientifically inaccurate results, such as mislabeled substances, incorrect reaction phenomena, or structurally flawed molecular diagrams. Biology-related tasks (Sec.~\ref{sec:biology}) further expose limitations in domain-specific understanding, with frequent errors in microbial coloring, ecosystem modeling, and biomedical segmentation.
Similarly, in mathematics (Sec.~\ref{sec:math}), \modelname struggles with representing precise spatial and logical relationships, often misplacing annotations or violating geometric constraints. Even in agriculture (Sec.~\ref{sec:agriculture}), where the visual domain is more natural, the model still fails to associate correct visual semantics with textual prompts, generating wrong crop types, incorrect labels, or hallucinated scene details. These findings indicate that despite its strong general vision-language capabilities, \modelname lacks the structured scientific grounding and symbolic reasoning needed to serve as a reliable knowledge-driven visual generator.

In spatially-aware generation tasks (Sec.~\ref{sec:spatial}), the model fails to consistently preserve structural layouts and spatial constraints. It often disregards reference layouts, misrepresents pose orientations, or alters semantic regions during inpainting or segmentation. These errors imply that \modelname does not maintain a coherent internal representation of space, which is essential for reasoning about object positions, geometry, and physical feasibility.

In temporally-aware generation (Sec.~\ref{sec:temporal}), \modelname shows limited understanding of motion continuity and causality. It struggles to infer plausible future or past frames in dynamic scenes and often produces temporally inconsistent actions or object trajectories. Such behavior highlights the absence of robust temporal modeling mechanisms within the generation process.

Taken together, these observations indicate that \modelname, despite its impressive multimodal generation ability, cannot yet be regarded as a world model. It lacks grounded representations of physics, commonsense, and time—key attributes necessary for coherent and causally consistent world understanding. Closing this gap will likely require new inductive biases, structured world knowledge integration, and training objectives beyond pattern learning.

\subsection{Limitations in Generation Process Control}
\label{sec:lowlevel_limitation}

Despite its strong generative capabilities, \modelname exhibits clear limitations in controlling low-level visual properties, such as image resolution, aspect ratio, and pixel-wise numerical constraints.

As observed in Sec.~\ref{sec:res} and Sec.~\ref{sec:aspect}, the model cannot accurately follow user-specified resolutions or aspect ratios. It is restricted to producing images in only three predefined resolutions, making it impossible to generate outputs at arbitrary sizes or layouts. Even when different resolutions are explicitly requested, the model defaults to the closest fixed resolution, which restricts its usability in applications requiring strict layout or scaling control.

Furthermore, as shown in Sec.~\ref{sec:numerical}, \modelname struggles to handle precise numerical conditions. It fails to produce exact RGB values when asked to generate black or grayscale images, and introduces unexpected noise or variability in what should be pixel-accurate segmentation outputs. This behavior indicates that the model optimizes for perceptual plausibility rather than low-level numerical accuracy.

These findings highlight a fundamental limitation: \modelname lacks mechanisms for fine-grained control over structural and quantitative properties in image generation. This restricts its effectiveness in scientific, industrial, or graphics applications where pixel-level consistency, spatial accuracy, or numerical fidelity are critical.

\subsection{Limitations in Spatial Alignment}
\label{sec:spatial_limitation}

One of the critical limitations observed in \modelname is its difficulty in handling tasks that require precise spatial alignment between visual inputs and outputs. While the model demonstrates strong semantic understanding and plausible image generation capabilities, it often struggles with faithfully preserving spatial structures, positional constraints, and layout consistency across various tasks.

This limitation becomes evident in tasks such as spatial control (Sec.~\ref{sec:spatial_control}), low-level synthesis(Sec.~\ref{sec:ir}), and segmentation (Sec.~\ref{sec:seg}). For example, in layout-to-image generation, the model frequently ignores object positions or produces misaligned placements even when the layout is explicitly defined. In sketch-based and pose-based generation, although the model captures high-level semantics, it often fails to strictly follow spatial cues—such as body orientation or object alignment—resulting in images that are semantically correct but spatially inconsistent.

Similarly, in segmentation tasks, \modelname often misinterprets the visual instruction. It may assign inconsistent colors to the same class in semantic segmentation or fail to preserve instance boundaries in instance segmentation. The generated outputs tend to resemble stylized interpretations rather than precise spatial mappings, suggesting that the model prioritizes visual plausibility over pixel-level accuracy.

These findings suggest that \modelname lacks strong inductive bias for spatial alignment, which is essential for tasks that require precise spatial grounding. In contrast, many recent vision-language models designed for unified multimodal perception\cite{jiao2024lumen,xia2024gsva,zhang2024omg}, such as those trained with spatially-aligned outputs like segmentation masks or bounding boxes, tend to incorporate architectural components (e.g., spatial token decoders, vision backbones with explicit positional encoding) that enforce spatial consistency. Similarly, models like ControlNet~\cite{zhang2023adding} integrate conditional control branches that directly guide the spatial generation process, leading to improved alignment in tasks such as edge-to-image or pose-to-image generation.
The general-purpose architecture of \modelname, while effective for high-level visual understanding and instruction following, may lack such spatial priors at the structural level. As a result, it may require either significantly more aligned training data, or architectural augmentation (e.g., external control modules or auxiliary spatial heads) to improve its ability to faithfully align outputs with spatial inputs in generation tasks.

\subsection{Limitations in Instruction Alignment}
\label{sec:instruction_alignment}

Another key limitation observed in \modelname is its inconsistent alignment with provided instructions, particularly when simultaneously conditioned on both text and image prompts. While the model demonstrates strong multimodal comprehension in general, it often fails to accurately reflect all details or constraints from the input instructions in the generated outputs.

This misalignment appears across a variety of tasks. For instance, in image editing (Sec.~\ref{sec:edit}) and personalization (Sec.~\ref{sec:personalization}) scenarios, specific attributes described in the prompt may be ignored or altered (e.g., missing accessories, color mismatches). Similarly, in temporal generation (Sec.~\ref{sec:temporal}) and story visualization (Sec.~\ref{sec:story}), the model sometimes drops key narrative elements or misrepresents the intended progression. Furthermore, in discriminative tasks (Sec.~\ref{sec:discrimative}), \modelname often fails to follow task-level instructions defined by either task names or natural language prompts. Despite efforts to specify the expected task type, such as “semantic segmentation,” “object detection,” or “pose estimation”, the model frequently produces mismatched outputs or ignores the instruction altogether. This behavior also persists when in-context learning is applied, indicating that the model lacks robust task disambiguation and flexible prompt conditioning (Sec.~\ref{sec:seg}).

These inconsistencies suggest that while \modelname can interpret rich instructions, it lacks a robust grounding mechanism to ensure strict adherence. This may be due to limited alignment supervision during training or the model's tendency to prioritize global plausibility over fine-grained control. Improving instruction alignment will likely require more structured data or enhanced training objectives.

\clearpage
\section{Conclusion}
In this paper, we conducted a comprehensive qualitative evaluation of \modelname's image generation capabilities, covering a wide spectrum of tasks from traditional synthesis to advanced multimodal reasoning. Our evaluation framework is structured around six key dimensions, and we summarize the model’s performance as follows:

\begin{itemize}
\item \textbf{Overall Characteristics.} As discussed in Sec.~\ref{sec:overall}, \modelname exhibits clear limitations in generation process control. It fails to generate images at arbitrary resolutions, cannot flexibly adjust aspect ratios, and struggles to produce numerically accurate visual outputs such as precise RGB values or segmentation masks.

\item \textbf{Visual Synthesis and Conditional Alignment.} As shown in Sec.~\ref{sec:tradition}, \modelname performs strongly across a wide range of traditional image generation tasks. It shows solid capabilities in text-conditioned and multimodal-conditioned generation, stylization, and even low-level image processing tasks such as colorization and relighting.

\item \textbf{Visual Understanding.} In Sec.~\ref{sec:discrimative}, \modelname is evaluated on discriminative tasks with both visual and textual outputs. Although it can align outputs with general task goals, it often relies on global semantics rather than precise reasoning, leading to frequent hallucinations, spatial misalignments, and misunderstanding of task definitions.

\item \textbf{Knowledge and Commonsense-based Generation.} Sec.~\ref{sec:knowledge} and Sec.~\ref{sec:commonsense} evaluate the model's ability to encode domain-specific knowledge and commonsense reasoning. While \modelname shows surface-level competence, it frequently fails to follow scientific principles and factual accuracy, particularly in fields such as physics, chemistry, biology, and mathematics. This suggests that the model lacks the foundational grounding to act as a true world model (Sec.~\ref{sec:world_model_limitation}).

\item \textbf{Spatial Reasoning.} In Sec.~\ref{sec:spatial}, the model is tested on spatially-conditioned tasks such as layout-to-image, pose-to-image, and segmentation. Although \modelname captures high-level semantics, it struggles to maintain pixel-level spatial alignment. These issues likely stem from a lack of architectural inductive biases for structured spatial reasoning (Sec.~\ref{sec:spatial_limitation}).

\item \textbf{Temporal Reasoning.} In Sec.~\ref{sec:temporal}, the model is evaluated on temporally-aware generation tasks. The results indicate emerging abilities in handling sequence dynamics, but a lack of consistent scene continuity, causality modeling, and temporal memory remains a significant bottleneck.
\end{itemize}

Despite its remarkable performance in general synthesis and instruction following, \modelname still faces key limitations in four major aspects: (1) its inability to model real-world knowledge across scientific domains and temporal processes (\textit{Sec.~\ref{sec:world_model_limitation}}), (2) its lack of fine-grained generation control (\textit{Sec.~\ref{sec:lowlevel_limitation}}), (3) insufficient spatial alignment in structure-aware tasks (\textit{Sec.~\ref{sec:spatial_limitation}}), and (4) difficulty in strictly following task instructions and prompts (\textit{Sec.~\ref{sec:instruction_alignment}}). We hope our findings not only provide insights into the current limitations of \modelname but also inspire future research directions toward more controllable, grounded, and structured multimodal generation systems.
%\section*{Acknowledgment}
%We express our gratitude to all contributors from OpenAI for their technical efforts on the GPT-4V project~\cite{gpt4,gpt4v,gpt4vcontribution,gpt4vblog}, and we are profoundly thankful to OpenAI for granting early access to their remarkable tool. Our sincere appreciation goes to Misha Bilenko for his invaluable guidance and support. We also extend heartfelt thanks to our Microsoft colleagues for their insights, with special acknowledgment to John Montgomery, Marco Casalaina, Gregory Buehrer, Nguyen Bach, Gopi Kumar, Luis Vargas, Kun Wu, Meenaz Merchant, Jianfeng Gao, Matt Lungren, Sheela Agarwal, Yumao Lu, Thomas Soemo, Fisayo Okikiolu, Ce Liu, Michael Zeng, Faisal Ahmed, Ehsan Azarnasab, and Lin Liang for their constructive feedback. We also thank Yingkai Yu for helping to create screenshots on GUI Navigation.

\clearpage
{
\bibliographystyle{plain}
\bibliography{egbib}
}

\end{document}